\documentclass[11pt,a4paper]{article}
\usepackage{times,latexsym}
\usepackage{url}
\usepackage[T1]{fontenc}

\usepackage{lipsum}  

\usepackage{enumitem}

\usepackage{multirow}
\usepackage{multicol}
\usepackage{booktabs}
\usepackage{tabularray}
\usepackage{soul}

\usepackage{dsfont}
\usepackage{algorithm}
\usepackage{algorithmic}

\usepackage[dvipsnames]{xcolor}
\newcommand{\changed}[1]{#1}

\usepackage{colortbl}
\definecolor{Gray}{gray}{0.9}
\usepackage{multirow,adjustbox}
\usepackage{diagbox}
\newcolumntype{M}[1]{>{\centering\arraybackslash}m{#1}}

\usepackage{bm,amsmath,amssymb,fancyhdr,graphicx, bbm}
\usepackage{mathtools}
\usepackage{hyperref}
\usepackage{graphicx}
\usepackage{subcaption}
\usepackage{tikz}
\usetikzlibrary{shapes, arrows, positioning}
\usepackage{soul}

\newcommand{\wit}[0]{{\tt WIT}}

\usepackage[acceptedWithA]{tacl2021v1}
\usepackage{xspace,mfirstuc,tabulary}

\newif\iftaclinstructions
\taclinstructionsfalse 
\iftaclinstructions
\renewcommand{\confidential}{}
\renewcommand{\anonsubtext}{(No author info supplied here, for consistency with
TACL-submission anonymization requirements)}
\newcommand{\instr}
\fi

\iftaclpubformat 

\else

\fi

\fancypagestyle{firstpagefooter}{%
  \fancyhf{}                              
  \fancyfoot[C]{
    \parbox{\textwidth}{%
        \centering
        \textit{This is a preprint of an article that has been accepted for publication in} \\ \textit{Transactions of the Association for Computational Linguistics (TACL)}
      }%
  }%
}

\title{On the Effect of Instruction Tuning Loss on Generalization}

\author{Anwoy Chatterjee$\color{red}^{*\bm{\dagger}}$ \\
  Dept. of Electrical Engineering \\
  Indian Institute of Technology Delhi \\
  {\small\texttt{anwoychatterjee@gmail.com}} \And
  H S V N S Kowndinya Renduchintala$\color{red}^{\bm{\dagger}}$ \\
  Media and Data Science Research \\
  Adobe Inc., India \\
  {\small\texttt{rharisrikowndinya333@gmail.com}}\AND
  Sumit Bhatia \\
  Media and Data Science Research \\ 
  Adobe Inc., India \\
  {\small\texttt{sumit.bhatia@adobe.com}}\And
  Tanmoy Chakraborty \\
  Dept. of Electrical Engineering \\
  Indian Institute of Technology Delhi \\
  {\small\texttt{tanchak@iitd.ac.in}}
}

\date{}

\begin{document}

\maketitle
\thispagestyle{firstpagefooter}
\begingroup
\renewcommand\thefootnote{\textcolor{red}{$\bm{\dagger}$}}
\footnotetext{These two authors contributed \textbf{\color{red}equally} to this work.}
\renewcommand\thefootnote{\textcolor{red}{*}} 
\footnotetext{Work done during internship at Media and Data Science Research (MDSR) Lab, Adobe Inc.}
\endgroup
\begin{abstract}
Instruction Tuning has emerged as a pivotal post-training paradigm that enables pre-trained language models to better \textit{follow} user instructions. Despite its significance, little attention has been given to optimizing the loss function used. A fundamental, yet often overlooked, question is whether the conventional auto-regressive objective -- where loss is computed only on response tokens, excluding prompt tokens -- is truly optimal for instruction tuning. In this work, we systematically investigate the impact of differentially weighting prompt and response tokens in instruction tuning loss, and propose \textbf{W}eighted \textbf{I}nstruction \textbf{T}uning (\wit) as a better alternative to conventional instruction tuning. Through extensive experiments on five language models of different families and scale, three finetuning datasets of different sizes, and \changed{five} diverse evaluation benchmarks, we show that the standard instruction tuning loss often yields suboptimal performance and limited robustness to input prompt variations. We find that a low-to-moderate weight for prompt tokens coupled with a moderate-to-high weight for response tokens yields the best-performing models across settings \changed{and also serve as better starting points for the subsequent preference alignment training.} These findings highlight the need to reconsider instruction-tuning loss and offer actionable insights for developing more robust and generalizable models. Our code is open-sourced \href{https://github.com/kowndinya-renduchintala/WIT}{here}.
\end{abstract}

\section{Introduction}
\label{sec:introduction}
Transformer-based language models (LMs) pre-trained using just an auto-regressive objective over massive text corpora~\citep{brown2020language,touvron2023llama} demonstrate remarkable performance across a range of NLP tasks~\citep{zhao2021calibrate,wang2022self,wan-etal-2023-gpt,sun-etal-2023-text}. However, they often struggle to reliably follow user instructions as they are essentially \textit{text-completion} models, whose pre-training objective, i.e., next-token prediction, has a fundamental mismatch with the goal of instruction following. 

Instruction tuning aims to bridge this gap by finetuning an LM on a diverse collection of task instances phrased as instructions~\citep{wei2022finetuned,sanh2022multitask,ouyang2022training}, where each task instance consists of a task description (i.e., the instruction), an optional input, a corresponding output, and in some cases, a few demonstrations. Instruction tuning has been shown to significantly improve instruction following capability and generalization of LMs to unseen tasks~\cite{wang-etal-2022-super, wei2022finetuned, sanh2022multitask, chung2024scaling}, and hence has emerged as a widely adopted method in adapting pre-trained LMs to better follow user instructions.

\begin{figure*}[t!]
    \centering
    \includegraphics[width=0.8\textwidth]{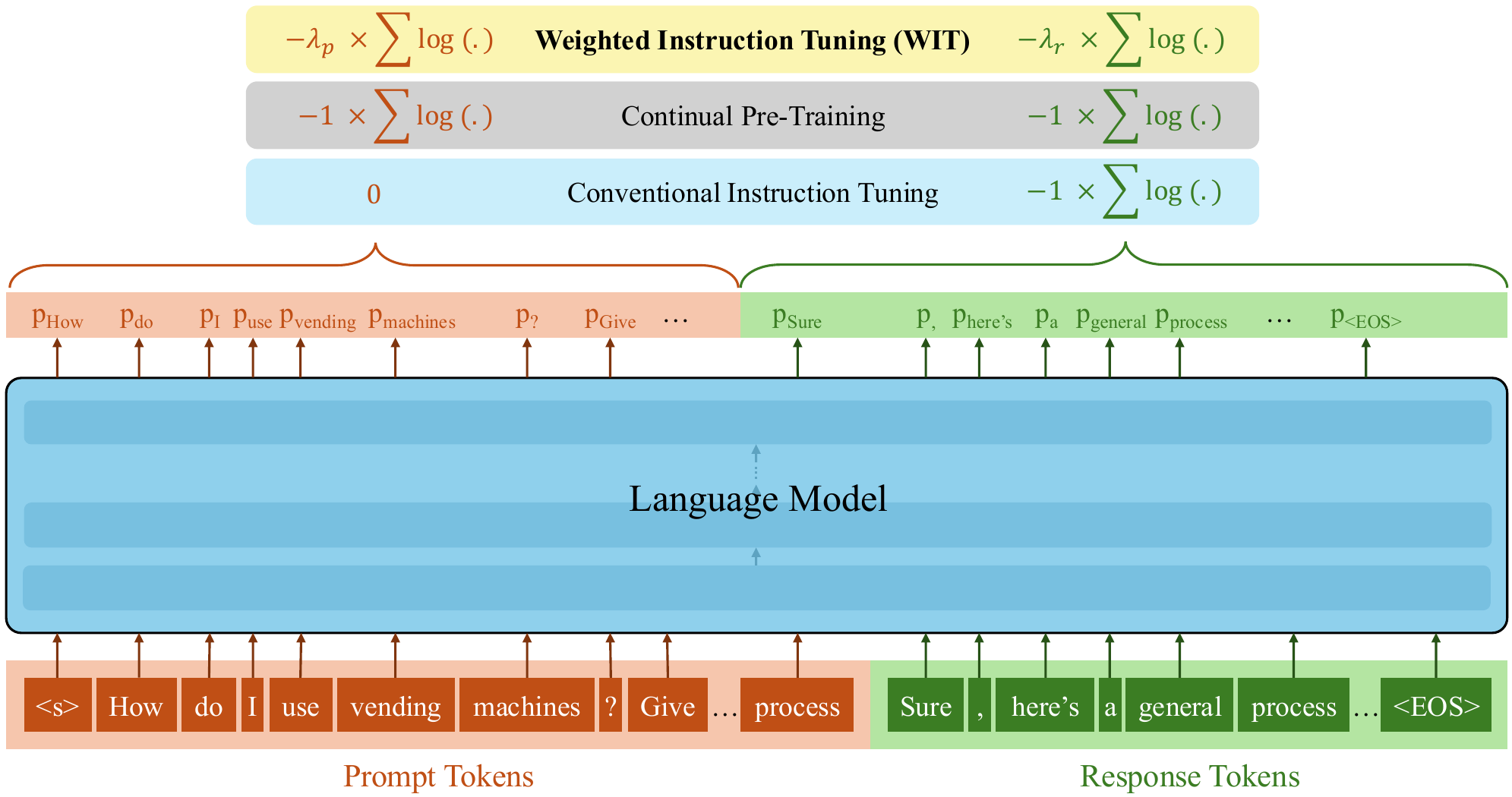}
    \caption{Conventional Instruction Tuning zeroes out the loss on prompt tokens, while continual pre-training treats prompt and response tokens equally. We find that both approaches are suboptimal and introduce \textbf{Weighted Instruction Tuning} (\wit), which assigns different weights $\lambda_p$ and $\lambda_r$ ($0\leq\,\lambda_p,\, \lambda_r\,\leq1$) to prompt and response token losses respectively, as a better alternative.}
    \label{fig:intro}
\end{figure*}

While many studies have shown that the effectiveness of instruction tuning is heavily contingent on various factors such as task composition~\citep{wang2023far,dong-etal-2024-abilities,renduchintala-etal-2024-smart}, data quality~\citep{zhou2023lima,ding-etal-2023-enhancing}, data quantity~\citep{ji2023exploring,yuan2023scaling} and training dynamics~\citep{mukherjee2023orca,pareja2024unveiling}, a very fundamental yet under-explored factor is the loss function itself. The most commonly utilized loss function for instruction tuning is an auto-regressive objective where loss on prompt tokens is zeroed out~\citep{aribandiext5,li2024self,touvron2023llama,chiang2023vicuna,mitra2023orca}
, thereby backpropagating only on response tokens. Although the conventional loss function has been shown to be effective in practice, it is not clear \textit{why} this should be the optimal choice, and to the best of our knowledge, there has not been a comprehensive study on the choice of the loss function to be used for instruction tuning.

Although a couple of recent works \citep{huerta-enochian-ko-2024-instruction, shi2025instruction} explored alternative instruction tuning loss formulations, they still leave out a lot of open questions. For instance, ~\citet{shi2025instruction} proposed Instruction Modelling, which does not zero out the loss on prompt tokens and instead employs the same auto-regressive objective used in the pre-training step -- effectively treating instruction tuning as continual pre-training. However, this is only found to be beneficial when lengthy prompts are coupled with brief responses or when only a small number of training examples are involved. Similarly,~\citet{huerta-enochian-ko-2024-instruction} proposed using a small non-zero weight on prompt tokens, called prompt loss weight (PLW). The authors found that a non-zero PLW is beneficial when working with instruction-tuning data containing short completions and that it can safely be ignored when working with instruction-tuning data containing longer completions. However, its applicability across diverse training and evaluation datasets remains unexplored. Moreover, the extent to which prompt token weights should depend solely on the relative length of completions to prompts remains unclear.

While both these approaches offer some promising directions, they also reveal a deeper issue: the conventional loss function treats prompt and response tokens in a binary fashion -- ignoring the former entirely during loss computation and giving full weight to response tokens. Prompts carry critical task-specific cues and implicit instructions that shape the model’s response. Ignoring their learning signal may deprive the model of valuable contextual guidance, while fully emphasizing response tokens can lead to overfitting on response patterns. Recent concerns about models memorizing response patterns~\citep{neftune, shi2025instruction, chu2025sftmemorizesrlgeneralizes} further highlight the need for a more flexible loss formulation for instruction tuning. We hypothesize that by differentially weighting prompts and responses, we can better balance the contributions of contextual understanding and response generation, thereby fostering improved generalization.

To this end, we propose \textbf{Weighted Instruction Tuning} (\wit) as an alternative to the conventional instruction tuning loss that assigns different weights to prompt and response tokens, enabling more fine-grained control of what the model learns. Figure~\ref{fig:intro} illustrates this notion of differential weighting and shows how it differs from standard approaches of instruction tuning and continual pre-training. We perform extensive finetuning experiments using this new loss function, by training {$525$ models} with different weights on prompt and response tokens across different model families, model sizes and instruction tuning datasets. \changed{Furthermore, in order to investigate the transferability of gains from \wit\ to preference alignment phase, we carry out an additional $525$ training runs on top of these models using the Direct Preference Optimization (DPO) algorithm~\citep{rafailov2023direct}.}  We evaluate the models on popular benchmarks like MMLU~\citep{hendrycks2021measuring} and BBH~\citep{suzgun-etal-2023-challenging} to measure knowledge and reasoning capabilities, IFEval~\citep{zhou2023instruction} to objectively evaluate instruction-following ability, AlpacaEval~\citep{alpaca_eval} and \changed{MT-Bench~\citep{zheng2023judging}} for judging conversational proficiency. The key insights from our study are as follows:

\begin{itemize}[nosep, wide, labelwidth=!, labelindent=0pt]
    \item The conventional instruction tuning loss \emph{rarely} yields the best-performing model across different configurations.
    \item Assigning a low-to-moderate weight ($0$\:–\:$0.5$) to prompt tokens and a moderate-to-high weight ($0.5$\:–\:$1$) to response tokens consistently results in the best-performing models across various settings -- with optimal configuration of prompt and response token weights achieving an average relative gain of \changed{$\sim 6.55\%$} over the conventional loss.
    \item \changed{The gains from using \wit-loss also transfer to the subsequent preference alignment training using the DPO algorithm, i.e., \wit-finetuned models are \emph{better starting points compared to conventional instruction-tuned models}, for DPO.}
    \item A relatively moderate response-token weight not only enhances performance on standard benchmarks, but also improves model robustness to minor prompt variations.
    \item In many cases (although not always), finetuning solely on prompts also enhances instruction following compared to the base model, suggesting the possibility of instruction tuning the model even in the absence of response annotations.
\end{itemize}

\changed{We also present a post hoc analysis of how prompt characteristics -- like length and diversity -- correlate with optimal prompt-token weights, offering insights into factors influencing the choice of token weights. We also examine how \wit~ reshapes prompt and response probability distributions, highlighting its impact on model behavior. Our findings aim to aid the research in development of more robust and generalizable models.}

\section{Proposed Formulation}
\label{sec:proposed_formulation}
Let $\mathcal{D}=\{(\mathbf{P}_i, \mathbf{R}_i)\}_{i=1}^{N_{\mathcal{T}}}$ be an instruction tuning dataset consisting of $N_{\mathcal{T}}$ $(prompt, response)$ pairs, where each prompt $\mathbf{P}_i$ consists of an instruction (implicit or explicit) and an optional input, while $\mathbf{R}_i$ represents the expected ground-truth response. If $|\mathbf{S}|$ denotes the number of tokens in sequence $\mathbf{S}$, then $\mathbf{P}_i$ and $\mathbf{R}_i$ can be expanded as:  
\[
\mathbf{P}_i = \{p_i^{(1)}, p_i^{(2)}, \ldots, p_i^{(|\mathbf{P}_i|)}\}, \]
\[
\mathbf{R}_i = \{r_i^{(1)}, r_i^{(2)}, \ldots, r_i^{(|\mathbf{R}_i|)}\}
\]
The conventional instruction tuning, which is an auto-regressive objective that zeroes out the loss on prompt tokens, is given by:
\begin{small}
\begin{equation}
    \mathcal{L}_{IT} = \frac{- \sum\limits_{i=1}^{N_{\mathcal{T}}}\sum\limits_{j=1}^{|\mathbf{R}_i|} \log\; \mathbb{P}_{\mathcal{M}}\left(r_i^{(j)} \;|\; \mathbf{P}_i, r_i^{(1)},\ldots, r_i^{(j-1)}\right)}{\sum\limits_{i=1}^{N_{\mathcal{T}}} |\mathbf{R}_i|}
\end{equation}
\end{small}
Here, $\mathbb{P}_{\mathcal{M}}(.)$ denotes the probability assigned by the language model $\mathcal{M}$. 

As discussed in Section~\ref{sec:introduction}, ignoring learning signals corresponding to the prompts may lead the model to struggle with comprehending novel prompts, while assigning full weight on response tokens can hamper generalization ability by potentially overfitting on common response patterns in the instruction tuning data. Hence, we propose Weighted Instruction Tuning (\wit), which assigns differential weights to the prompt and response tokens, as an alternative to the conventional instruction tuning loss. It is given by:

\begin{small}
\begin{multline}
    \mathcal{L}_{\wit} = \frac{-1}{\sum\limits_{i=1}^{N_{\mathcal{T}}} \Big(\mathds{I}{(\lambda_p \neq 0)}\cdot| \mathbf{P}_i| +  \mathds{I}{(\lambda_r \neq 0)}\cdot |\mathbf{R}_i|\Big)} \times
    \\ \sum_{i=1}^{N_{\mathcal{T}}} \bigg[ \lambda_p \sum\limits_{j=1}^{|\mathbf{P}_i|} \log\; \mathbb{P}_{\mathcal{M}}\left(p_i^{(j)} \;|\; p_i^{(1)},\ldots, p_i^{(j-1)}\right)
    \\ + \lambda_r \sum\limits_{j=1}^{|\mathbf{R}_i|} \log\; \mathbb{P}_{\mathcal{M}}\left(r_i^{(j)} \;|\; \mathbf{P}_i, r_i^{(1)},\ldots, r_i^{(j-1)}\right) \bigg]
\end{multline}
\end{small}
where the weighting factors $\lambda_p$ and $\lambda_r$, denote the \textit{prompt} and \textit{response token weights}, respectively, while $\mathds{I}{(.)}$ is the indicator function. \changed{$\mathcal{L}_{\wit}$ computes the weighted sum of log-probabilities -- scaling the log-probabilities of prompt tokens by $\lambda_p$ and those of response tokens by $\lambda_r$ -- and then normalizes by the count of tokens with non-zero weight. The indicator function ($\mathds{I}$)} \changed{ensures that the weighted sum is divided exactly by those tokens whose weight is non-zero.} Note that the conventional instruction tuning loss $\mathcal{L}_{IT}$ is a special case of $\mathcal{L}_{\wit}$ for $(\lambda_p, \lambda_r) = (0,1)$. 

\section{Experimental Setup}\label{sec:exp_setup}
\label{sec:exp_setup}

\subsection{Finetuning Data}
\label{sec:finetuning_data}
\subsubsection*{Instruction Tuning}
We considered the following three commonly used diverse instruction tuning datasets to study the role of prompt and response token weights: 
\begin{itemize}[nosep, wide, labelwidth=!, labelindent=0pt]

\item[\textbf{(i)}] \textbf{LIMA}~\cite{zhou2023instruction} is a carefully curated set of $1K$ high-quality $(prompt, response)$ pairs from sources such as Stack Exchange, wikiHow and Reddit, along with some manually authored examples. 

\item[\textbf{(ii)}] \textbf{Alpaca-Cleaned} is a filtered version of the original Alpaca dataset~\cite{taori2023stanford} after removing problematic instances, with $52K$ $(prompt, response)$ pairs generated by \texttt{text-davinci-003}.

\item[\textbf{(iii)}] \textbf{T\"ulu-v2}~\cite{ivison2023camels} is a data mixture with instances from diverse sources such as FLAN-v2~\citep{longpre2023flan}, Open Assistant~\citep{kopf2023openassistant}, GPT4-Alpaca~\citep{peng2023instruction}, and Open-Orca~\citep{lian2023openorca}, containing \(326K\) $(prompt, response)$ pairs in total, from which we randomly select \(150K\) samples to reduce overall experiment cost and runtime.
\end{itemize}

The above choice of three datasets together covers a small dataset (LIMA), a moderately-sized dataset (Alpaca-Cleaned) and a large dataset (T\"ulu-v2). Furthermore, they also differ in other characteristics such as response length, prompt length and diversity, etc (Section~\ref{sec:prompt_weight_corr}).

\subsubsection*{Preference Alignment Training}
\changed{For preference alignment training, we use a binarized version of the \textbf{UltraFeedback} dataset~\citep{cui2310ultrafeedback}, consisting of around \(60K\) $(prompt$, $chosen\_response$, $rejected\_response)$ tuples.}

\subsection{Finetuning Procedure}
\label{sec:finetuning_procedure}

For our experiments, we consider five models spanning different model families and sizes -- Llama-3.2-1B, Gemma-2-2B, Llama-3.2-3B, Mistral-7B, and Llama-3-8B. We finetune each model for \(1\) epoch on T\"ulu-v2, for \(2\) epochs on Alpaca-Cleaned and for \(5\) epochs on LIMA.
Following \citet{touvron2023llama, phased-it} and other contemporary works, we use a learning rate of $5\times 10^{-6}$ for Mistral-7B and a learning rate of $2\times 10^{-5}$ for all other models, with batch size 64, weight decay 0.1, and cosine learning rate decay with linear warmup over the first 1\% of steps. \changed{For preference alignment phase, we apply DPO~\citep{rafailov2023direct}, similar to ~\citet{ivison2023camels}, with a learning rate of $5\times 10^{-7}$, batch size 32, weight decay 0.0, and 0.1 warmup ratio, finetuning each model for 2 epochs.} We ran all the experiments on \(8\) NVIDIA A100-SXM4-80GB GPUs, utilizing Flash Attention $2.0$~\citep{dao2023flashattention} 
and for larger models like Mistral-7B and Llama-3-8B, we use the full-sharded data parallel functionality in PyTorch\footnote{\href{https://pytorch.org/docs/stable/notes/fsdp.html}{https://pytorch.org/docs/stable/notes/fsdp.html}}.
The code to reproduce all our results is open-sourced \href{https://github.com/kowndinya-renduchintala/WIT}{here}.

\begin{figure*}[ht!]
    \centering
    \renewcommand{\arraystretch}{1.2}  
    \begin{tblr}{@{}p{1.2em}@{} @{\hskip 0.3em} c @{\hskip 0.3em} c @{\hskip 0.3em} c @{\hskip 0.3em} c @{\hskip 0.3em} c@{}}  
        \rotatebox{90}{\parbox{2.5cm}{\centering {T\"ulu-v2}}} &  
        \begin{subfigure}[b]{0.19\textwidth}
            \caption*{\centering {Llama-3-1B}}
            \includegraphics[width=\textwidth]{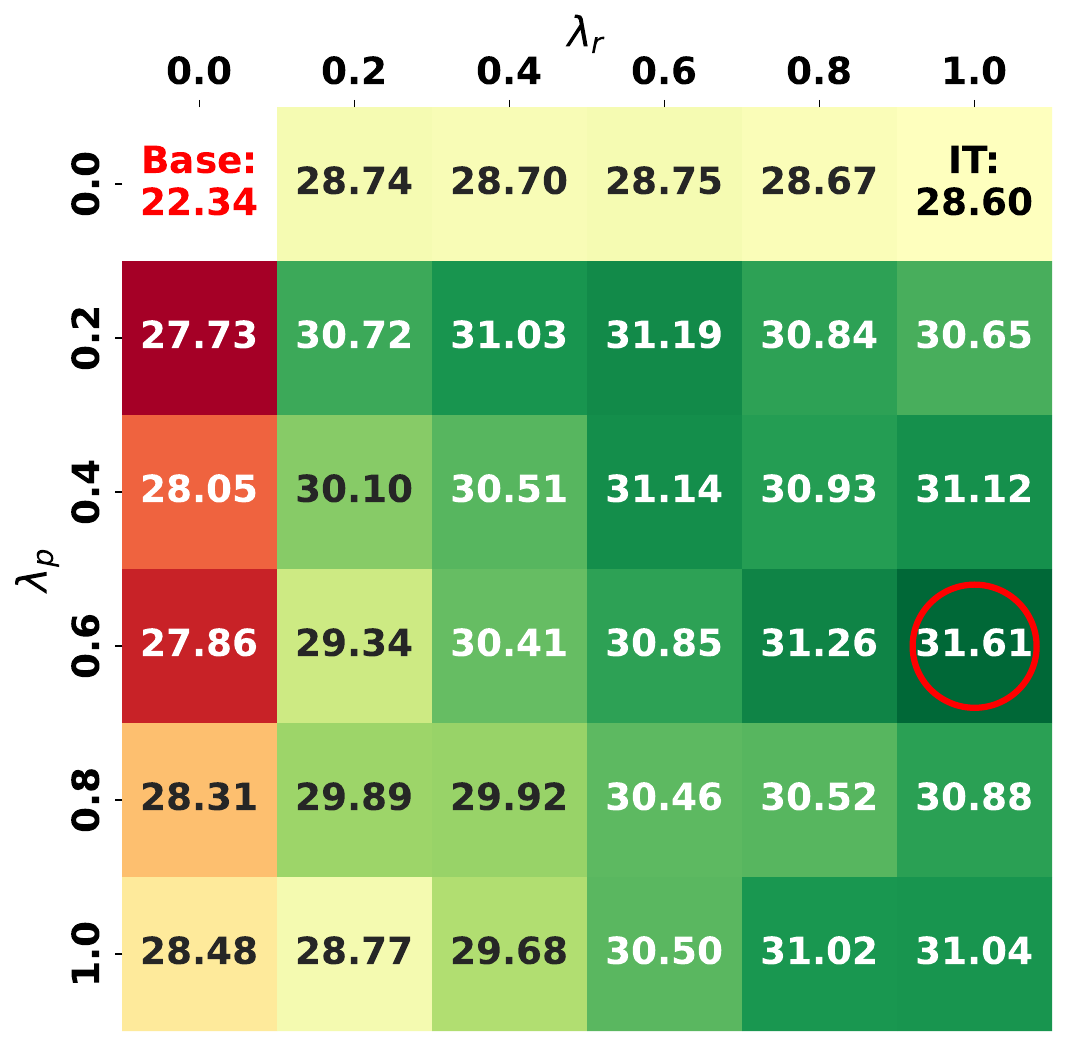} 
        \end{subfigure} &
        \begin{subfigure}[b]{0.19\textwidth}
            \caption*{\centering {Gemma-2-2B}}
            \includegraphics[width=\textwidth]{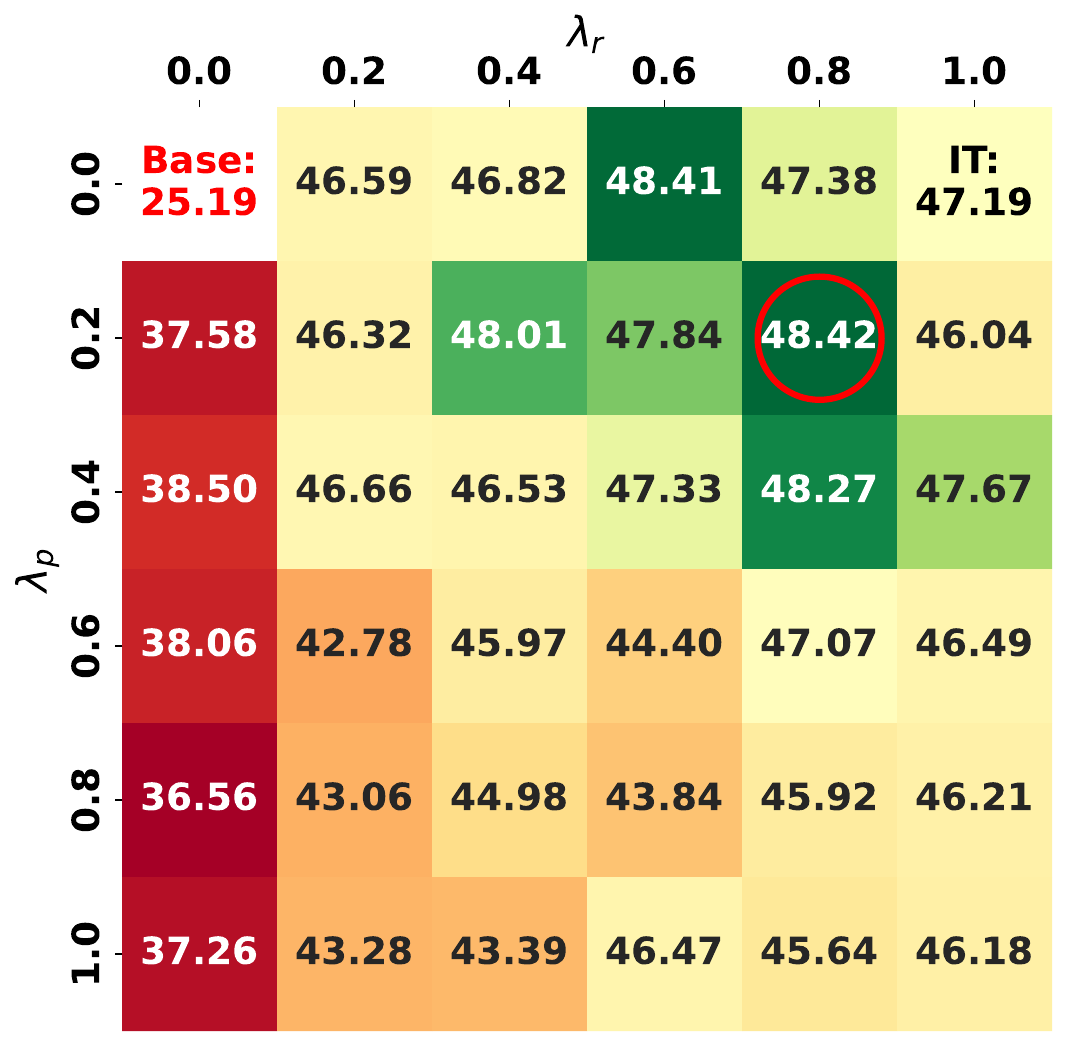}
            
        \end{subfigure} &
        \begin{subfigure}[b]{0.19\textwidth}
            \caption*{\centering {Llama-3-3B}}
            \includegraphics[width=\textwidth]{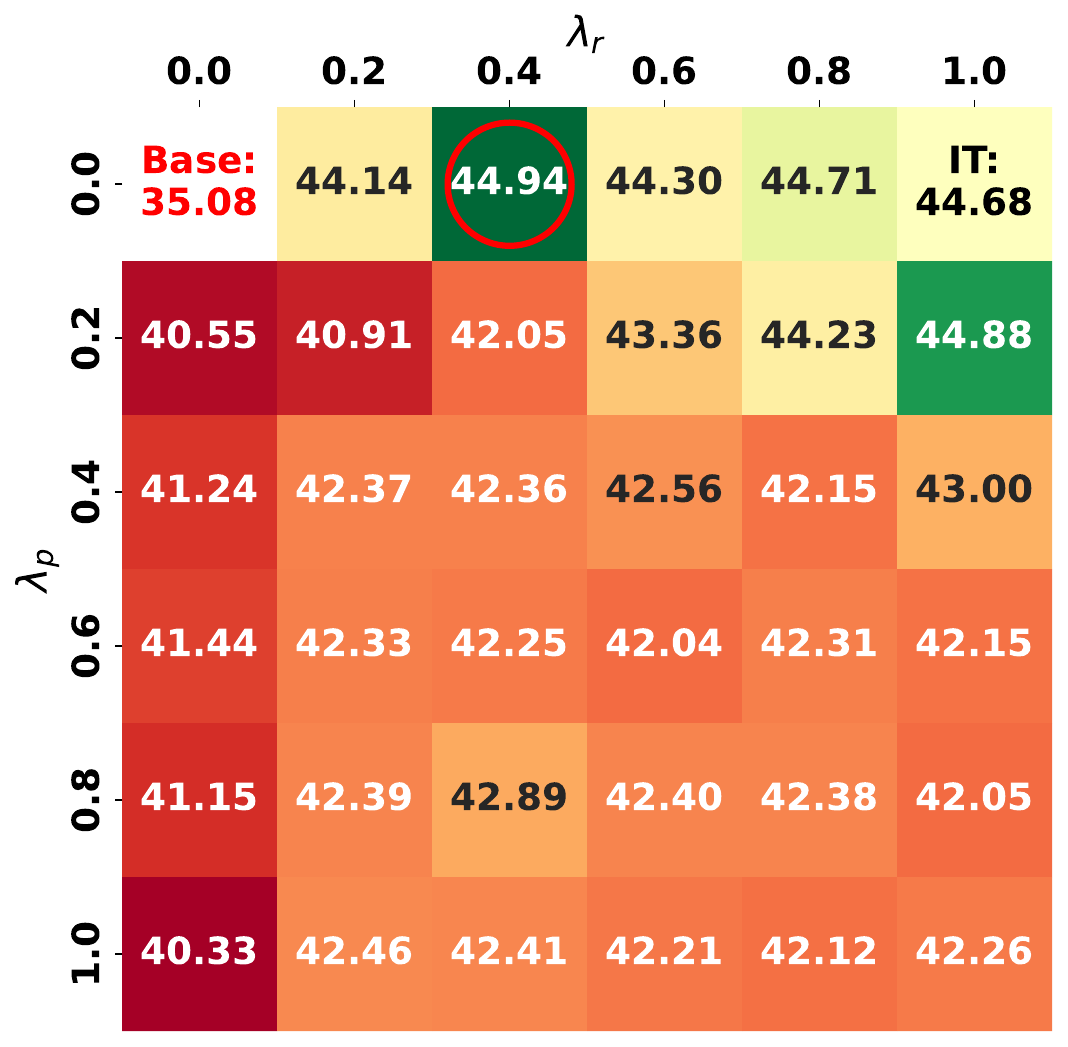}
            
        \end{subfigure} &
        \begin{subfigure}[b]{0.19\textwidth}
            \caption*{\centering {Mistral-7B}}
            \includegraphics[width=\textwidth]{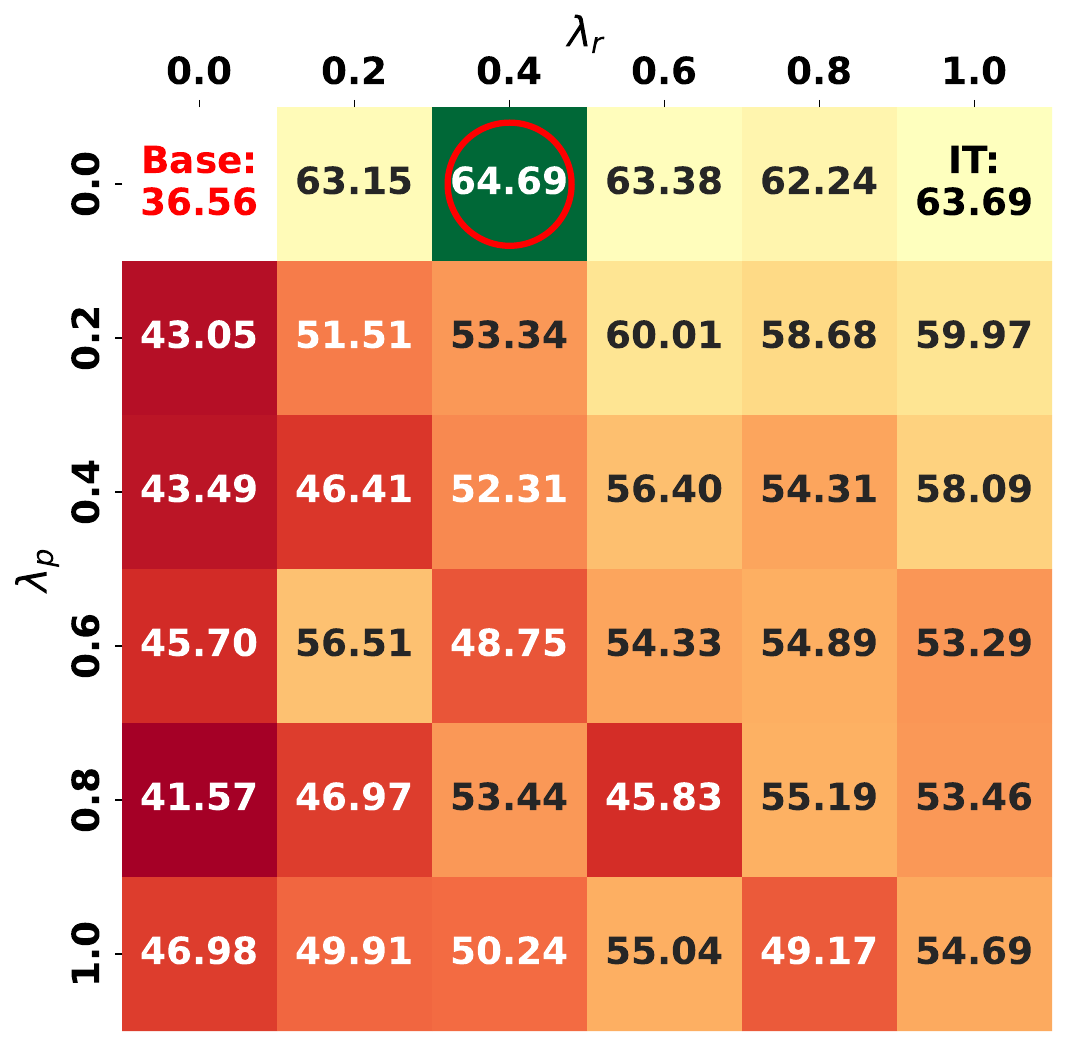}
            
        \end{subfigure} &
        \begin{subfigure}[b]{0.19\textwidth}
            \caption*{\centering {Llama-3-8B}}
            \includegraphics[width=\textwidth]{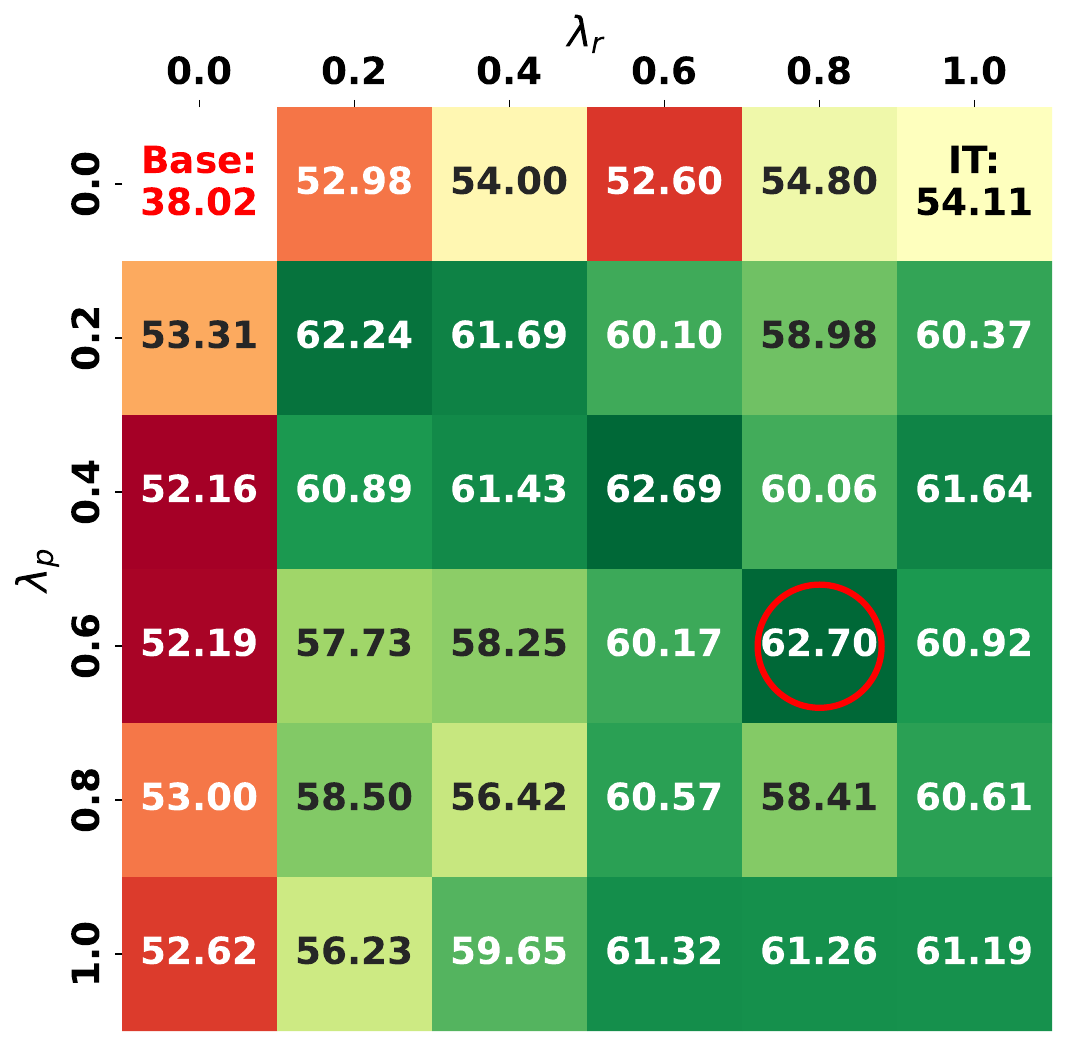}
            
        \end{subfigure} \\
        
        \rotatebox{90}{\parbox{2.5cm}{\centering {Alpaca-Cleaned}}} &  
        \begin{subfigure}[b]{0.19\textwidth}
            \includegraphics[width=\textwidth]{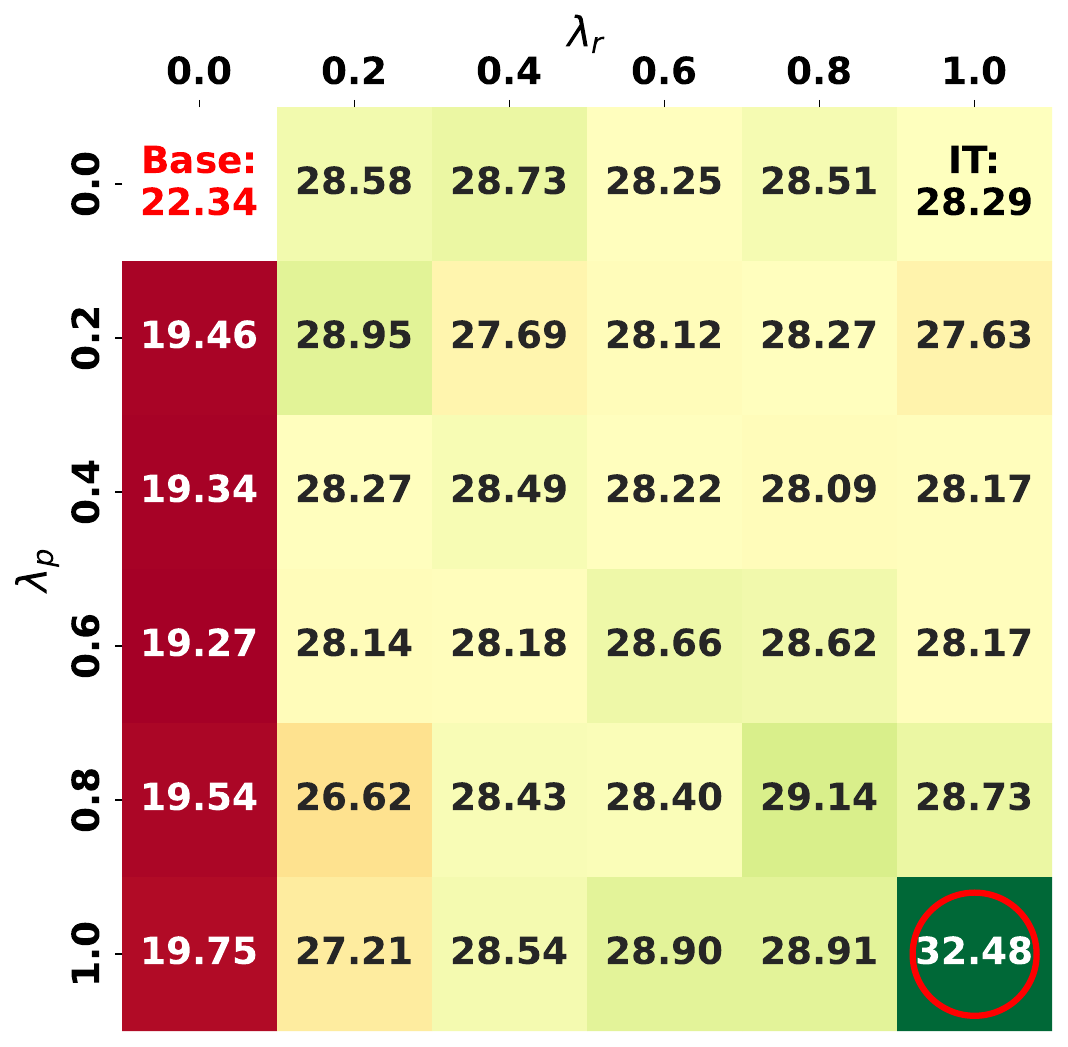}
        \end{subfigure} &
        \begin{subfigure}[b]{0.19\textwidth}
            \includegraphics[width=\textwidth]{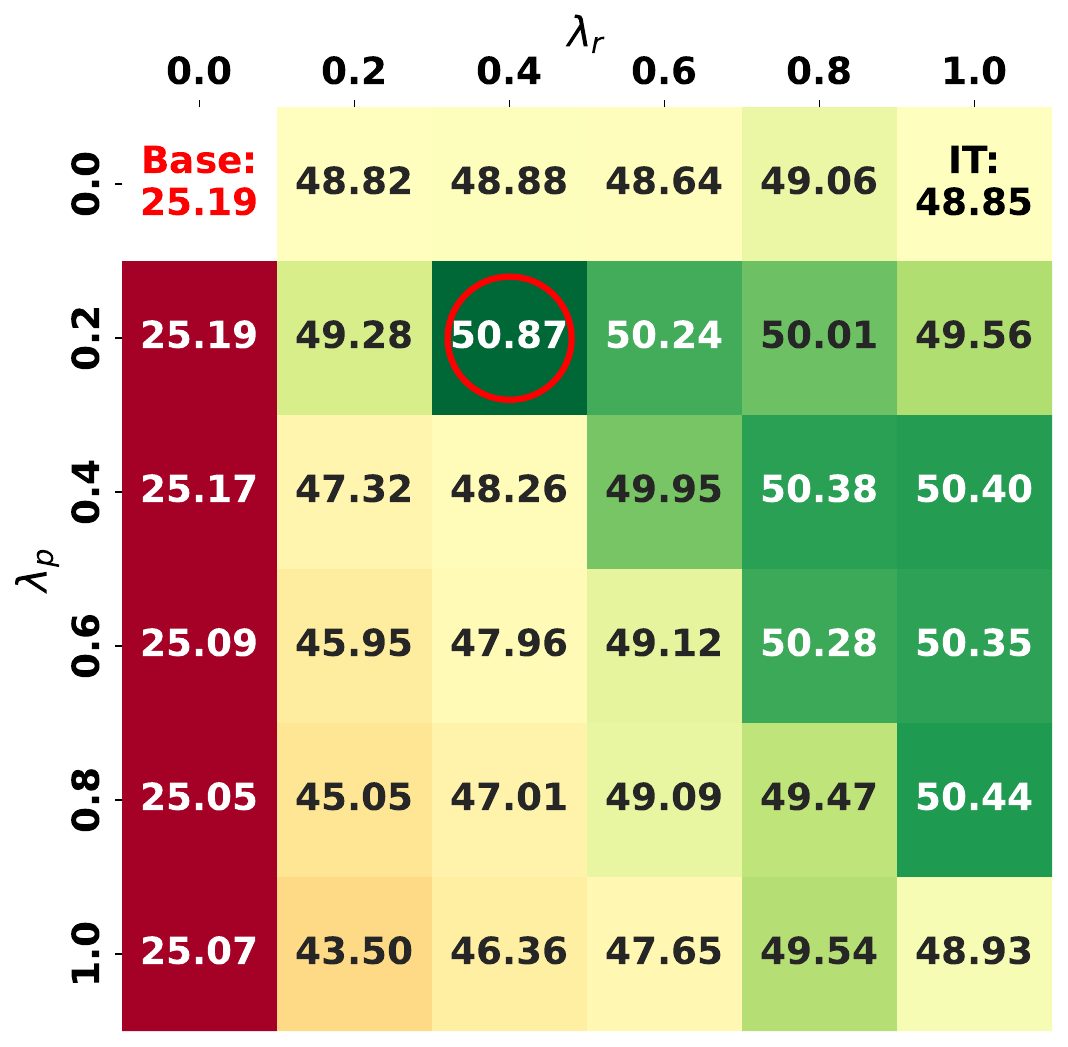}
        \end{subfigure} &
        \begin{subfigure}[b]{0.19\textwidth}
            \includegraphics[width=\textwidth]{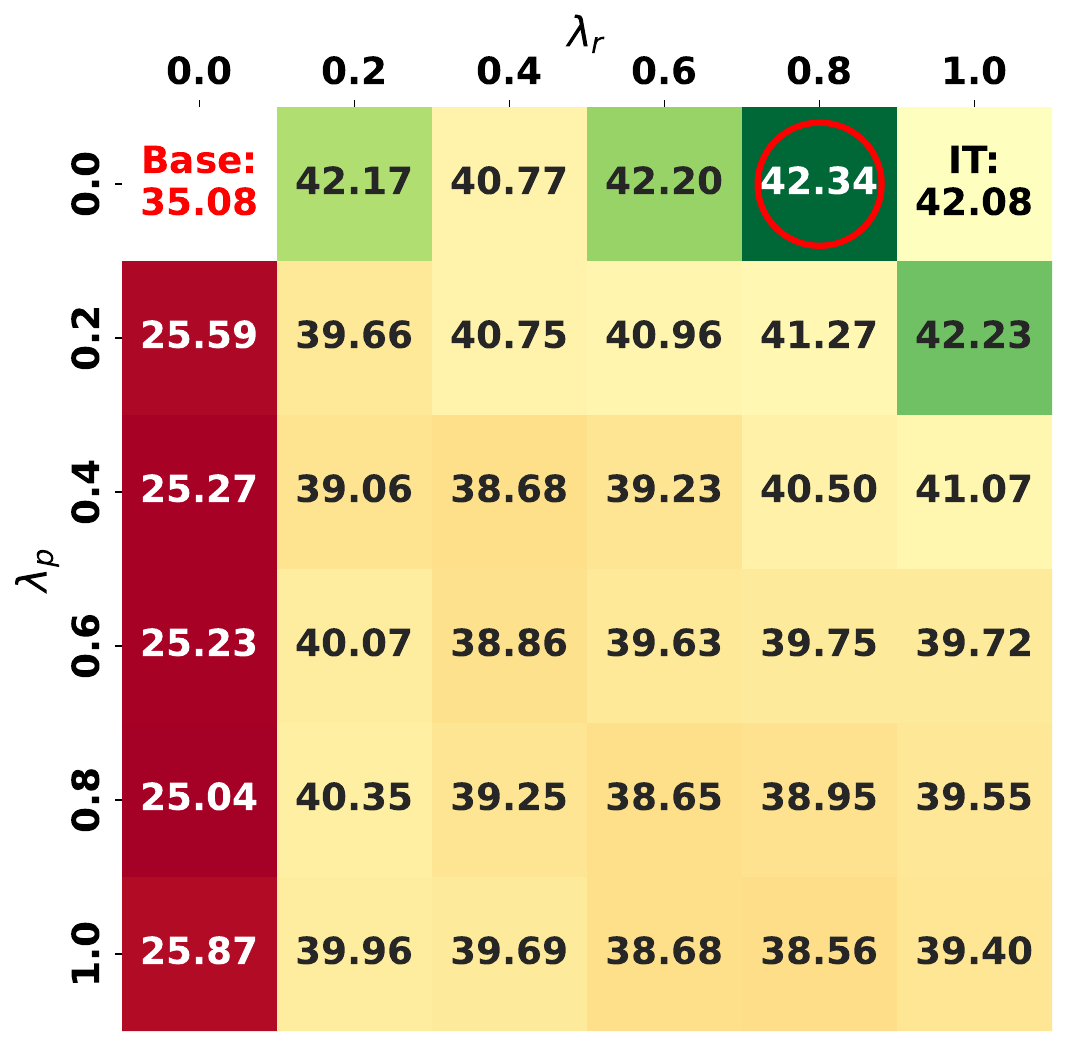}
        \end{subfigure} &
        \begin{subfigure}[b]{0.19\textwidth}
            \includegraphics[width=\textwidth]{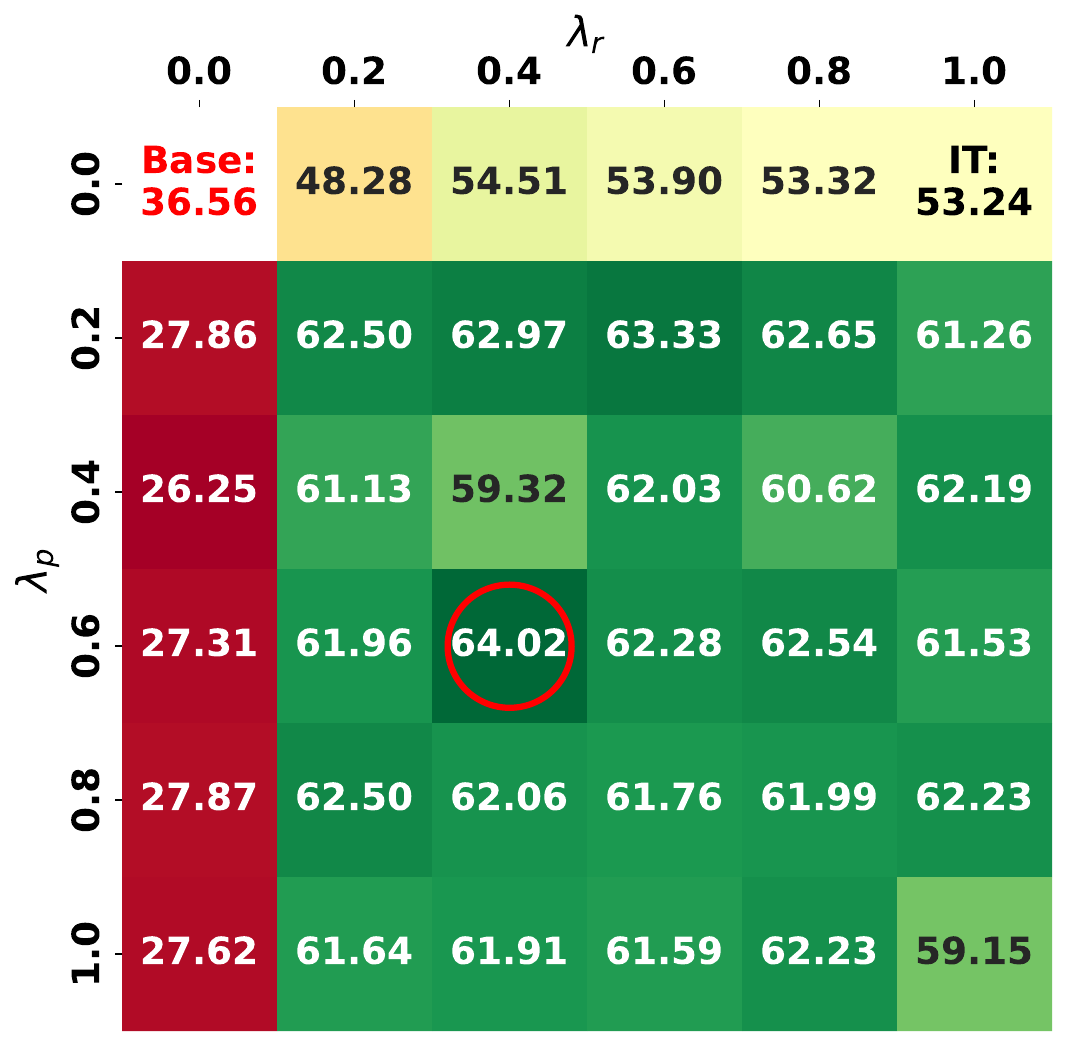}
        \end{subfigure} &
        \begin{subfigure}[b]{0.19\textwidth}
            \includegraphics[width=\textwidth]{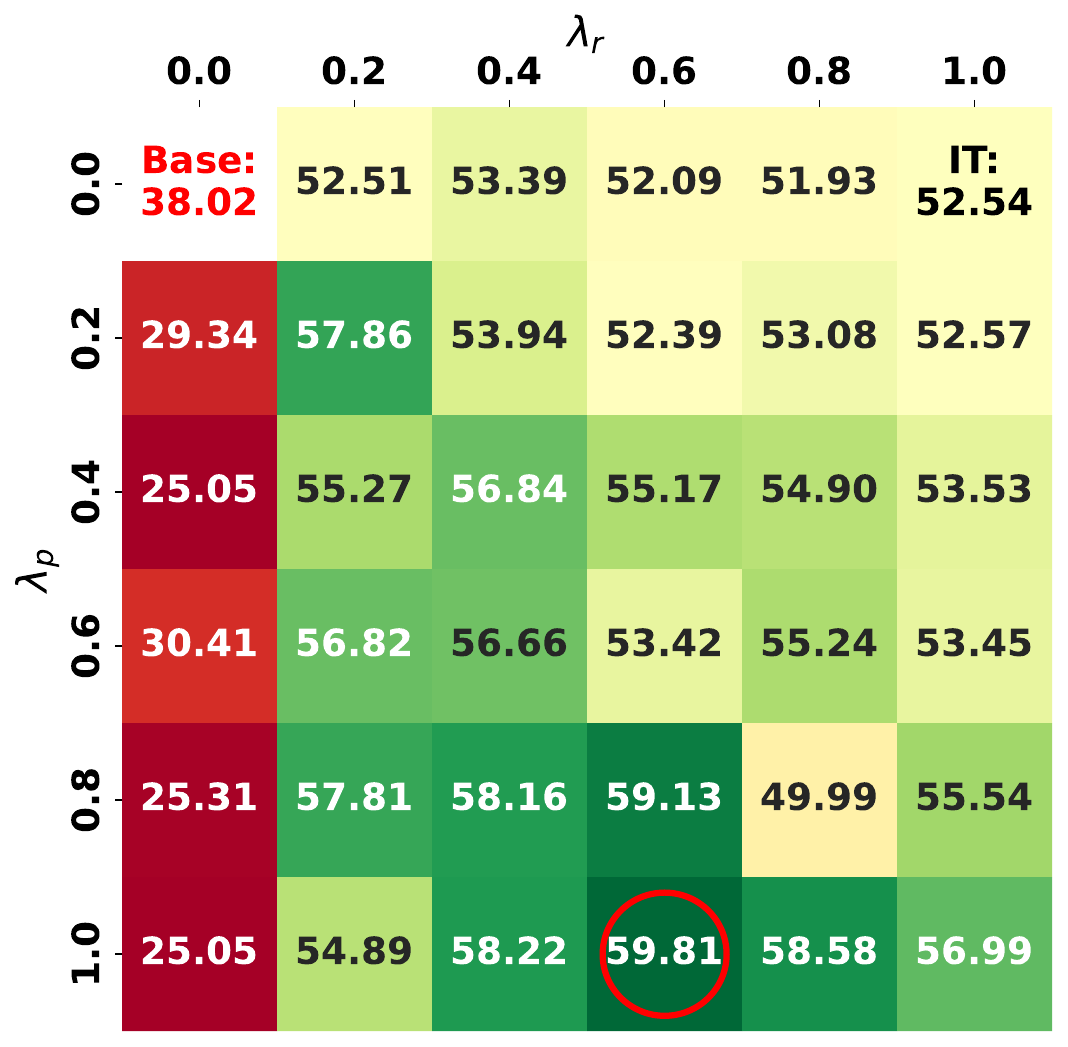}
        \end{subfigure} \\

        \rotatebox{90}{\parbox{2.5cm}{\centering {LIMA}}} &  
        \begin{subfigure}[b]{0.19\textwidth}
            \includegraphics[width=\textwidth]{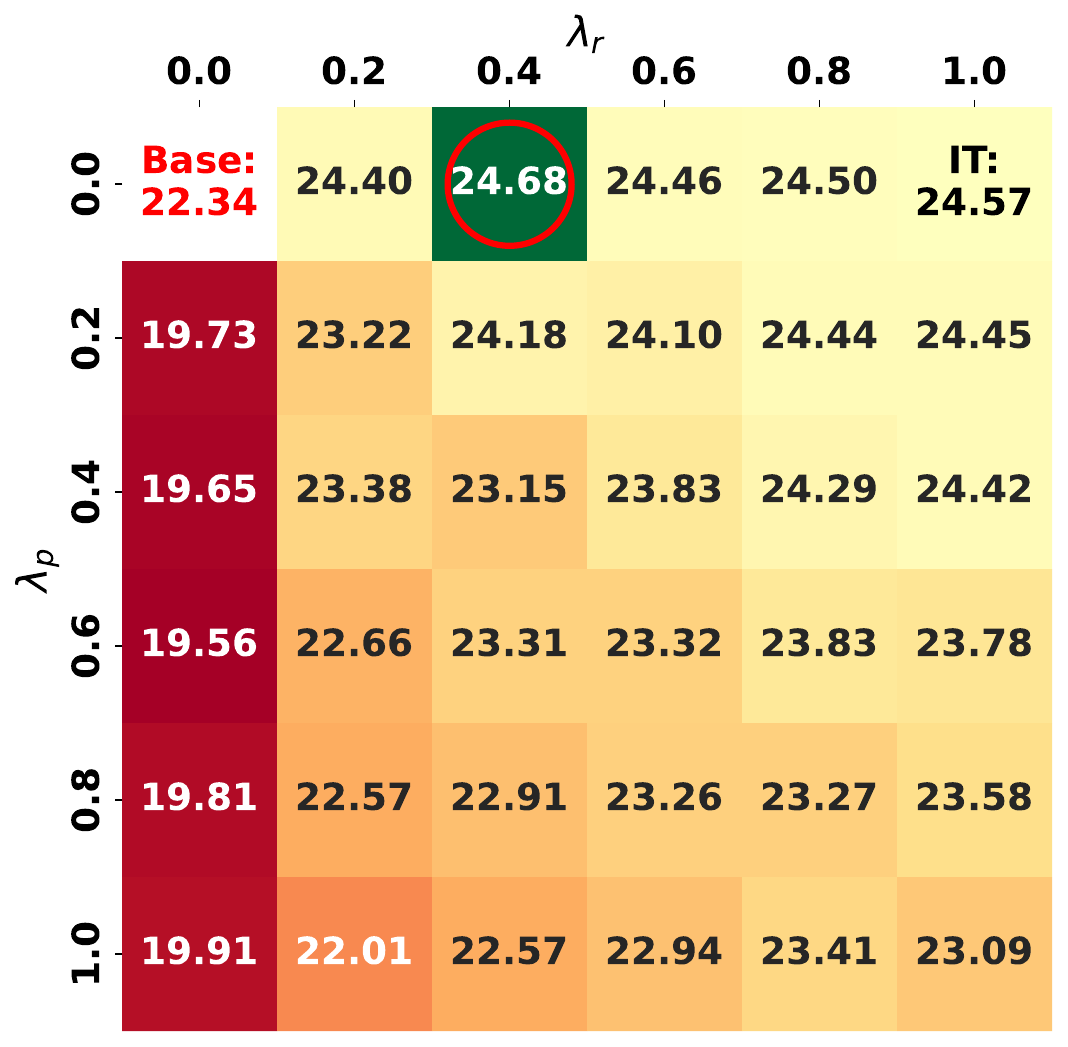}
        \end{subfigure} &
        \begin{subfigure}[b]{0.19\textwidth}
            \includegraphics[width=\textwidth]{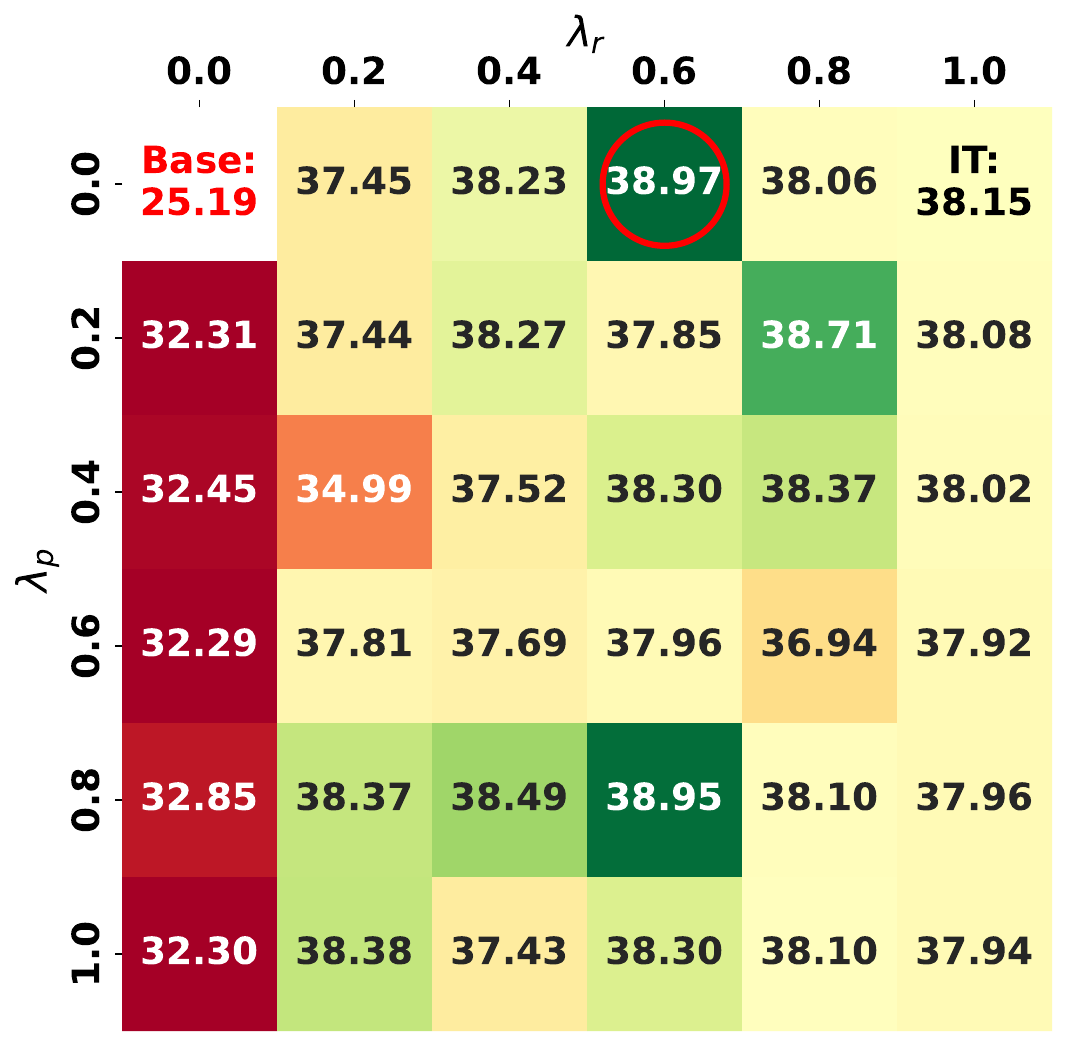}
        \end{subfigure} &
        \begin{subfigure}[b]{0.19\textwidth}
            \includegraphics[width=\textwidth]{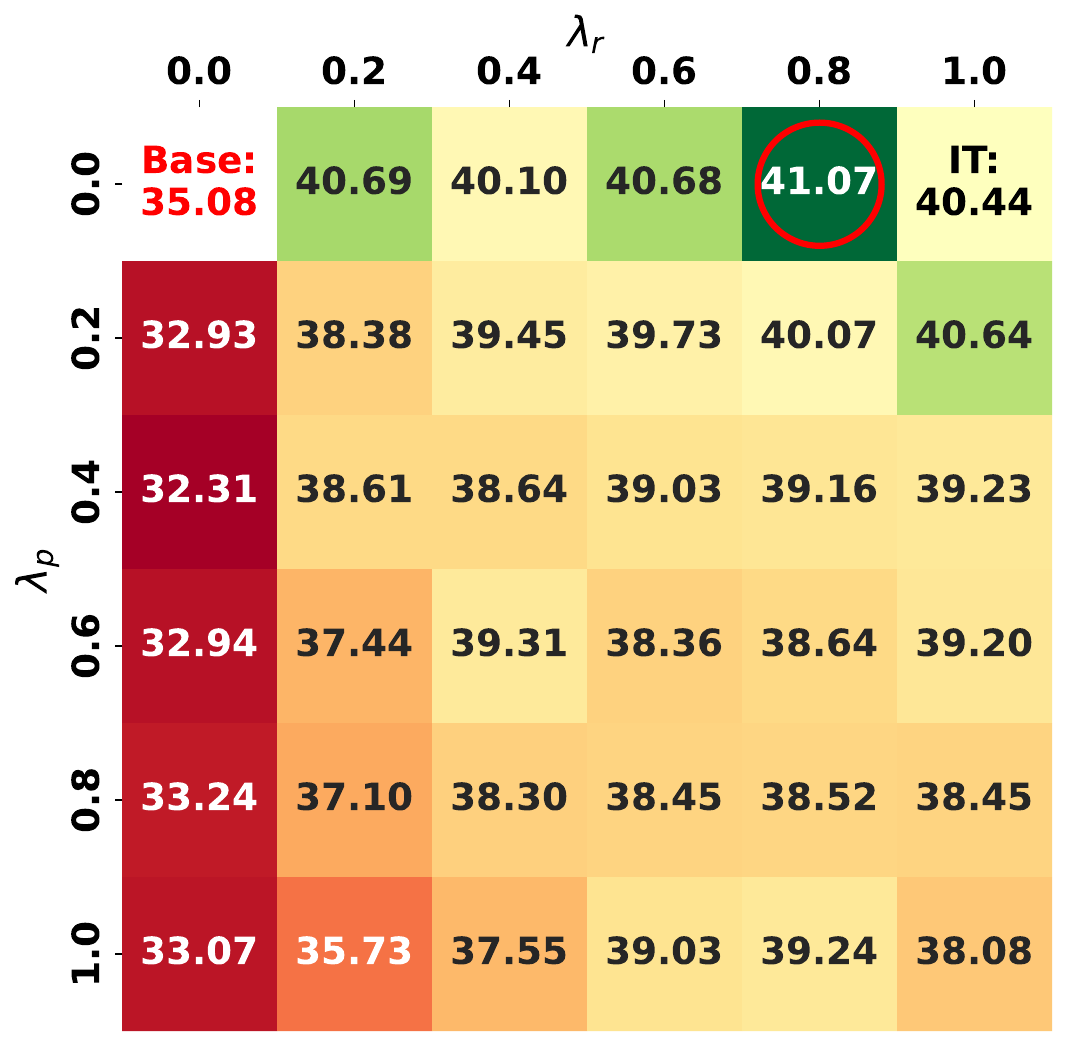}
        \end{subfigure} &
        \begin{subfigure}[b]{0.19\textwidth}
            \includegraphics[width=\textwidth]{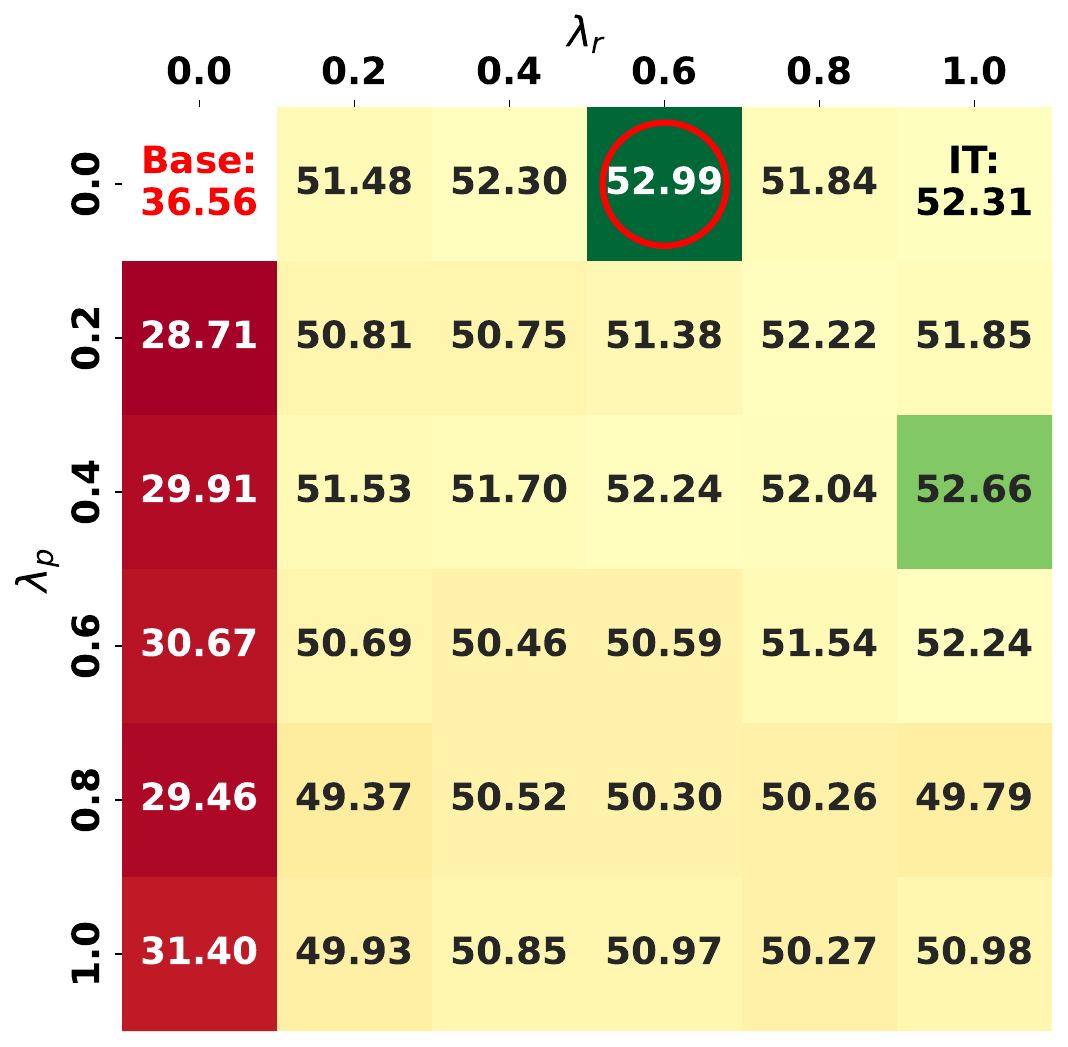}
        \end{subfigure} &
        \begin{subfigure}[b]{0.19\textwidth}
            \includegraphics[width=\textwidth]{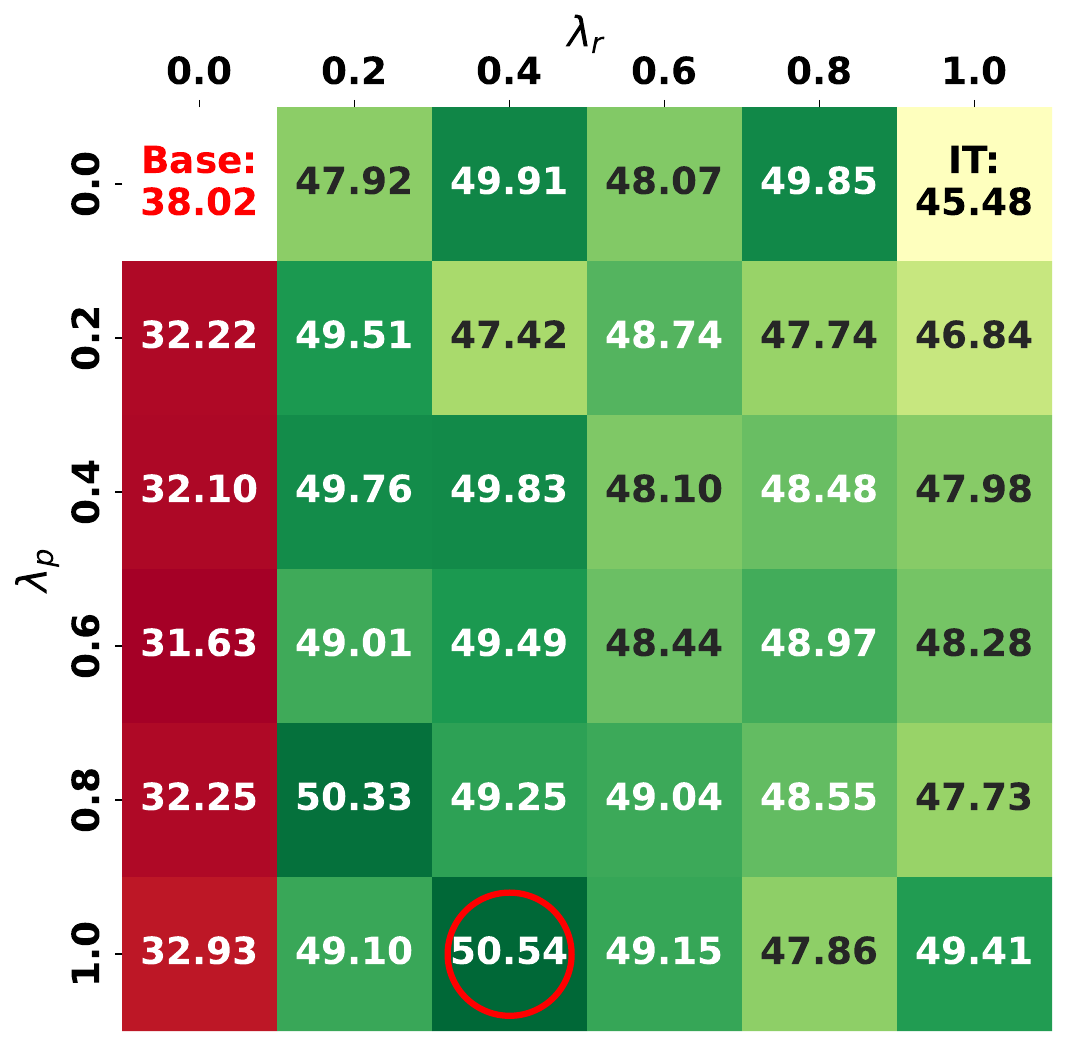}
        \end{subfigure} \\
    \end{tblr}

    \caption{Heatmaps depicting average performance across \changed{five} benchmarks (MMLU, BBH, AlpacaEval, IFEval \changed{and MT-Bench}) for different configurations of $(\lambda_p, \lambda_r)$ and for different models finetuned on T\"ulu-v2, Alpaca-Cleaned, and LIMA. Best performing configuration is highlighted with a red circle. The color map is based on relative gain with respect to conventional instruction tuning. Rows correspond to prompt token weights ($\lambda_p$) and columns correspond to response token weights ($\lambda_r$). Conventional instruction tuning is marked with \texttt{IT} and base model performance is marked with \texttt{Base}.}
    \label{fig:main_heatmap}
\end{figure*}

\subsection{Evaluation Protocol}
\label{sec:evaluation_protocol}

We assess the performance of our instruction-tuned models across various dimensions by employing the following evaluation suites:
\begin{itemize}[nosep, wide, labelwidth=!, labelindent=0pt]
\item[\textbf{(i)}] 

\textbf{MMLU (Massive Multitask Language Understanding)}~\citep{hendrycks2021measuring} is a benchmark spanning \(57\) tasks across humanities, STEM, and social sciences, with approximately \(14K\) multiple-choice \((prompt, response)\) pairs. We evaluate models in a \textit{zero-shot setting} using flexible exact match, following LM Evaluation Harness~\cite{eval-harness}.

\item[\textbf{(ii)}] \textbf{BBH (Big-Bench Hard)}~\citep{suzgun-etal-2023-challenging} is a challenging subset of the BIG-Bench benchmark, comprising \(23\) tasks with \(6.5K\) examples requiring logical deduction and multi-step reasoning. We evaluate BBH in a \textit{zero-shot setting without chain-of-thought (CoT) prompting}, using flexible exact match as the evaluation metric, similar to MMLU.

\item[\textbf{(iii)}] 
 \textbf{AlpacaEval}~\citep{alpaca_eval} is an LLM-based evaluation framework with \(805\) prompts designed to assess conversational ability. Using AlpacaEval-1.0, we report model win rates against \texttt{text-davinci-003}, judged by GPT-4o-mini. Unlike MMLU and BBH, which emphasize correctness, AlpacaEval provides a holistic measure by evaluating both response quality and relevance in instruction-tuned models.

\item[\textbf{(iv)}] \textbf{IFEval}~\citep{zhou2023instruction} is an instruction-following evaluation benchmark that focuses on a set of "verifiable instructions" offering an automated yet objective evaluation of instruction-following capability, unlike LLM-as-a-judge.

\item[\textbf{(v)}] \changed{\textbf{MT-Bench}~\citep{zheng2023judging} evaluates multi-turn conversational and instruction-following abilities using $80$ high-quality multi-turn questions. We adopt the single-answer grading scheme, with $160$ responses rated from $1$ to $10$ by an LLM judge\footnote{We scale the rating by 10 while averaging with other benchmarks.}. For our experiments, we use Llama-3.3-70B as the judge.}

\end{itemize}

\section{Results}\label{sec:results}
To study the role of prompt and response tokens in instruction tuning, we finetune five language models of different scales (Section~\ref{sec:finetuning_procedure}) on the three instruction tuning datasets (Section~\ref{sec:finetuning_data}) by varying the prompt and response weight configurations $(\lambda_p, \lambda_r)$ in $\{0, \,0.2, \,0.4, \,0.6, \,0.8, \,1.0\}$. We then evaluate the generalization capability of these instruction-tuned models across \changed{five} diverse benchmarks -- MMLU, BBH, IFEval, AlpacaEval \changed{and MT-Bench}. Figure~\ref{fig:main_heatmap} depicts the average performance of models across all benchmarks whereas Figures~\ref{fig:tulu_all}, \ref{fig:alpaca-cleaned_all}, and \ref{fig:lima_all} in the Appendix contain the individual benchmark performances for T\"ulu-v2, Alpaca-Cleaned, and LIMA as training data, respectively.

As illustrated in Figure~\ref{fig:main_heatmap}, conventional instruction tuning, i.e., $\lambda_p \!= \!0$ and $\lambda_r \!= \!1$, is \changed{never} the optimal choice. This underscores the critical role of loss function design in instruction tuning. Similarly, $\lambda_p \!=\!1$ and $\lambda_r \!=\!1$, which corresponds to the same auto-regressive objective used in pre-training step, i.e., instruction tuning treated as continual pre-training as suggested by \citet{shi2025instruction}, also performs suboptimally. In fact, it yields optimal average performance in exactly $1$ out of $15$ $(model$, $training\_dataset$) combinations, reinforcing the need to reconsider loss weighting strategies to enhance performance and generalization.

Building on these observations, we further quantify the \textit{relative} performance gains of \wit~ compared to the conventional instruction tuning. As summarized in Table~\ref{tab:relative_gain}, \wit~ yields consistent improvements in average benchmark performance, achieving an average relative gain of around $6.55\%$. These findings underscore the value of assigning different weights to prompt and response tokens. In some cases, the benefits are especially pronounced -- for example, finetuning Mistral-7B on the AlpacaCleaned dataset with $\lambda_p=0.6$ and $\lambda_r=0.4$ achieves a relative performance gain of approximately $20.25\%$.

We now present our key \textit{empirical} findings based on the trends observed across different configurations.

\begin{table}[t]
\small
    \centering
    \resizebox{\columnwidth}{!}{
        \begin{tabular}{ccccc}
            \hline
            \textbf{Model} & \begin{tabular}{c}\textbf{Training}\\\textbf{Data}\end{tabular} & \begin{tabular}{c}\textbf{Conventional}\\\textbf{Loss}\end{tabular} & \begin{tabular}{c}\textbf{\wit~Loss}\\\textbf{(Optimal} $\mathbf{\lambda_p, \lambda_r}$\textbf{)}\end{tabular} & \begin{tabular}{c}\textbf{Relative}\\\textbf{Gain}\end{tabular} \\ \hline
            \multirow{3}{*}{Llama-3.2-1B} & T\"ulu-v2 & 28.60 & 31.61 & +10.49\% \\
            & AlpacaCleaned & 28.29 & 32.48 & +14.81\% \\
            & LIMA & 24.57 & 24.68 & +0.45\% \\ \hline
            \multirow{3}{*}{Gemma-2-2B} & T\"ulu-v2 & 47.19 & 48.42 & +2.61\% \\
            & AlpacaCleaned &  48.85 & 50.87 & +1.04\% \\
            & LIMA & 38.15 & 38.97 & +2.15\%\\ \hline
            \multirow{3}{*}{Llama-3.2-3B} & T\"ulu-v2 & 44.68 & 44.94 & +0.58\% \\
            & AlpacaCleaned & 42.08 & 42.34 & +0.62\%\\
            & LIMA & 40.44 & 41.07 & +1.56\% \\ \hline
            \multirow{3}{*}{Mistral-7B} & T\"ulu-v2 & 63.69 & 64.69 & +1.57\% \\
            & AlpacaCleaned & 53.24 & 64.02 & +20.25\% \\
            & LIMA & 52.31 & 52.99 & +1.3\% \\ \hline
            \multirow{3}{*}{Llama-3-8B} & T\"ulu-v2 & 54.11 & 62.70 & +15.88\% \\
            & AlpacaCleaned & 52.54 & 59.81 & +13.84\% \\
            & LIMA & 45.48 & 50.54 & +11.13\% \\  \hline
            & & & \begin{tabular}{c}\textbf{Average}\\\textbf{Relative Gain}\end{tabular} \textbf{=} & \changed{+6.55\%} \\\hline
        \end{tabular}
    }
    \caption{Relative percentage gain of \wit~(for optimal prompt and response token weights) over conventional instruction tuning on downstream tasks.}
    \label{tab:relative_gain}
\end{table}

\subsection{Key Observations}

\paragraph{Low-to-Moderate Prompt-Token Weight Yields Best Performing Models.}

While the optimal prompt-token weight varies based on the specific setting, i.e., the particular combination of model, training dataset and evaluation benchmark (as also demonstrated in Figures~\ref{fig:tulu_all}, \ref{fig:alpaca-cleaned_all}, and \ref{fig:lima_all} in the Appendix), \changed{we find that in approximately $81\%$ of the cases, i.e., $61$ out of the $75$ $(model$, $training\_dataset$, $evaluation\_benchmark)$ combinations that we considered, the best performance is achieved with a low-to-moderate prompt-token weight in the range of $0$ to $0.6$. Furthermore, in $56\%$ of the cases, i.e., $43$ out of the $75$ settings, the optimal prompt-token weight is non-zero}, strongly suggesting that ignoring prompt tokens for instruction tuning is suboptimal. 

\paragraph{Moderate-to-High Response-Token Weight Yields Optimal Models.}
Existing instruction-tuning approaches~\cite{shi2025instruction, huerta-enochian-ko-2024-instruction} assign maximal weight to response tokens (i.e., $\lambda_r=$1). \changed{However, our experiments reveal that $\lambda_r=1$ is the optimal configuration in only $24\%$ of the cases, i.e., $18$ out of the $75$ $(model$, $training\_dataset$, $evaluation\_benchmark)$ combinations. And in all the remaining $76\%$ of the cases, $\lambda_r<1$ yields the best performance. Furthermore, in $73.33\%$ of the cases, i.e., $55$ out of $75$ settings, a moderate-to-high response-token weight, in the range of $0.4$ to $1$, yielded the best performance.} These findings further reinforce that conventional instruction tuning, i.e., $(\lambda_p, \lambda_r)=(0, 1)$, leads to suboptimal performance. We hypothesize that an extreme response-token weight might encourage memorization of response patterns and hurt generalization, as also noted by \citet{neftune} and \citet{shi2025instruction}. 

\begin{figure*}[ht!]
    \centering
    \renewcommand{\arraystretch}{1.2}  
    \begin{tblr}{@{}p{1.2em}@{} @{\hskip 0.3em} c @{\hskip 0.3em} c @{\hskip 0.3em} c @{\hskip 0.3em} c @{\hskip 0.3em} c@{}}  
        \rotatebox{90}{\parbox{2.5cm}{\centering {T\"ulu-v2}}} &  
        \begin{subfigure}[b]{0.19\textwidth}
            \caption*{\centering {Llama-3-1B}}
            \includegraphics[width=\textwidth]{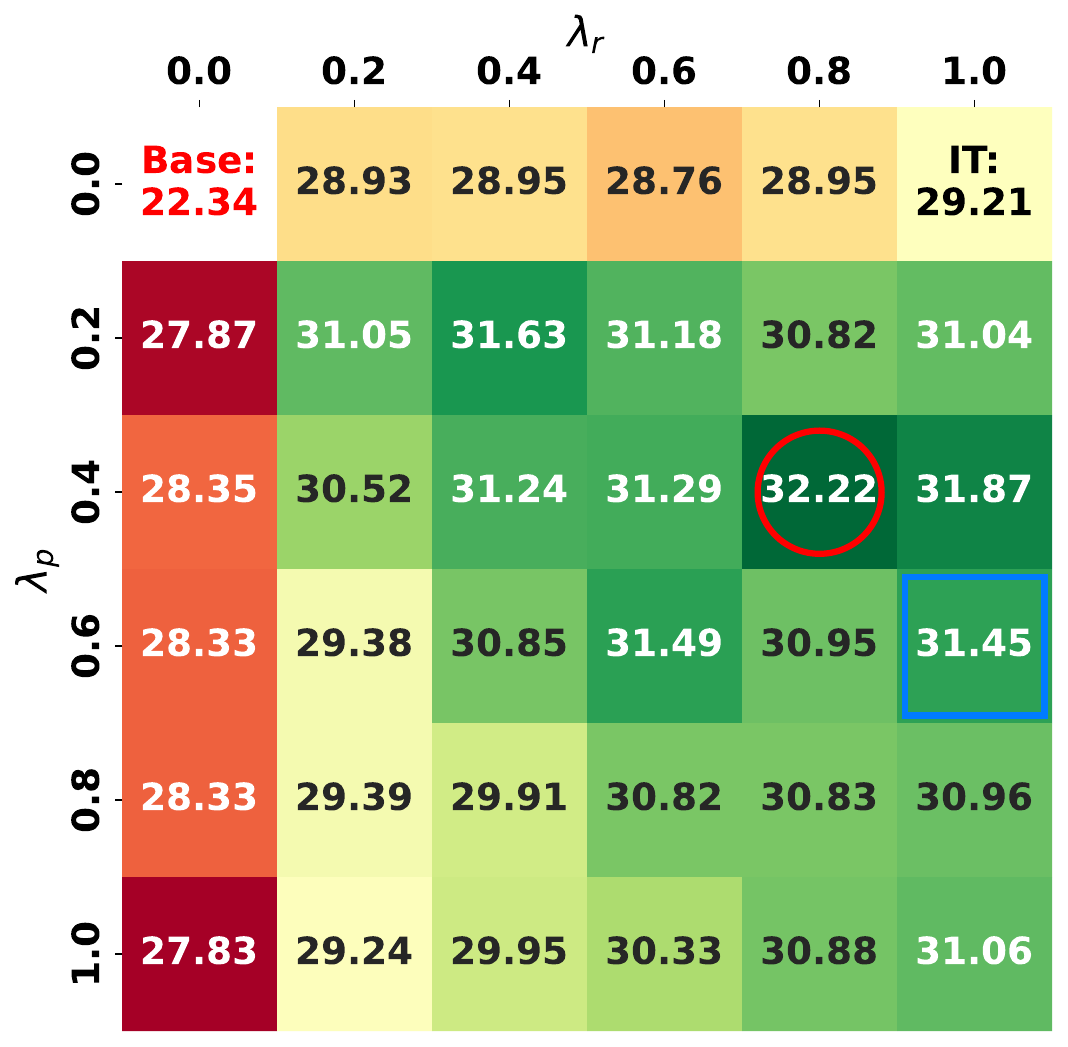} 
        \end{subfigure} &
        \begin{subfigure}[b]{0.19\textwidth}
            \caption*{\centering {Gemma-2-2B}}
            \includegraphics[width=\textwidth]{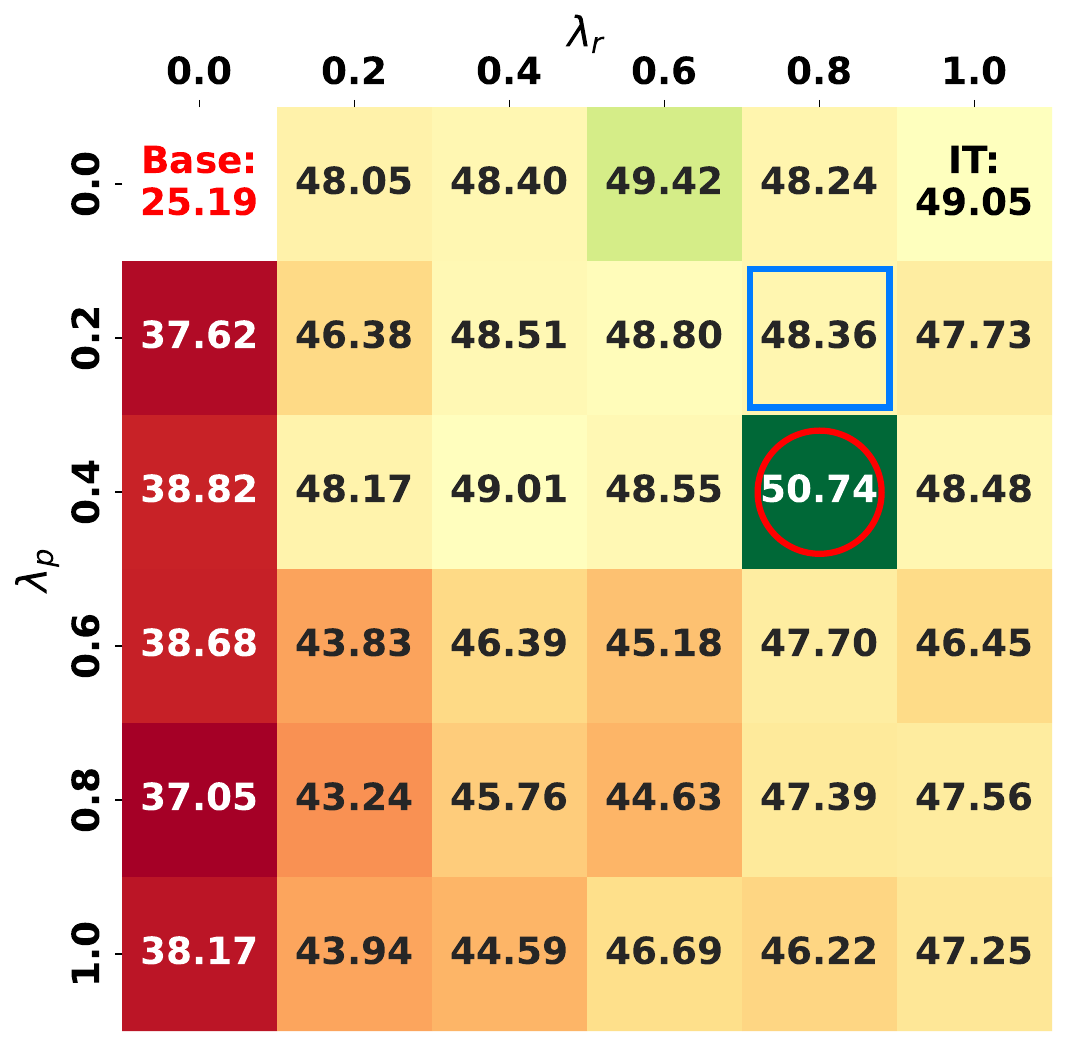}
            
        \end{subfigure} &
        \begin{subfigure}[b]{0.19\textwidth}
            \caption*{\centering {Llama-3-3B}}
            \includegraphics[width=\textwidth]{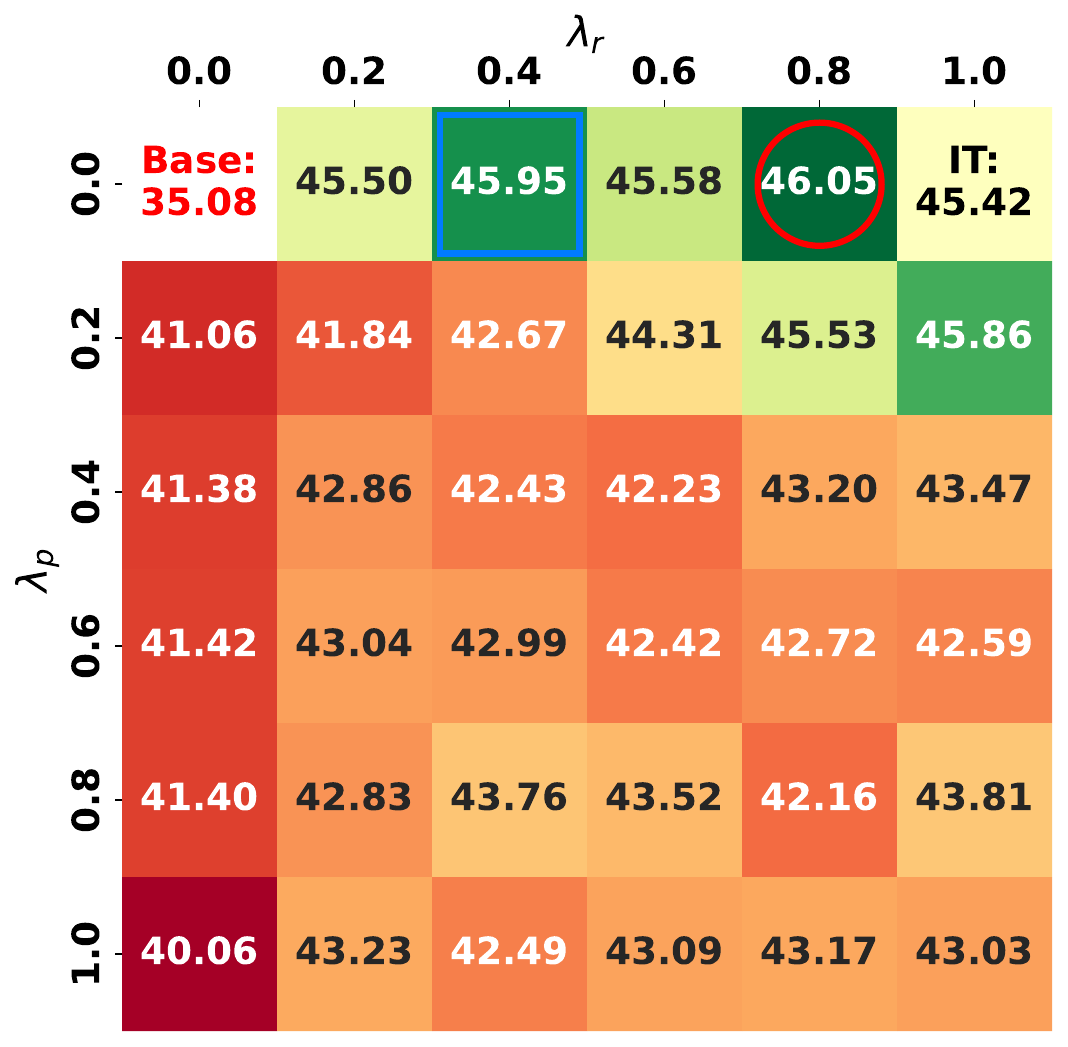}
            
        \end{subfigure} &
        \begin{subfigure}[b]{0.19\textwidth}
            \caption*{\centering {Mistral-7B}}
            \includegraphics[width=\textwidth]{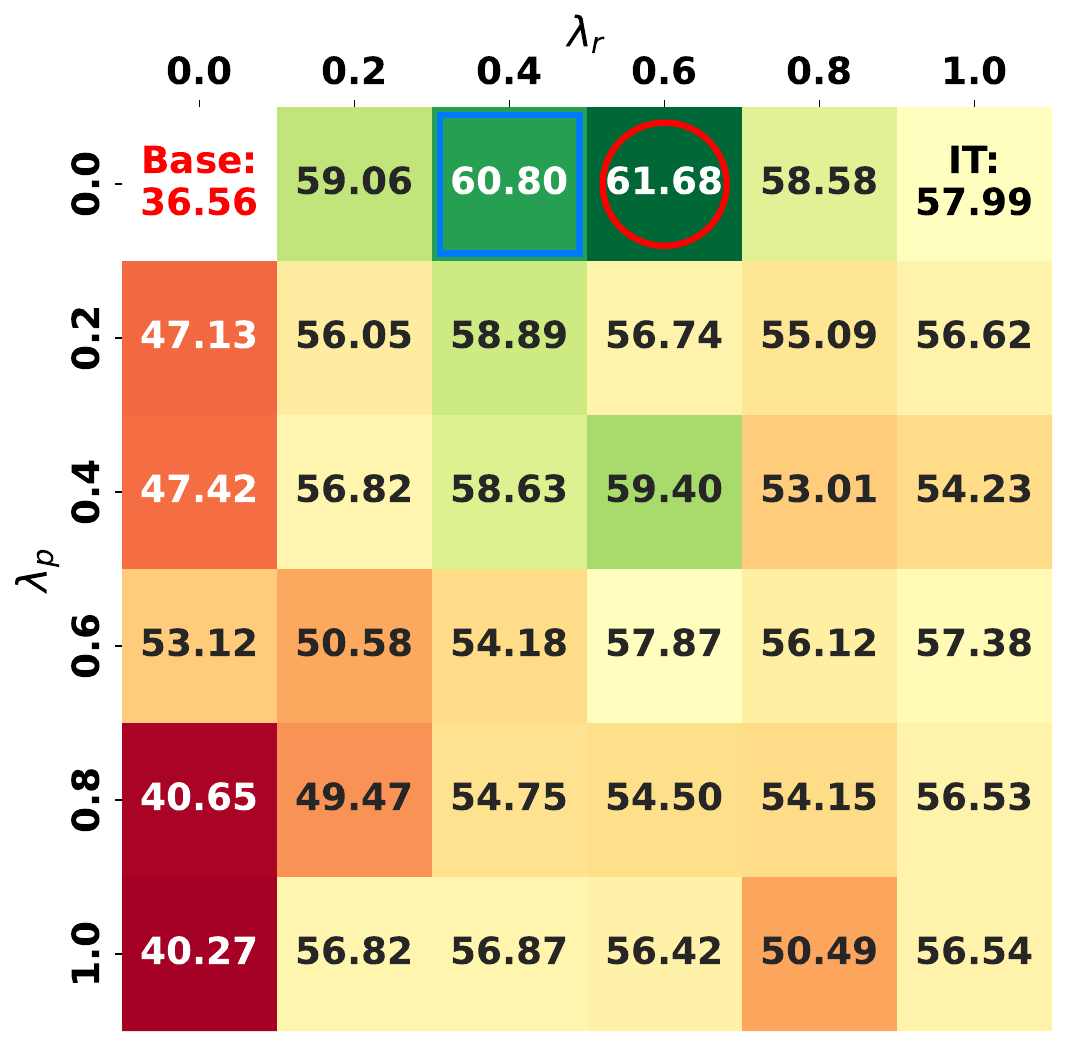}
            
        \end{subfigure} &
        \begin{subfigure}[b]{0.19\textwidth}
            \caption*{\centering {Llama-3-8B}}
            \includegraphics[width=\textwidth]{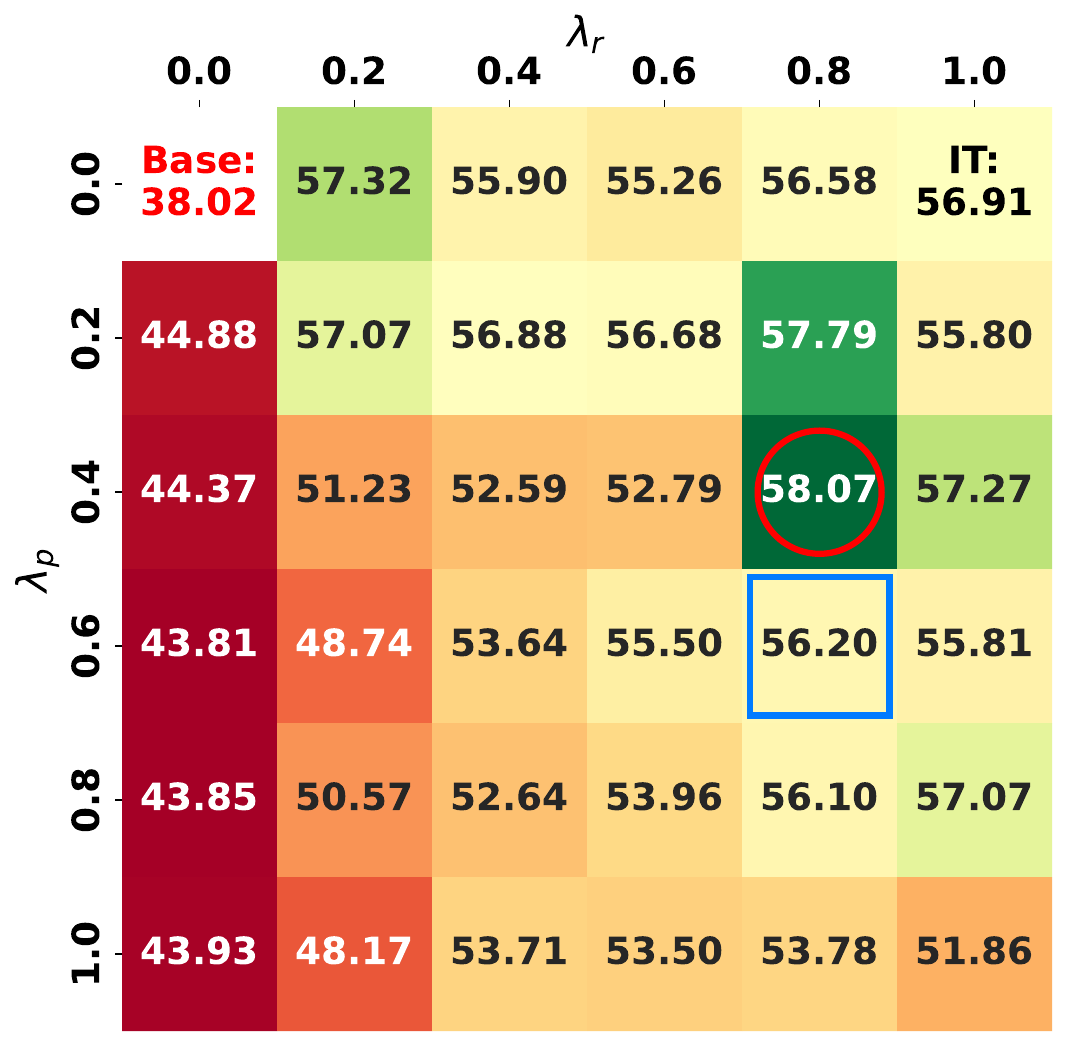}
            
        \end{subfigure} \\
        
        \rotatebox{90}{\parbox{2.5cm}{\centering {Alpaca-Cleaned}}} &  
        \begin{subfigure}[b]{0.19\textwidth}
            \includegraphics[width=\textwidth]{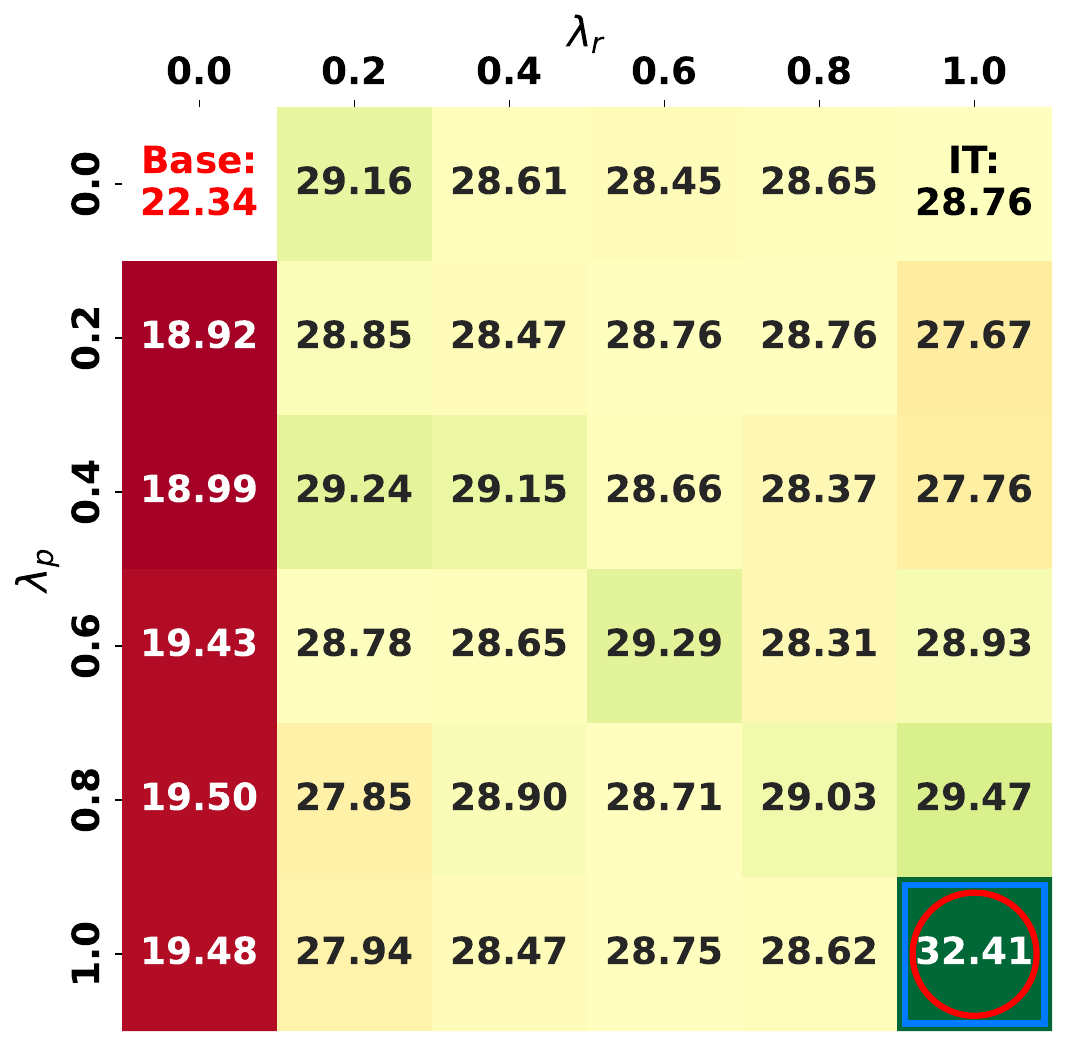}
        \end{subfigure} &
        \begin{subfigure}[b]{0.19\textwidth}
            \includegraphics[width=\textwidth]{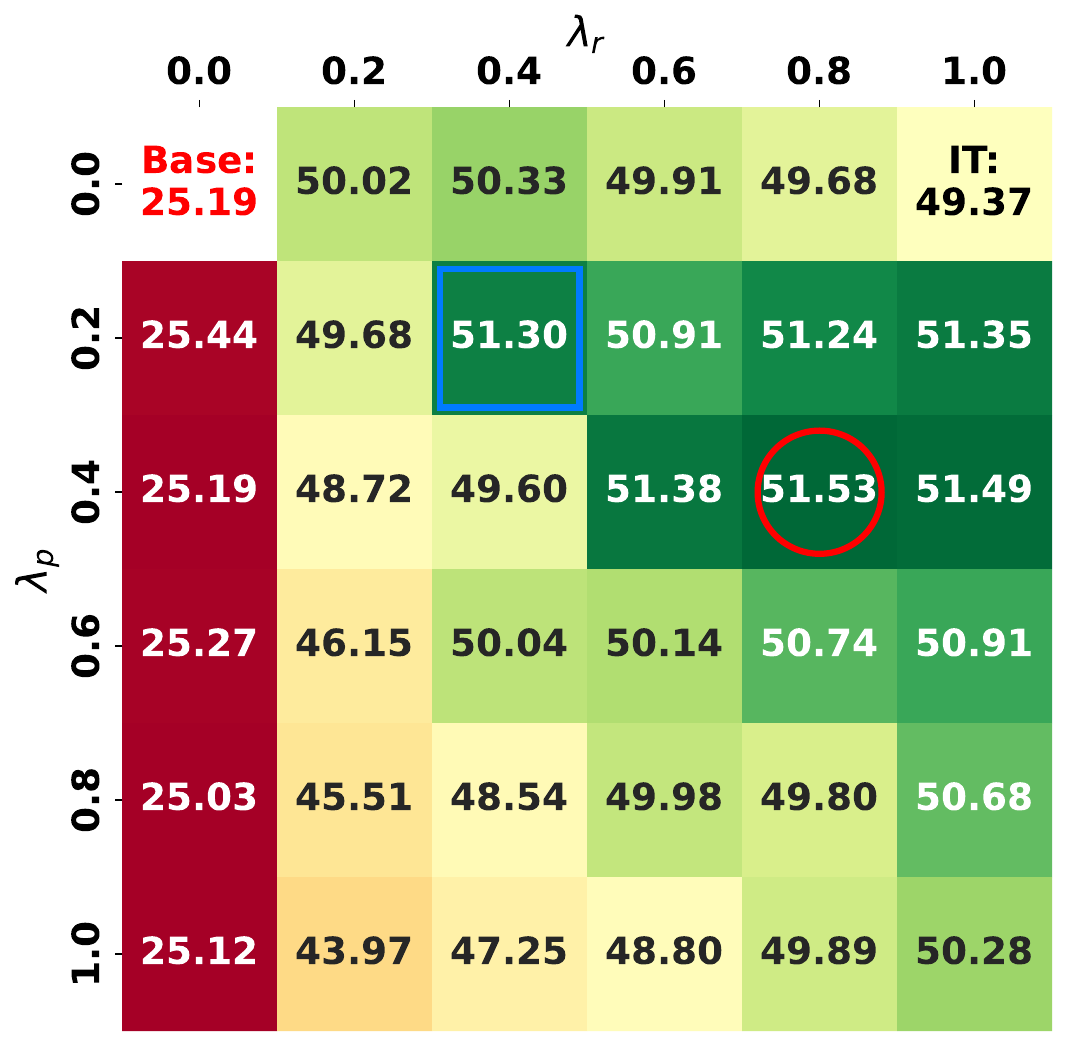}
        \end{subfigure} &
        \begin{subfigure}[b]{0.19\textwidth}
            \includegraphics[width=\textwidth]{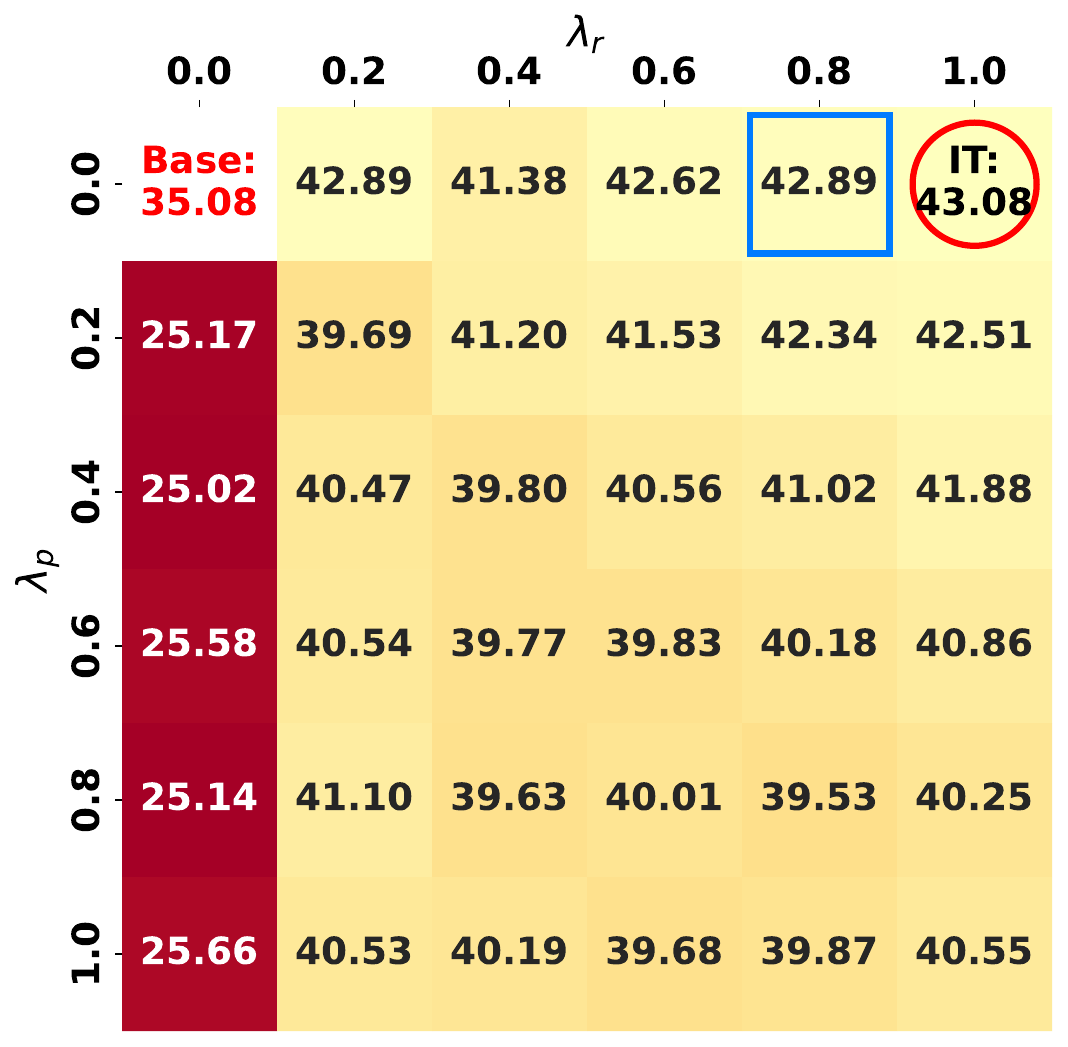}
        \end{subfigure} &
        \begin{subfigure}[b]{0.19\textwidth}
            \includegraphics[width=\textwidth]{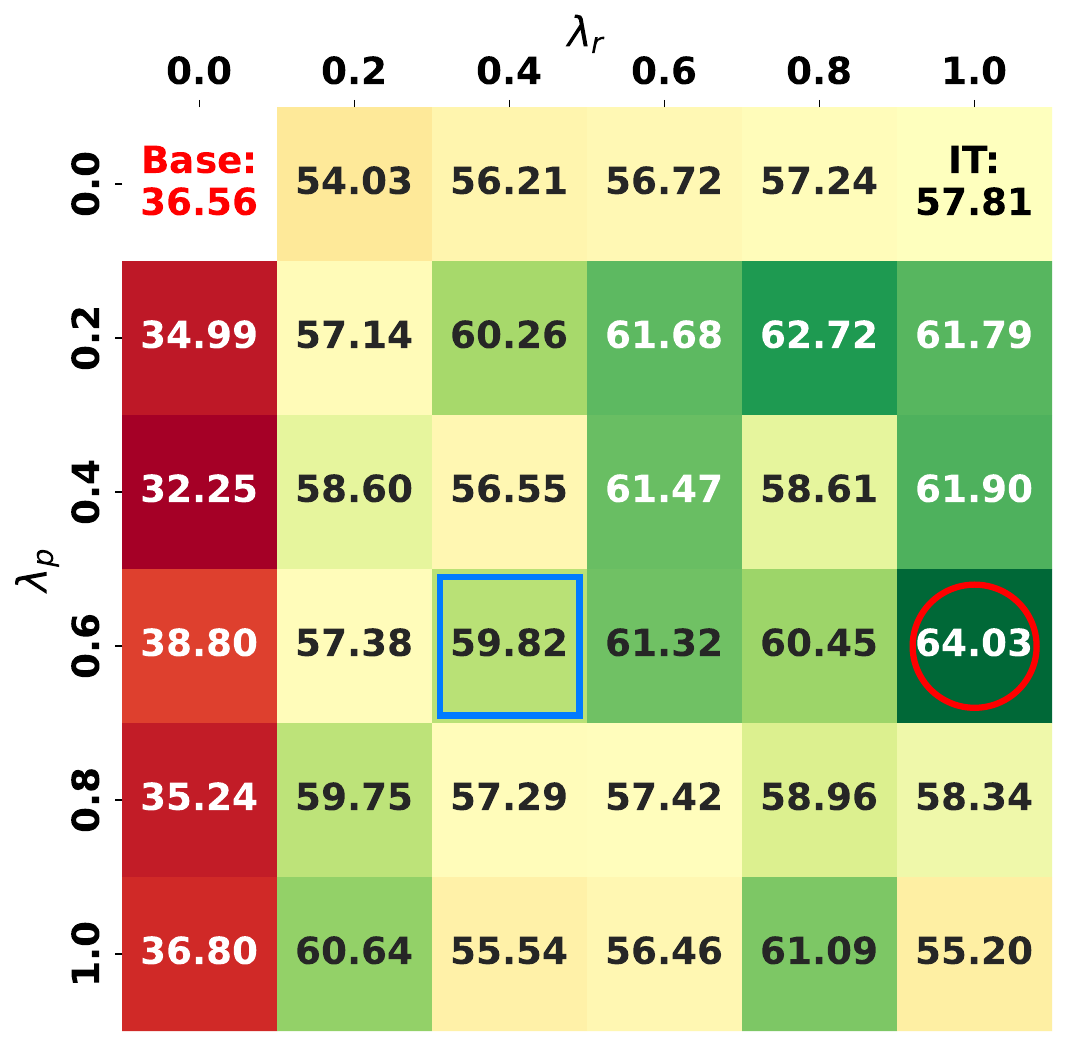}
        \end{subfigure} &
        \begin{subfigure}[b]{0.19\textwidth}
            \includegraphics[width=\textwidth]{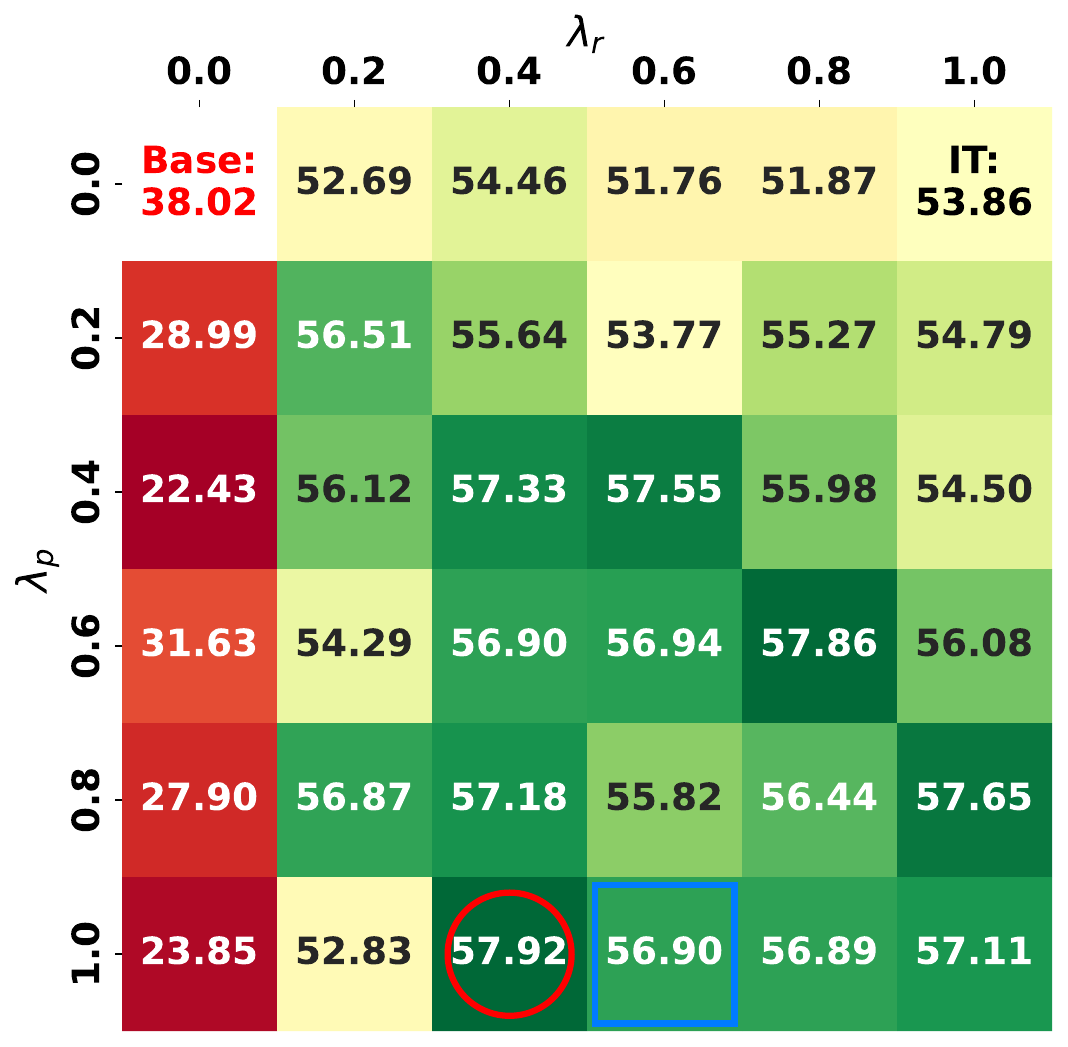}
        \end{subfigure} \\

        \rotatebox{90}{\parbox{2.5cm}{\centering {LIMA}}} &  
        \begin{subfigure}[b]{0.19\textwidth}
            \includegraphics[width=\textwidth]{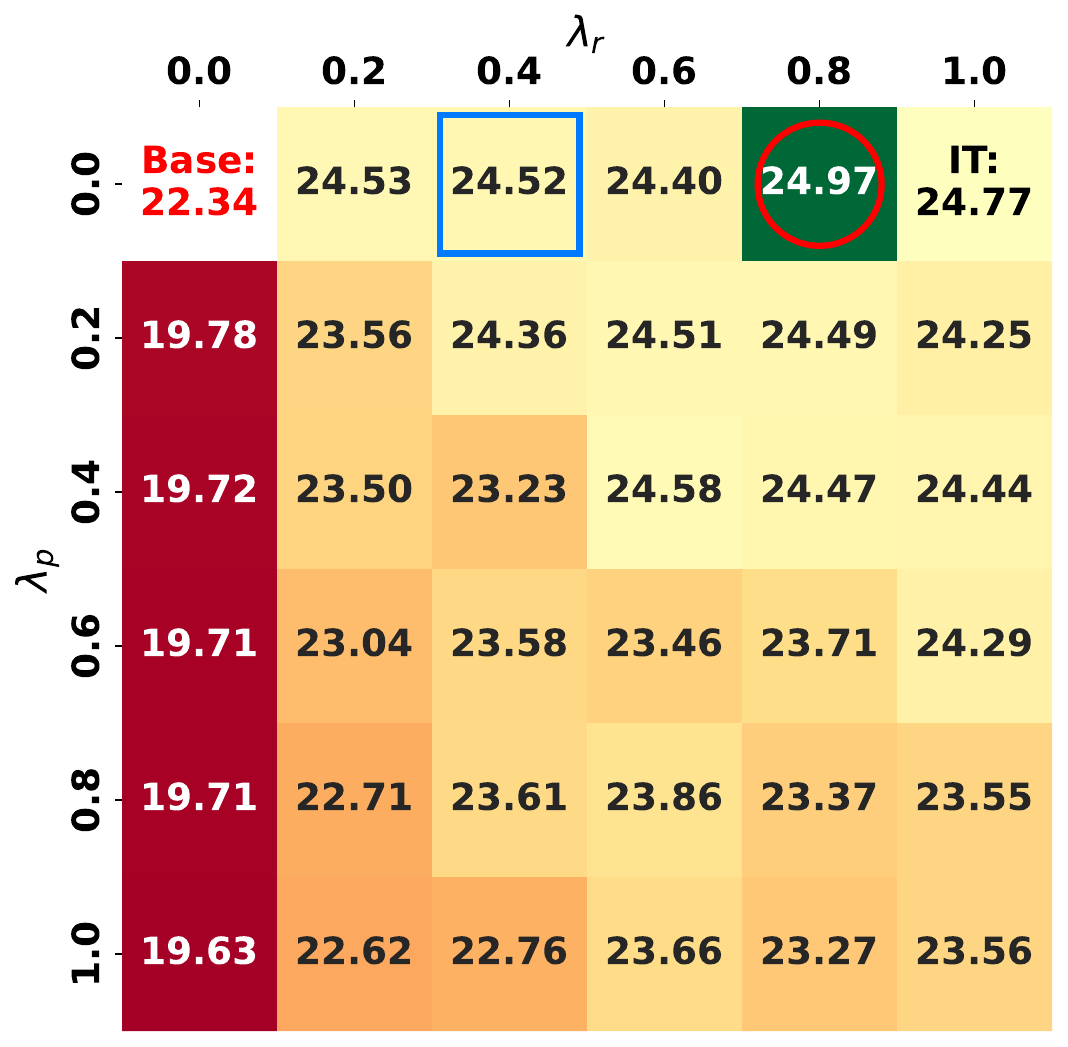}
        \end{subfigure} &
        \begin{subfigure}[b]{0.19\textwidth}
            \includegraphics[width=\textwidth]{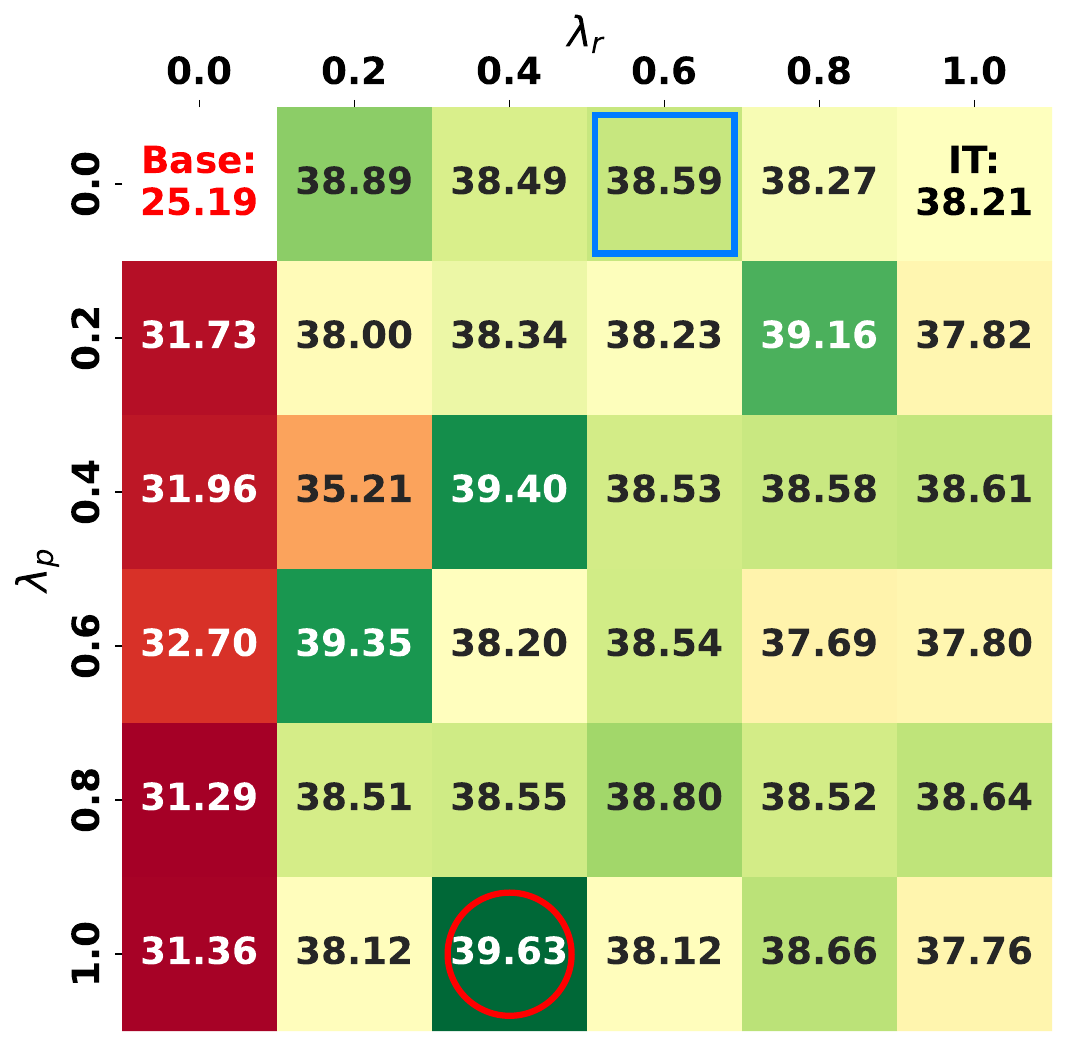}
        \end{subfigure} &
        \begin{subfigure}[b]{0.19\textwidth}
            \includegraphics[width=\textwidth]{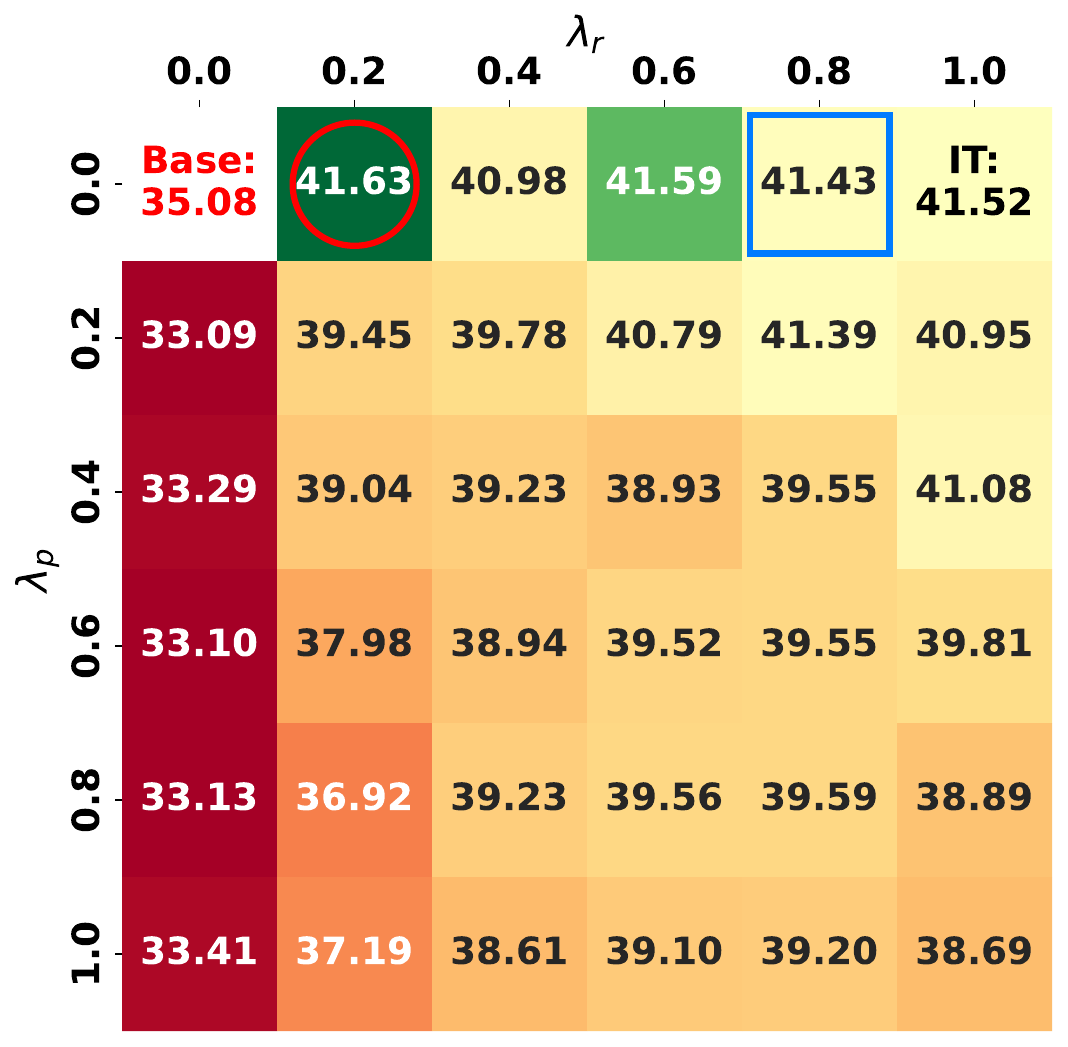}
        \end{subfigure} &
        \begin{subfigure}[b]{0.19\textwidth}
            \includegraphics[width=\textwidth]{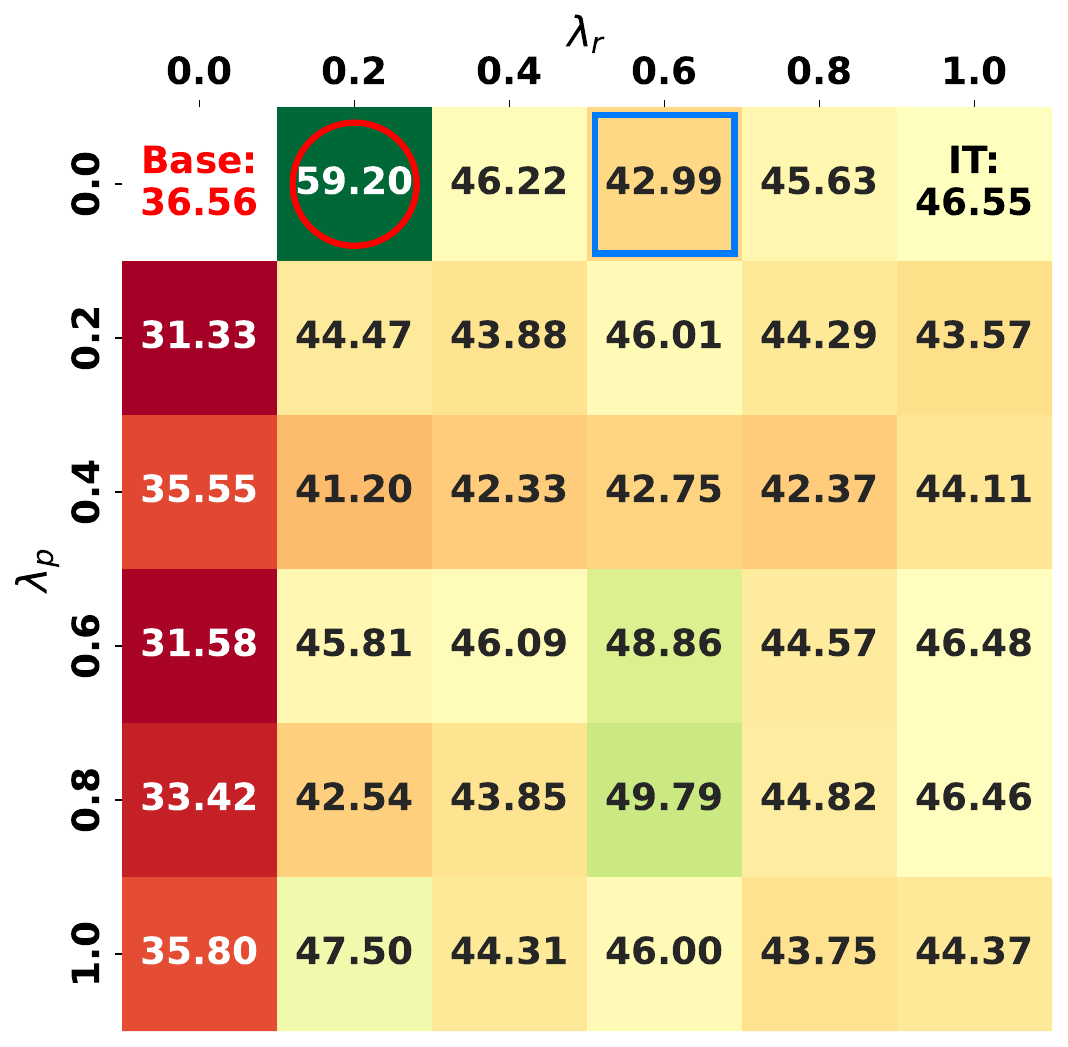}
        \end{subfigure} &
        \begin{subfigure}[b]{0.19\textwidth}
            \includegraphics[width=\textwidth]{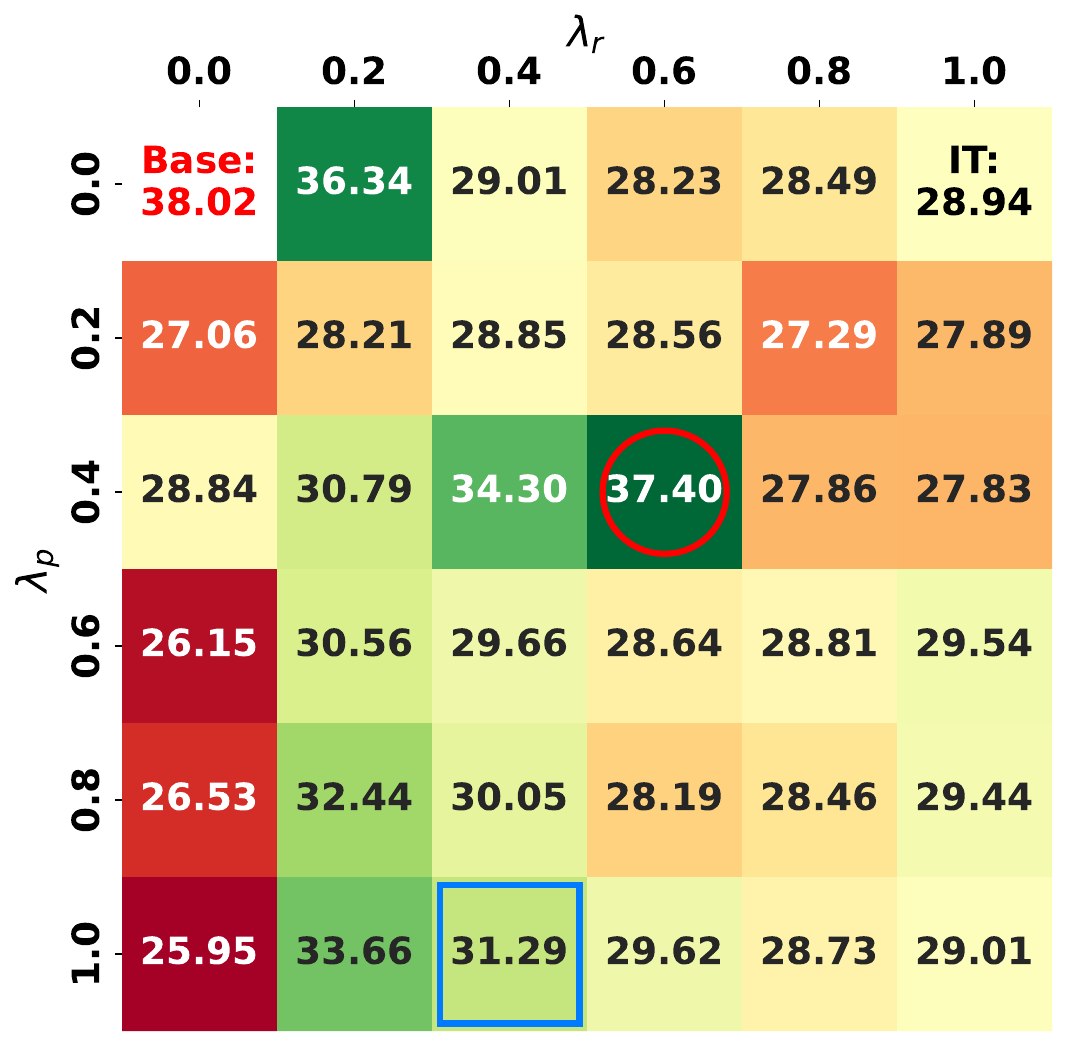}
        \end{subfigure} \\
    \end{tblr}

    \caption{Heatmaps depicting average performance across \changed{five} benchmarks (MMLU, BBH, AlpacaEval, IFEval and MT-Bench) for different configurations of $(\lambda_p, \lambda_r)$ and for different instruction-tuned models which underwent DPO on UltraFeedback dataset. Best performing configuration after DPO is highlighted with a red circle and best performing configuration from before DPO is highlighted with a blue square. The color map is based on relative gain with respect to conventional instruction tuning. Rows correspond to prompt token weights ($\lambda_p$) and columns correspond to response token weights ($\lambda_r$). Conventional instruction tuning is marked with \texttt{IT} and base model performance is marked with \texttt{Base}.}
    \label{fig:dpo_heatmap}
\end{figure*}

\begin{table}[t!]
    \small
    \centering
    \scalebox{0.8}{
    \begin{tabular}{ccc}
    \hline
        \begin{tabular}{c}\textbf{Evaluation}\\\textbf{Benchmark}\end{tabular} & \begin{tabular}{c}\textbf{Average}\\\textbf{Optimal} $\lambda_p$\end{tabular} & \begin{tabular}{c}\textbf{Average}\\\textbf{Optimal} $\lambda_r$\end{tabular} \\
        \hline
        MMLU & 0.28 & 0.56 \\
        BBH & 0.17 & 0.61 \\
        AlpacaEval & 0.36 & 0.64 \\
        IFEval & 0.48 & 0.43 \\
        MT-Bench & 0.23 & 0.55 \\
        \hline
    \end{tabular}}
    \caption{Optimal prompt-token weight ($\lambda_p$) and response-token weight ($\lambda_r$) for various evaluation benchmarks, averaged across different $(model$, $training\_dataset)$ combinations. The optimal response-token weight varies from moderate to high, with values ranging from $0.43$ for IFEval to $0.64$ for AlpacaEval, while the optimal prompt-token weight varies from low to moderate, from $0.17$ for BBH to $0.48$ for IFEval.}
    \label{tab:benchmark_trends}
\end{table}

\paragraph{Varying Effects of Response-Token Weight on Instruction Adherence and Conversational Fluency.}
The results on IFEval, AlpacaEval, and MT-Bench across different models and training datasets, as observed in Figures ~\ref{fig:tulu_all}, \ref{fig:alpaca-cleaned_all} and \ref{fig:lima_all}, reveal a \textit{trade-off} between instruction adherence and conversational fluency. \changed{For IFEval, which measures instruction-following ability, lower response weights are favoured -- in $60\%$ of cases, i.e., $9$ out of $15$ $(model$, $training\_dataset)$ combinations, $\lambda_r \le 0.4$ is optimal. In contrast, conversational fluency benchmarks -- AlpacaEval and MT-Bench -- prefer relatively higher response weights -- in $60\%$ of settings, i.e., $18$ out of $30$ combinations, $\lambda_r \ge 0.6$ is optimal, and in $80\%$ cases, $\lambda_r \ge 0.4$ yields best performance. Table~\ref{tab:benchmark_trends} also reflects this trend -- the average optimal $\lambda_r$ is relatively lower for IFEval ($0.43$) compared to AlpacaEval ($0.64$) and MT-Bench ($0.55$).} These findings underscore the importance of tailoring prompt and response weighting in \wit~to align with the intended downstream behaviour of instruction-tuned models.

\begin{figure*}[ht!]
    \centering
    \renewcommand{\arraystretch}{1.2}  
    \begin{tabular}{@{}p{1.2em}@{} @{\hskip 0em} c@{\hskip 0em} c@{\hskip 0em} c@{\hskip 0em} c@{\hskip 0em} c@{}}
    
        \rotatebox{90}{\parbox{2.5cm}{\centering T\"ulu-v2}} &  
        \begin{subfigure}[b]{0.19\textwidth}
            \caption*{\centering Llama-3-1B}
            \includegraphics[width=\textwidth]{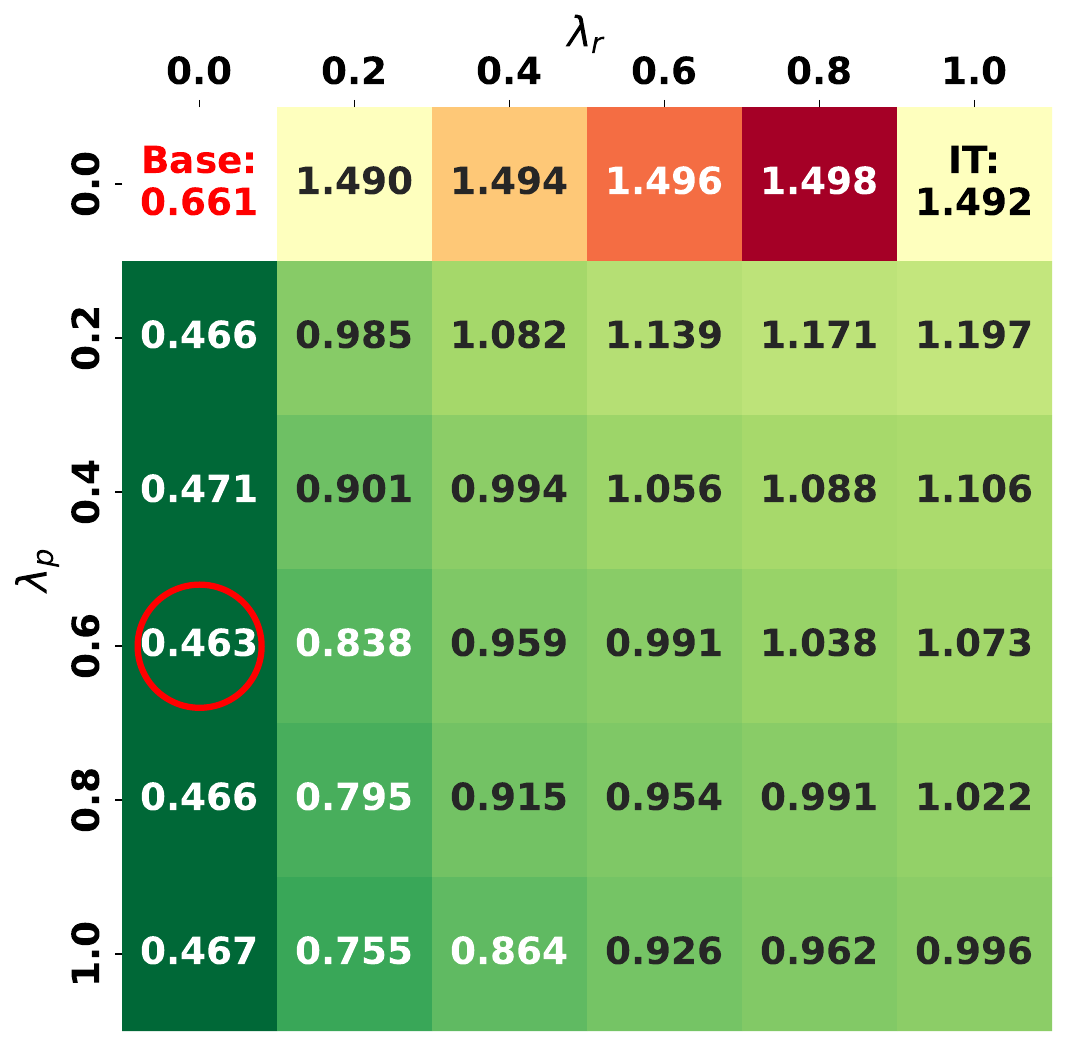}
        \end{subfigure} &
        \begin{subfigure}[b]{0.19\textwidth}
            \caption*{\centering Gemma-2-2B}
            \includegraphics[width=\textwidth]{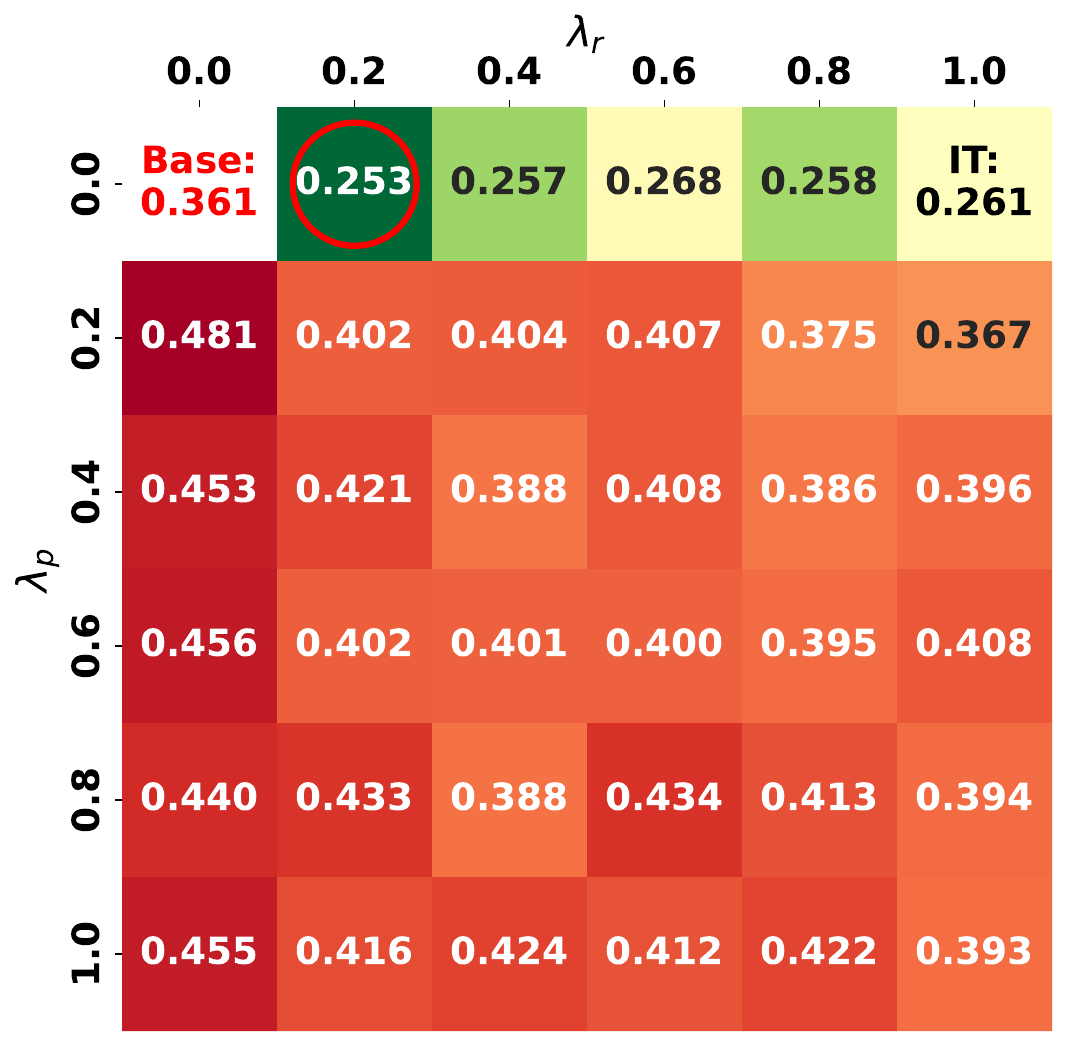}
        \end{subfigure} &
        \begin{subfigure}[b]{0.19\textwidth}
            \caption*{\centering Llama-3-3B}
            \includegraphics[width=\textwidth]{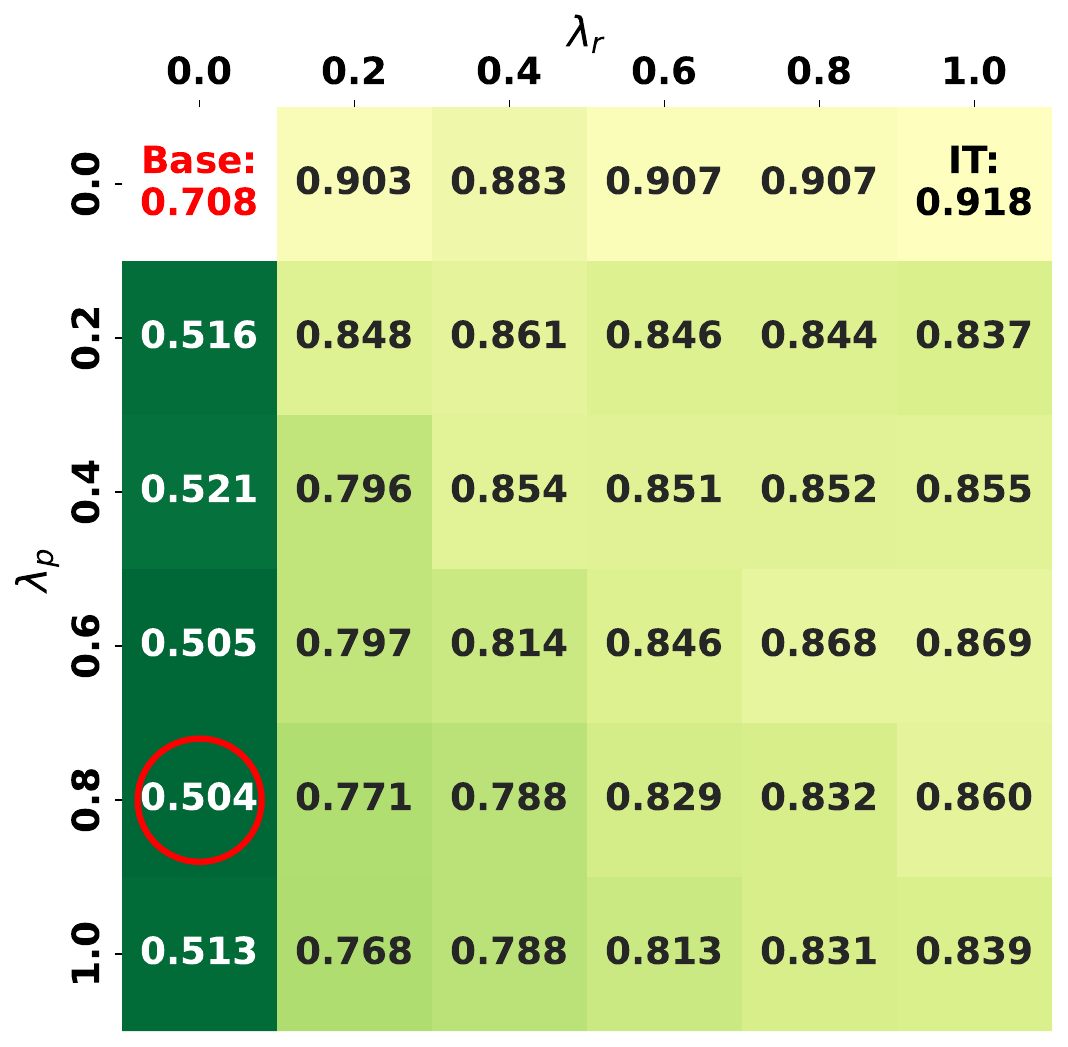}   
        \end{subfigure} &
        \begin{subfigure}[b]{0.19\textwidth}
            \caption*{\centering Mistral-7B}
            \includegraphics[width=\textwidth]{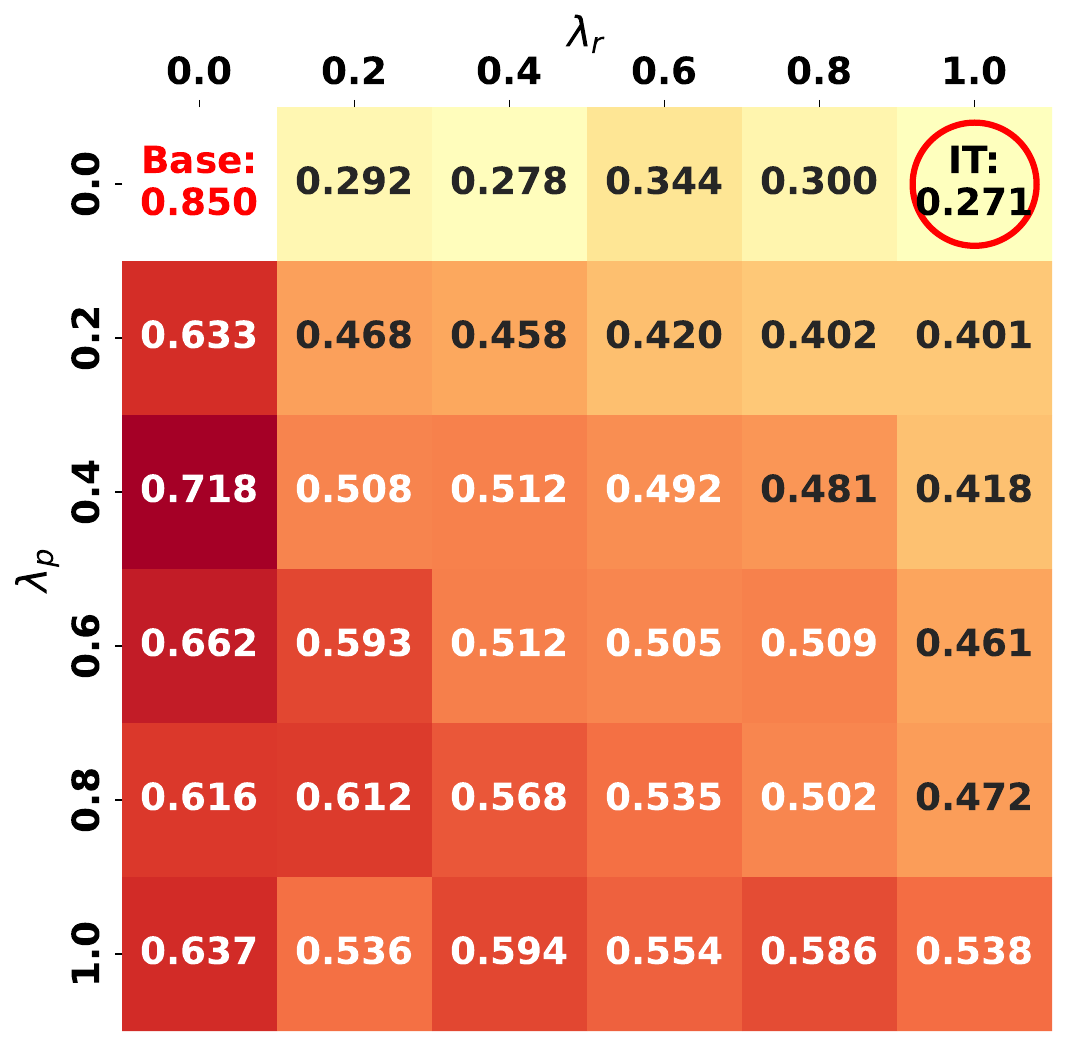}
        \end{subfigure} &
        \begin{subfigure}[b]{0.19\textwidth}
            \caption*{\centering Llama-3-8B}
            \includegraphics[width=\textwidth]{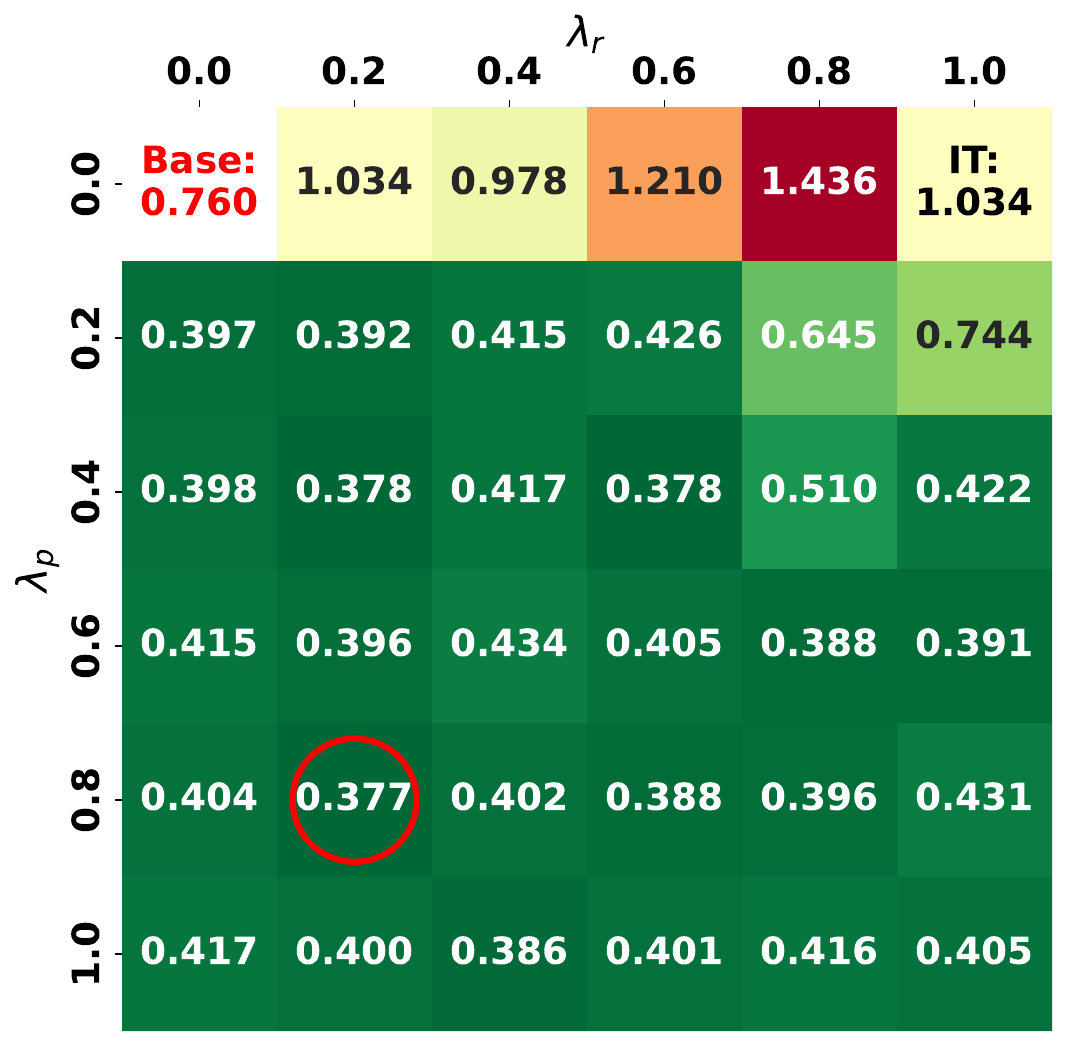}
        \end{subfigure} \\
        
        \rotatebox{90}{\parbox{2.5cm}{\centering {Alpaca-Cleaned}}} &  
        \begin{subfigure}[b]{0.19\textwidth}
            \includegraphics[width=\textwidth]{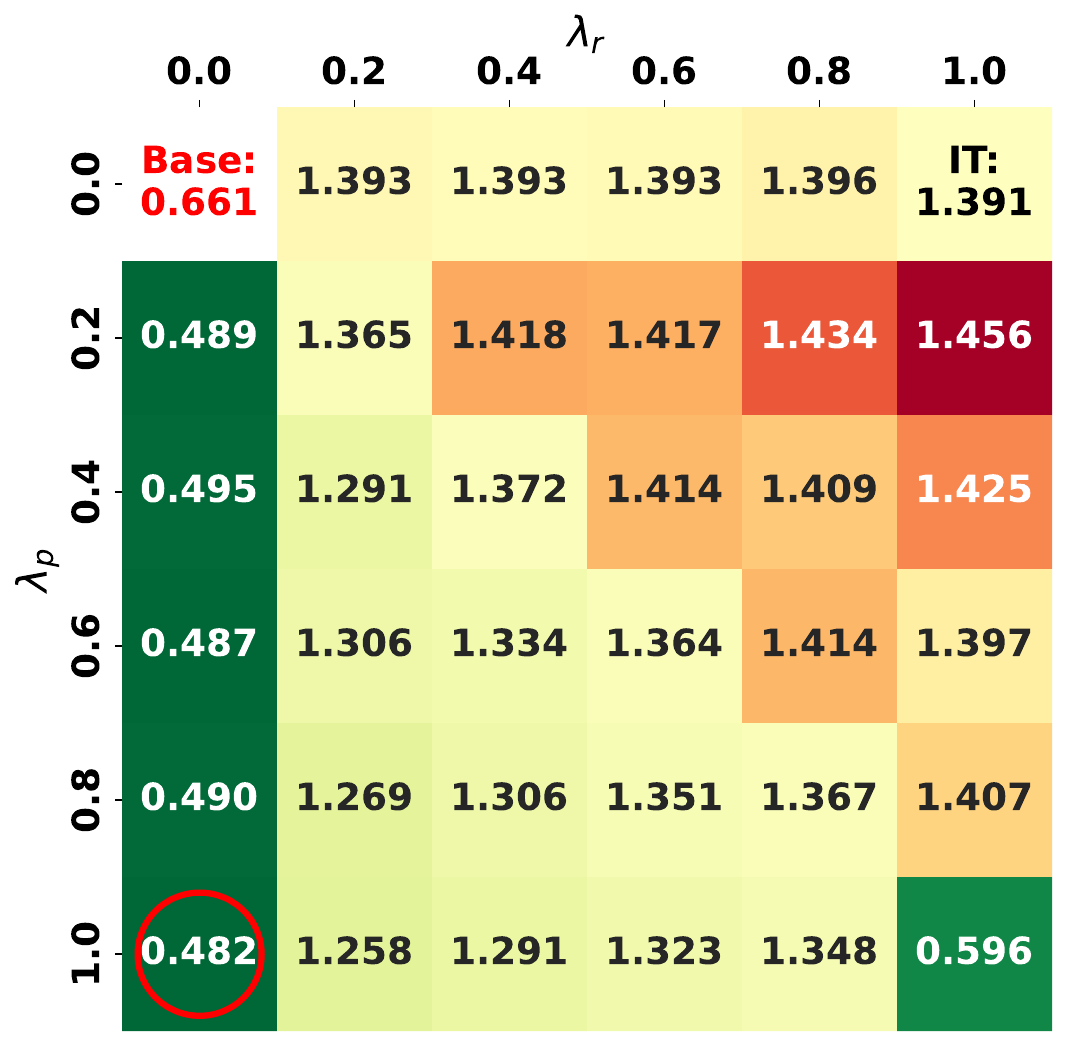}
        \end{subfigure} &
        \begin{subfigure}[b]{0.19\textwidth}
            \includegraphics[width=\textwidth]{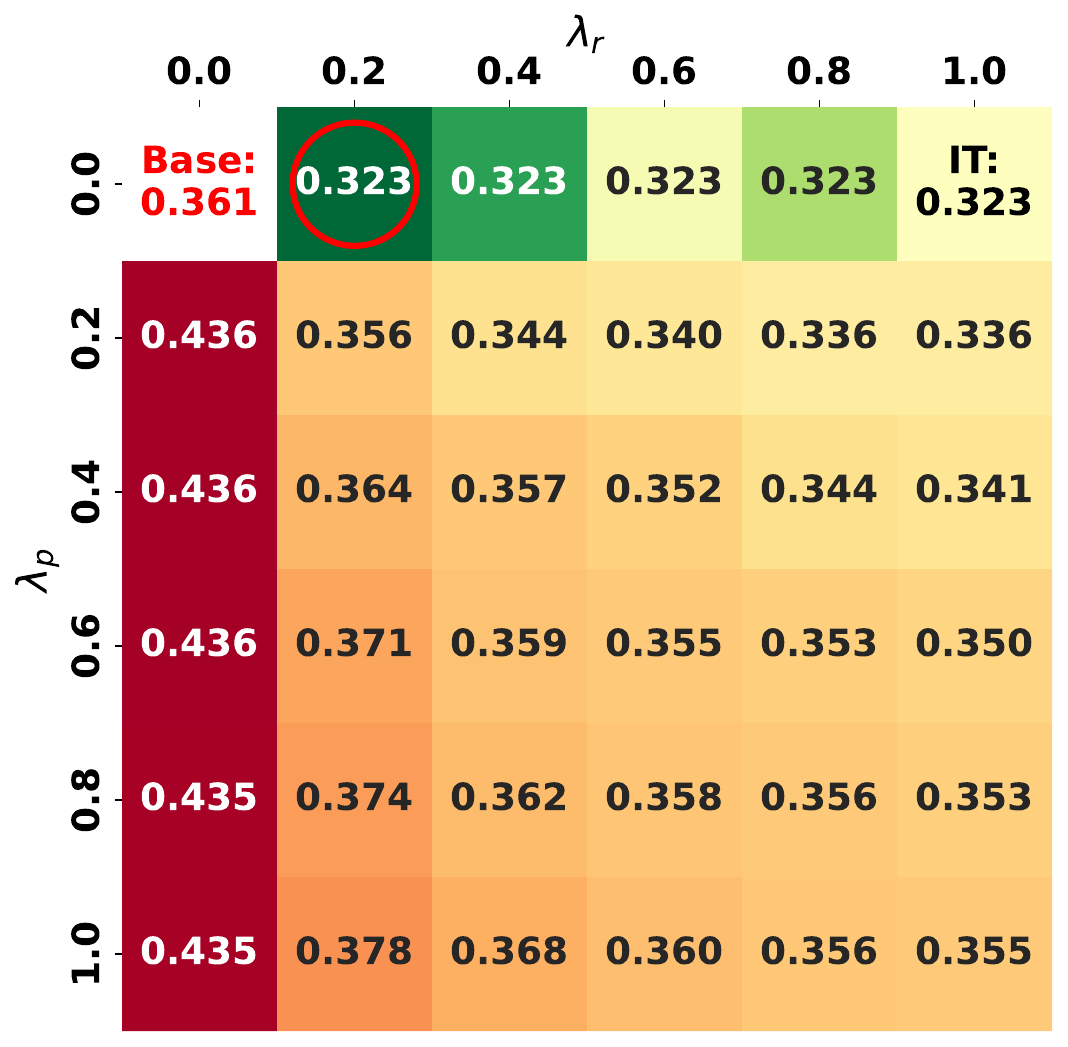}
        \end{subfigure} &
        \begin{subfigure}[b]{0.19\textwidth}
            \includegraphics[width=\textwidth]{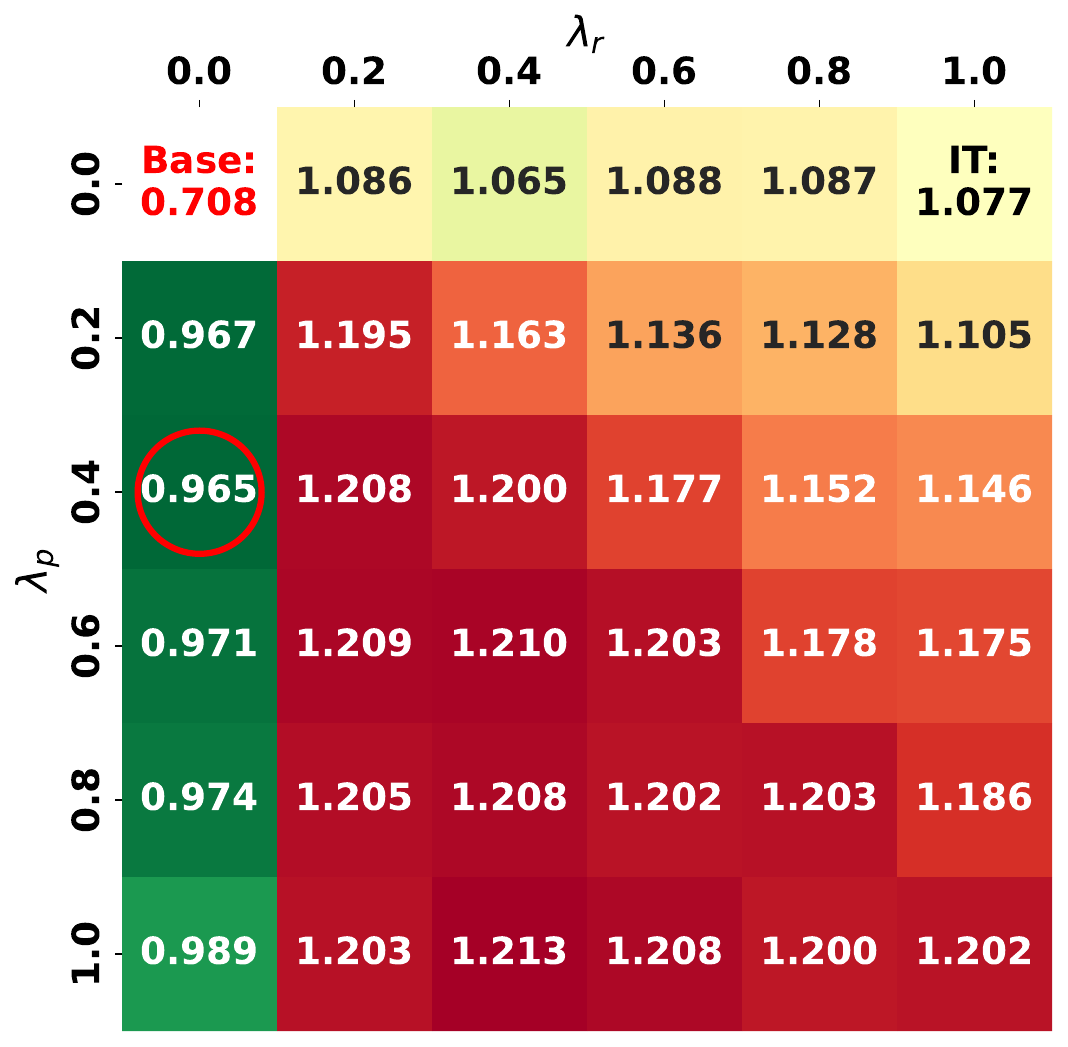}
        \end{subfigure} &
        \begin{subfigure}[b]{0.19\textwidth}
            \includegraphics[width=\textwidth]{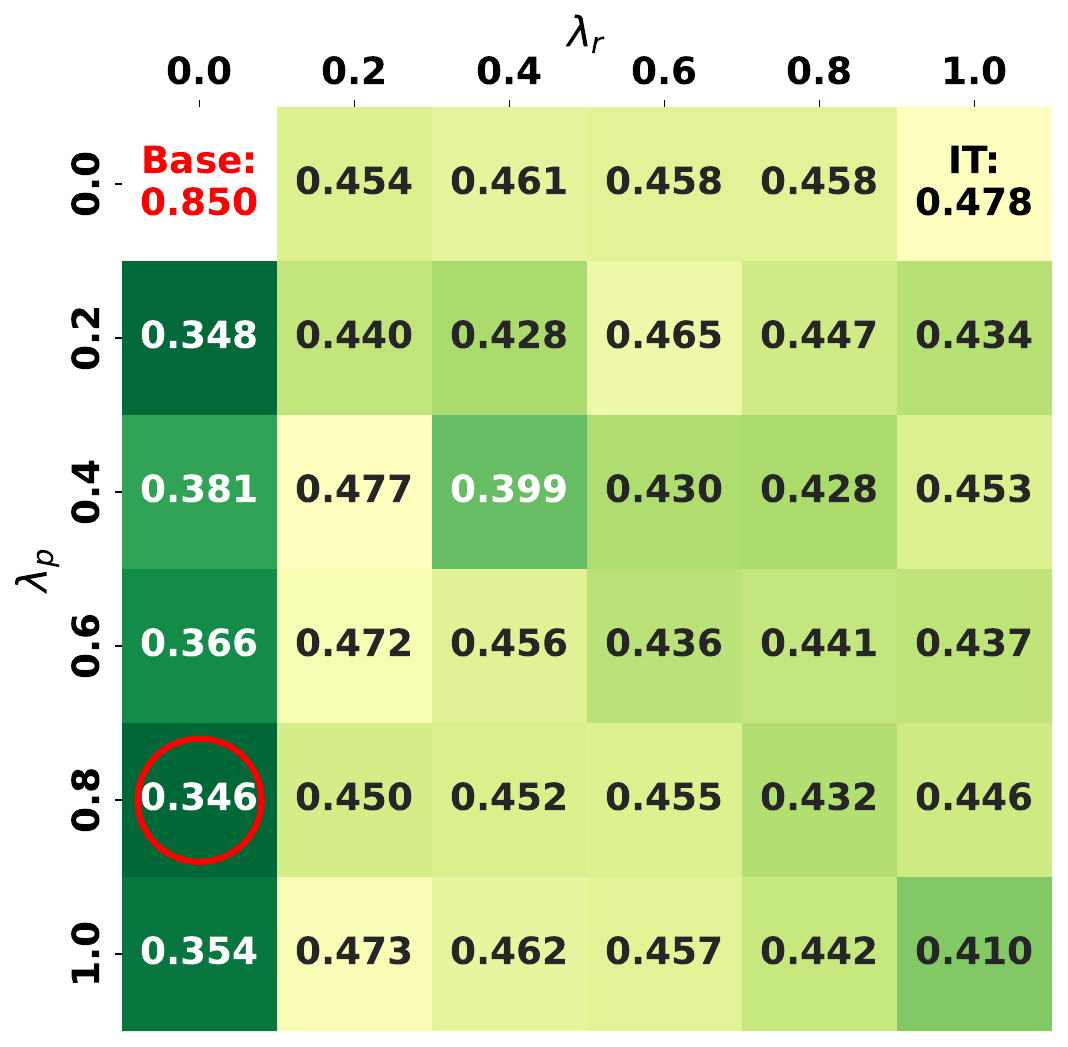}
        \end{subfigure} &
        \begin{subfigure}[b]{0.19\textwidth}
            \includegraphics[width=\textwidth]{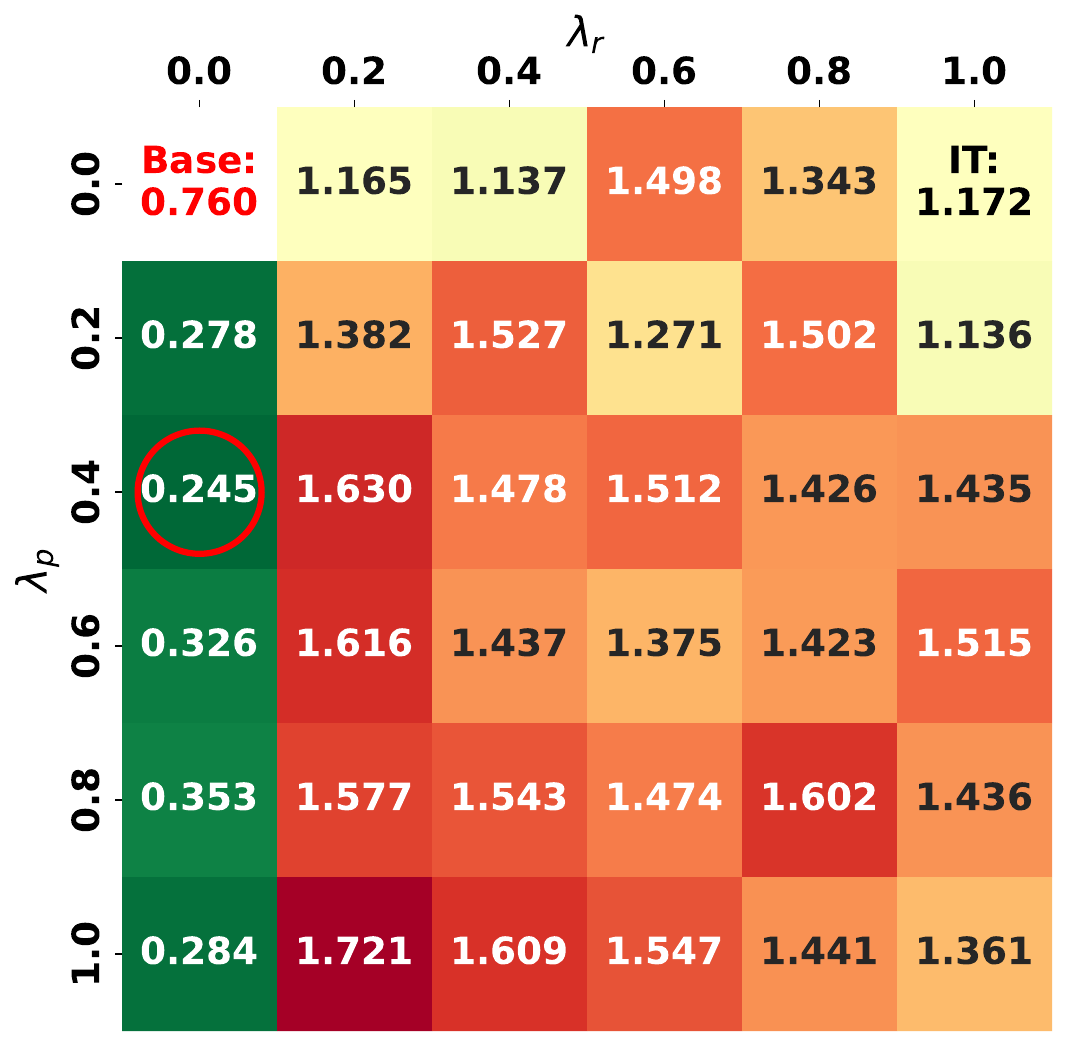}
        \end{subfigure} \\

        \rotatebox{90}{\parbox{2.5cm}{\centering {LIMA}}} &  
        \begin{subfigure}[b]{0.19\textwidth}
            \includegraphics[width=\textwidth]{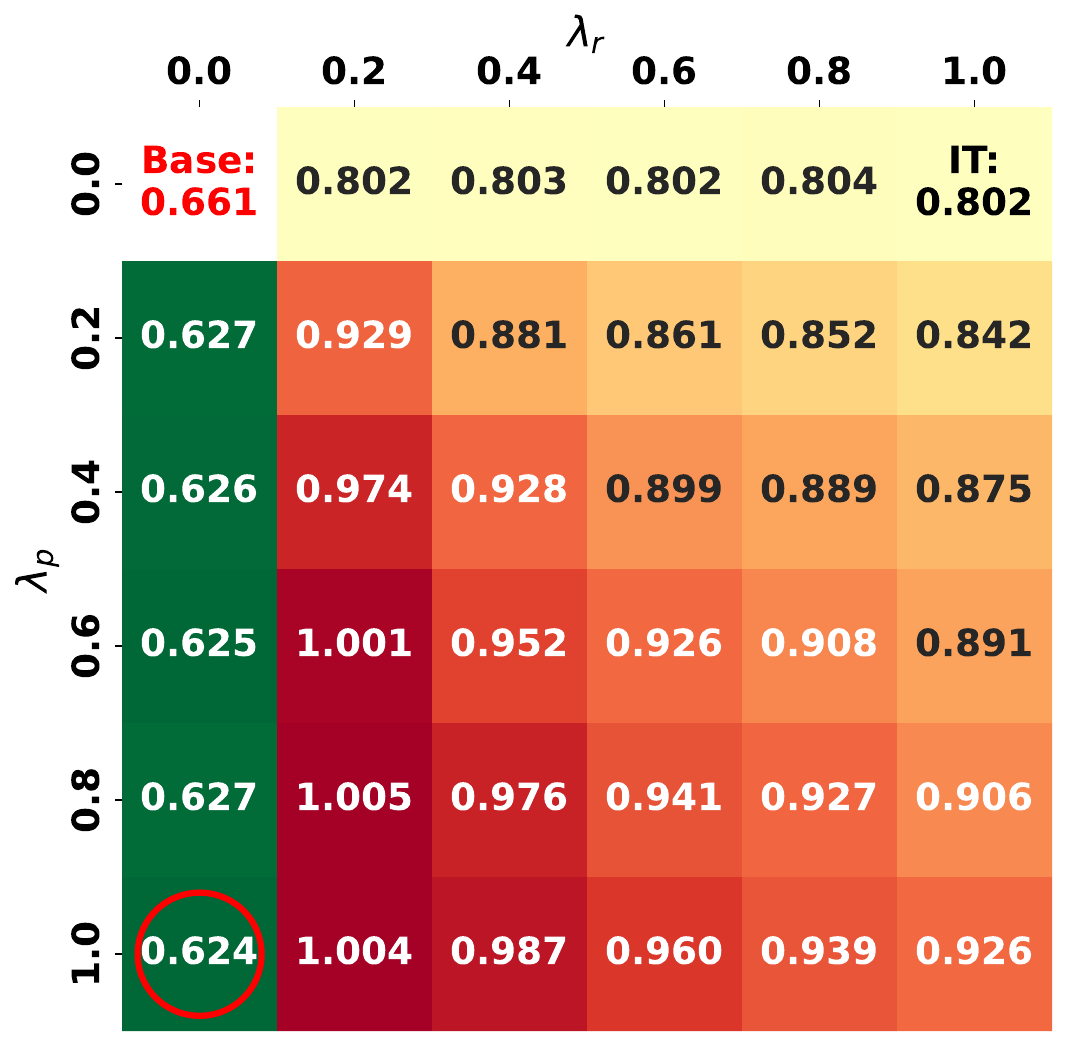}
        \end{subfigure} &
        \begin{subfigure}[b]{0.19\textwidth}
            \includegraphics[width=\textwidth]{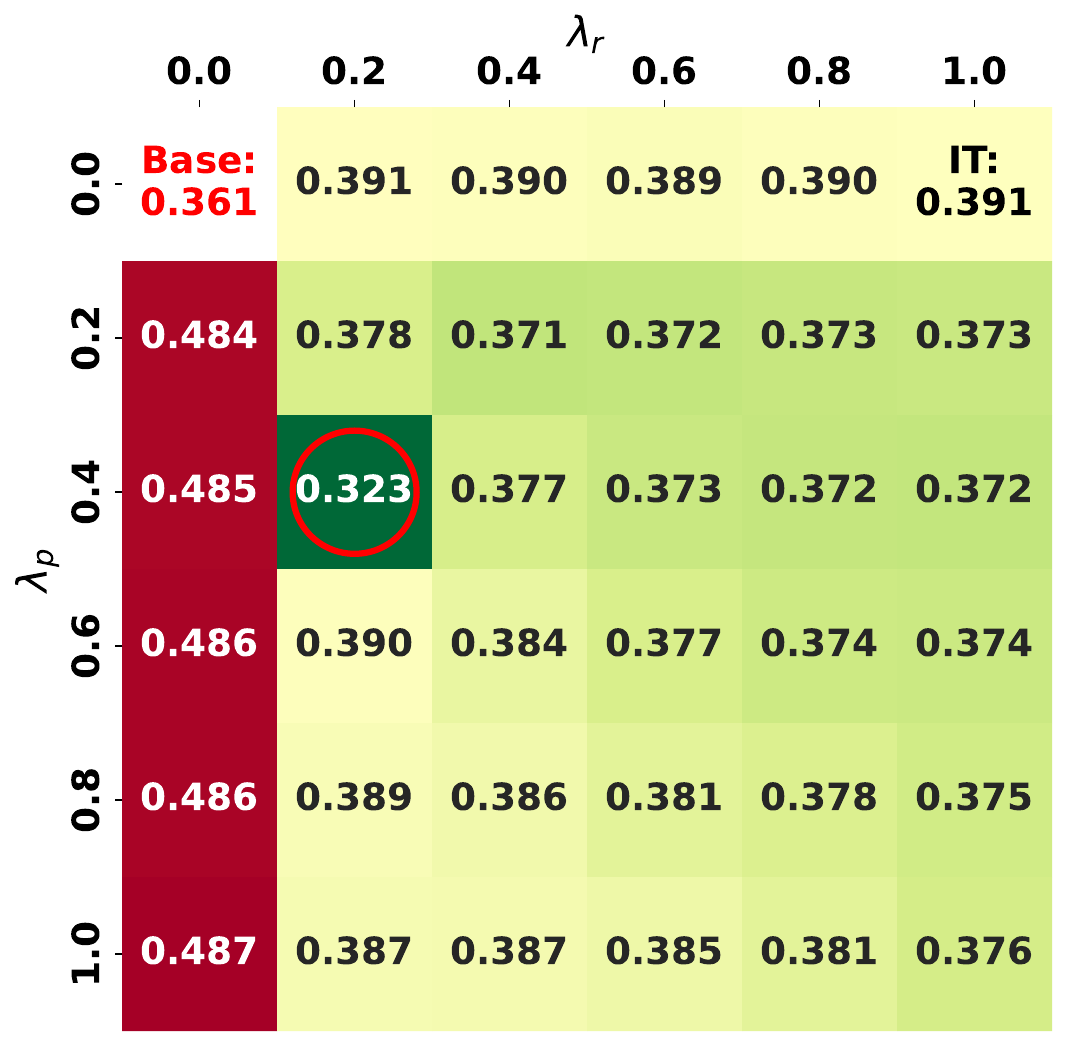}
        \end{subfigure} &
        \begin{subfigure}[b]{0.19\textwidth}
            \includegraphics[width=\textwidth]{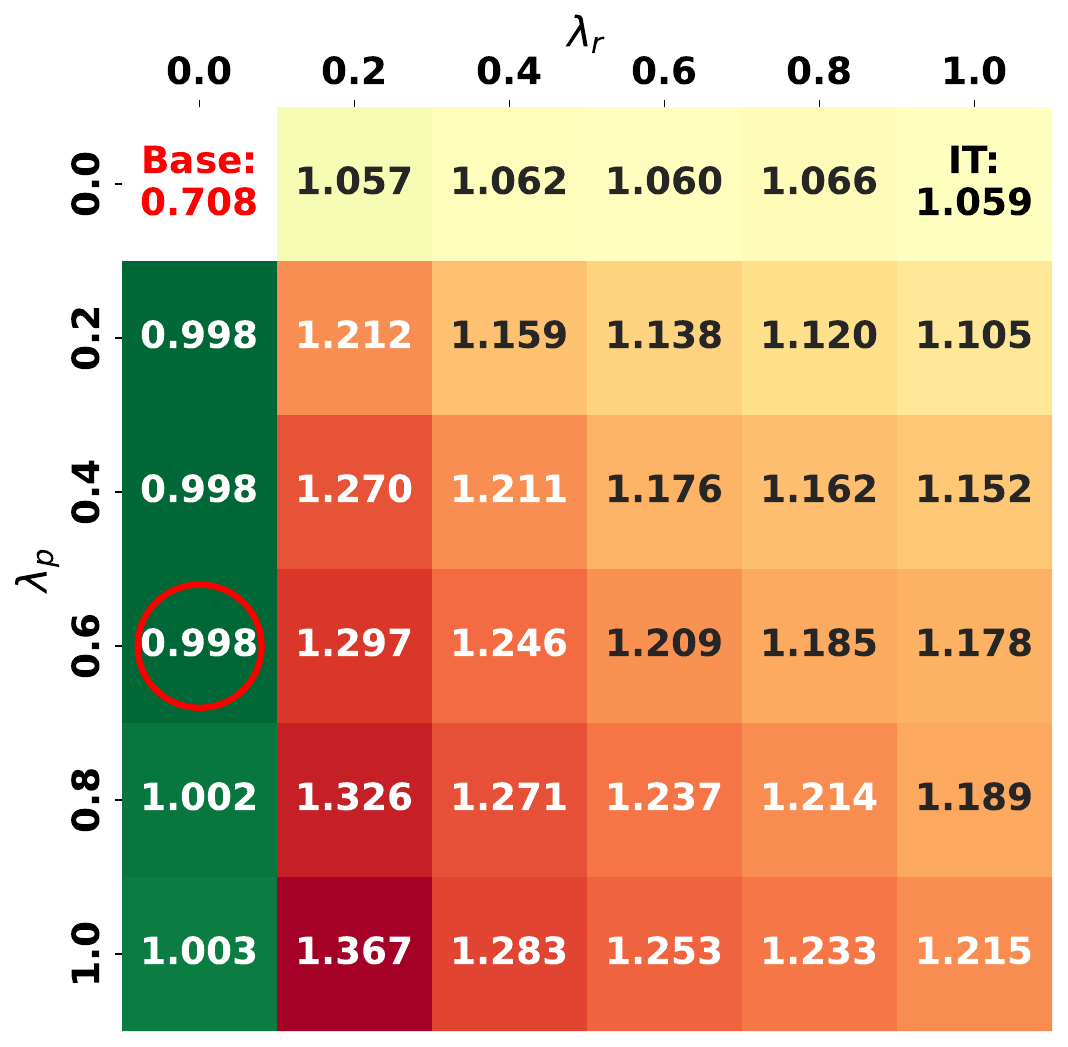}
        \end{subfigure} &
        \begin{subfigure}[b]{0.19\textwidth}
            \includegraphics[width=\textwidth]{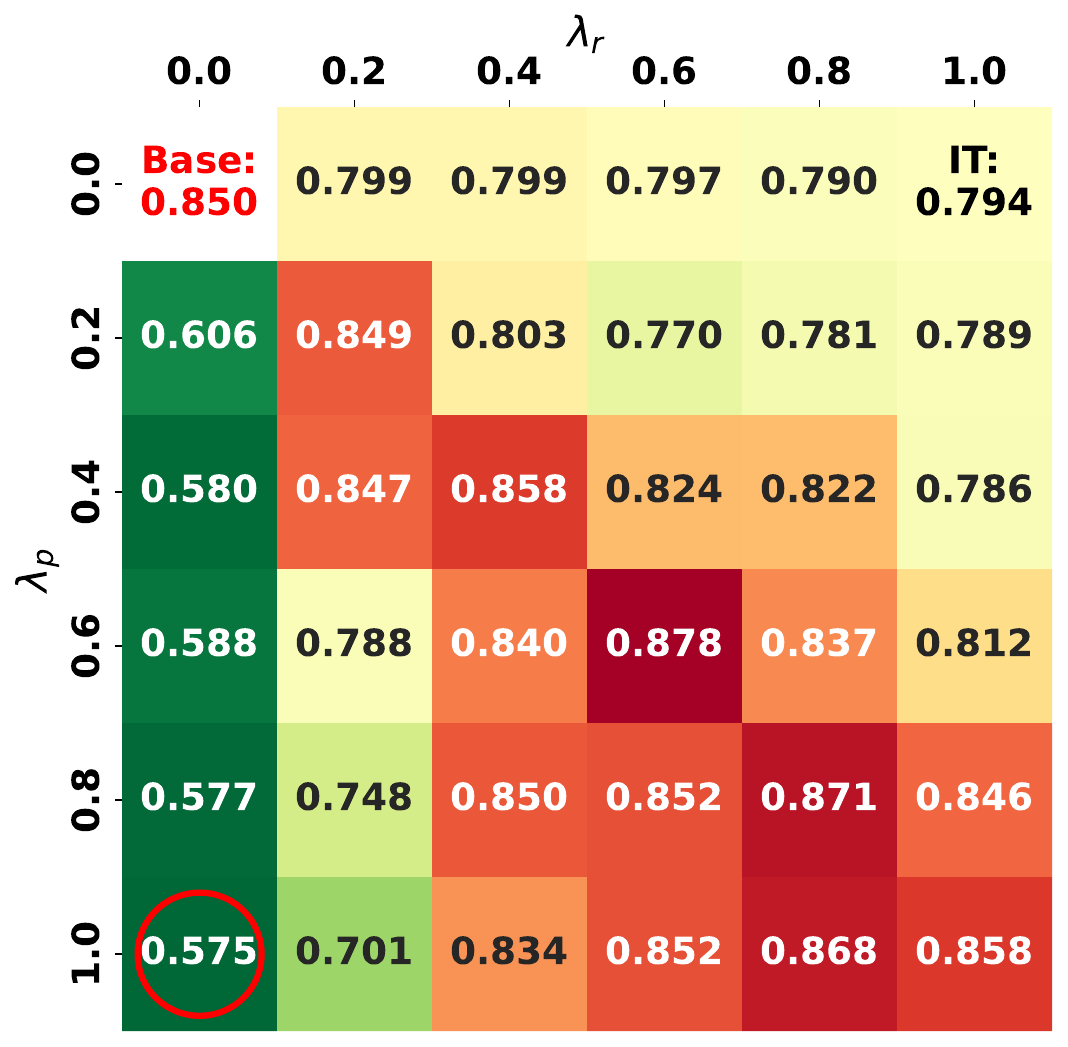}
        \end{subfigure} &
        \begin{subfigure}[b]{0.19\textwidth}
            \includegraphics[width=\textwidth]{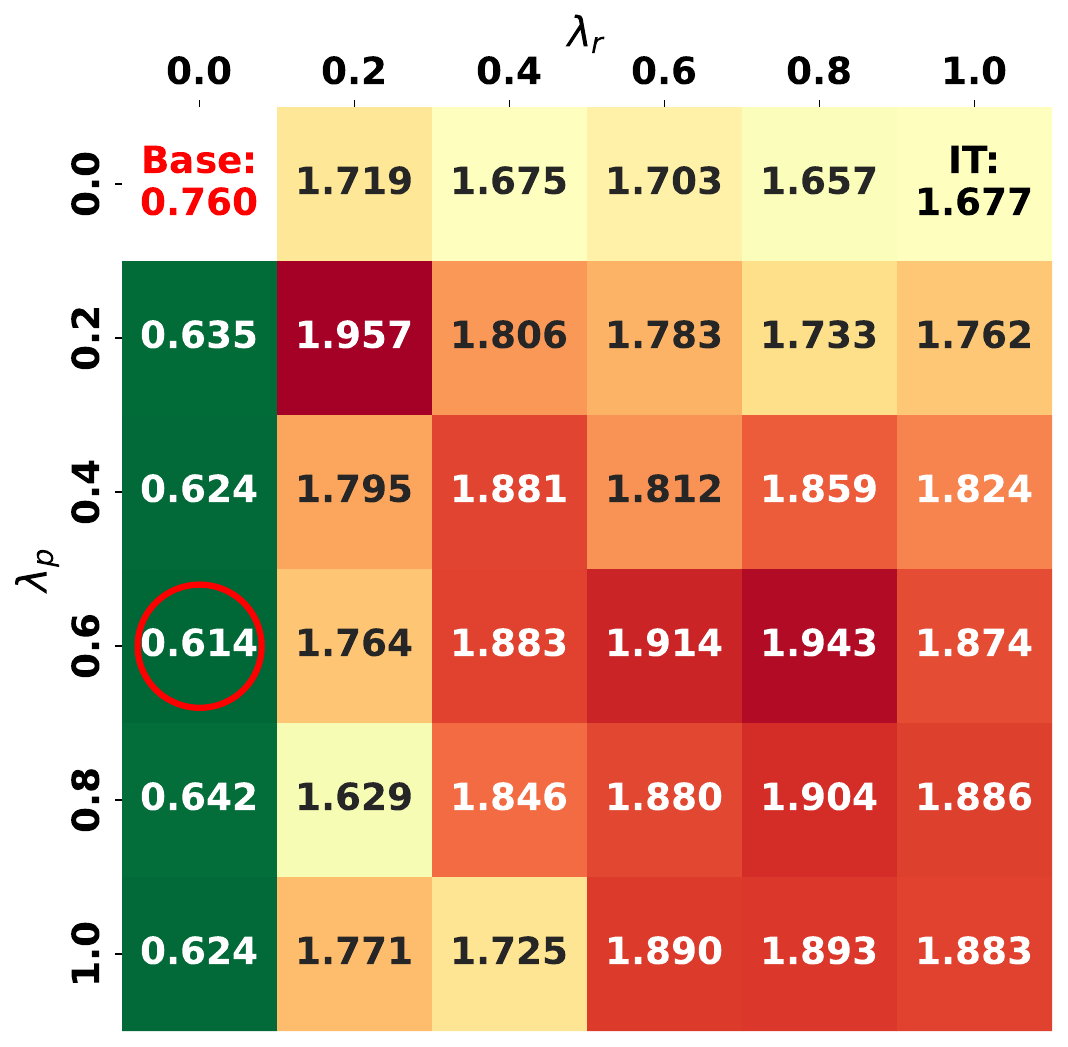}
        \end{subfigure} \\
    \end{tabular}

    \caption{Heatmaps depicting Prompt Sensitivity Index (POSIX) for various ($\lambda_p$, $\lambda_r$) across models finetuned on T\"ulu-v2, Alpaca-Cleaned and LIMA. Least prompt sensitivity configuration is highlighted with a red circle. The color map is based on relative gain with respect to conventional instruction tuning. Rows correspond to prompt token weights ($\lambda_p$) and columns correspond to response token weights ($\lambda_r$). Conventional instruction tuning is marked with \texttt{IT} and base model performance is marked with \texttt{Base}.}
    \label{fig:posix_heatmap}
\end{figure*}

\paragraph{Prompt-Only Tuning Also Enhances Base Model Capabilities.}
As depicted in Figures~\ref{fig:tulu_all}, \ref{fig:alpaca-cleaned_all} and \ref{fig:lima_all}, training with $\lambda_r=0$, i.e., computing loss only on prompt tokens, still leads to notable improvements over the base LM across all benchmarks, except AlpacaEval, when using the large and diverse T\"ulu-v2 dataset for finetuning. In contrast, for smaller datasets like Alpaca-Cleaned and LIMA, improvements appear primarily on IFEval. Thus, even without direct optimization on response tokens, prompt-only finetuning enhances instruction adherence, suggesting that training on unannotated prompts can also impart instruction-following. The observations also indicate that prompt-only tuning may require sufficiently large and diverse data to generalize effectively. Overall, the findings highlight the potential of leveraging large-scale unannotated datasets to boost instruction-following abilities without extensive labeled prompt-response pairs.

\subsection{Transferability of Gains from \wit~to Preference Alignment Phase}

\changed{To assess whether the gains from \wit~persist after preference alignment training phase, we performed DPO on models instruction-tuned with various prompt and response token weights. Figure~\ref{fig:dpo_heatmap} depicts the average benchmark performance of models that underwent DPO on top of the instruction-tuned models from Figure~\ref{fig:main_heatmap}. We find that DPO performed on top of conventional instruction-tuned models still yields suboptimal results when compared to DPO performed on top of weighted instruction tuned models. Table~\ref{tab:dpo_relative_gain} shows the relative performance gain on downstream tasks for DPO on top of \wit~(with optimal setting of prompt and response token weights) over DPO on conventional instruction tuning. We find that the optimal configuration of prompt and response token weights for DPO yields a relative gain of nearly $8\%$. Furthermore, while we note that the exact optimal $(\lambda_p, \lambda_r)$ configuration for DPO might be different compared to optimal $(\lambda_p, \lambda_r)$ of instruction-tuning, we find that DPO performed on top of optimal $(\lambda_p, \lambda_r)$ configuration for instruction tuning still yields $2.44\%$ relative gain in performance over DPO on top of conventional instruction tuned models. These findings highlight that models fine-tuned by \wit~serve as better starting points for the preference alignment training phase, and the performance gains transfer even after DPO training.}

\subsection{Robustness to Prompt Variations}
\begin{table}[t]
    \centering
    \resizebox{\columnwidth}{!}{
        \begin{tabular}{ccccc}
            \hline
            \textbf{Model} & \begin{tabular}{c}\textbf{Training}\\\textbf{Data}\end{tabular} & \begin{tabular}{c}\textbf{DPO on top of}\\\textbf{conventional}\\\textbf{instruction tuning}\end{tabular} & \begin{tabular}{c}\textbf{DPO on top of}\\\textbf{weighted}\\\textbf{instruction tuning}\\\textbf{(Optimal} $\mathbf{\lambda_p, \lambda_r}$\textbf{)}\end{tabular} & \begin{tabular}{c}\textbf{Relative}\\\textbf{Gain}\end{tabular} \\ \hline
            \multirow{3}{*}{Llama-3.2-1B} & T\"ulu-v2 & 29.21 & 32.22 & +10.31\% \\
            & AlpacaCleaned & 28.76 & 32.41  & +12.69\% \\
            & LIMA & 24.77 & 24.97 & +0.81\% \\ \hline
            \multirow{3}{*}{Gemma-2-2B} & T\"ulu-v2 & 49.05 & 50.74  & +3.45\% \\
            & AlpacaCleaned & 49.37 & 51.53 & +4.38\% \\
            & LIMA  & 38.21 & 39.63 & +3.72\% \\ \hline
            \multirow{3}{*}{Llama-3.2-3B} & T\"ulu-v2 & 45.42 & 46.05 & +1.39\% \\
            & AlpacaCleaned & 43.08 & 43.08 & 0.00\% \\
            & LIMA & 41.52 & 41.63 & +0.27\% \\ \hline
            \multirow{3}{*}{Mistral-7B} & T\"ulu-v2 & 57.99 & 61.68 & +6.36\% \\
            & AlpacaCleaned & 57.81 & 64.03 & +10.76\% \\
            & LIMA  & 46.55 & 59.2 & +27.18\% \\ \hline
            \multirow{3}{*}{Llama-3-8B} & T\"ulu-v2 & 56.91 & 58.01 & +2.03\% \\
            & AlpacaCleaned & 53.86 & 57.92  & +7.54\% \\
            & LIMA & 28.94 & 37.4 & +29.23\% \\  \hline
            & & & \begin{tabular}{c}\textbf{Average}\\\textbf{Relative Gain}\end{tabular} \textbf{=} & \changed{+8.01\%} \\\hline
        \end{tabular}
    }
    \caption{\changed{Relative percentage gain of DPO on top of \wit, for optimal $(\lambda_p, \lambda_r)$, over DPO on conventional instruction tuning.}}
    \label{tab:dpo_relative_gain}
\end{table}

Beyond achieving high performance on various evaluation benchmarks, a key desirable property of an instruction-tuned LM is its \textit{robustness} to prompt variations. Prior work has shown that these models are often sensitive to minor changes in prompts~\cite{arora2023ask, leidinger-etal-2023-language, voronov-etal-2024-mind, mizrahi-etal-2024-state, sclar2024quantifying}. To quantify this, \citet{chatterjee-etal-2024-posix} introduced the Prompt Sensitivity Index (POSIX), which measures a model's sensitivity to \textit{intent-preserving} prompt variations, such as spelling errors, re-wordings or prompt format changes. Figure~\ref{fig:posix_heatmap} reports POSIX values for our models, using intent-preserving variants of $5K$ randomly sampled AlpacaCleaned prompts as provided by ~\citet{chatterjee-etal-2024-posix}.

In line with our observations in the case of performance on evaluation benchmarks, it can be noted from Figure~\ref{fig:posix_heatmap} that the models finetuned using the conventional instruction tuning loss almost never are the best in terms of prompt sensitivity (except for $1$ out of $15$ combinations), and are often more sensitive than even the corresponding base model (e.g., Llama-3-8B across all datasets). Also, lower response-token weights consistently lead to reduced sensitivity to input changes. Taken together with benchmark performance (Figures~\ref{fig:main_heatmap} and \ref{fig:posix_heatmap}), these results suggest that a moderate response-token weight offers the best trade-off between robustness and performance, further highlighting the limitations of extreme response weighting.

\section{Discussions}\label{sec:discussion}
\changed{Building on above empirical results, we discuss broader patterns and preliminary insights that could inspire future studies on the interplay between task characteristics and token weighting.}

\subsection{Prompt-Token Weight: When and Why?}
\label{sec:prompt_weight_corr}
\changed{As shown in Figures~\ref{fig:tulu_all}, \ref{fig:alpaca-cleaned_all}, and \ref{fig:lima_all} in the Appendix, the optimal prompt-token weight varies with the combination of language model, training dataset, and the evaluation benchmark. To gain insights that may help us understand when and why a non-zero prompt-token weight is beneficial, we conduct a correlation analysis between various prompt characteristics (e.g., prompt length) and the optimal prompt-token weight, by varying one variable at a time.}

\begin{table}[t!]
    \small
    \centering
    \scalebox{0.8}{
    \begin{tabular}{ccc}
    \hline
        \begin{tabular}{c}\textbf{Finetuning}\\\textbf{Data}\end{tabular} & \begin{tabular}{c}\textbf{Average}\\\textbf{Optimal} $\lambda_p$\end{tabular} & \begin{tabular}{c}\textbf{Average}\\\textbf{Optimal} $\lambda_r$\end{tabular} \\
        \hline
        T\"ulu-v2 & 0.20 & 0.58 \\
        Alpaca-Cleaned & 0.36 & 0.49 \\
        LIMA & 0.35 & 0.6 \\    
        \hline
    \end{tabular}}
    \caption{Optimal prompt-token weight ($\lambda_p$) and response-token weight ($\lambda_r$) for various training datasets averaged across different $(model$, $evaluation\_benchmark)$ combinations. A relatively low prompt-token weight, along with a relatively moderate response-token weight, yields the best performance for all three training datasets.}
    \label{tab:finetuning_data_trends}
\end{table}

\paragraph{Role of Finetuning Data in Selection of Prompt-Token Weight.}
\changed{Table~\ref{tab:finetuning_data_trends} reports the optimal prompt-token weight ($\lambda_p$) and response-token weight ($\lambda_r$) for different finetuning datasets averaged across various $(model, evaluation\_benchmark)$ combinations. This helps us study how the optimal prompt-token weight varies with finetuning data. While the average optimal prompt-token weight for all finetuning datasets is in the low-to-moderate range, it is comparatively lower for T\"ulu-v2 compared to Alpaca-Cleaned or LIMA. To better understand the possible dataset characteristics contributing to these trends, we study the prompt characteristics in the finetuning datasets, such as the average prompt length and the average generation ratio (i.e., the ratio of response length and prompt length) to capture the length characteristics, $n$-gram diversity~\citep{meister2023locally} of prompts to capture lexical diversity, and the average depth of prompts' dependency parse tree to capture syntactic complexity.}

Table~\ref{tab:corr_all_characteristics} shows that the average generation ratio is positively correlated with the optimal prompt-token weight, while the average prompt length exhibits a negative correlation. This indicates that higher prompt-token weights tend to be preferred when the finetuning data contains longer completions relative to prompts, but not necessarily when the prompts themselves are longer.
Furthermore, both lexical diversity, as measured by $n$-gram diversity, and syntactic complexity of the prompts are observed to negatively influence the optimal prompt-token weight.

\begin{table*}[ht!]
\centering
\resizebox{\textwidth}{!}{
\begin{tabular}{l@{\hskip 1pt}c@{\hskip 1pt}c@{\hskip 1pt}c@{\hskip 1pt}c@{\hskip 1pt}@{\hskip 1pt}c@{\hskip 1pt}c@{\hskip 1pt}c@{\hskip 1pt}@{\hskip 1pt}c@{\hskip 1pt}c@{\hskip 1pt}c}
\toprule
\multirow{2}{*}{Correlation} 
& \multicolumn{4}{c}{\textbf{Train Prompt Characteristics}} 
& \multicolumn{3}{c}{\textbf{Eval Prompt Characteristics}}
& \multicolumn{3}{c}{\textbf{Model Characteristics}} \\
\cmidrule(lr){2-5} \cmidrule(lr){6-8} \cmidrule(lr){9-11}
& \begin{tabular}{c}Avg. \\ Gen. \\Ratio\end{tabular} 
& \begin{tabular}{c}N-gram \\ Div. \end{tabular} 
& \begin{tabular}{c}Avg. Parse \\Tree Depth\end{tabular} 
& \begin{tabular}{c}Avg. \\ Prompt \\Len.\end{tabular} 
& \begin{tabular}{c}N-gram \\ Div.\end{tabular} 
& \begin{tabular}{c}Avg. Parse \\Tree Depth\end{tabular} 
& \begin{tabular}{c}Avg. \\ Prompt \\Len.\end{tabular} 
& \begin{tabular}{c}Model \\ Size\end{tabular} 
& \begin{tabular}{c}Avg. log-prob\\ of train prompt \\tokens\end{tabular} 
& \begin{tabular}{c}Avg. log-prob\\ of eval prompt \\tokens\end{tabular} \\
\midrule
Spearman 
& 0.50 & -0.50 & -0.50 & -0.50 
& 0.40 & -0.50 & -0.70 
& 0.20 & 0.50 & 0.50 \\
Kendall's $\tau$ 
& 0.33 & -0.33 & -0.33 & -0.33 
& 0.40 & -0.20 & -0.60 
& 0.20 & 0.20 & 0.20 \\
\bottomrule
\end{tabular}
}
\caption{\changed{Correlation coefficients (Spearman and Kendall's $\tau$) between the optimal prompt-token weight ($\lambda_p$) and various characteristics of the finetuning datasets, evaluation benchmarks and language models.}}
\label{tab:corr_all_characteristics}
\end{table*} 

\paragraph{Role of Evaluation Benchmark in Selection of Prompt-Token Weight.}
The optimal prompt- and response-token weights for different evaluation benchmarks averaged across various $(model, training\_dataset)$ combinations are presented in Table~\ref{tab:benchmark_trends}. This helps us study how the optimal prompt-token weight varies with evaluation benchmarks. We observe that the optimal prompt-token weight varies from low to moderate, ranging from $0.17$ for BBH to $0.48$ for IFEval. To investigate the possible underlying benchmark characteristics contributing towards the observed optimal prompt-token weights, we obtain prompt characteristics of evaluation benchmarks (similar to those extracted in the case of finetuning data) whose correlation with the optimal prompt-token weight is presented in Table~\ref{tab:corr_all_characteristics}. As with finetuning data, a lower prompt-token weight yields better performance on benchmarks with longer prompts; syntactic complexity of the prompts also has a negative correlation with optimal prompt-token weight. However, unlike with training data, we observe that the lexical diversity of evaluation benchmarks is positively correlated with the optimal prompt-token weight.

\begin{table}[t!]
    \small
    \centering
    \scalebox{0.8}{
    \begin{tabular}{ccc}
    \hline
        \begin{tabular}{c}\textbf{Language}\\\textbf{Model}\end{tabular} & \begin{tabular}{c}\textbf{Average}\\\textbf{Optimal} $\lambda_p$\end{tabular} & \begin{tabular}{c}\textbf{Average}\\\textbf{Optimal} $\lambda_r$\end{tabular} \\
        \hline
        Llama-3-1B & 0.33 & 0.63 \\
        Gemma-2-2B & 0.42 & 0.57 \\
        Llama-3-3B & 0.20 & 0.57 \\    
        Mistral-7B & 0.32 & 0.53 \\    
        Llama-3-8B & 0.35 & 0.50 \\    
        \hline
    \end{tabular}}
    \caption{Optimal prompt-token weight ($\lambda_p$) and response-token weight ($\lambda_r$) for various language models averaged across different $(training\_dataset, evaluation\_benchmark)$ combinations. A relatively lower prompt-token weight, coupled with a comparatively moderate response-token weight, yields the best performance for all five models.}
    \label{tab:language_model_trends}
\end{table}

\paragraph{Role of Language Model in Selection of Prompt-Token Weight.}
\changed{
To study how the optimal prompt-token weight varies across language models, Table~\ref{tab:language_model_trends} reports the optimal prompt-token weight ($\lambda_p$) and response-token weight ($\lambda_r$) for different language models, averaged across various $(training\_dataset, evaluation\_benchmark)$ combinations. We observe that the optimal prompt-token weight varies from low to moderate, ranging from $0.20$ for Llama-3-3B to $0.42$ for Gemma-2-2B. To better understand the potential factors contributing to these variations, we obtain model-dependent characteristics of train datasets and evaluation benchmarks, such as the average next-token log probabilities of prompts from finetuning datasets and evaluation benchmarks. The average next-token log probability is observed to be positively correlated with prompt-token weight (c.f. Table~\ref{tab:corr_all_characteristics}), suggesting that if a model has higher perplexity on prompts of a certain dataset, then a lower prompt-token weight can be more suitable. Furthermore, model size has a weak positive correlation with optimal $\lambda_p$.}

\paragraph{In Summary.} 
\changed{It is important to note that, as observed in our analysis, multiple factors influence the optimal prompt-token weight, often in different directions. Thus, considering the combined effect of these characteristics should be more effective than focusing on any single property when selecting prompt-token weights for \wit.}

\begin{figure*}[ht!]
    \centering
    \renewcommand{\arraystretch}{-1pt}  
    \begin{tblr}{@{}p{1.2em}@{} c c c}  
        \rotatebox{90}{\parbox{3cm}{\centering {~~~~~~Base Models}}} &  
        \begin{subfigure}[b]{0.3\textwidth}
            \centering
            \caption*{{Gemma-2-2B}}
            \includegraphics[width=\textwidth]{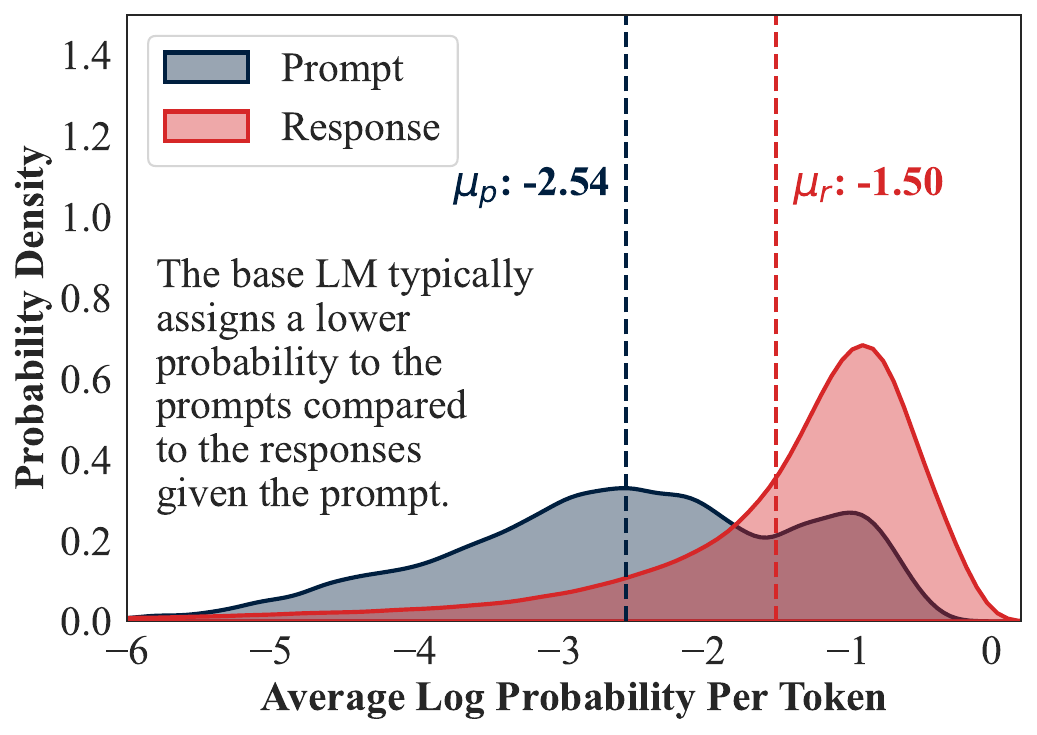}
        \end{subfigure} &
        \begin{subfigure}[b]{0.3\textwidth}
            \centering
            \caption*{{Mistral-7B}}
            \includegraphics[width=\textwidth]{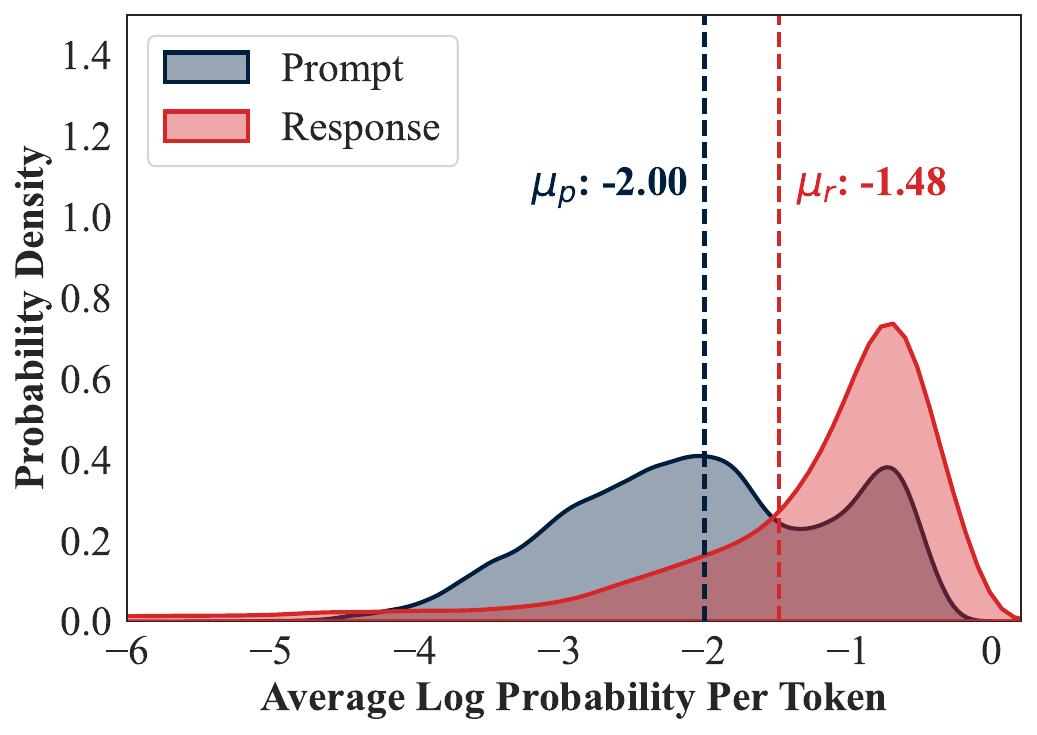}
        \end{subfigure} &
        \begin{subfigure}[b]{0.3\textwidth}
            \centering
            \caption*{{Llama-3-8B}}
            \includegraphics[width=\textwidth]{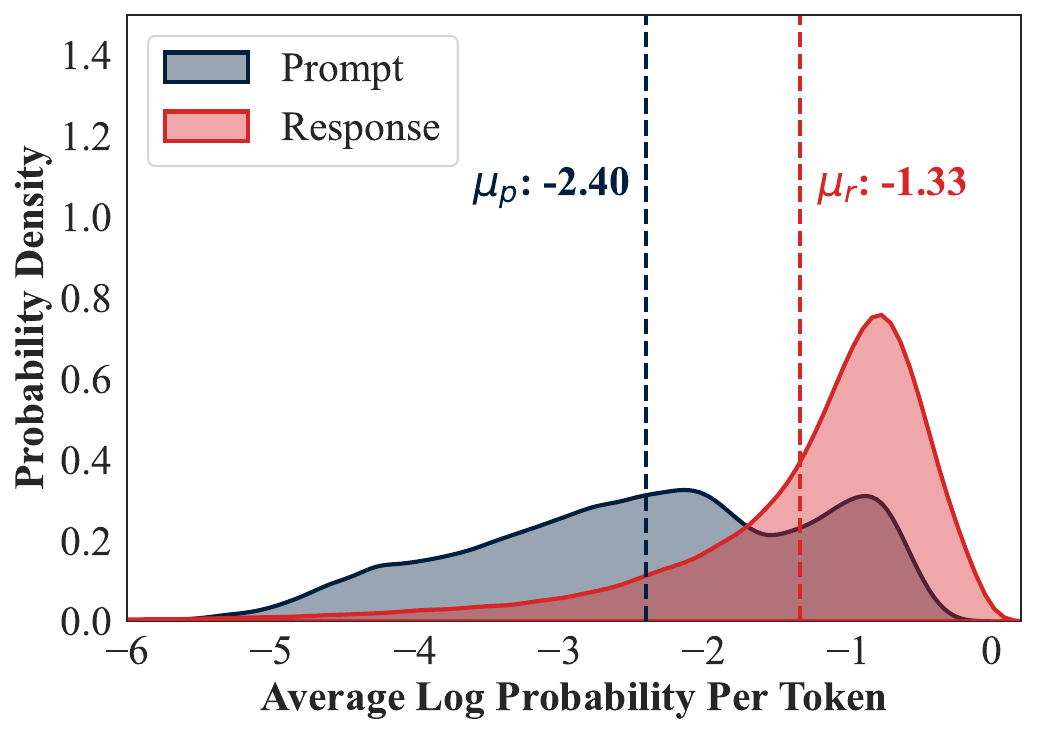}
        \end{subfigure} \\
        
        \rotatebox{90}{\parbox{3cm}{\centering {~Instruction Tuning}}} &  
        \begin{subfigure}[b]{0.3\textwidth}
            \centering
            \includegraphics[width=\textwidth]{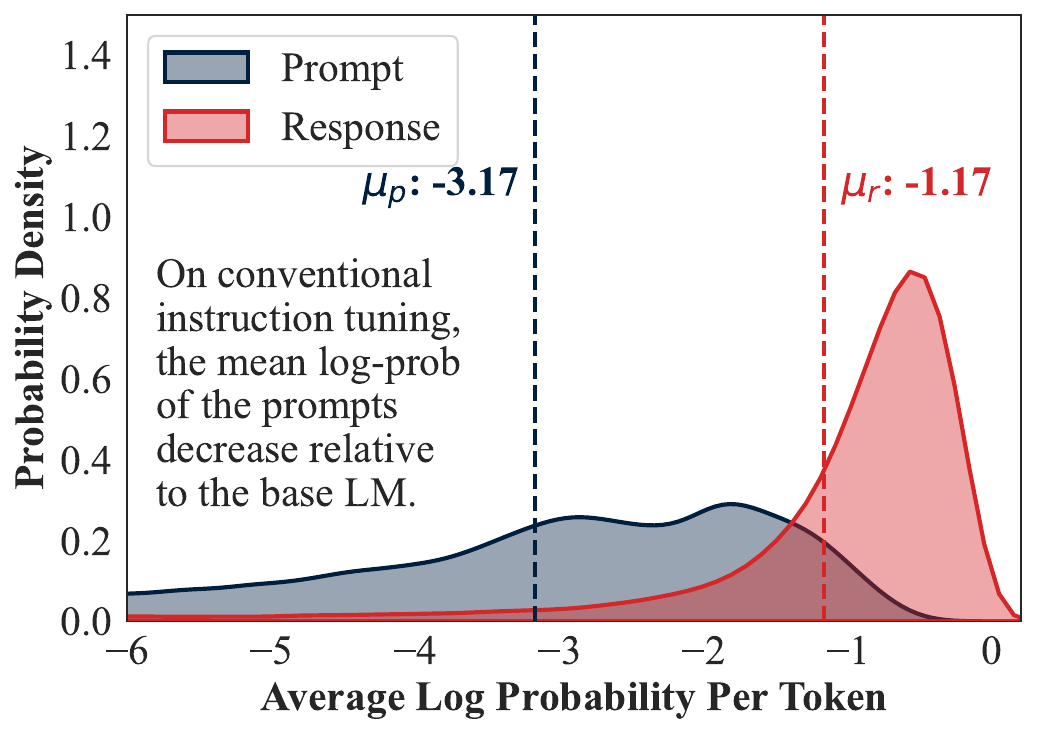}
        \end{subfigure} &
        \begin{subfigure}[b]{0.3\textwidth}
            \centering
            \includegraphics[width=\textwidth]{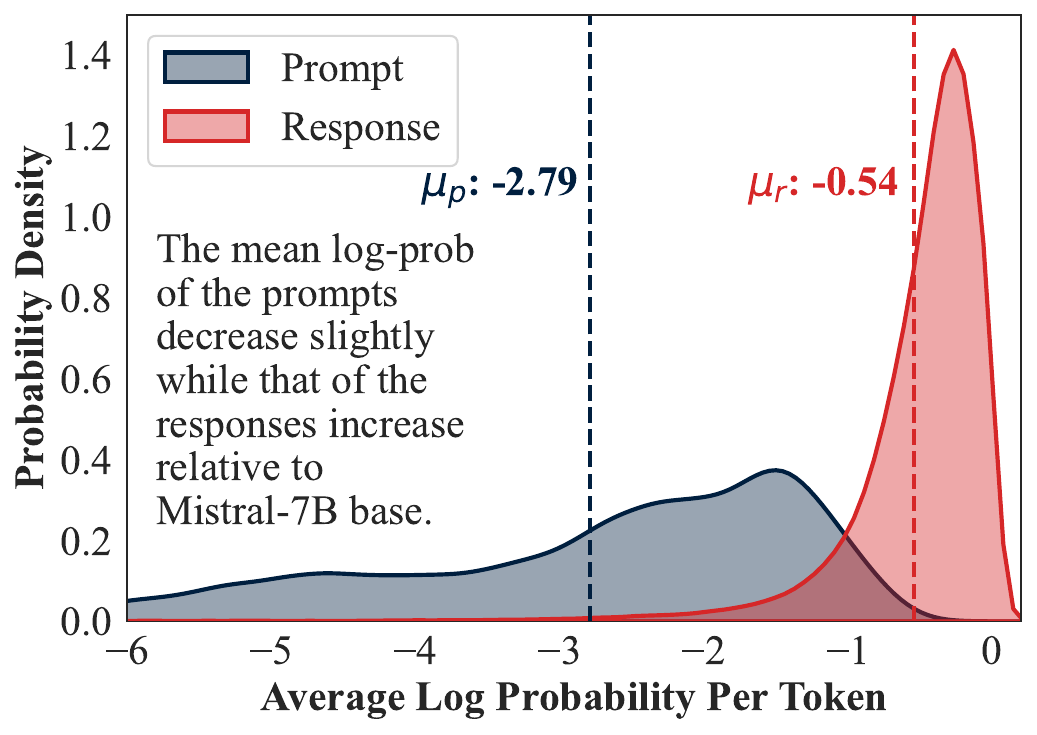}
        \end{subfigure} &
        \begin{subfigure}[b]{0.3\textwidth}
            \centering
            \includegraphics[width=\textwidth]{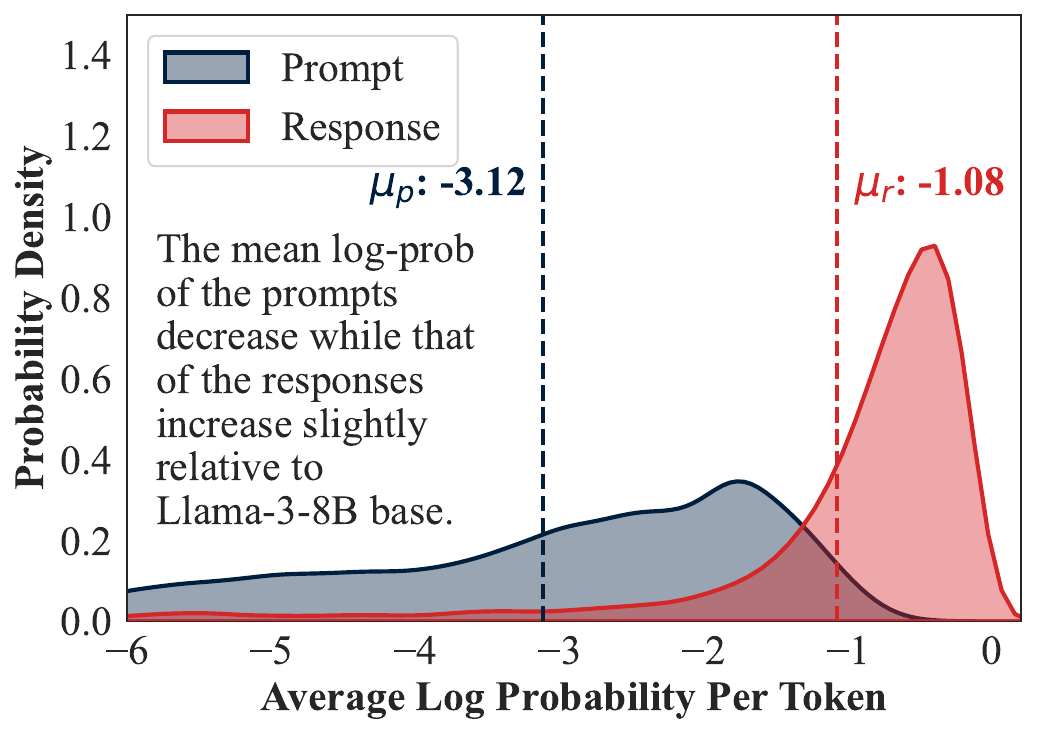}
        \end{subfigure} \\

        \rotatebox{90}{\parbox{3cm}{\centering {~~~~~~$\wit_{best}$}}} &  
        \begin{subfigure}[b]{0.3\textwidth}
            \centering
            \includegraphics[width=\textwidth]{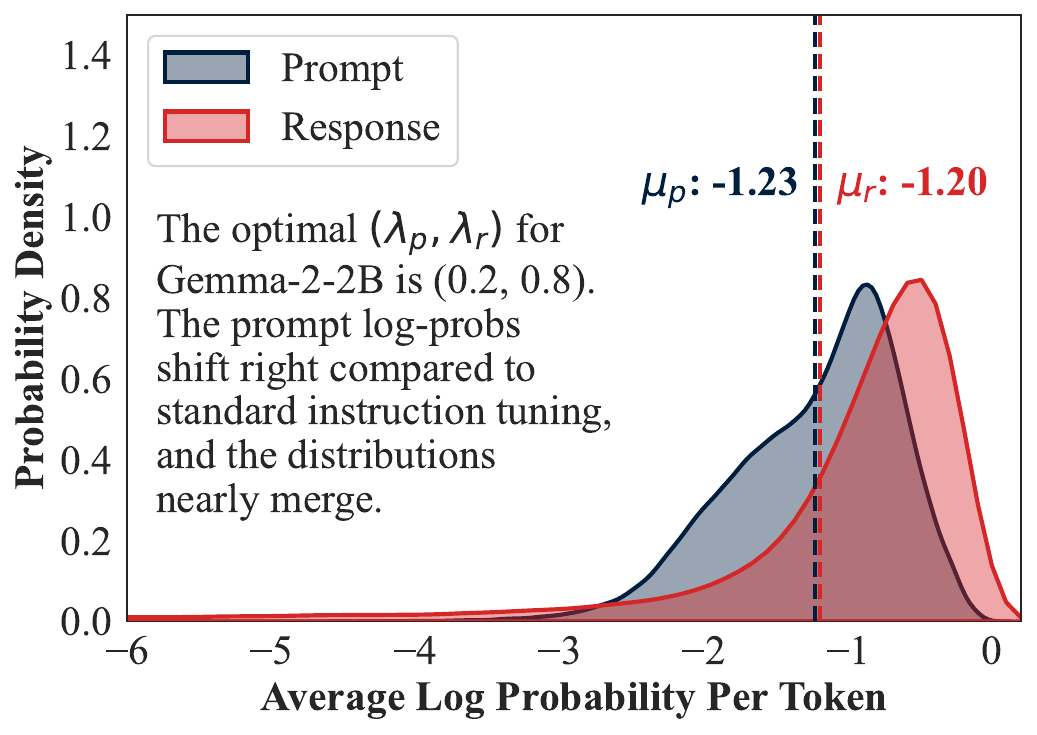}
        \end{subfigure} &
        \begin{subfigure}[b]{0.3\textwidth}
            \centering
            \includegraphics[width=\textwidth]{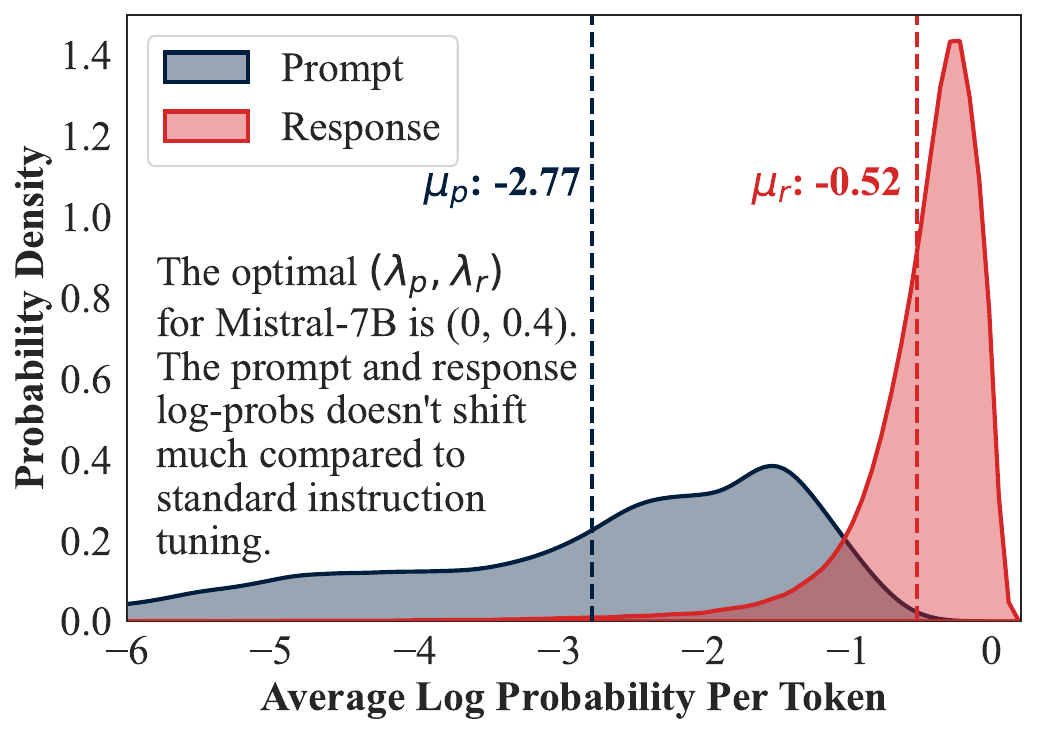}
        \end{subfigure} &
        \begin{subfigure}[b]{0.3\textwidth}
            \centering
            \includegraphics[width=\textwidth]{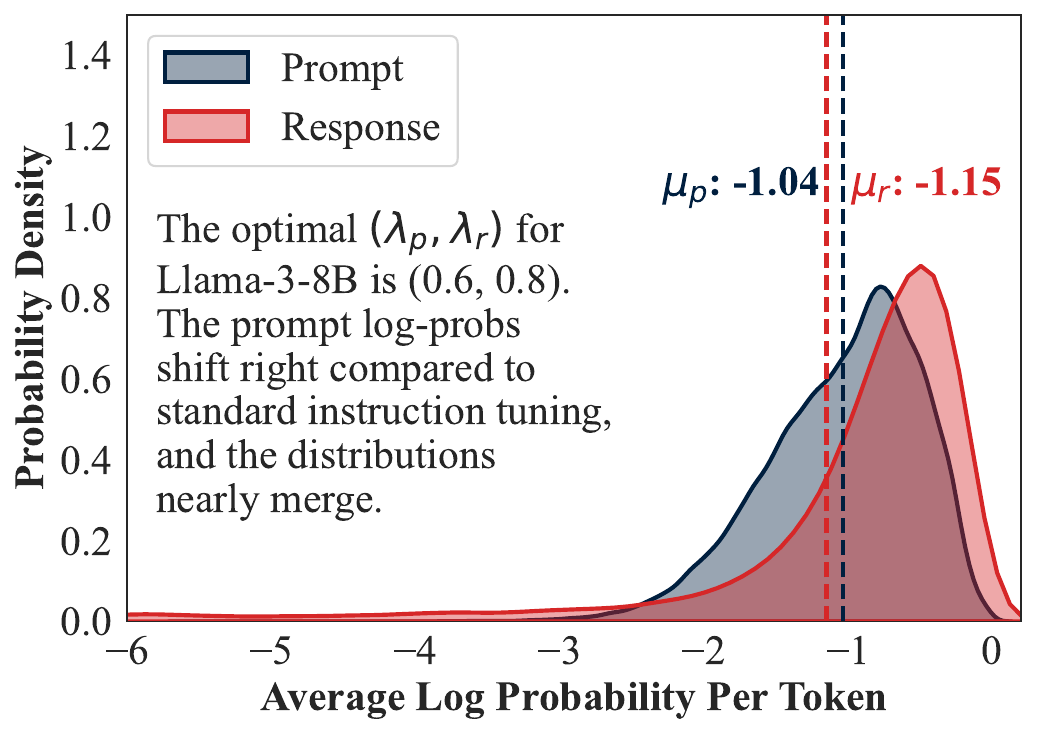}
        \end{subfigure} \\
    \end{tblr}
    
    \caption{Distribution of average log probabilities for prompts and responses (given the corresponding prompts) from training samples of T\"ulu-v2, comparing base models with their instruction-tuned counterparts trained using the conventional response-only loss and the \wit~loss (with optimal token weights).}
    \label{fig:avg_log_probs}
\end{figure*}

\subsection{Impact of Instruction Tuning on Prompt and Response Probabilities}

To assess how instruction tuning alters model behavior, we analyze the shifts in the log-probability distributions for prompt and response tokens. For this, we compute the length-normalized average log-probabilities for the training instances in T\"ulu-v2 across the base and two instruction-tuned variants of Gemma-2-2B, Mistral-7B, and Llama-3-8B (see Figure~\ref{fig:avg_log_probs}). The 1B and 3B variants of Llama-3 exhibit similar trends as the 8B model and are omitted for brevity. For each instance, we compute the \textit{average log-probability per token} for (i) the prompt, and (ii) the response given the prompt, enabling fair comparison across different sequence lengths.

\paragraph{Behaviour of the Base LMs.} 
Across all model families, we observe that base LMs assign lower probabilities to prompts in isolation compared to responses conditioned on prompts, as evidenced by a leftward shift in prompt probability distributions relative to responses (first row in Figure~\ref{fig:avg_log_probs}). This aligns with expectations, as models pretrained on naturally occurring text develop a stronger prior over plausible completions than over standalone queries.

\paragraph{Effect of Conventional Instruction Tuning.}
When models are instruction-tuned using the conventional response-only loss, we observe that while the probability distribution of responses remains largely unchanged compared to the base LM (except in the case of Mistral), the probability assigned to prompt tokens shifts further left, indicating a decrease in their likelihood (middle row in Figure~\ref{fig:avg_log_probs}). This reveals an interesting insight on the effect of response-only loss --- while the probability of the correct response given the prompt remains almost unchanged, the likelihood of the prompt itself decreases. Thus, the conventional instruction tuning loss, though it doesn't explicitly consider the prompt tokens, negatively affects the prediction of the input prompt tokens. We hypothesize that this degradation in prompt modelling might hurt the instruction comprehension ability of the models, potentially leading to a drop in performance on instruction-following benchmarks like IFEval, as observed in Figure~\ref{fig:tulu_all}.

\paragraph{Effect of~\wit.}

When trained with the \wit~ loss using optimal prompt-response weights, the prompt probability distribution shifts rightward and aligns closely with that of the responses, especially for Llama and Gemma models (bottom row in Figure~\ref{fig:avg_log_probs}). For Mistral, however, this shift is negligible as the optimal \wit~setting involves a null prompt weight. 
These observations indicate that \wit~encourages the model to assign relatively higher likelihood to prompts, while the average log-likelihoods of responses remain similar or, in some cases, even decrease relative to conventional instruction tuning, likely improving instruction comprehension and mitigating overfitting on response patterns. This balanced treatment of prompts and responses contributes to better generalization across downstream tasks as well as enhanced robustness, as demonstrated in Figures~\ref{fig:main_heatmap} and~\ref{fig:posix_heatmap}. 

\section{Related Work}
We review the prior work on instruction tuning across three main dimensions -- instruction tuning algorithms, finetuning data, and evaluation. 

\paragraph{Instruction Tuning Algorithms.} 
Conventional instruction tuning uses an auto-regressive objective with loss zeroed on prompt tokens -- a practice that, as recent work suggests, can encourage overfitting to response patterns~\citep{neftune, shi2025instruction}. To mitigate this, \citet{neftune} proposed \textit{NEFTune}, which adds noise to input embeddings to improve response quality, but offers no gains on OpenLLM benchmarks. Another approach, introduced by \citet{shi2025instruction} as \textit{Instruction Modelling}, is akin to continual pre-training and applies loss to both prompt and response tokens; this benefits low-resource settings but underperforms on OpenLLM benchmarks. Assigning a small weight to prompt-token loss has also shown promise for datasets with short responses~\citep{huerta-enochian-ko-2024-instruction}, though its effectiveness has primarily been validated on Alpaca variants. Other works leverage large proprietary models for phased training or fine-tuning on GPT-4-generated completions~\citep{phased-it, xie2024non}. Recent findings even suggest that instruction-following can emerge from response-only training~\citep{hewitt2024instruction, an2024response}, though this requires further validation.

\paragraph{Instruction Tuning Data.} 

The effectiveness of instruction tuning has been found to heavily depend upon task composition~\citep{wang2023far,dong-etal-2024-abilities,renduchintala-etal-2024-smart}, data quality~\citep{zhou2023lima,ding-etal-2023-enhancing} and data quantity~\citep{ji2023exploring,yuan2023scaling}. Notable instruction tuning datasets include FLAN \cite{wei2022finetuned}, Super-Natural Instructions \cite{wang-etal-2022-super}, Alpaca \cite{taori2023stanford}, 
T\"ulu \cite{ivison2023camels}, Dolly \cite{DatabricksBlog2023DollyV2}, and LIMA \cite{zhou2023lima} to name a few. For a more comprehensive review of data management for instruction tuning, we refer the reader to the survey by ~\citet{wang2023data}.

\paragraph{Evaluation of Instruction Tuned Models.}
Evaluation of instruction tuned models can be broadly classified into two categories: close-ended and open-ended evaluations. Close-ended evaluations offer more objective evaluations -- these include multiple-choice questions (MCQs) based benchmarks like MMLU~\citep{hendrycks2021measuring}, BBH-Hard~\citep{suzgun-etal-2023-challenging} as well as benchmarks like IFEval~\citep{zhou2023instruction} which contain verifiable prompts, that can be evaluated using a program for instance. Open-ended evaluations, on the other hand, attempt to assess the quality of the output. A most common method is to use LLM-as-a-judge, where an LLM like GPT-4 is used to perform comparisons of responses to assess their quality. AlpacaEval~\citep{alpaca_eval} is one such approach. For a comprehensive review of evaluation methods, we refer the reader to the survey by ~\citet{zhang2023instruction}.

\section{Conclusions}
We proposed \wit~as an alternative to conventional instruction tuning and analyzed the effects of differentially weighting prompt and response token losses. Our experiments on various models, datasets, and benchmarks show that both conventional instruction tuning and continual pre-training are generally suboptimal. While prior work~\citep{wei2022finetuned, ivison2023camels, zhou2023lima, shi2025instruction, huerta-enochian-ko-2024-instruction} consistently assigns maximal weight to response tokens, our results highlight the advantages of reducing response-token loss and including prompt-token loss. This overlooked balance offers new directions for robust instruction tuning. We also observe that the gains with \wit~transfer even to the preference alignment phase. 
Moreover, we find that finetuning solely on prompts -- though not always optimal -- can still impart instruction following ability, highlighting potential for instruction tuning without response annotations.

Beyond performance, our findings suggest that instruction tuning loss functions influence model robustness and may shape biases. This highlights loss function design as a potential tool for aligning LMs with ethical and safety objectives, mitigating adversarial vulnerabilities, and improving reliability in real-world applications.

\section*{Limitations}
One limitation of our approach is the use of fixed weights, i.e., one for all prompt tokens and another for all response tokens, throughout training. However, our preliminary analysis shows that optimal weights likely depend on factors like prompt and response likelihood from the lens of the model, which evolve during training. Moreover, no universal values of optimal prompt and response token weights exist across models or datasets. 
Future work exploring adaptive loss weighting strategies that dynamically adjust based on model predictions or training dynamics may be key to developing more robust and generalizable models.

\section*{Acknowledgements}
We thank the reviewers and the action editor for their time and effort in reviewing our manuscript and for their thoughtful feedback, which considerably improved the clarity and overall presentation of our work. Anwoy Chatterjee gratefully acknowledges the support of the Google PhD Fellowship. Tanmoy Chakraborty acknowledges the Rajiv Khemani Young Faculty
Chair Professorship in AI.

\bibliography{tacl2021}
\bibliographystyle{acl_natbib}

\appendix

\begin{figure*}[ht!]
  \centering
  \setlength{\tabcolsep}{2pt}      
  \renewcommand{\arraystretch}{1}  

  \begin{tabular}{@{} >{\centering\arraybackslash}m{2.1mm} *{5}{c} @{}}
      & Llama-3-1B & Llama-3-3B & Llama-3-8B & Mistral-7B & Gemma-2-2B \\[2pt]

      \raisebox{1.3cm}[0pt][0pt]{\rotatebox{90}{\makebox[0pt][c]{MMLU}}} &
      \includegraphics[width=.19\textwidth]{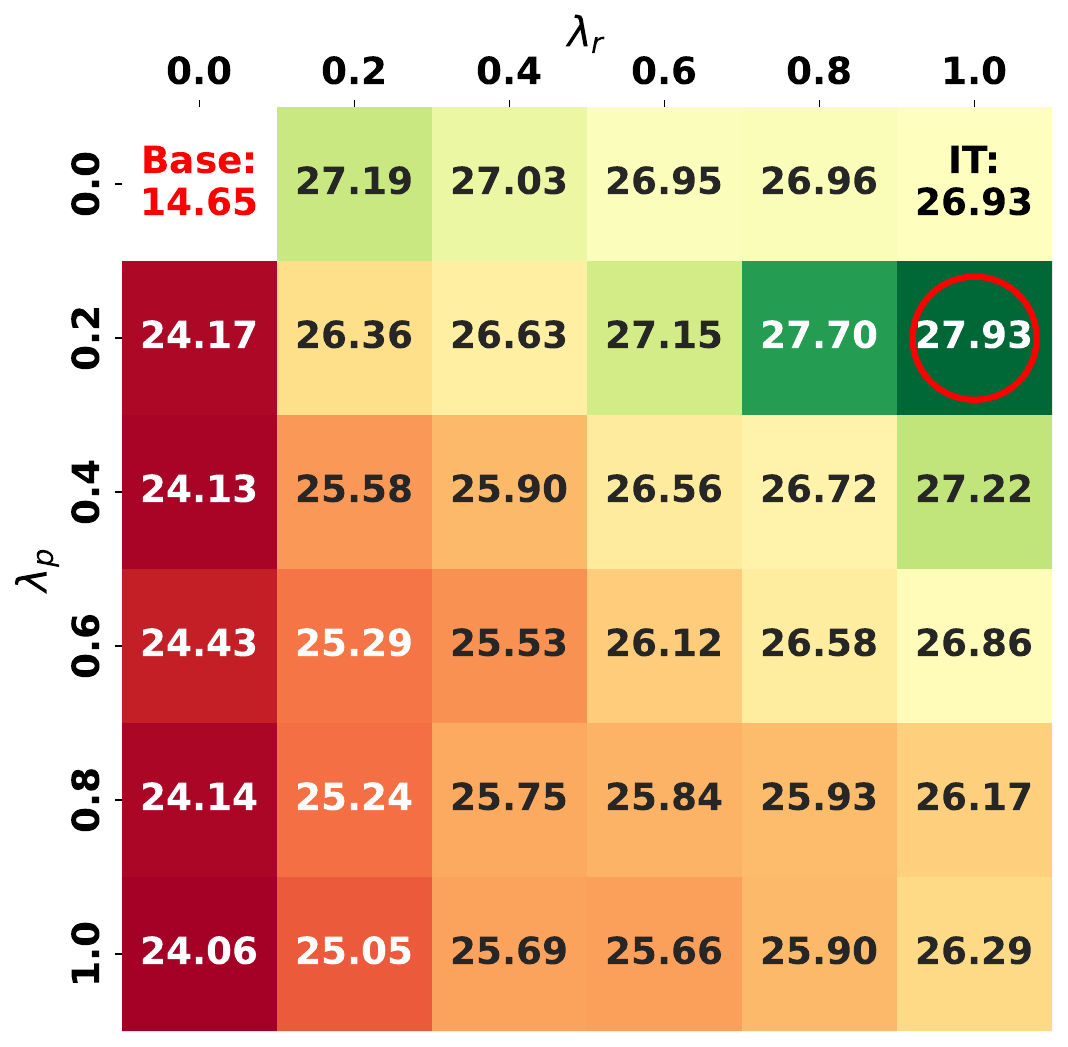} &
      \includegraphics[width=.19\textwidth]{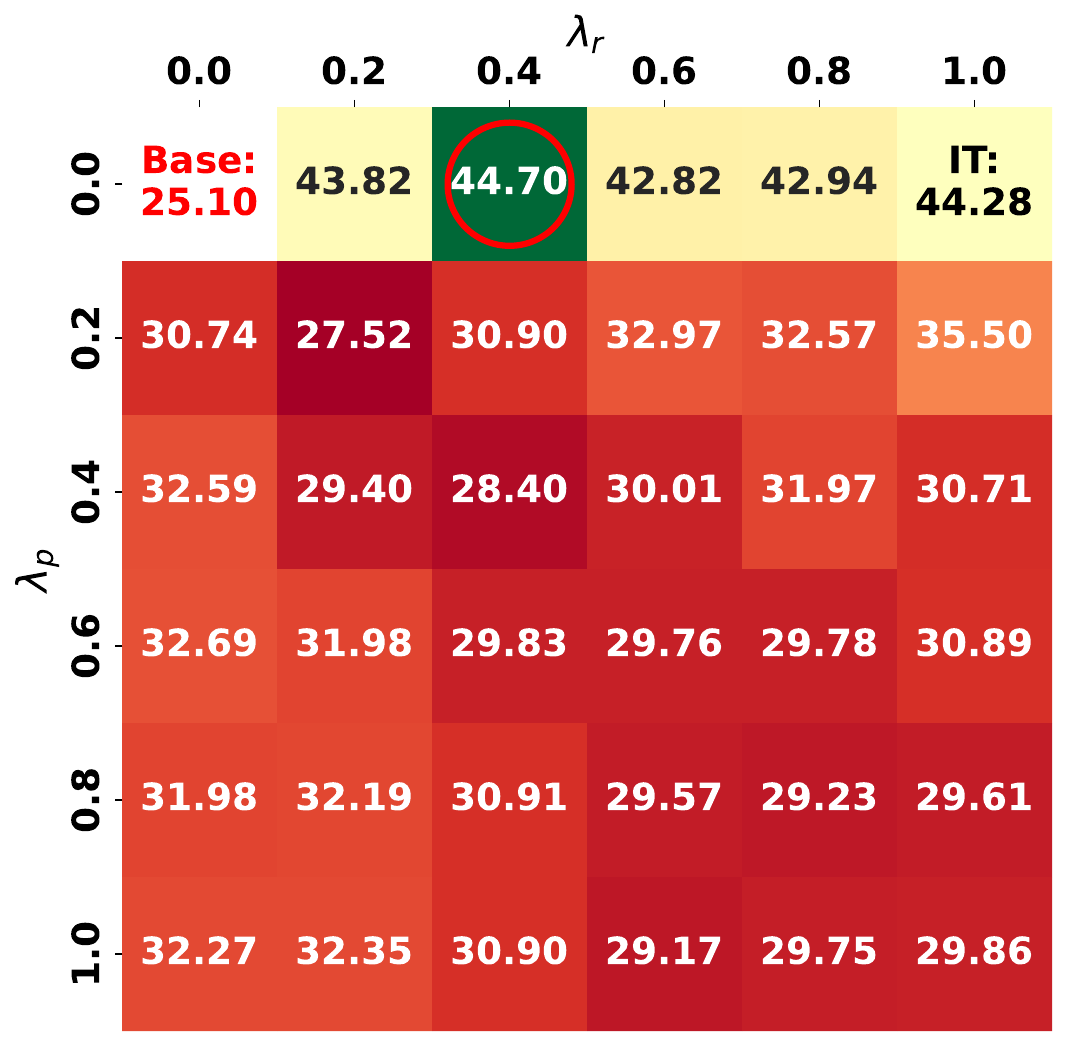} &
      \includegraphics[width=.19\textwidth]{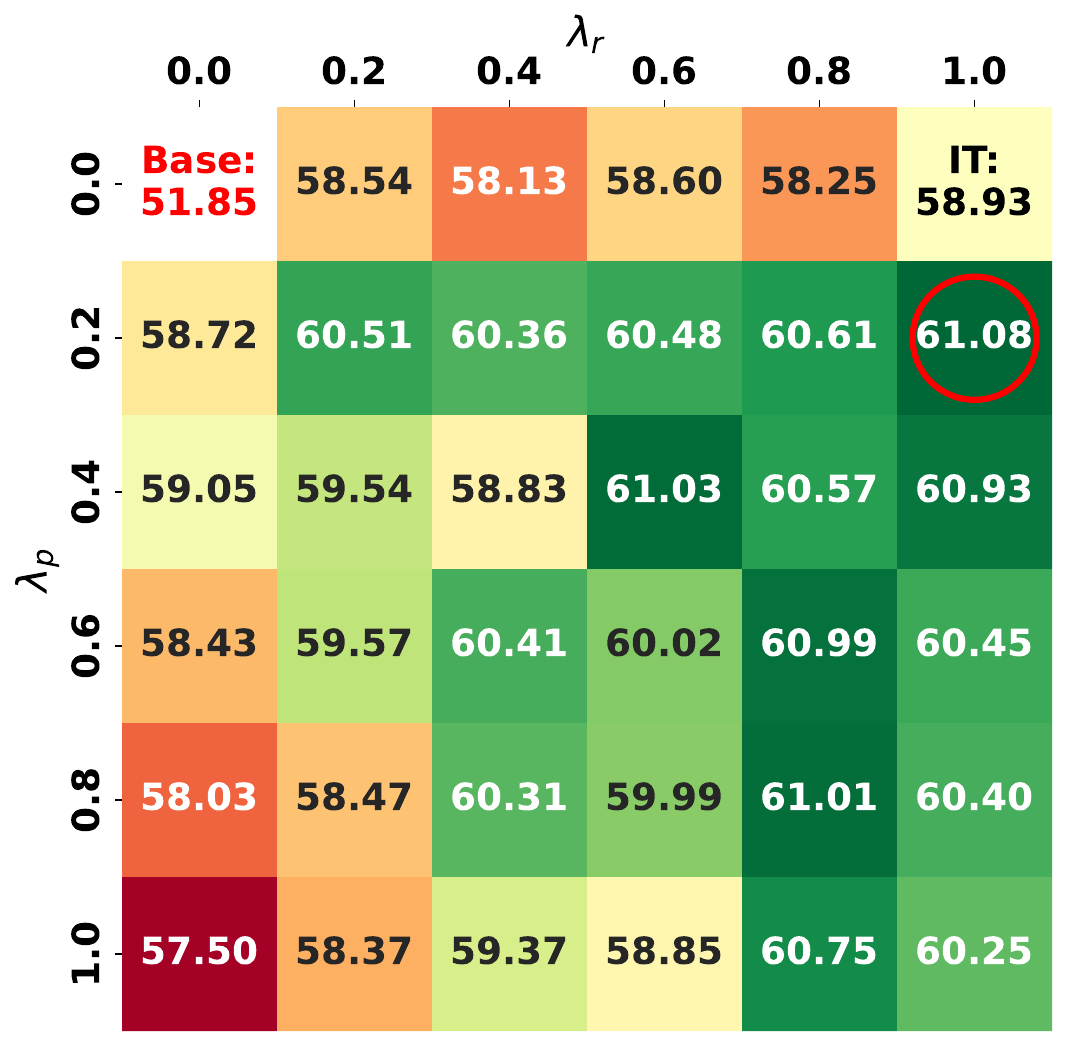} &
      \includegraphics[width=.19\textwidth]{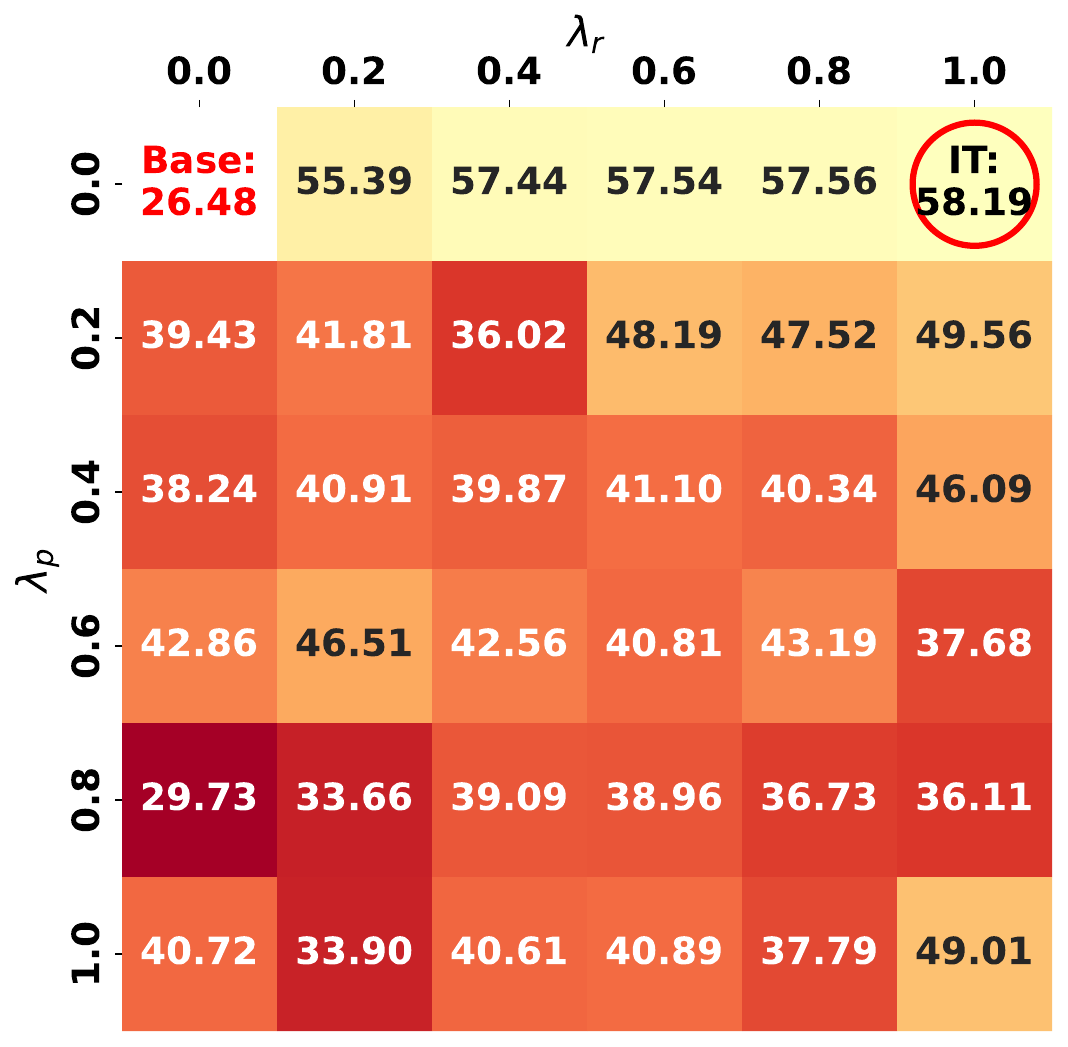} &
      \includegraphics[width=.19\textwidth]{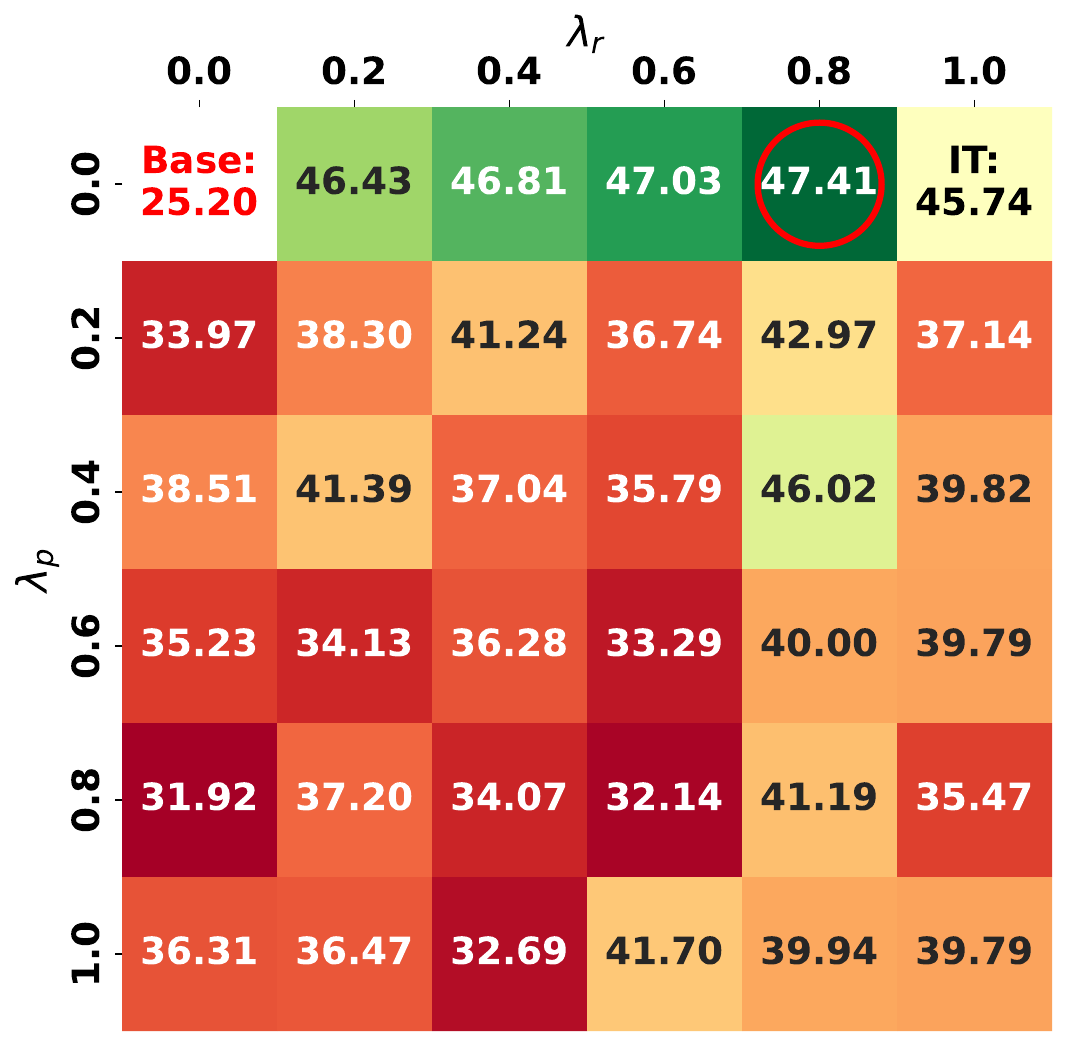} \\[4pt]

        \raisebox{1.3cm}[0pt][0pt]{\rotatebox{90}{\makebox[0pt][c]{BBH}}} &
      \includegraphics[width=.19\textwidth]{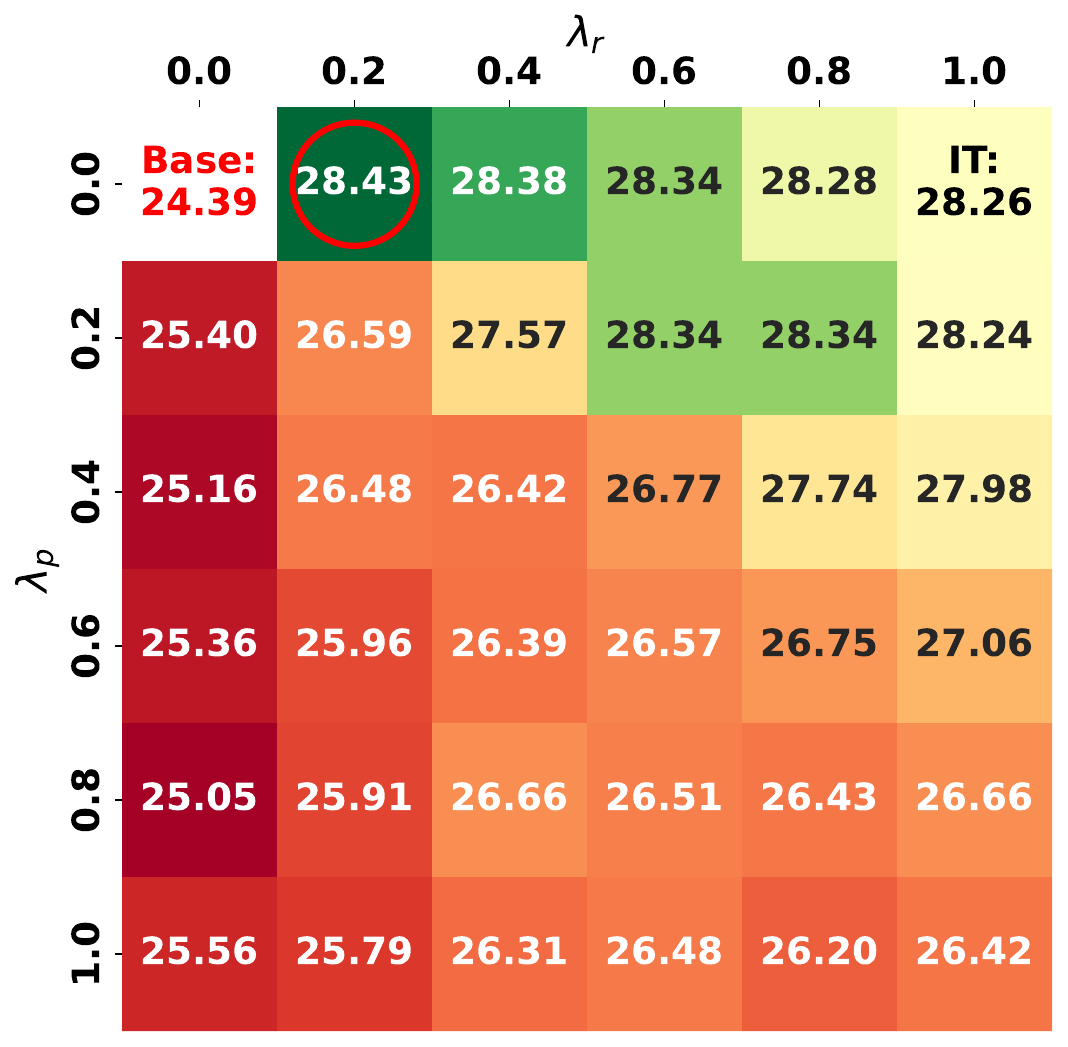} &
      \includegraphics[width=.19\textwidth]{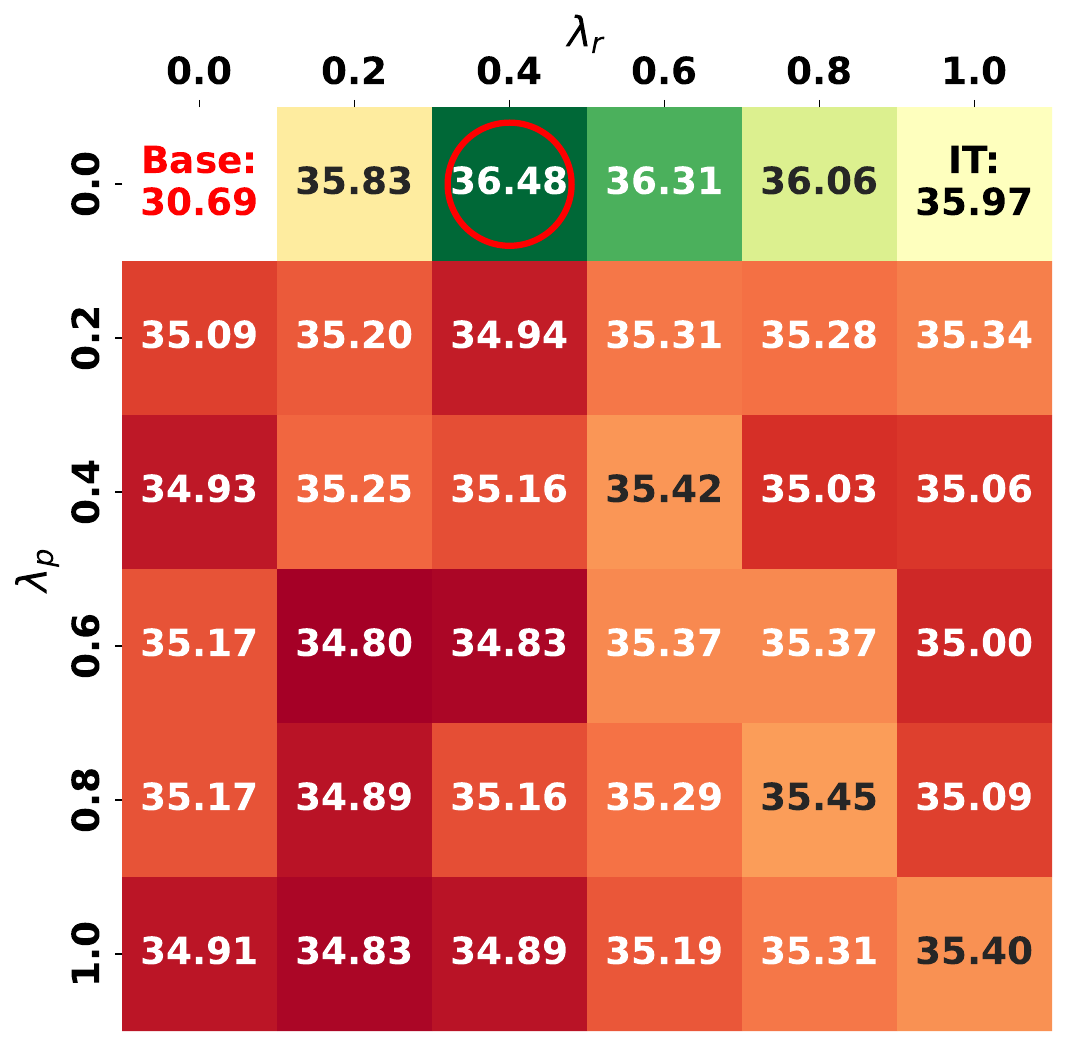} &
      \includegraphics[width=.19\textwidth]{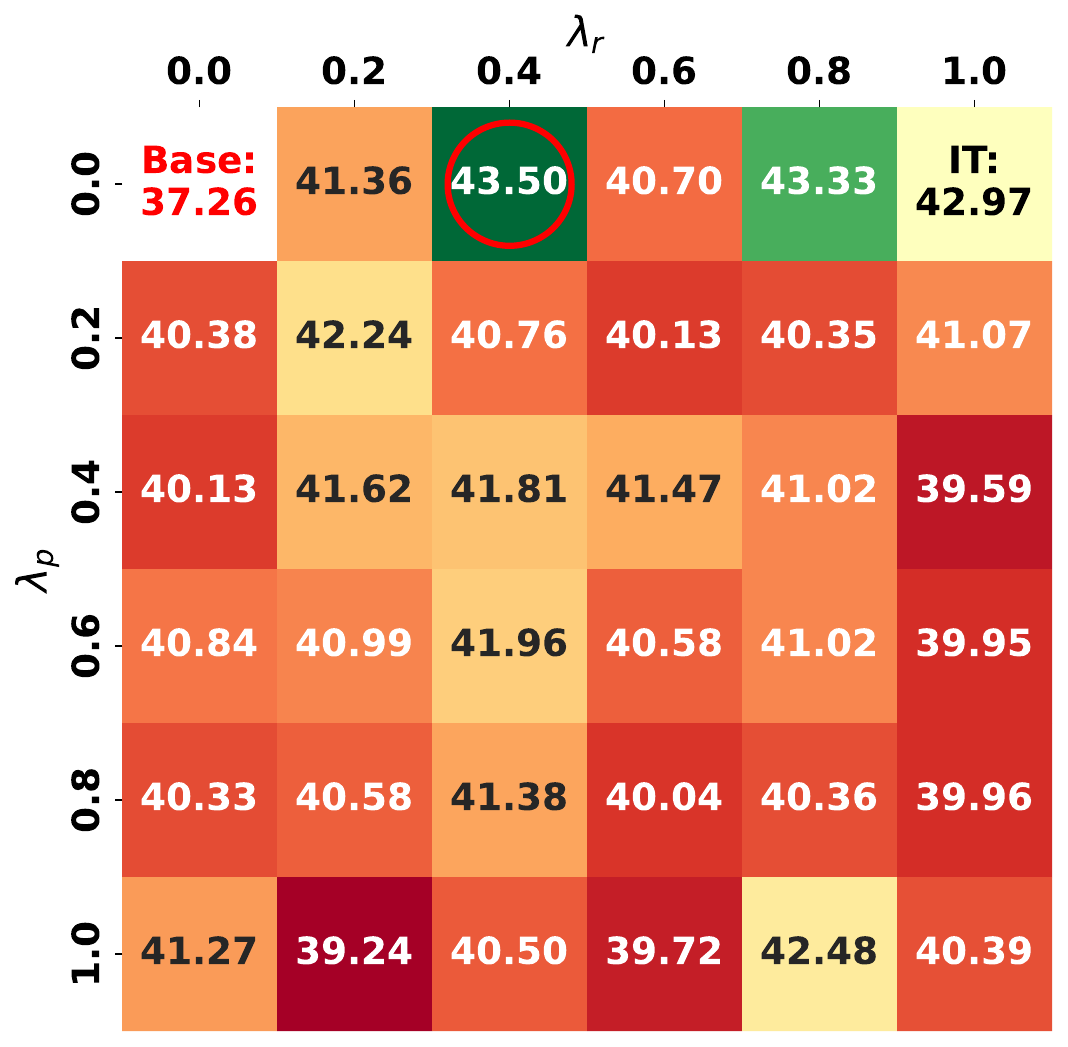} &
      \includegraphics[width=.19\textwidth]{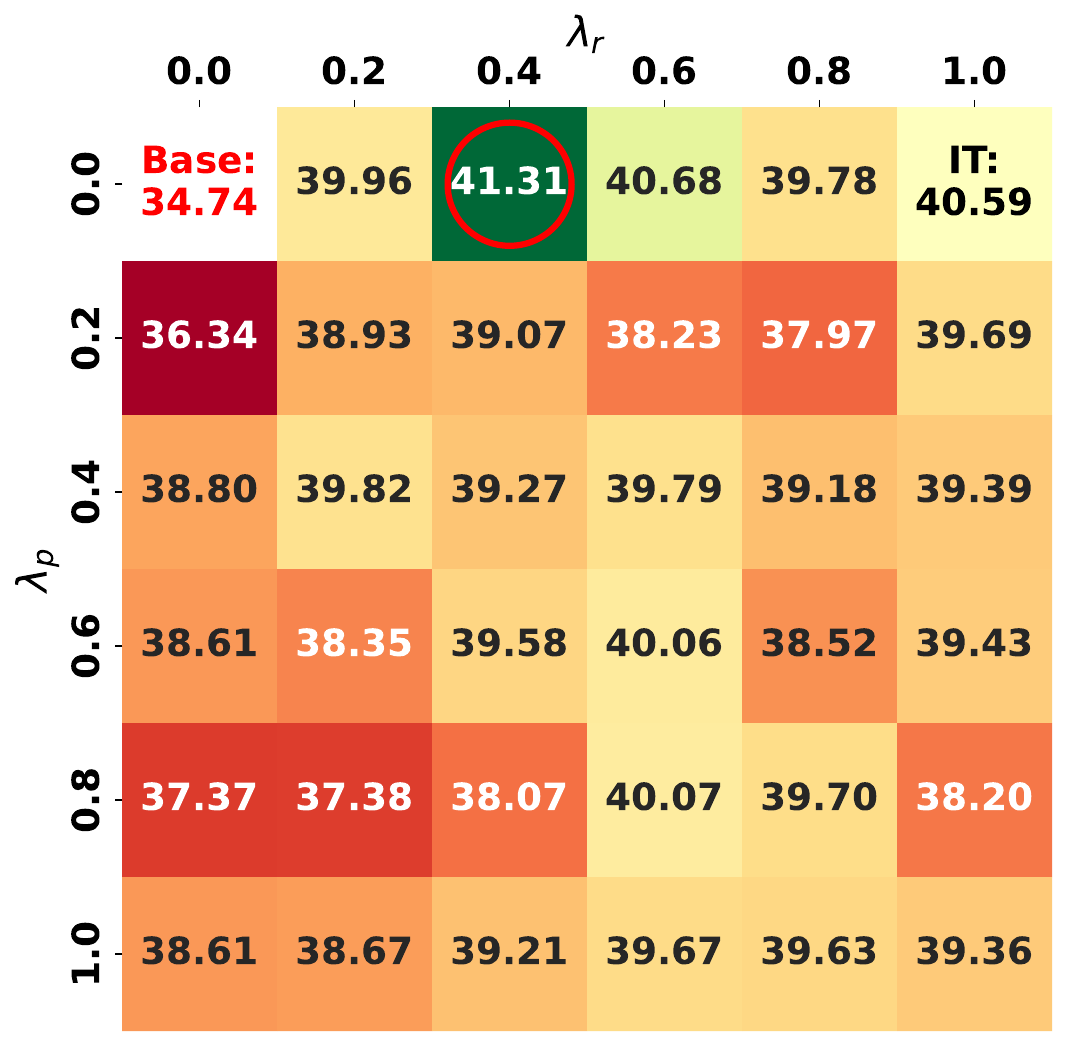} &
      \includegraphics[width=.19\textwidth]{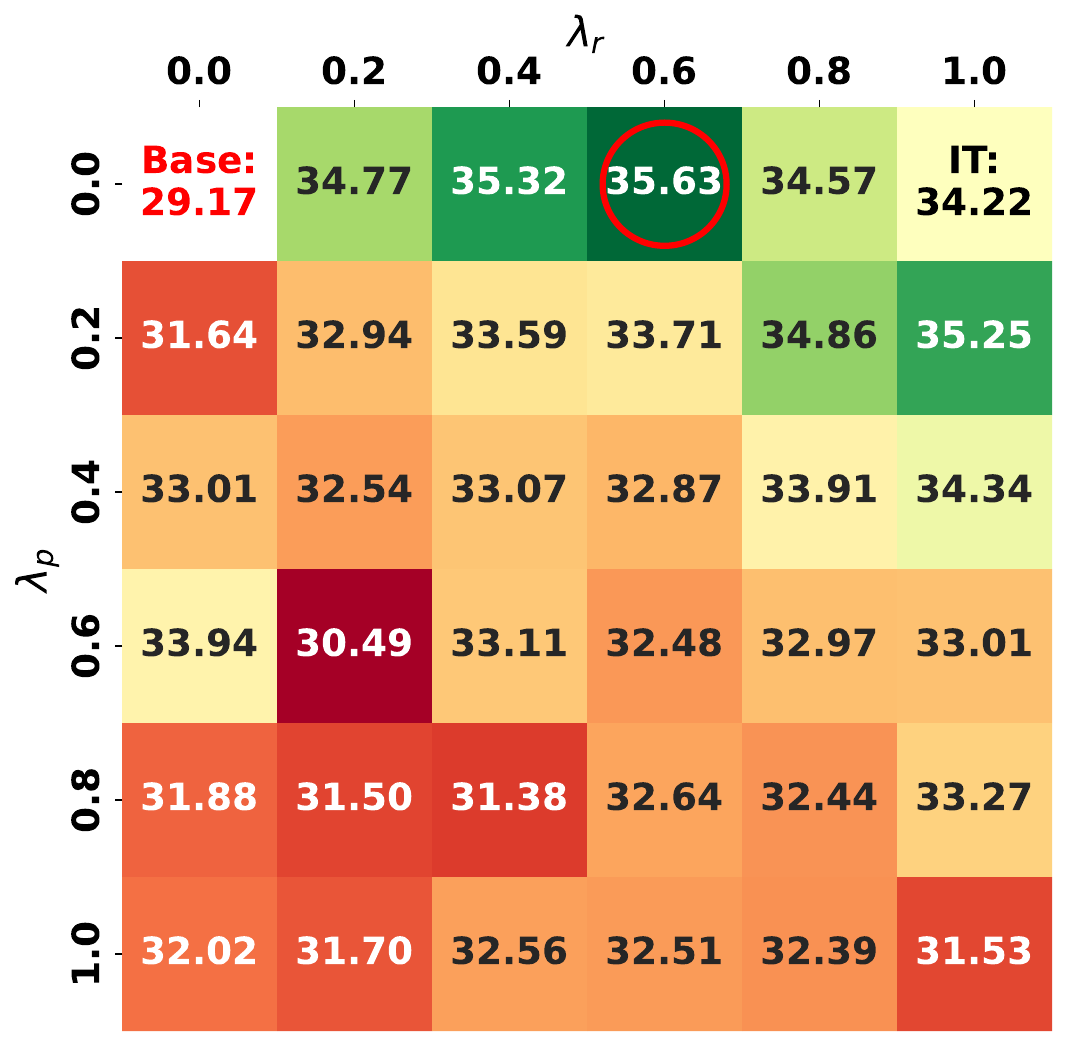} \\[4pt]

      \raisebox{1.3cm}[0pt][0pt]{\rotatebox{90}{\makebox[0pt][c]{AlpacaEval}}} &
      \includegraphics[width=.19\textwidth]{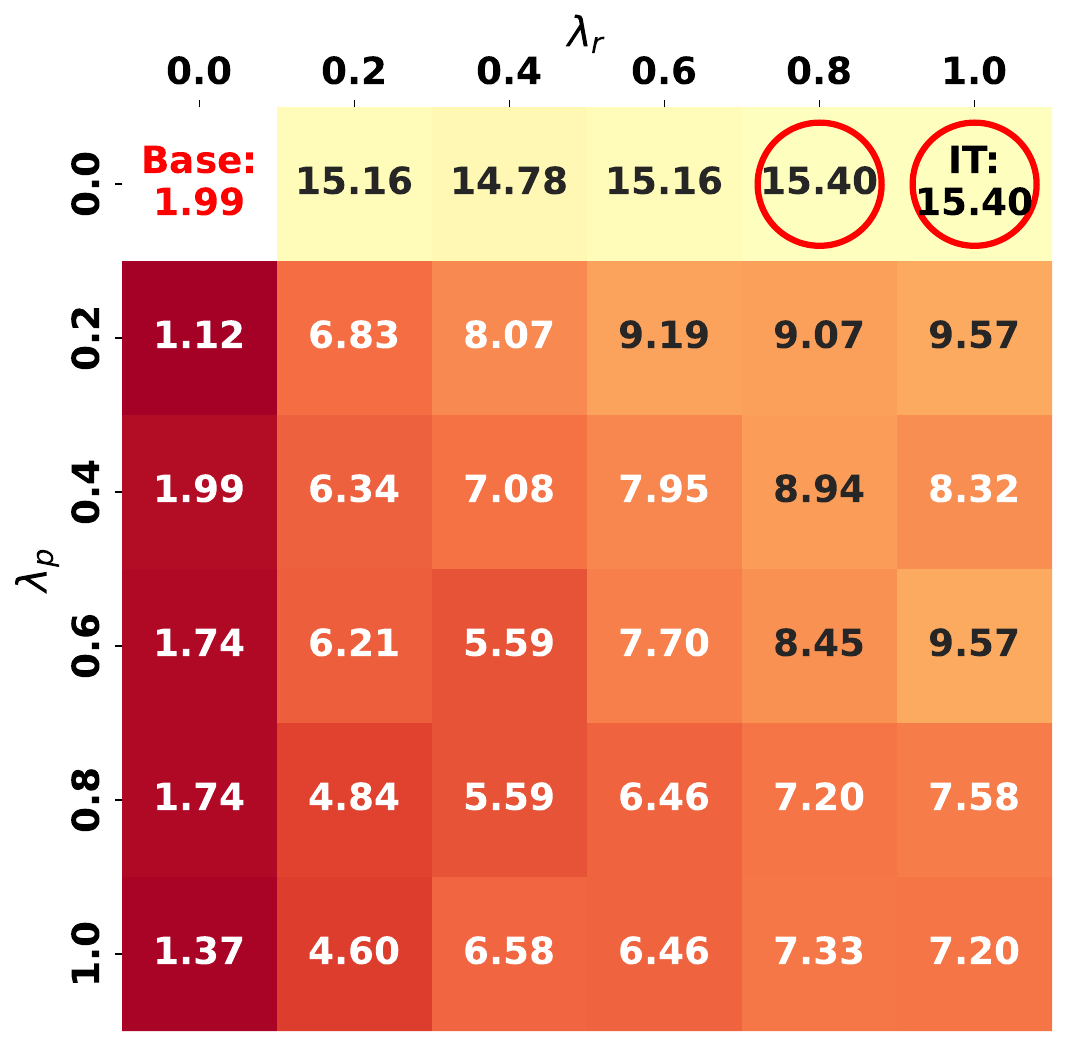} &
      \includegraphics[width=.19\textwidth]{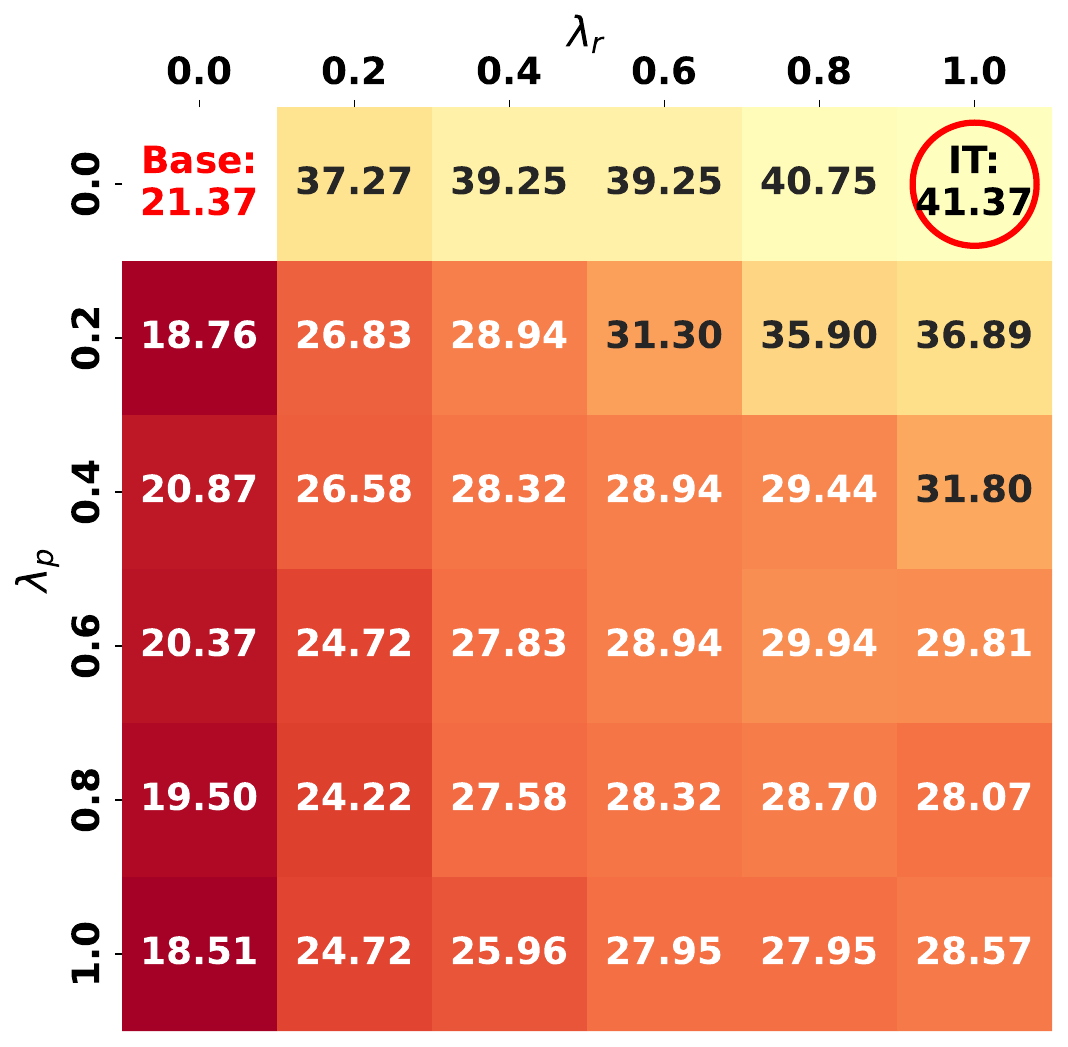} &
      \includegraphics[width=.19\textwidth]{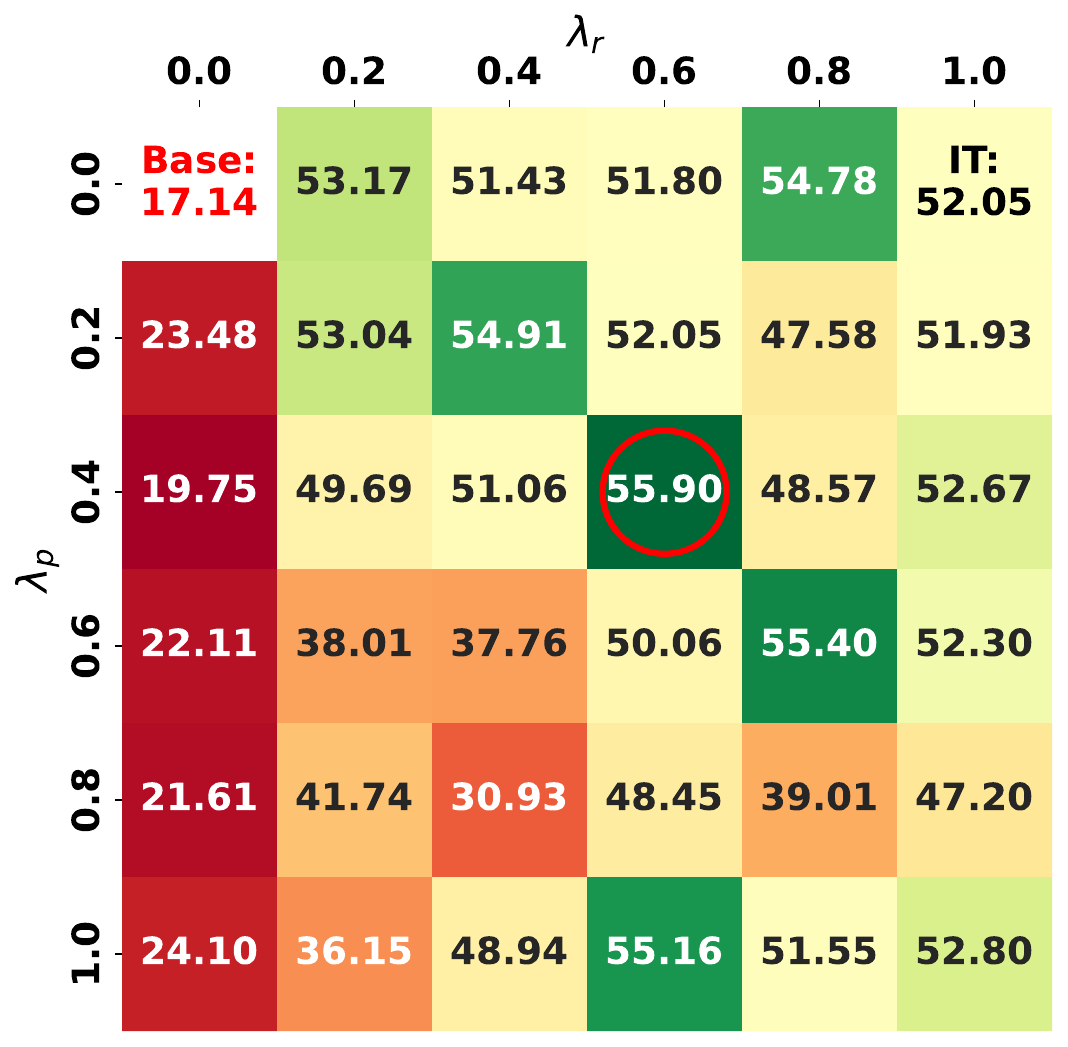} &
      \includegraphics[width=.19\textwidth]{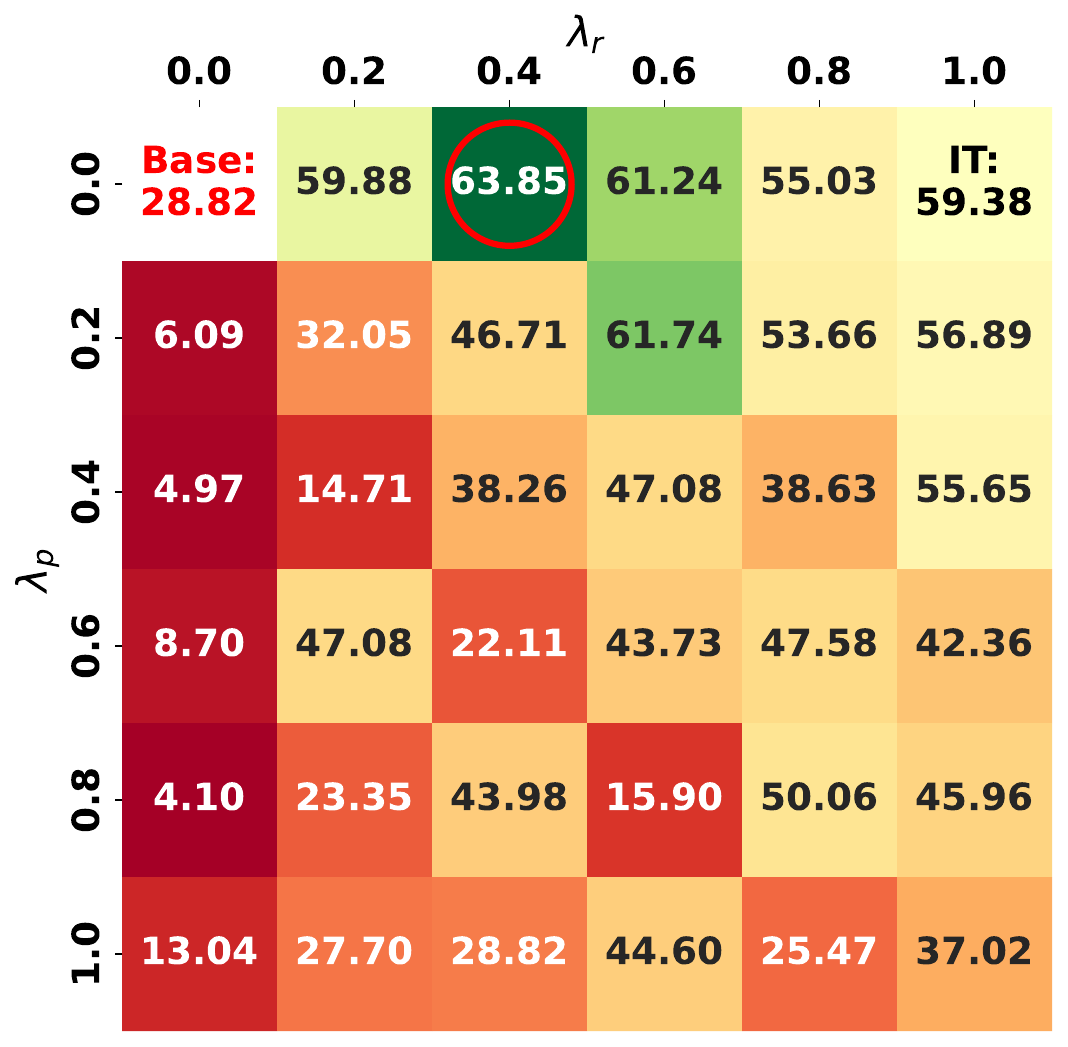} &
      \includegraphics[width=.19\textwidth]{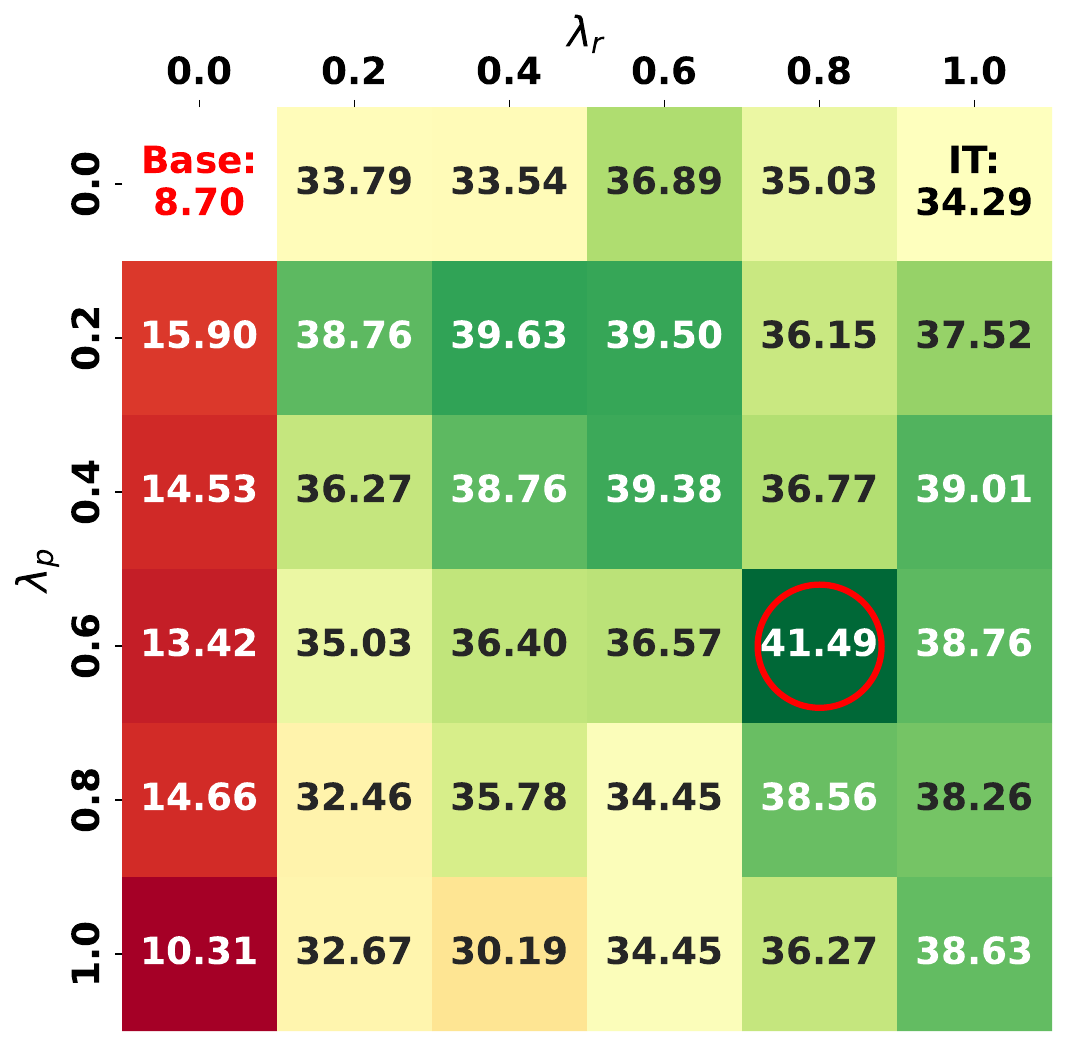} \\[4pt]

      \raisebox{1.3cm}[0pt][0pt]{\rotatebox{90}{\makebox[0pt][c]{IFEval}}} &
      \includegraphics[width=.19\textwidth]{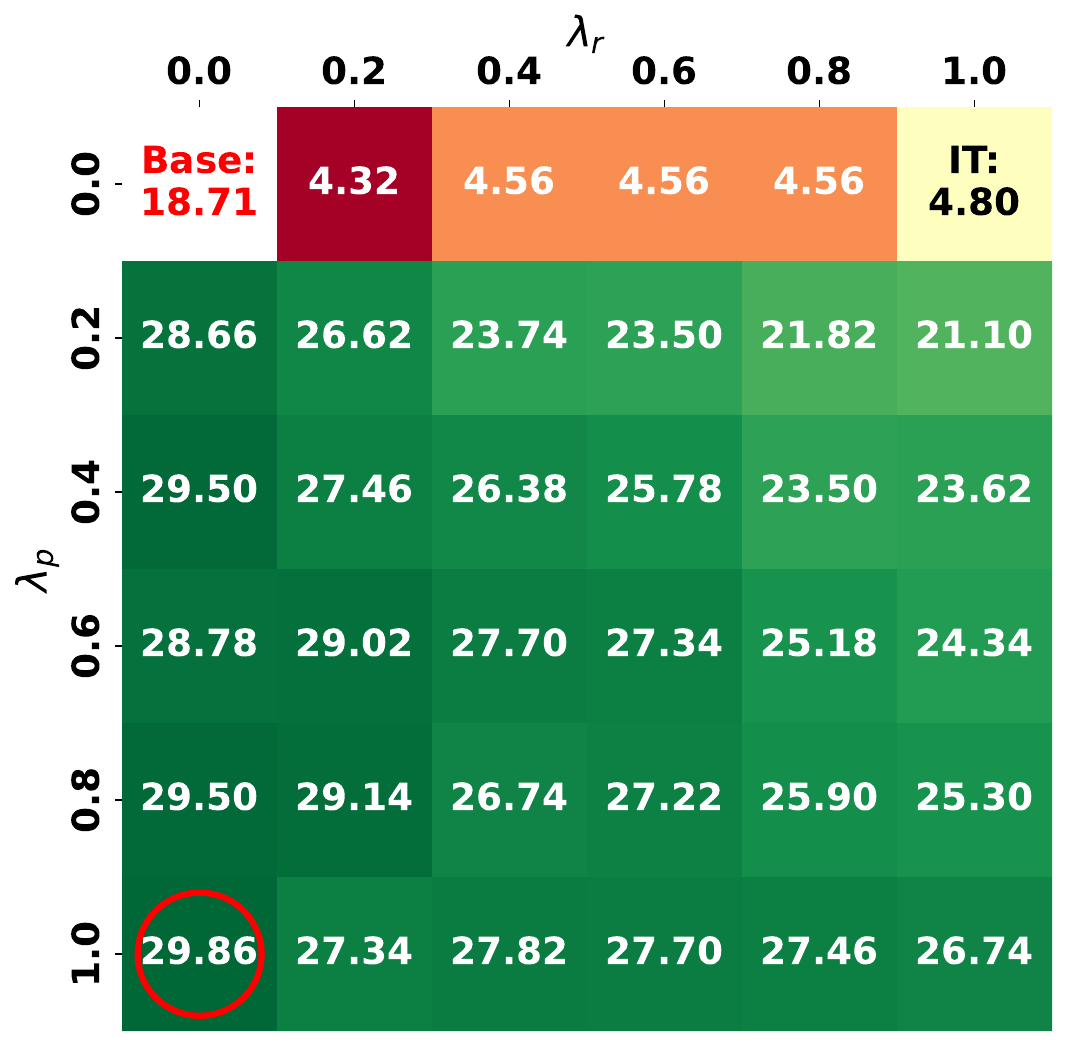} &
      \includegraphics[width=.19\textwidth]{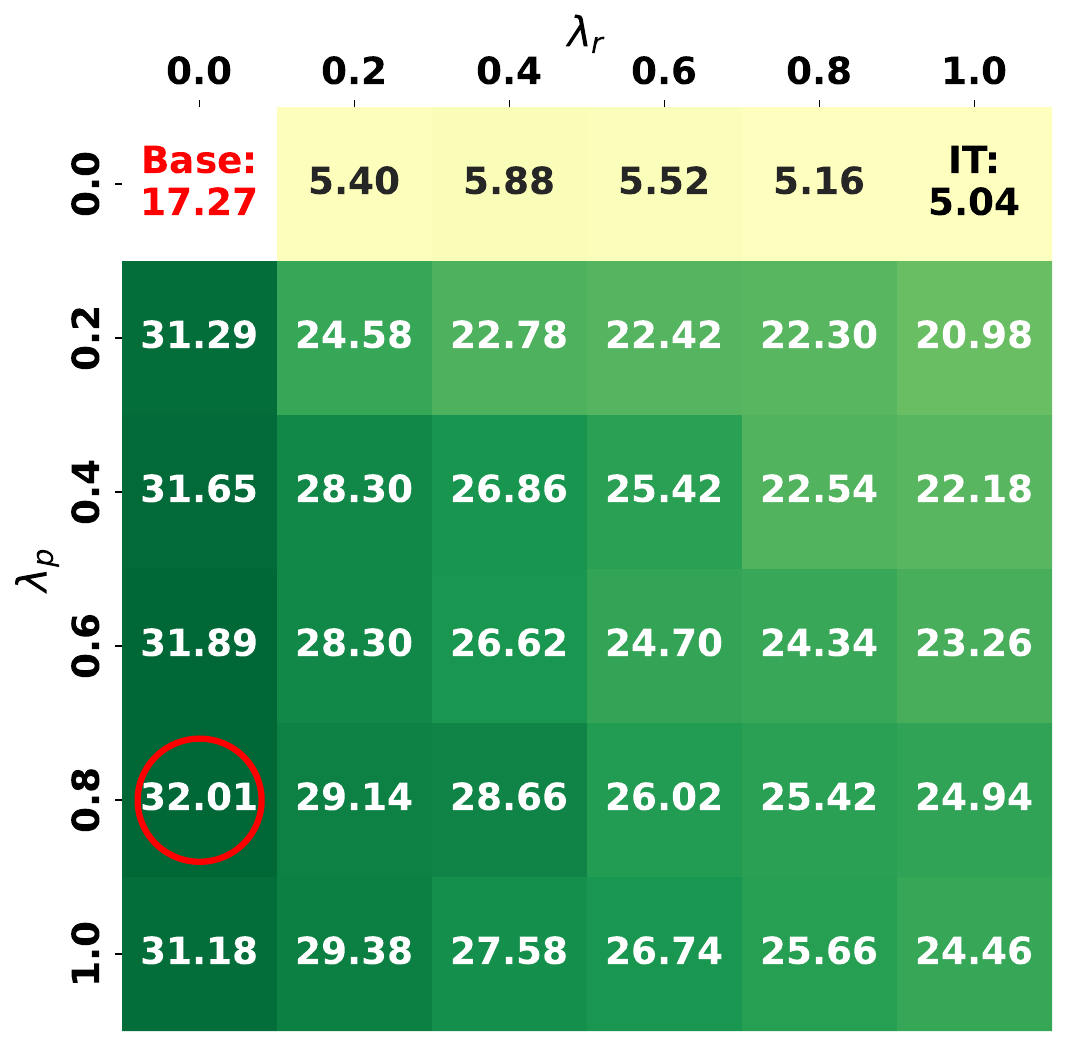} &
      \includegraphics[width=.19\textwidth]{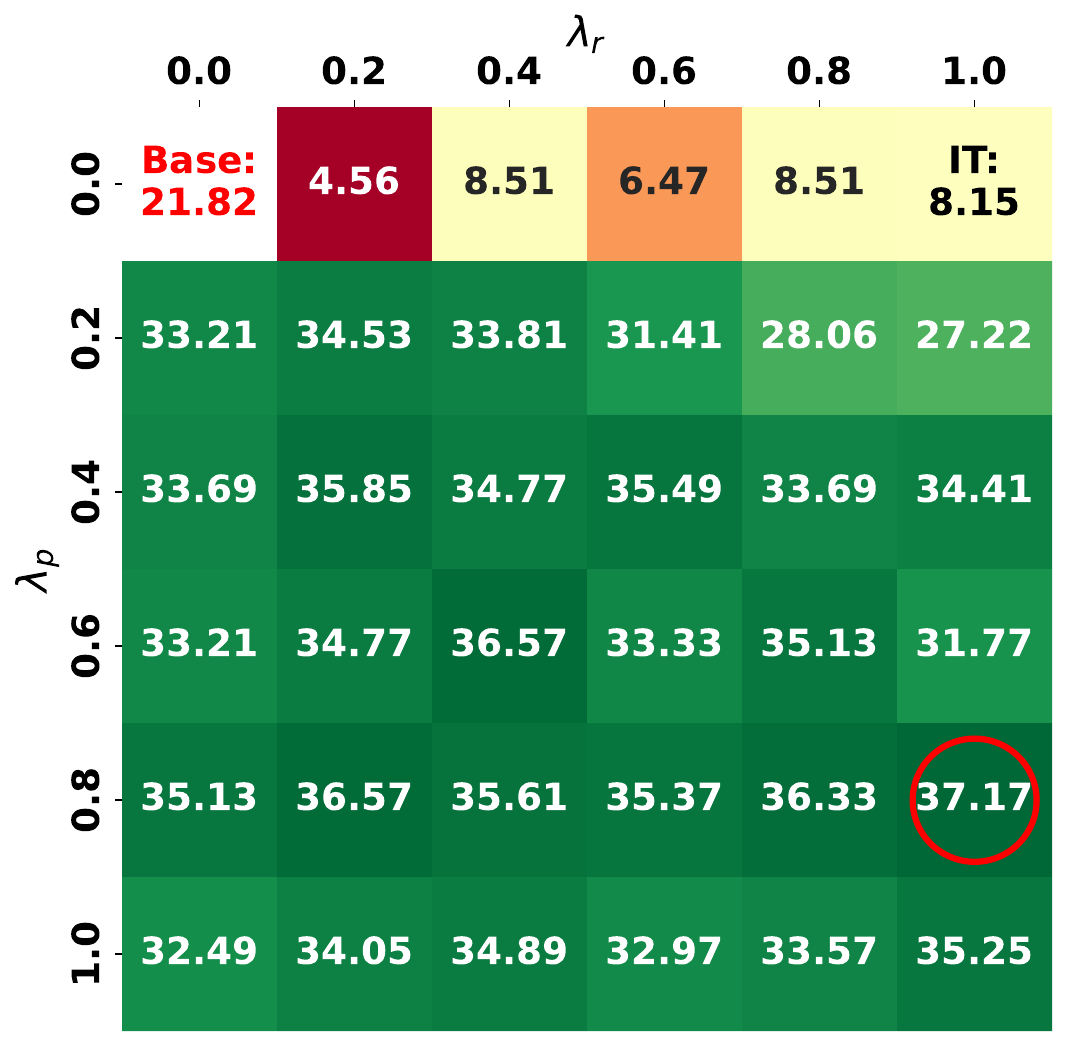} &
      \includegraphics[width=.19\textwidth]{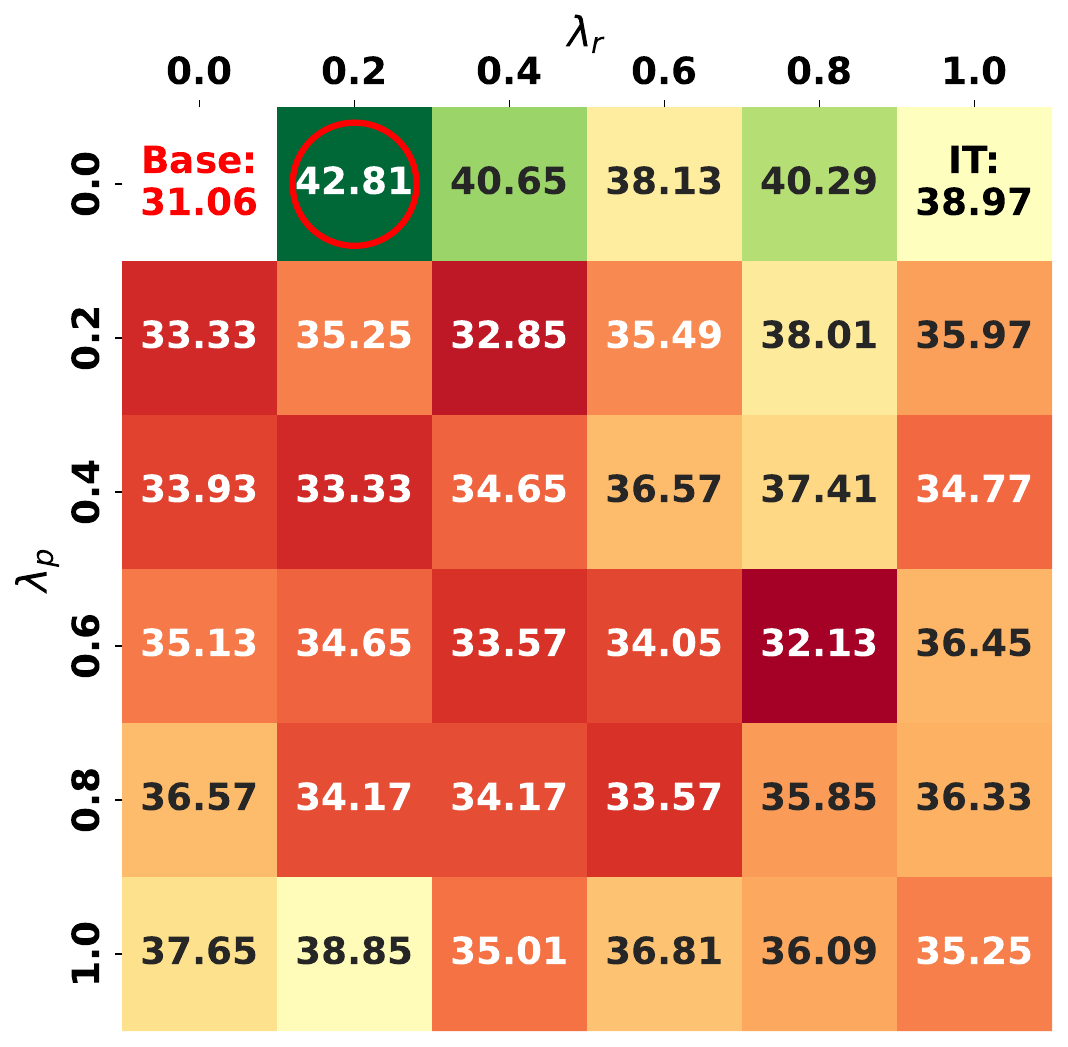} &
      \includegraphics[width=.19\textwidth]{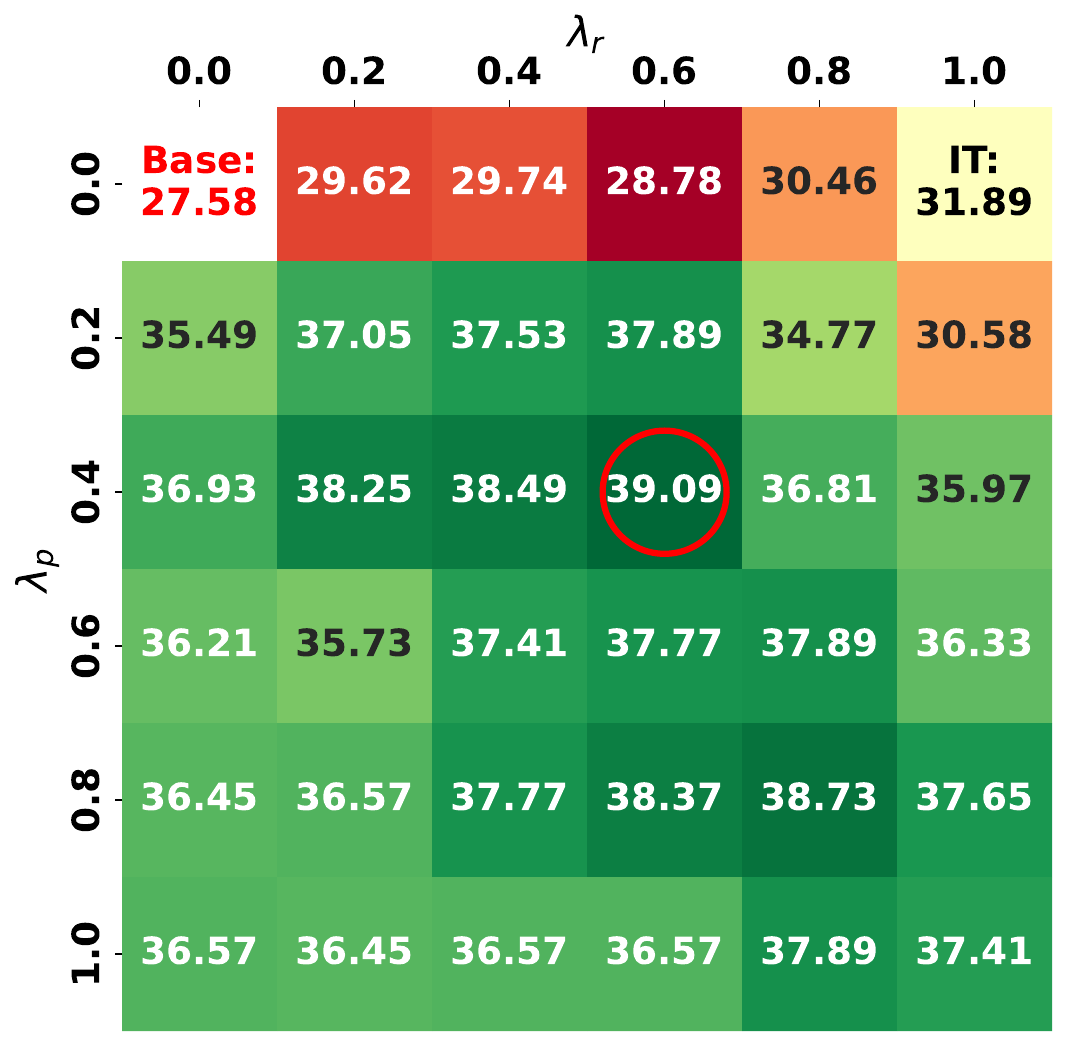} \\[4pt]

      \raisebox{1.3cm}[0pt][0pt]{\rotatebox{90}{\makebox[0pt][c]{MT-Bench}}} &
      \includegraphics[width=.19\textwidth]{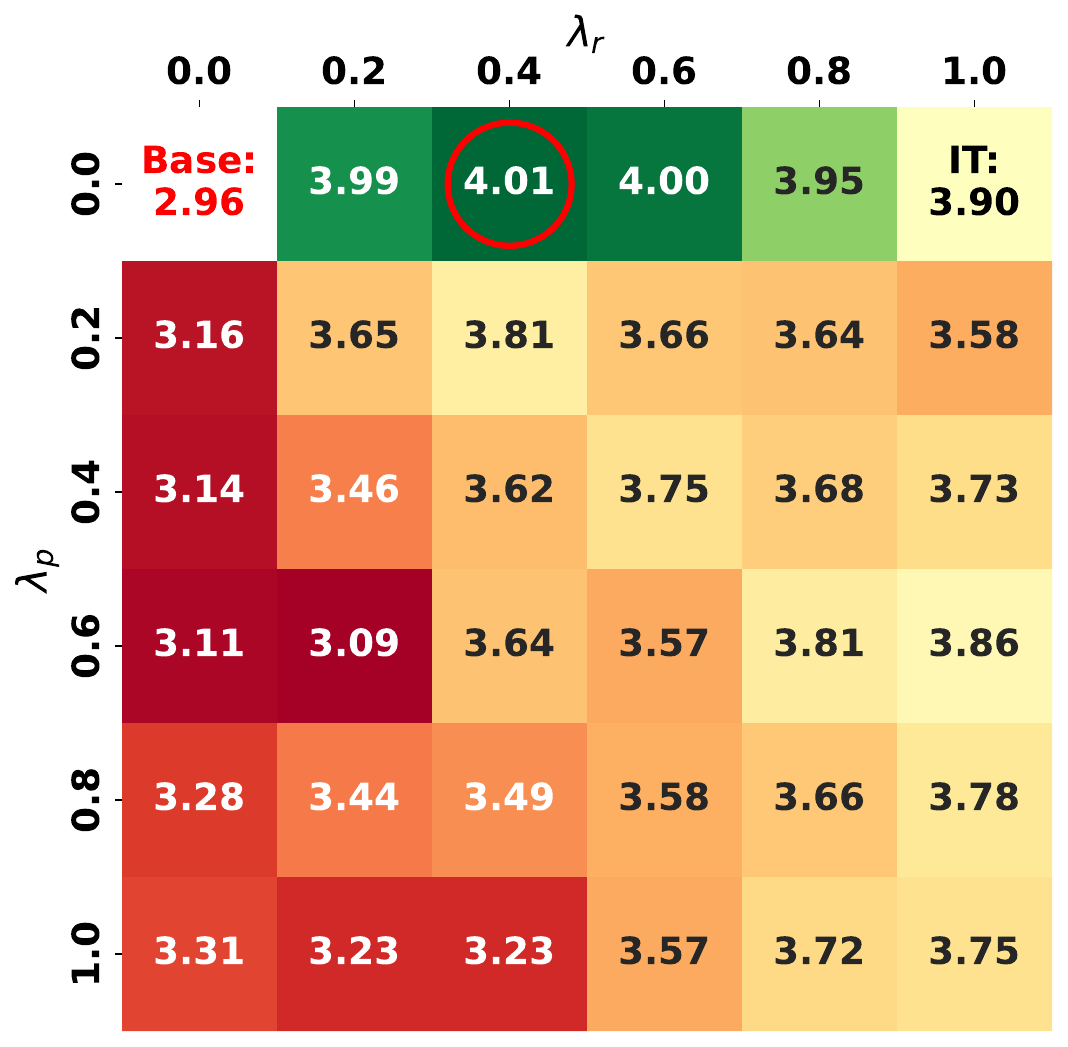} &
      \includegraphics[width=.19\textwidth]{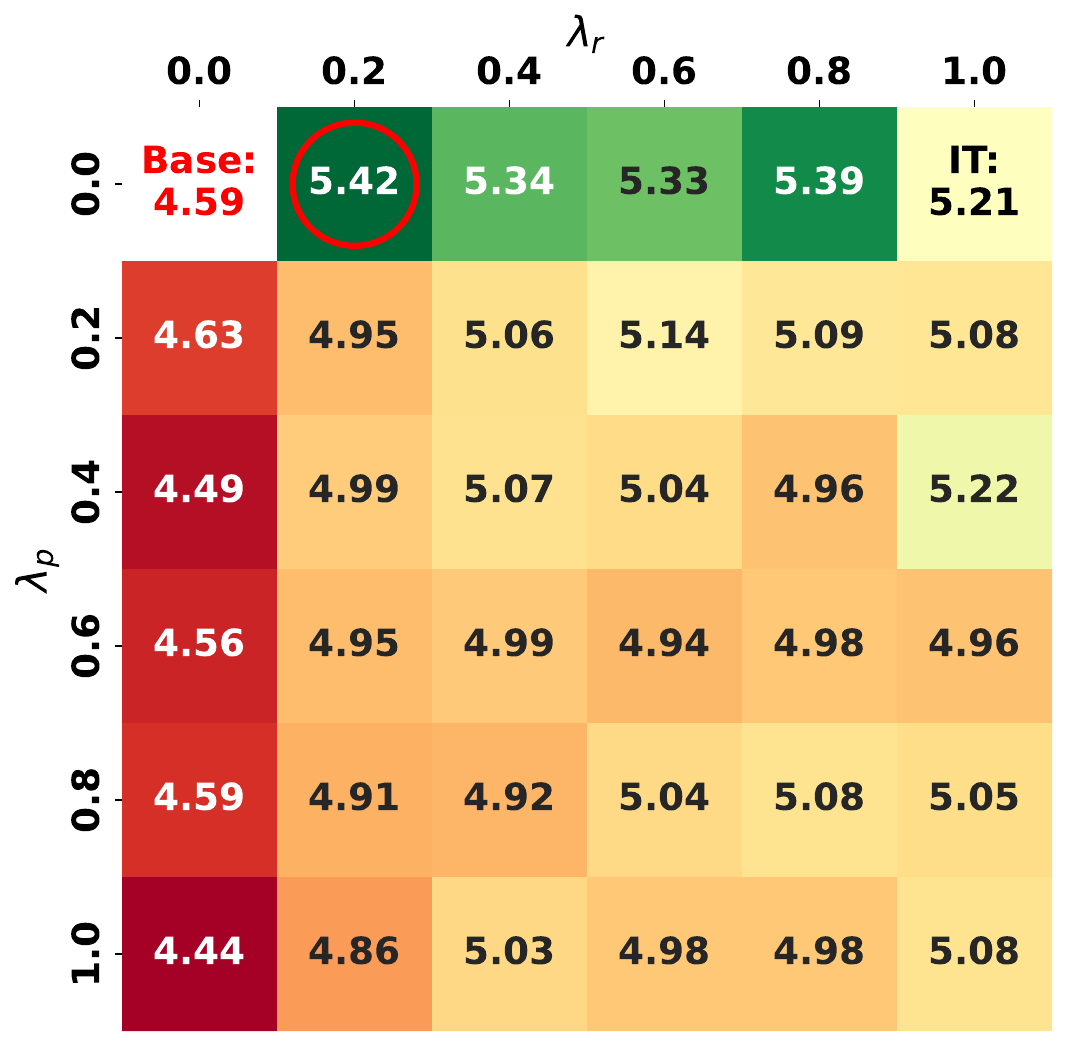} &
      \includegraphics[width=.19\textwidth]{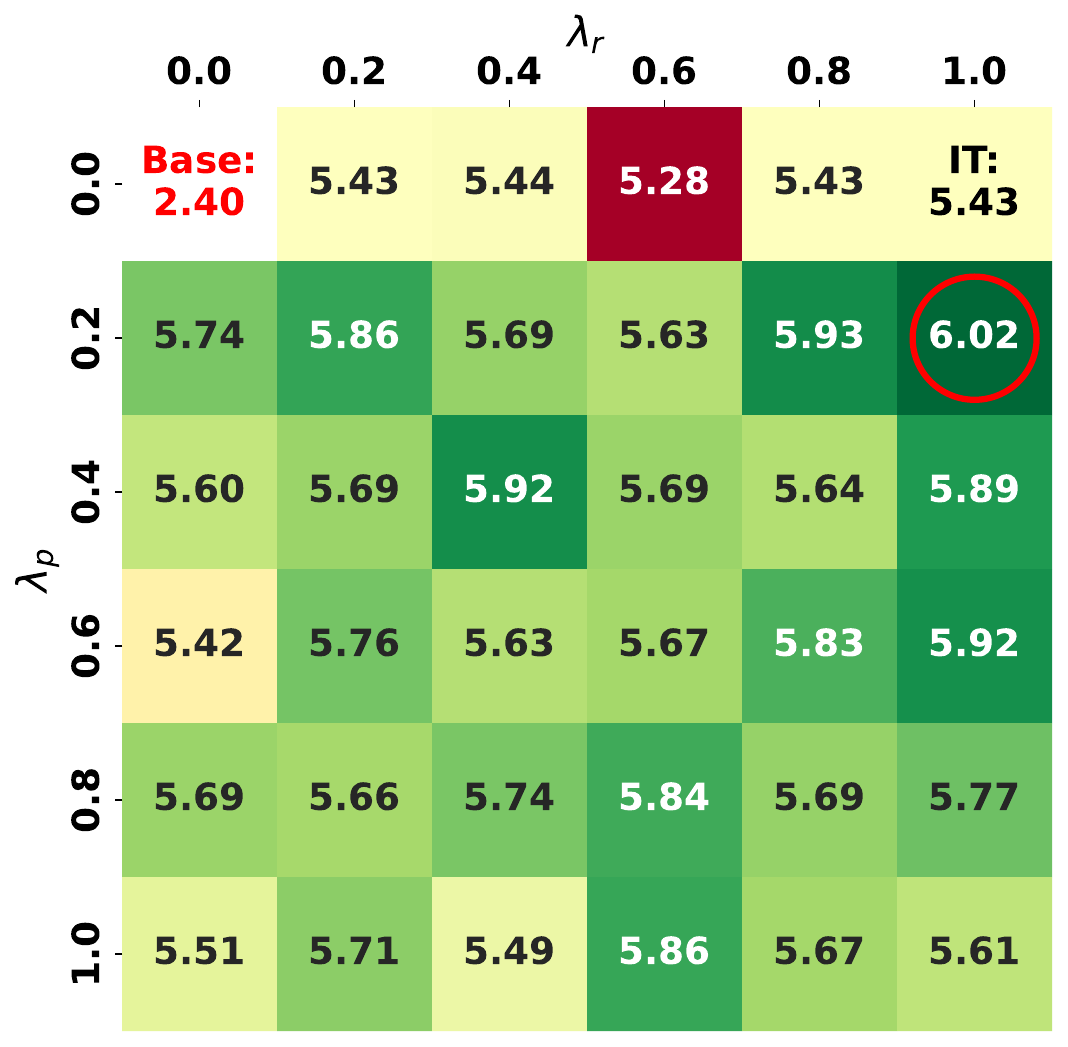} &
      \includegraphics[width=.19\textwidth]{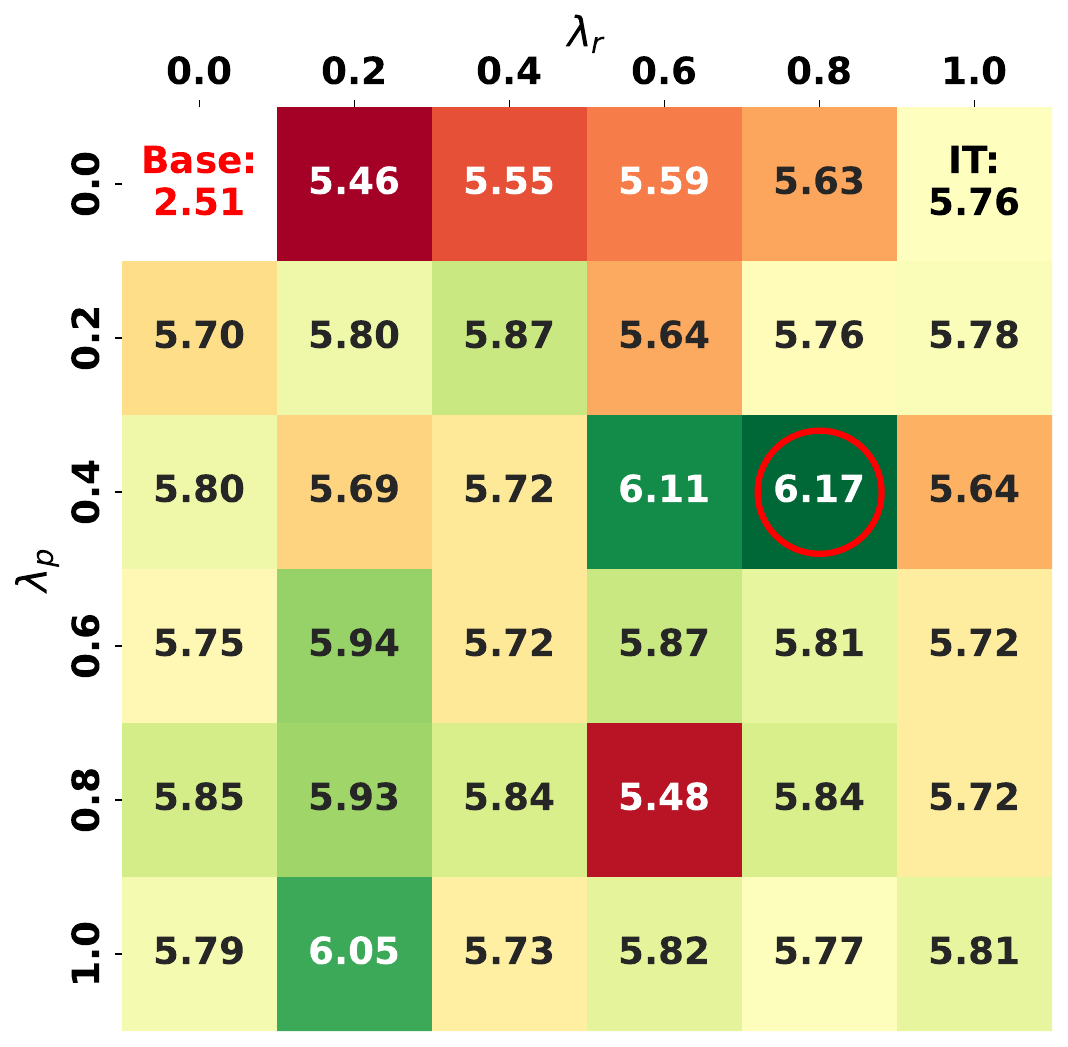} &
      \includegraphics[width=.19\textwidth]{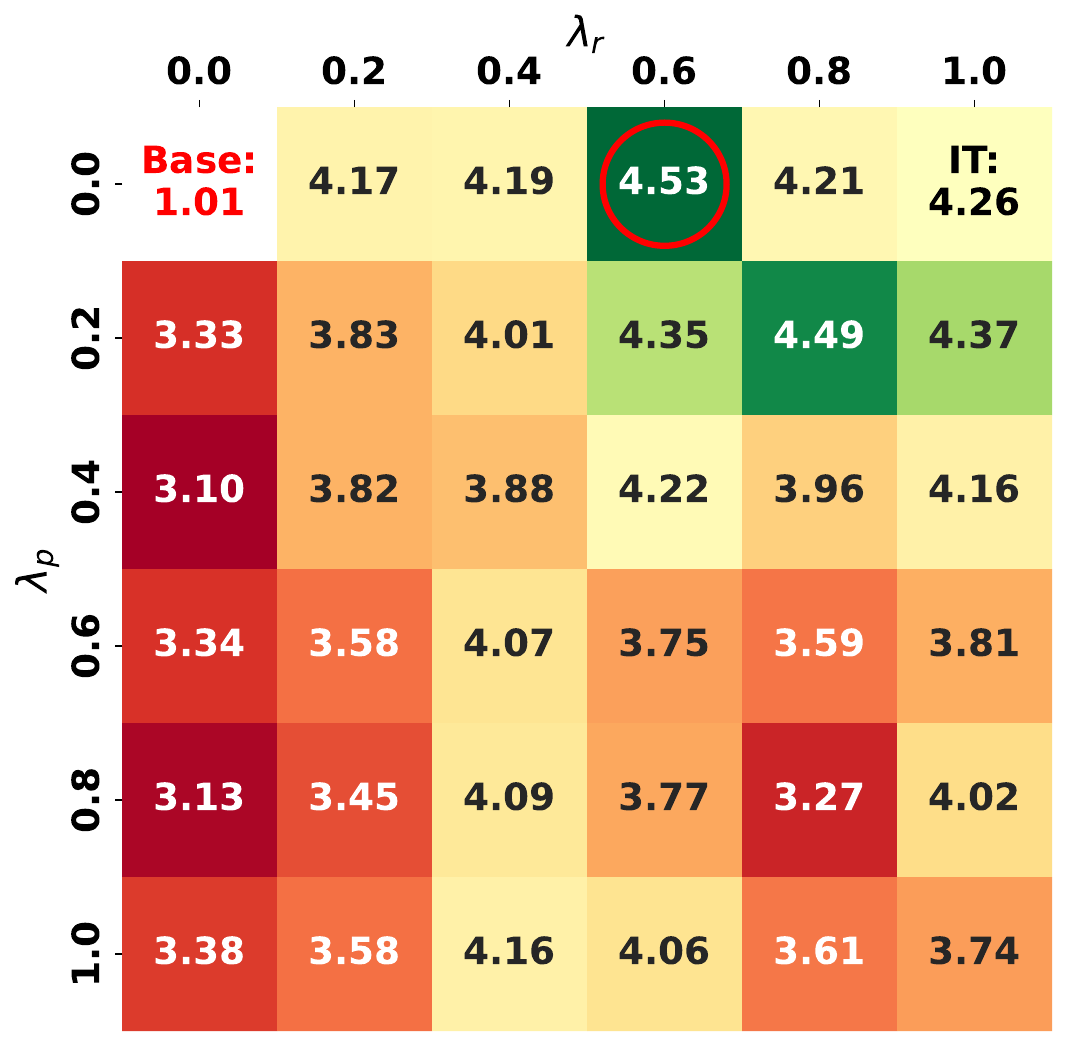} \\
  \end{tabular}

    \caption{Heatmaps depicting performance on MMLU (first row), BBH (second row), AlpacaEval (third row), IFEval (fourth row), and MT-Bench (fifth row) for different configurations of ($\lambda_p$, $\lambda_r$) and for different models finetuned on \textbf{T\"ulu-v2}. In each heatmap, the best performance is highlighted with a red circle. The color map is based on relative gain with respect to conventional instruction tuning. Each row of a heatmap corresponds to a prompt-token weight, and each column corresponds to a response-token weight. Conventional instruction tuning is marked with \texttt{IT}, and base model performance is marked with \texttt{Base}.}
    \label{fig:tulu_all}
\end{figure*}

\begin{figure*}[ht!]
  \centering
  \setlength{\tabcolsep}{2pt}      
  \renewcommand{\arraystretch}{1}  

  \begin{tabular}{@{} >{\centering\arraybackslash}m{2.1mm} *{5}{c} @{}}
      & Llama-3-1B & Llama-3-3B & Llama-3-8B & Mistral-7B & Gemma-2-2B \\[2pt]

      \raisebox{1.3cm}[0pt][0pt]{\rotatebox{90}{\makebox[0pt][c]{MMLU}}} &
      \includegraphics[width=.19\textwidth]{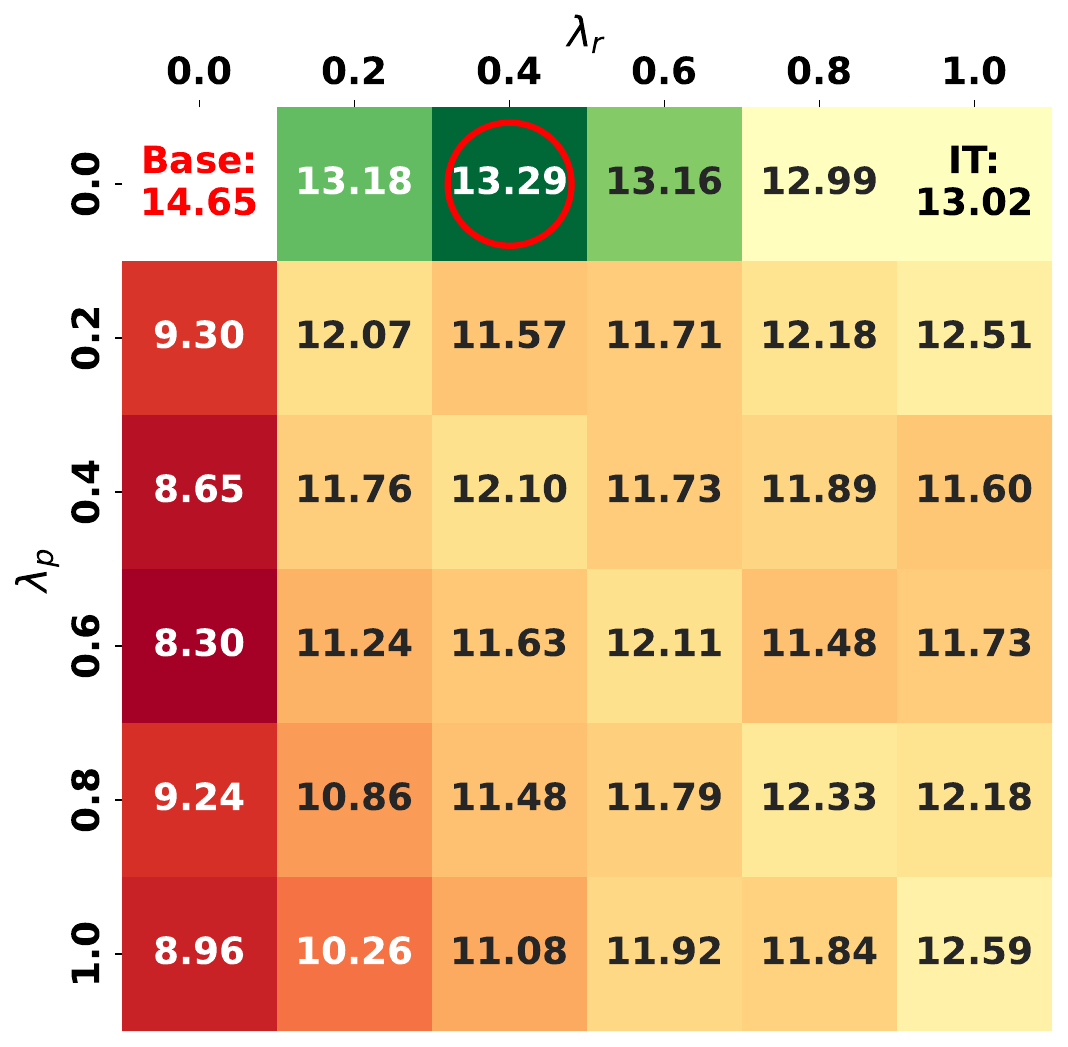} &
      \includegraphics[width=.19\textwidth]{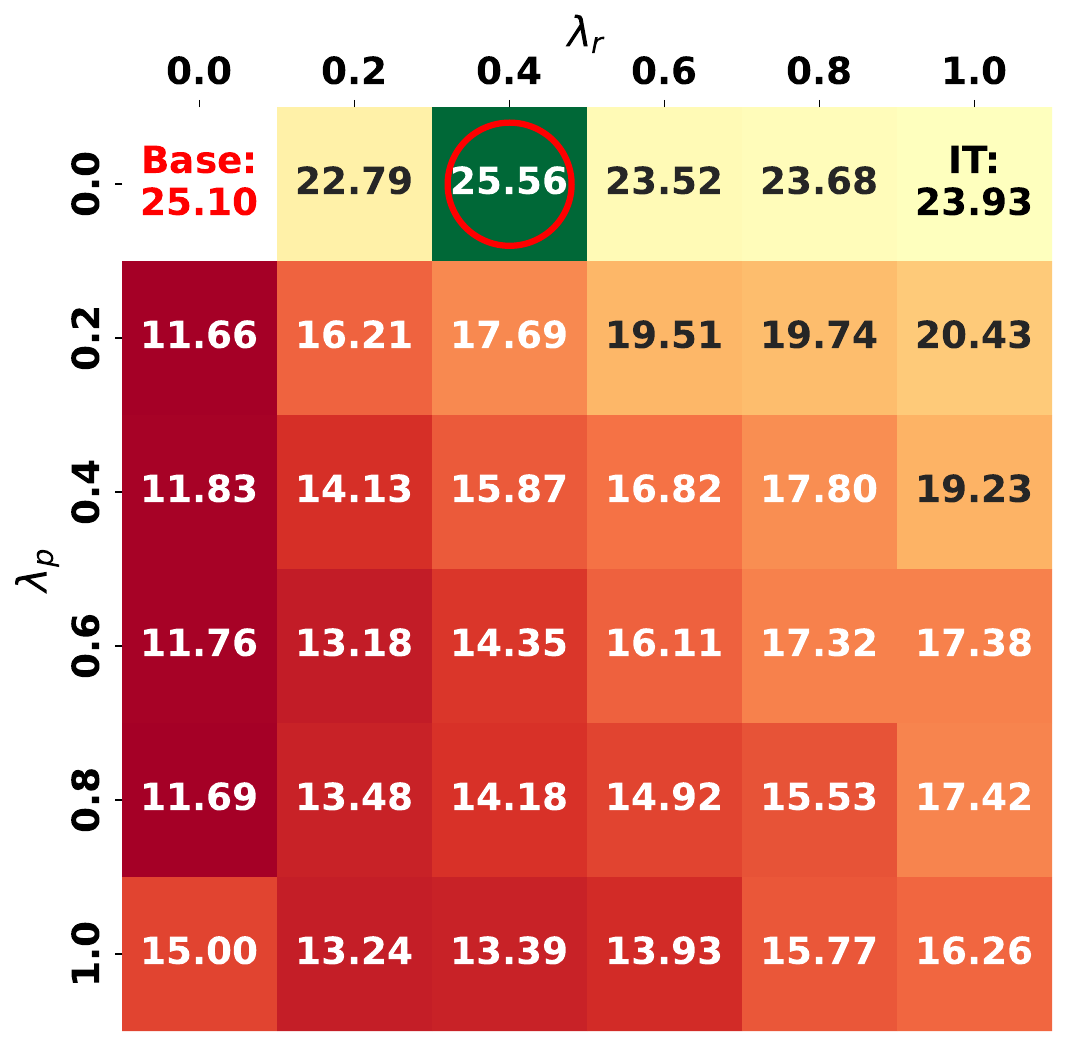} &
      \includegraphics[width=.19\textwidth]{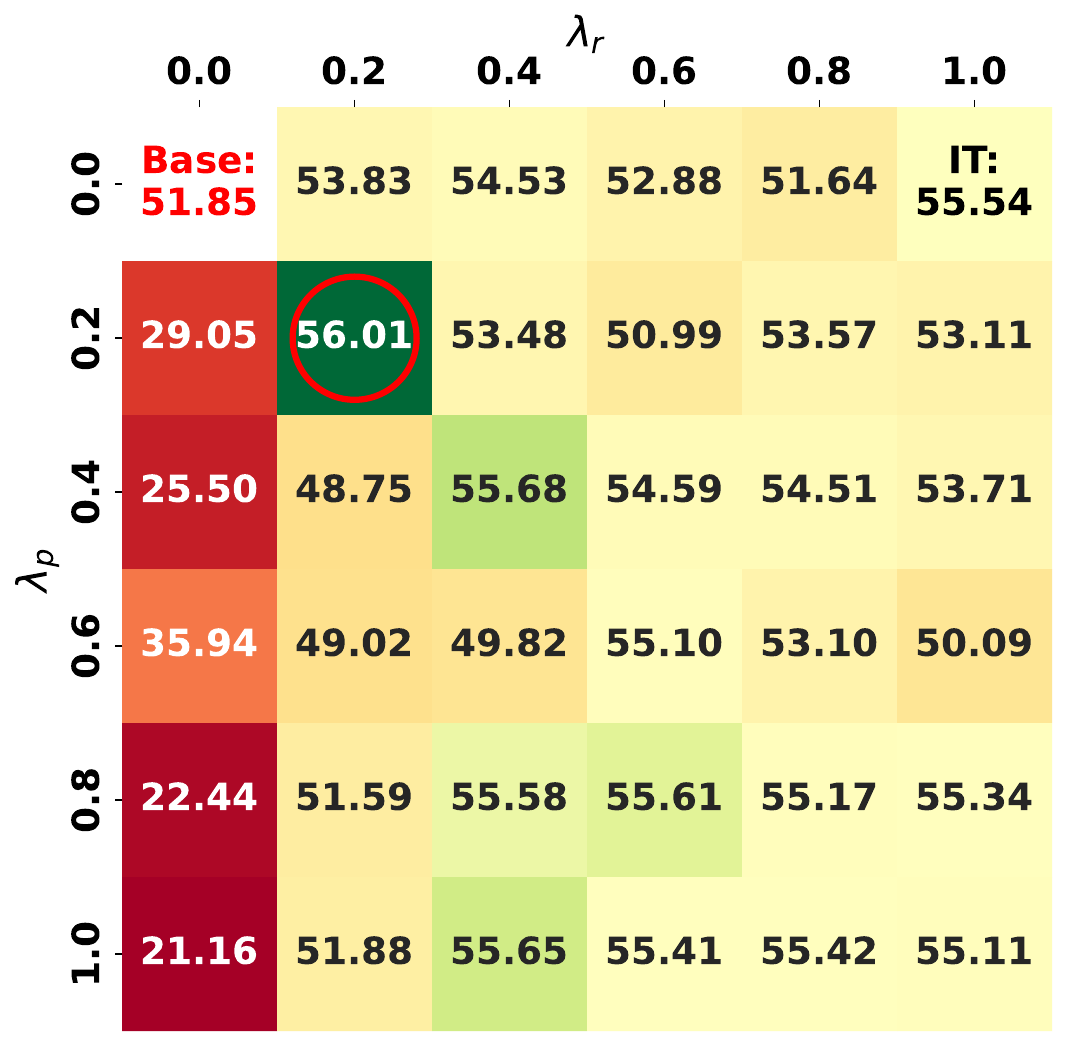} &
      \includegraphics[width=.19\textwidth]{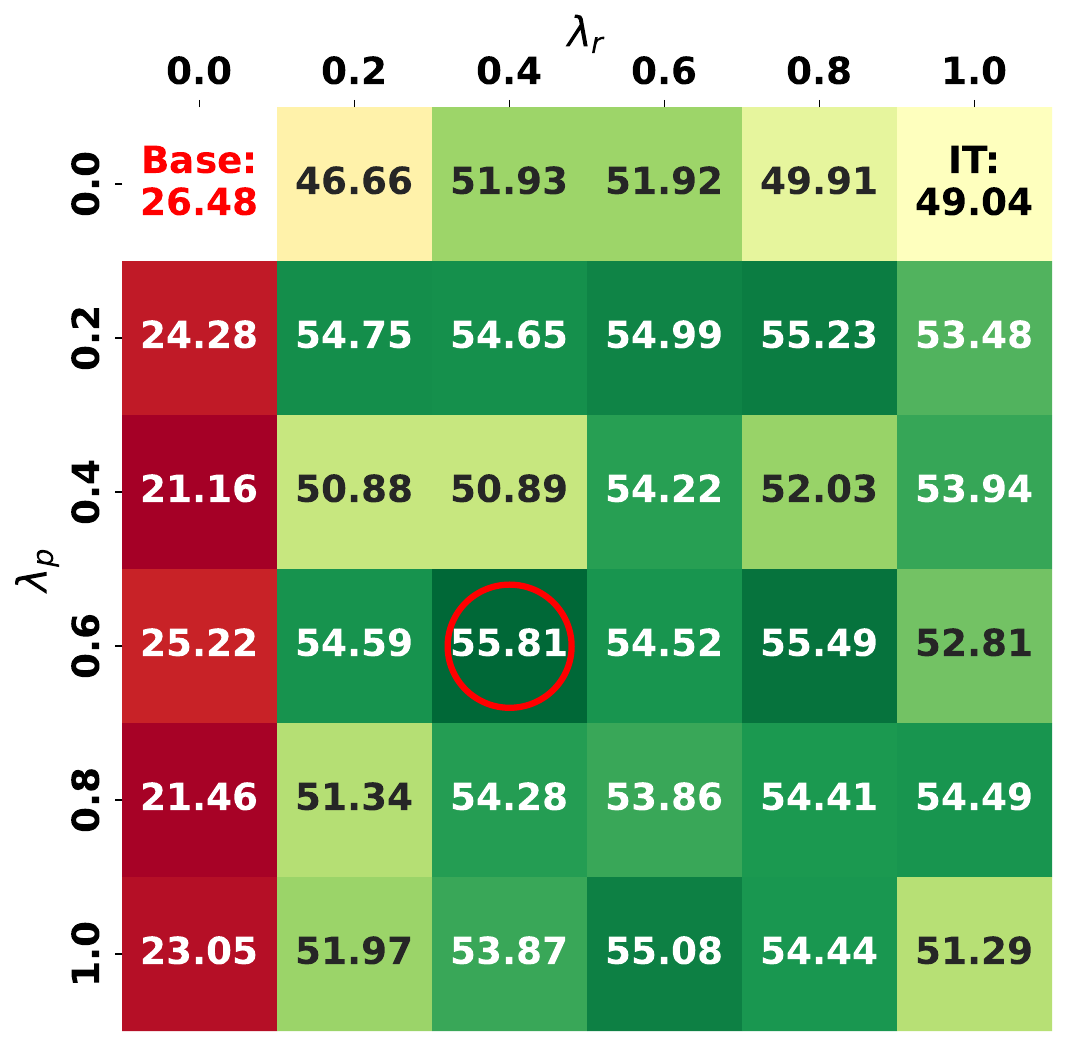} &
      \includegraphics[width=.19\textwidth]{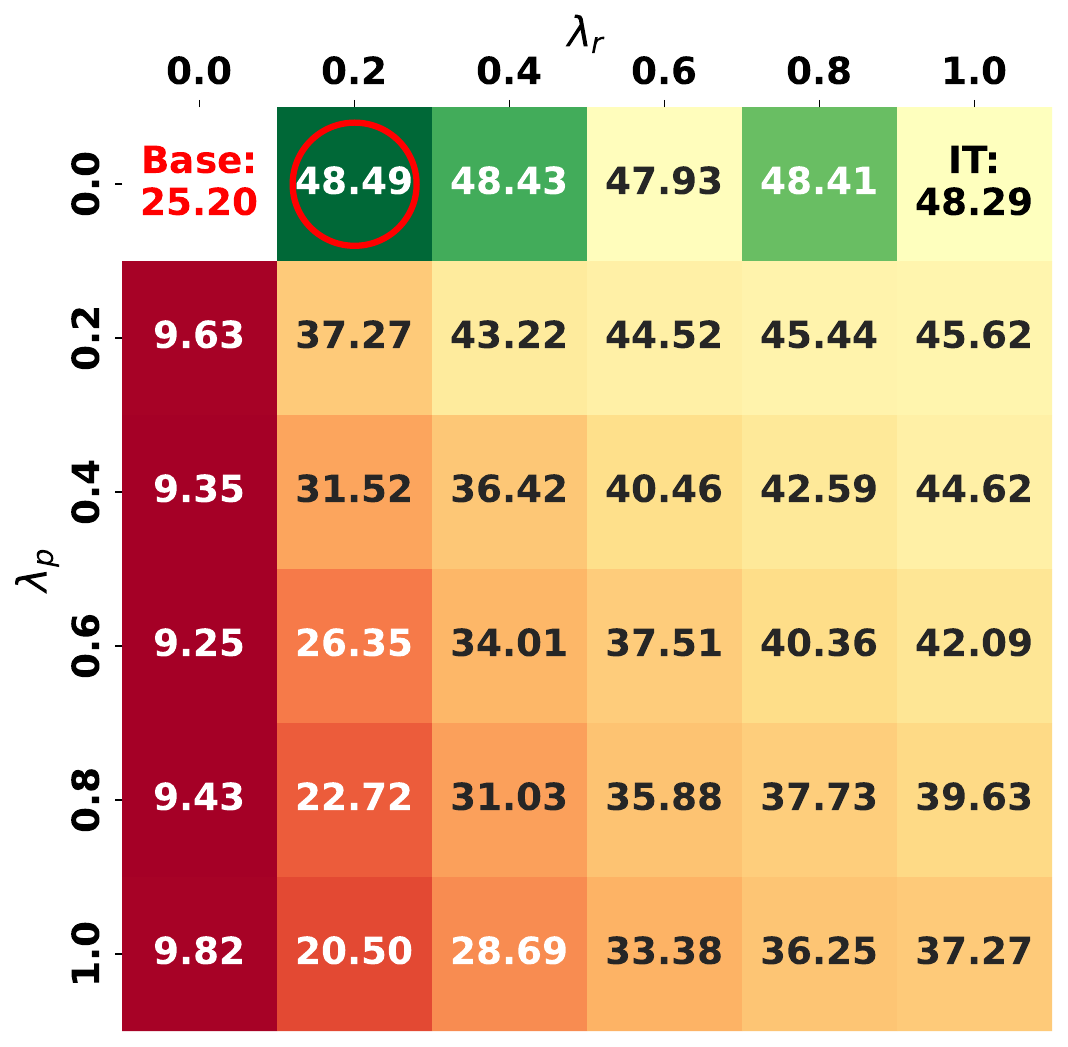} \\[4pt]

        \raisebox{1.3cm}[0pt][0pt]{\rotatebox{90}{\makebox[0pt][c]{BBH}}} &
      \includegraphics[width=.19\textwidth]{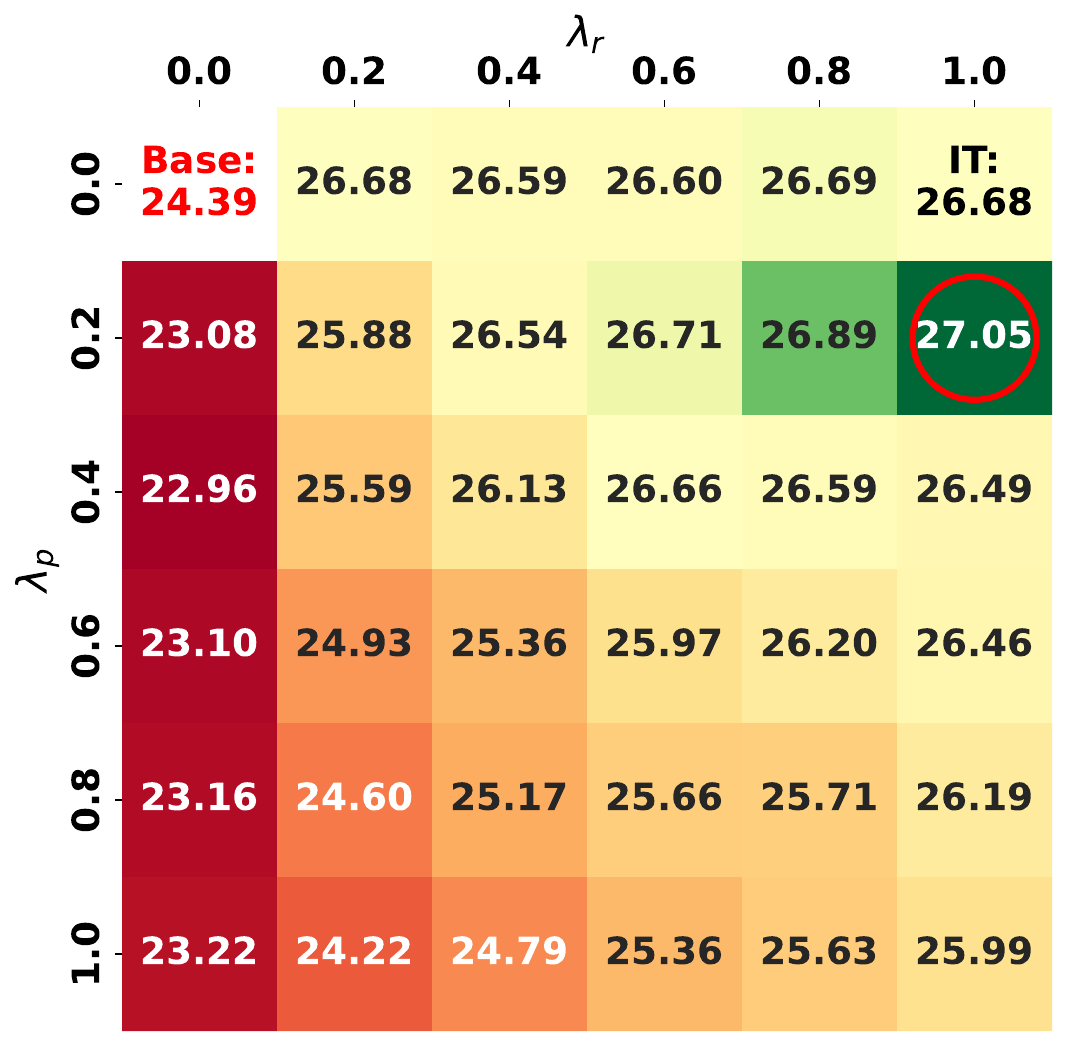} &
      \includegraphics[width=.19\textwidth]{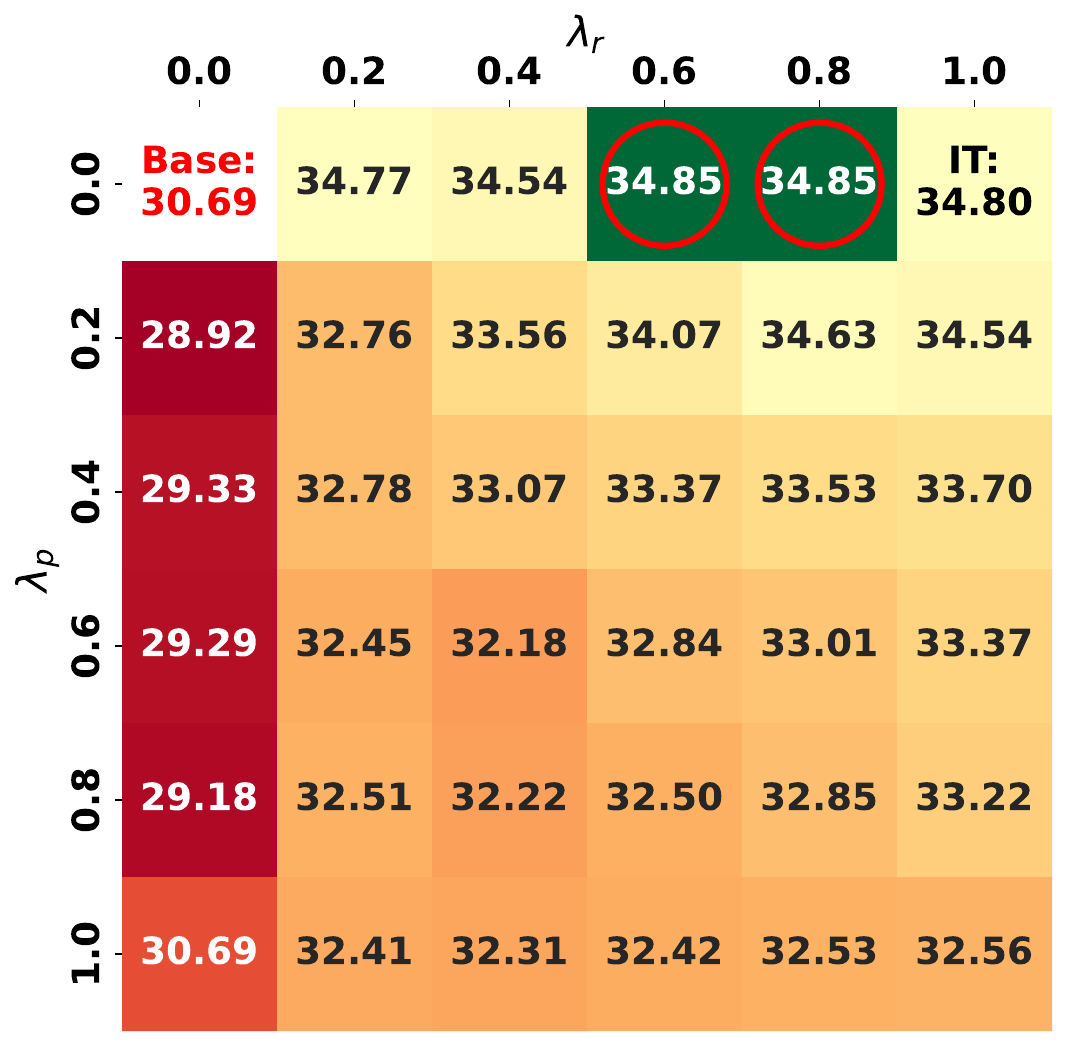} &
      \includegraphics[width=.19\textwidth]{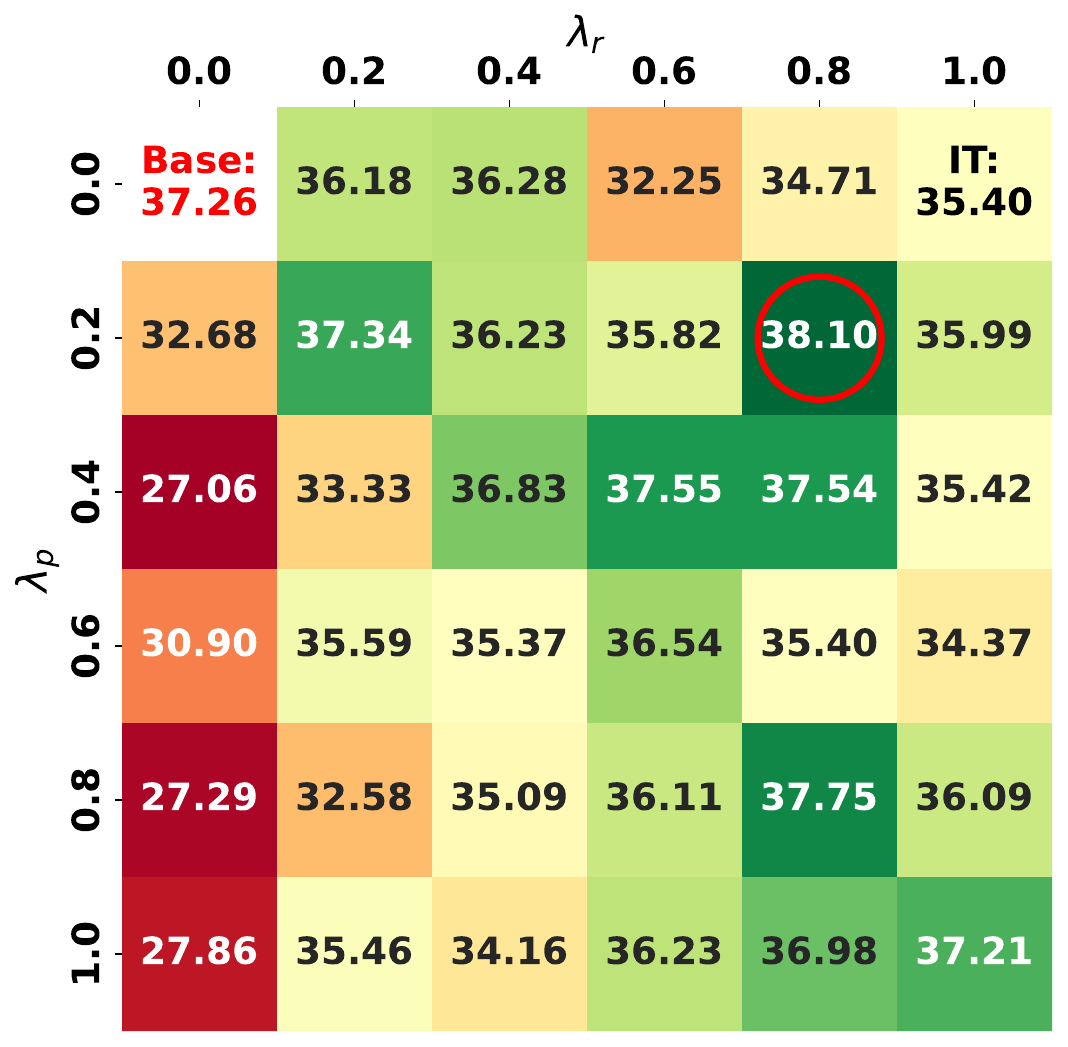} &
      \includegraphics[width=.19\textwidth]{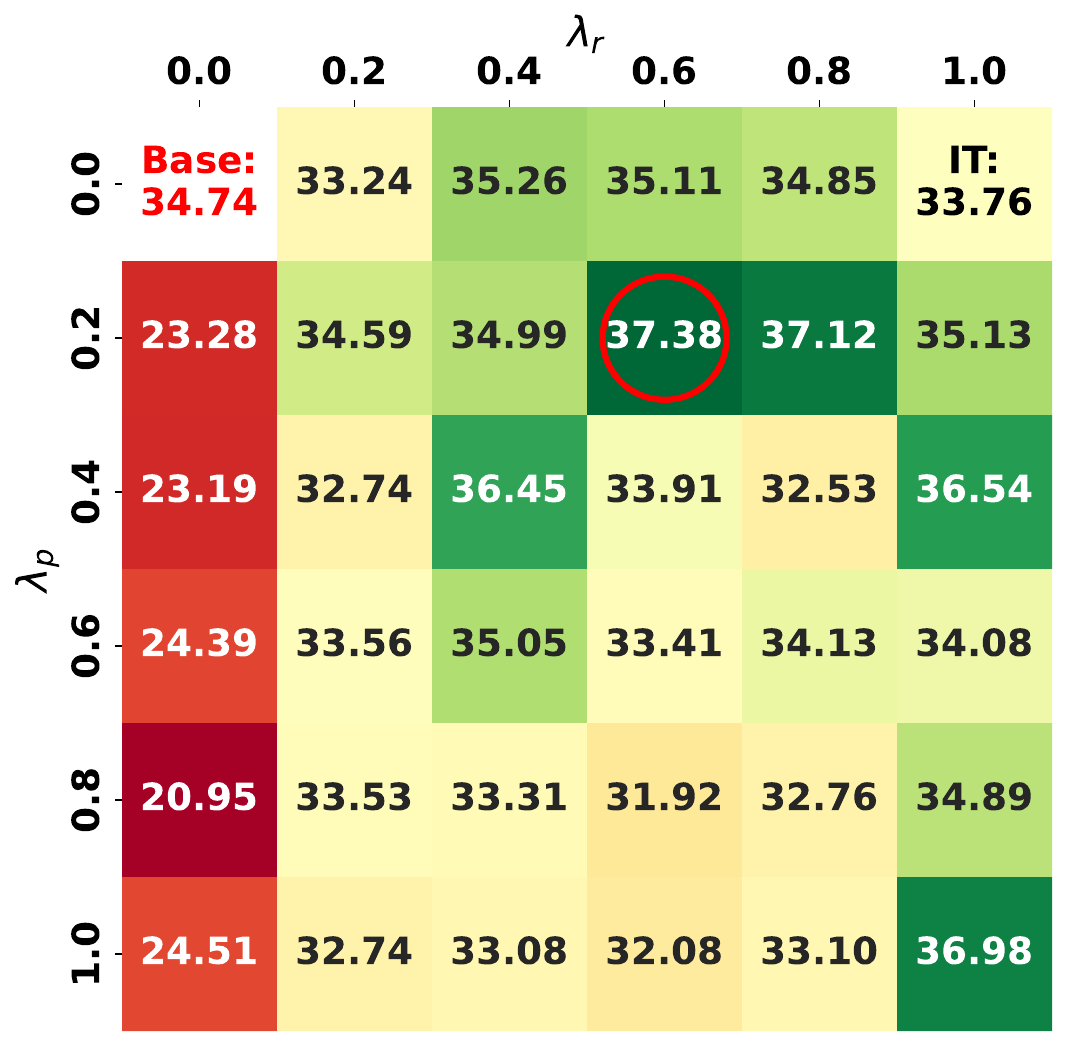} &
      \includegraphics[width=.19\textwidth]{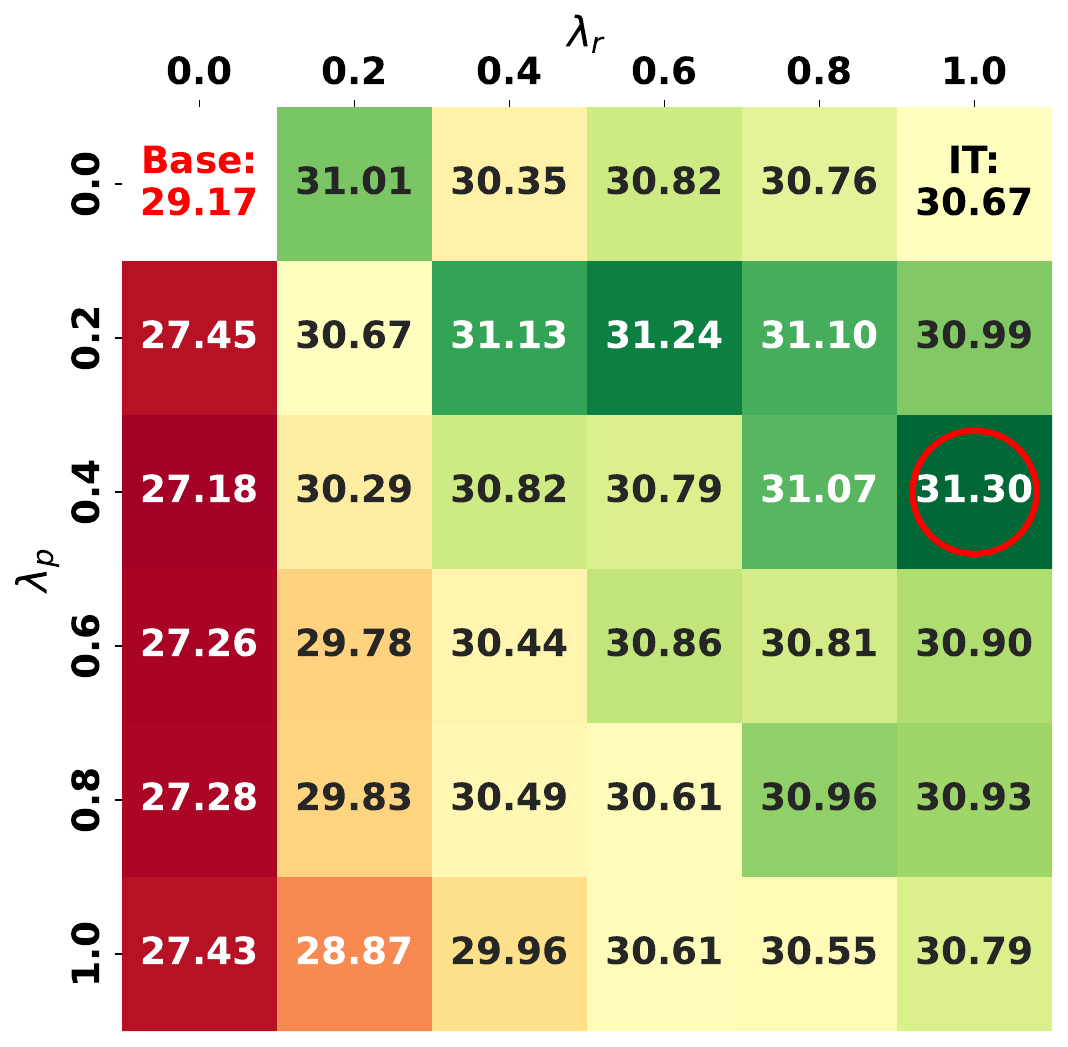} \\[4pt]

      \raisebox{1.3cm}[0pt][0pt]{\rotatebox{90}{\makebox[0pt][c]{AlpacaEval}}} &
      \includegraphics[width=.19\textwidth]{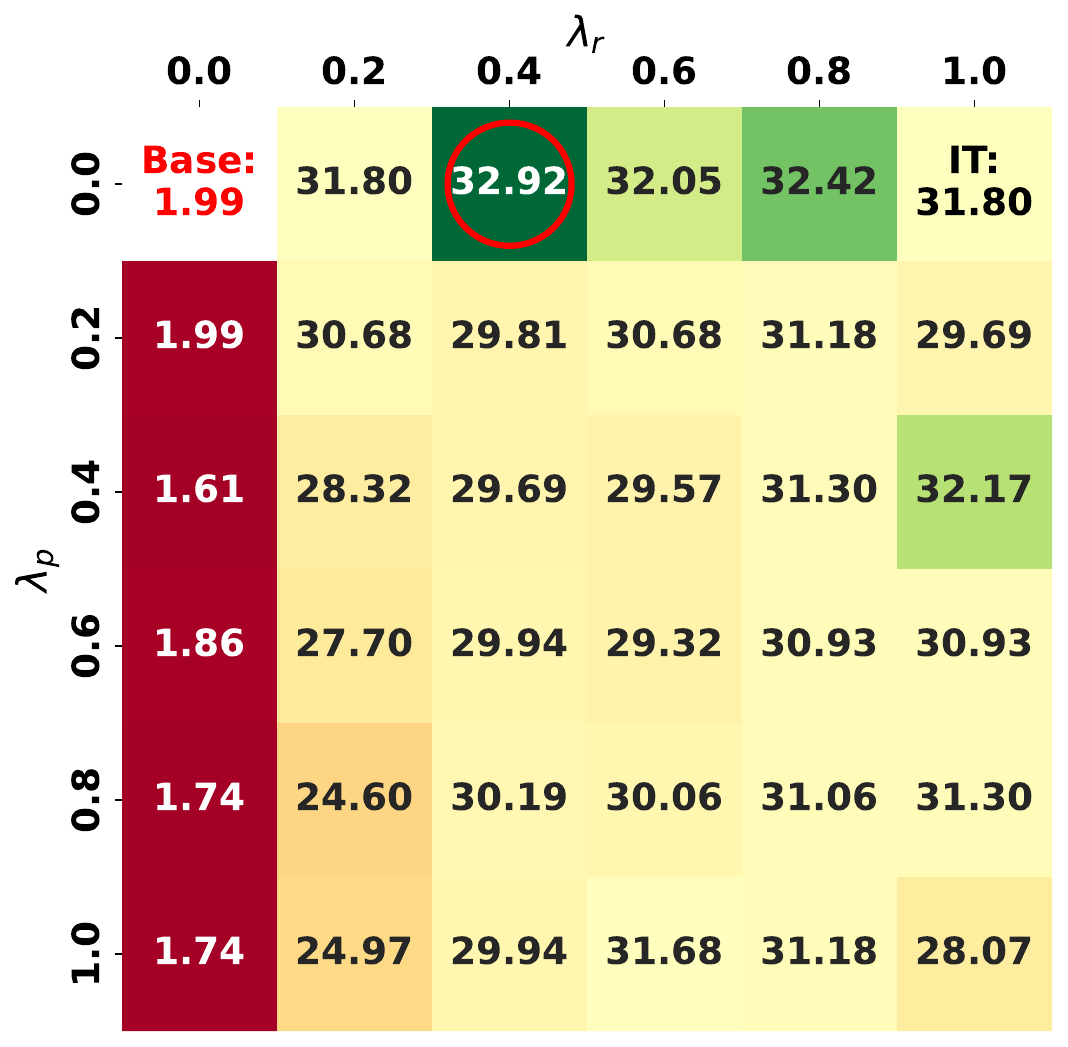} &
      \includegraphics[width=.19\textwidth]{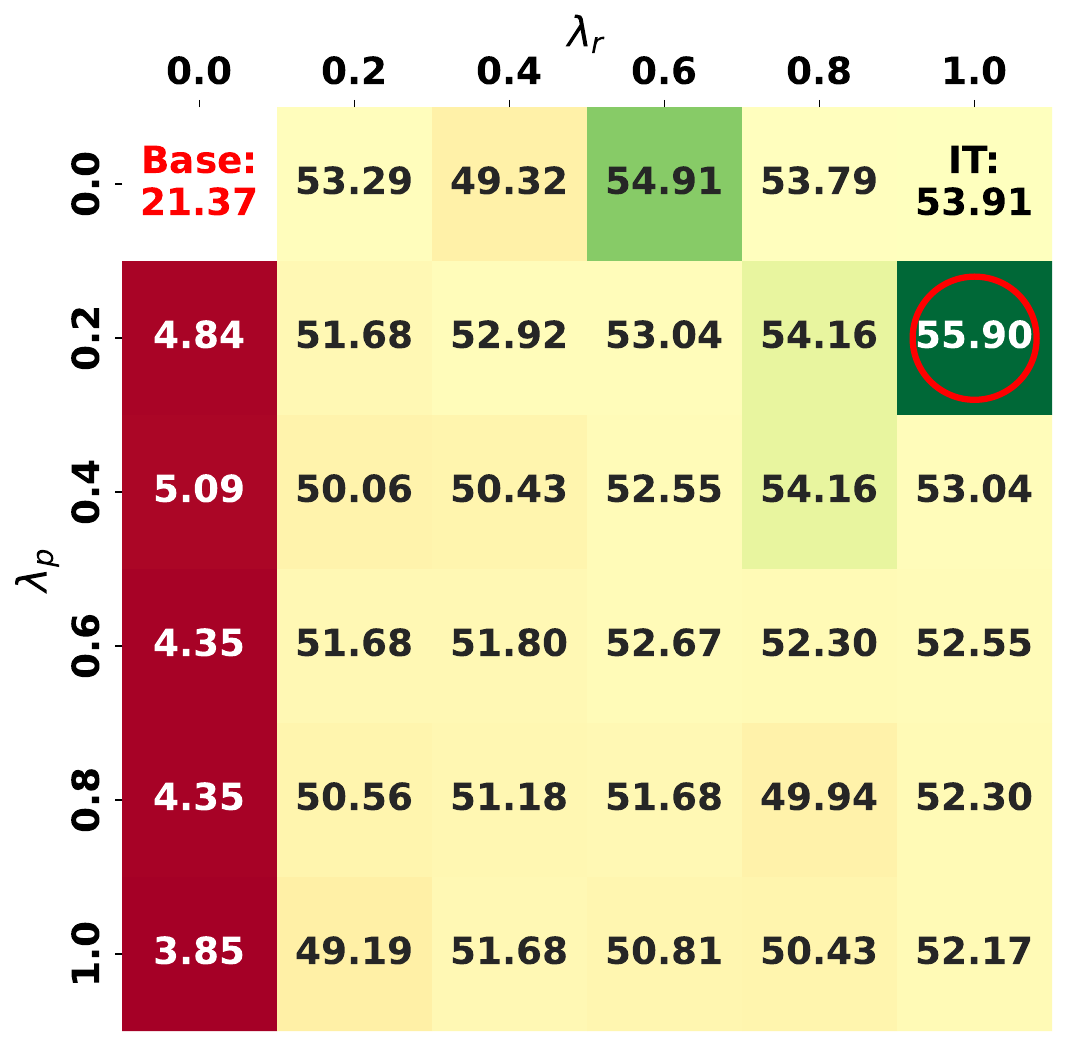} &
      \includegraphics[width=.19\textwidth]{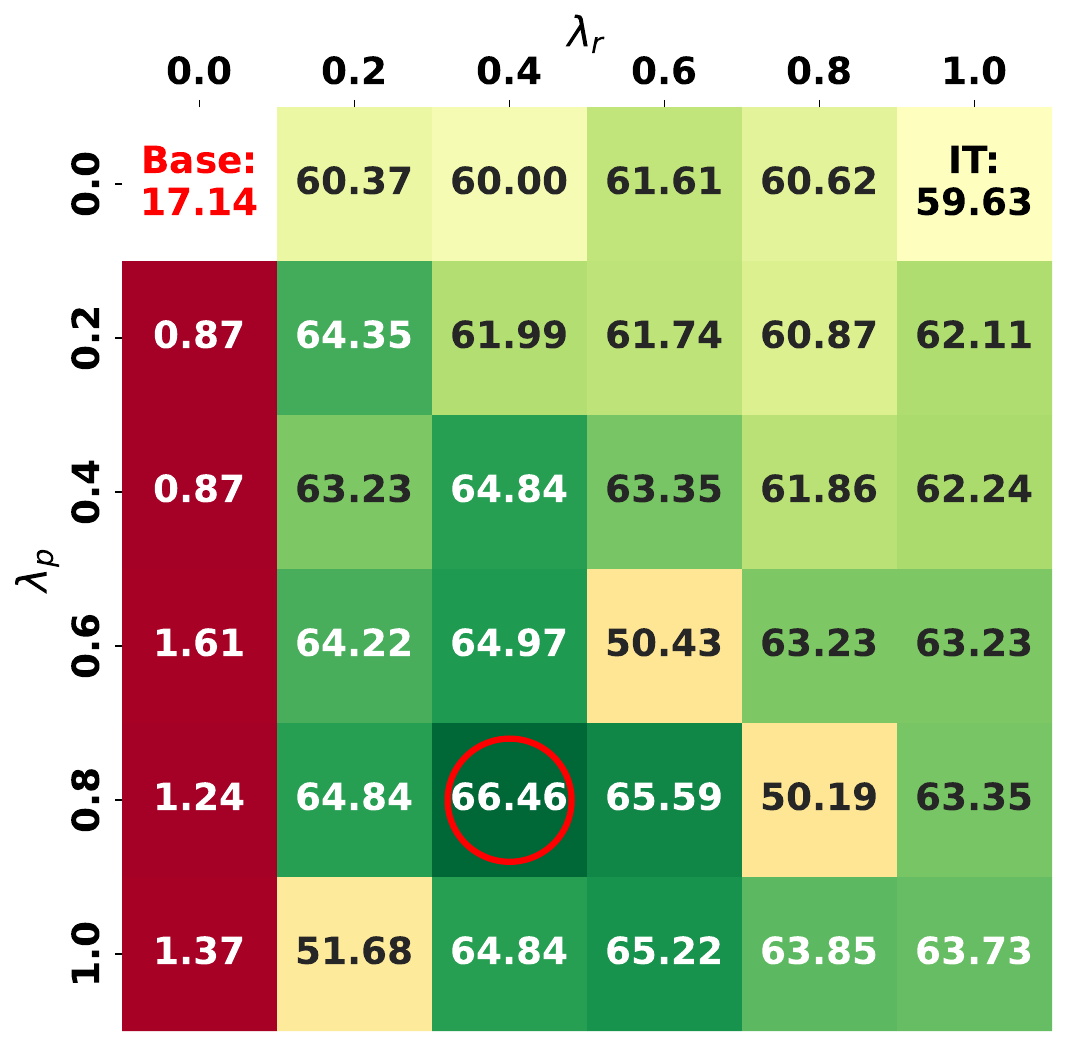} &
      \includegraphics[width=.19\textwidth]{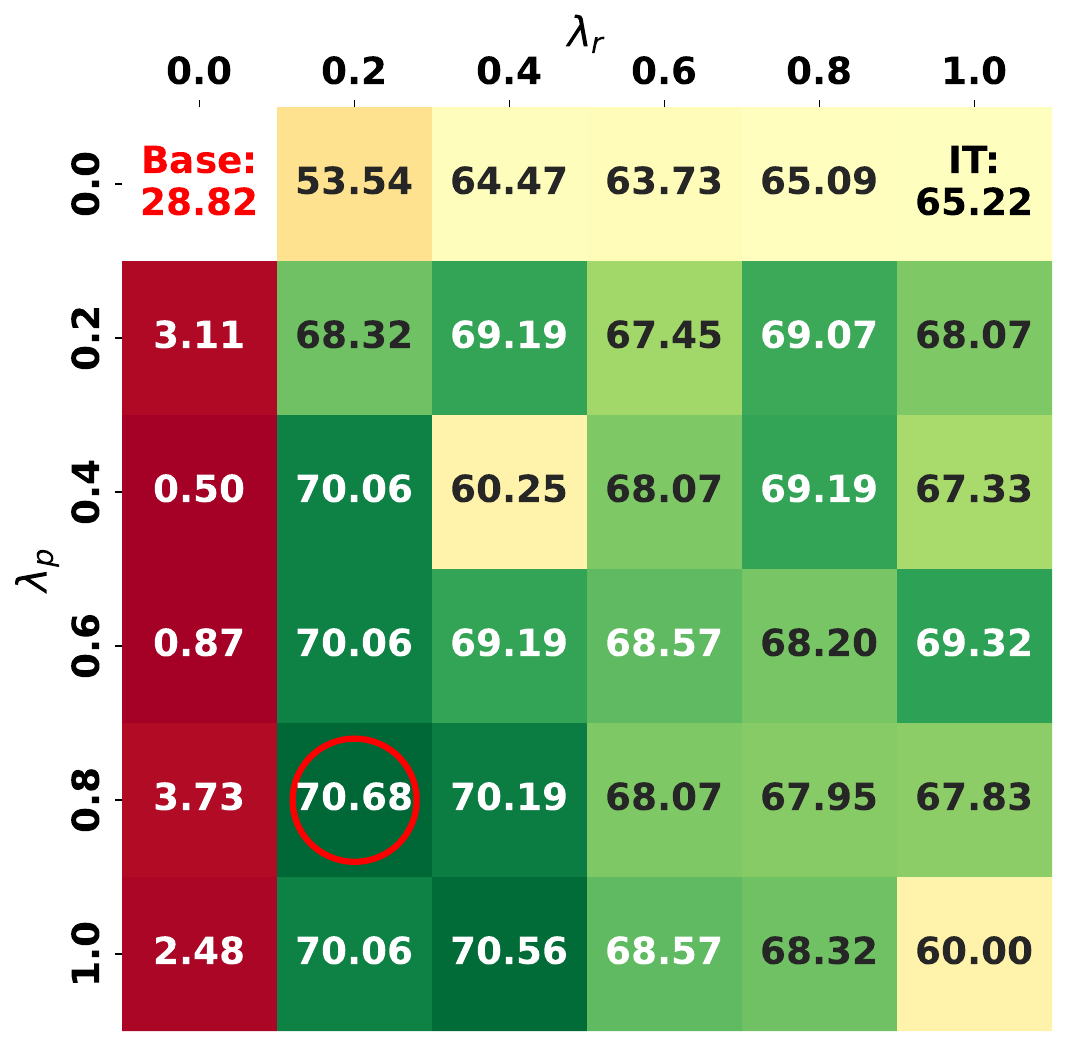} &
      \includegraphics[width=.19\textwidth]{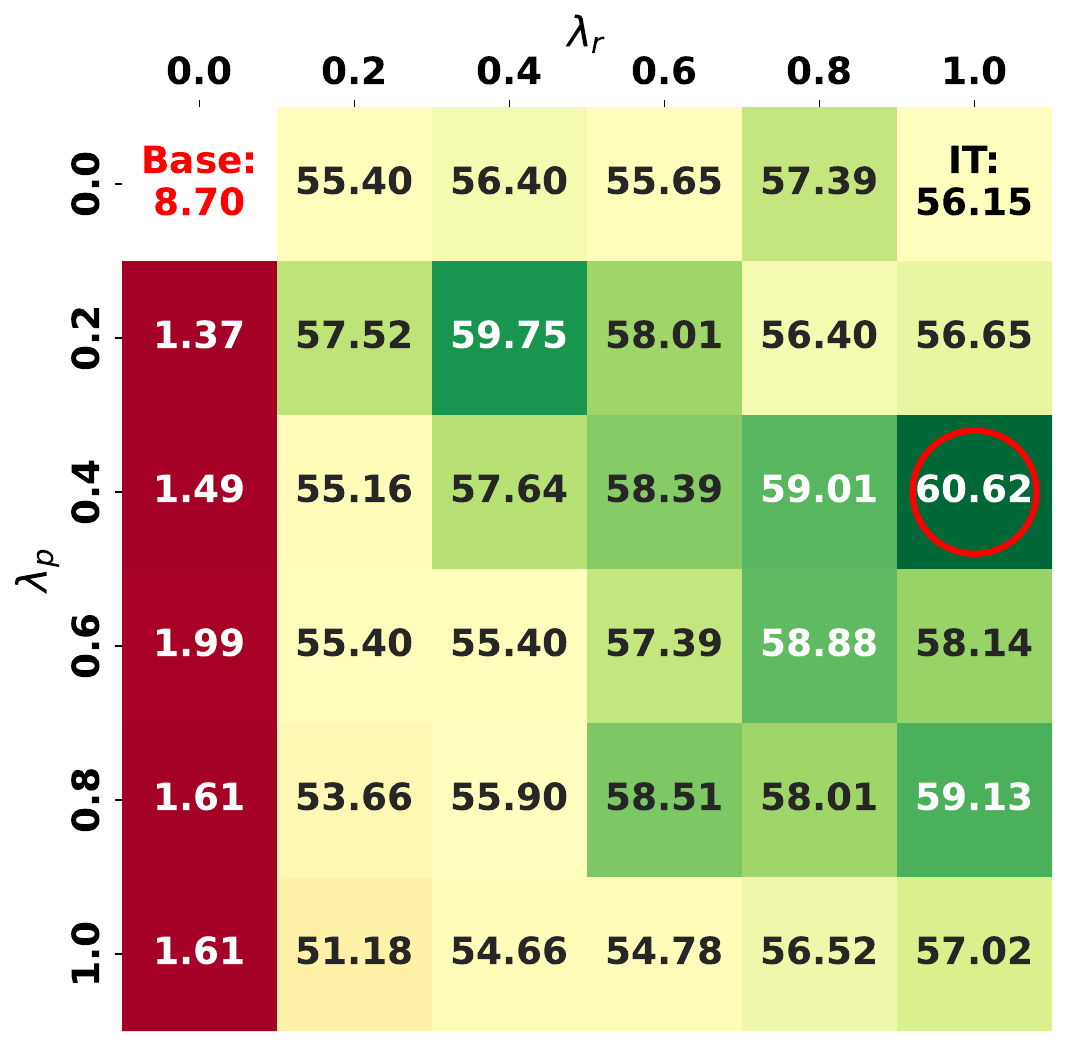} \\[4pt]

      \raisebox{1.3cm}[0pt][0pt]{\rotatebox{90}{\makebox[0pt][c]{IFEval}}} &
      \includegraphics[width=.19\textwidth]{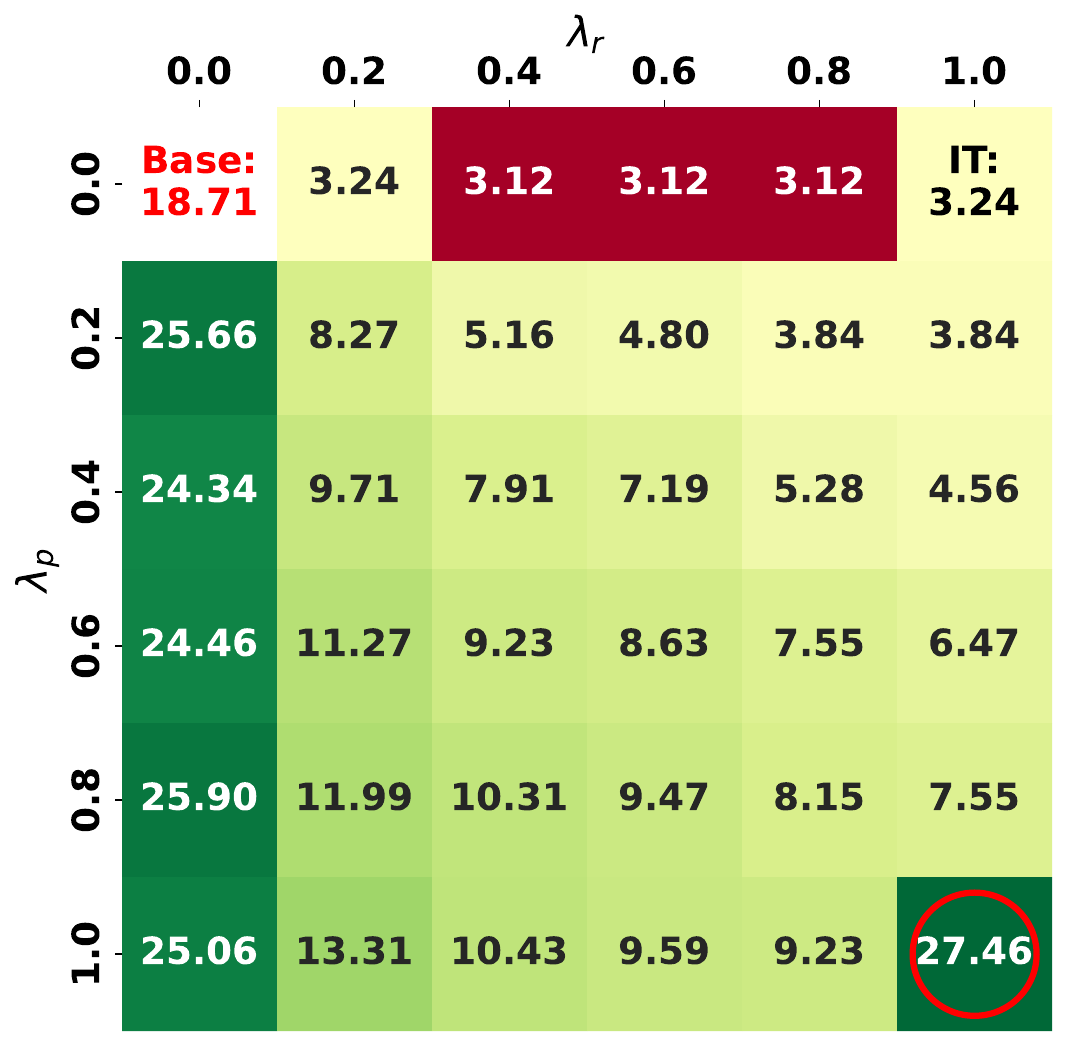} &
      \includegraphics[width=.19\textwidth]{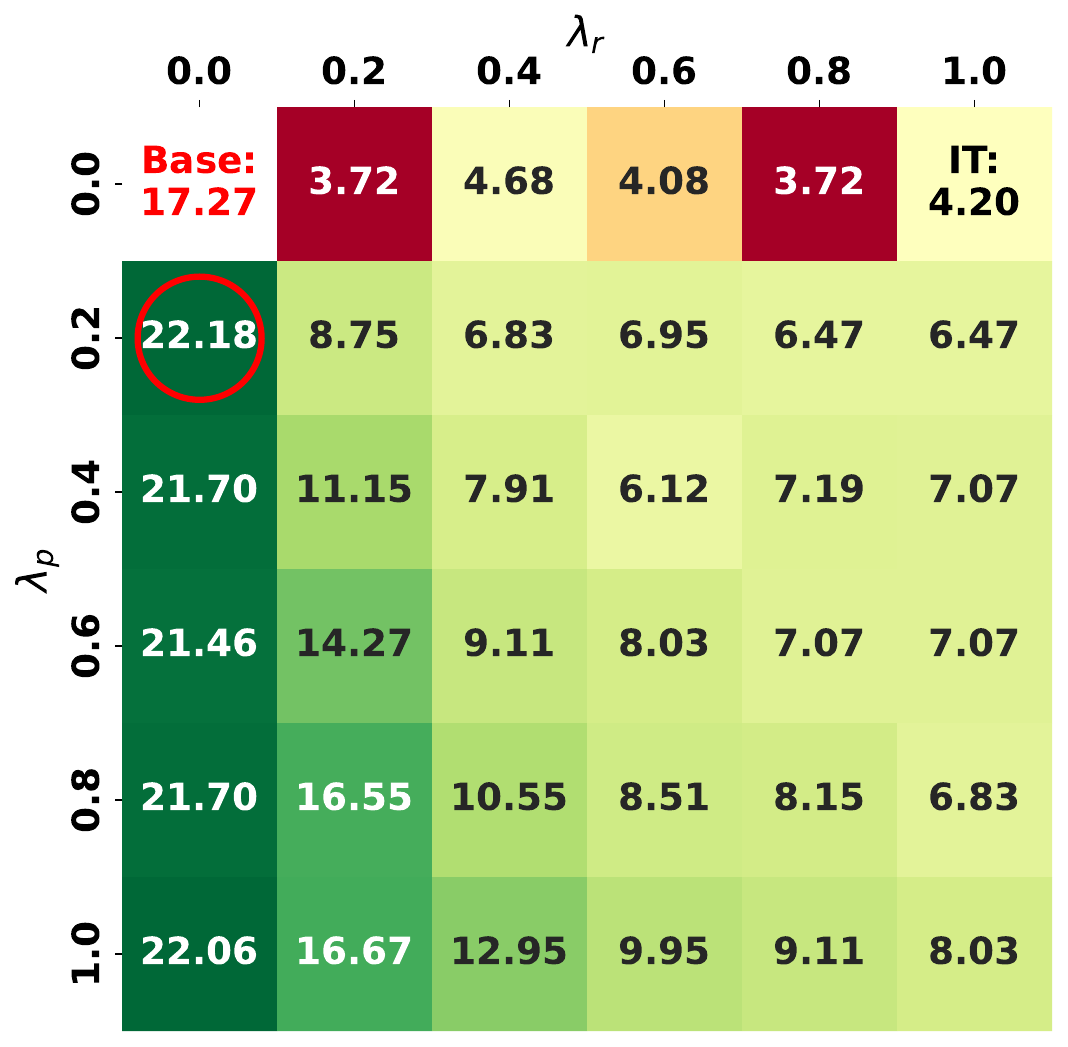} &
      \includegraphics[width=.19\textwidth]{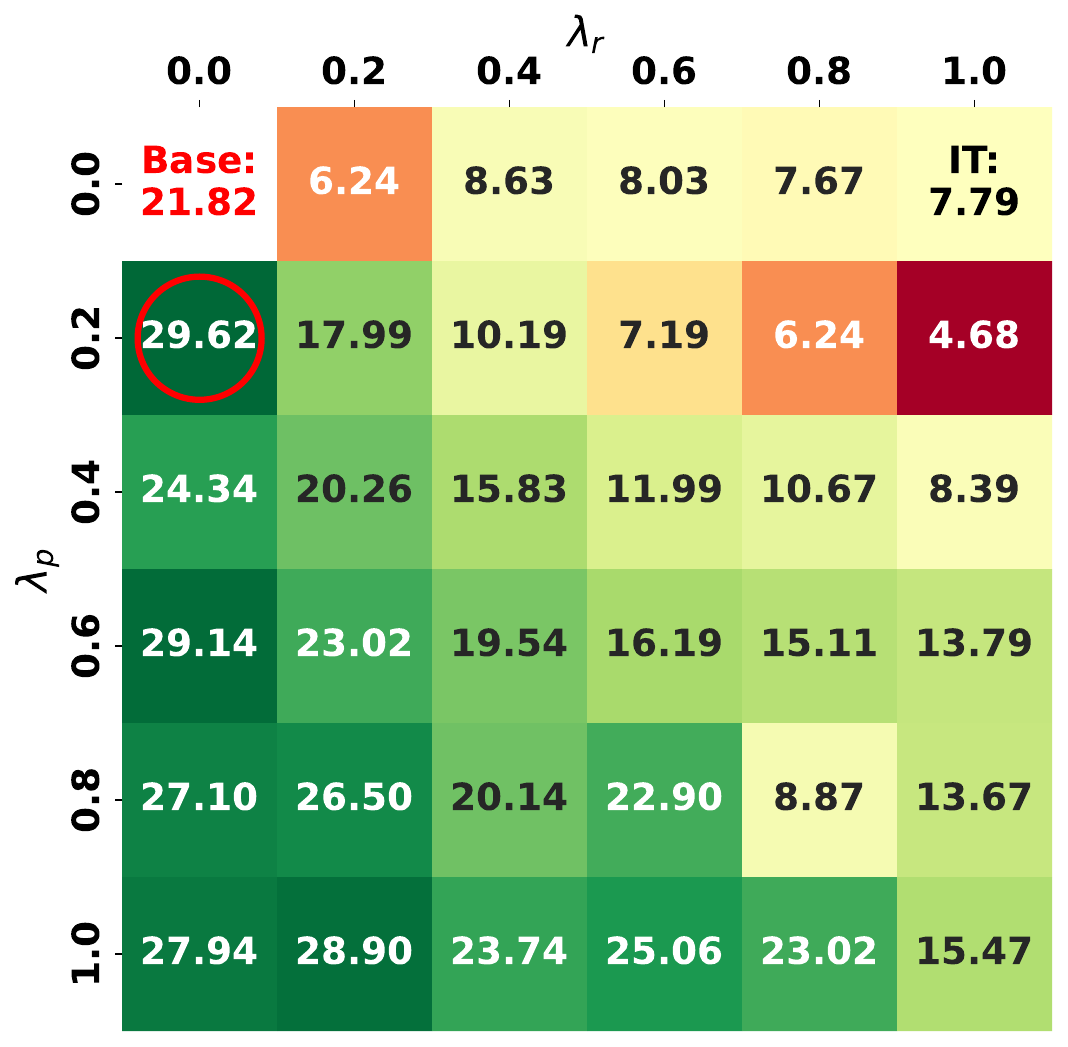} &
      \includegraphics[width=.19\textwidth]{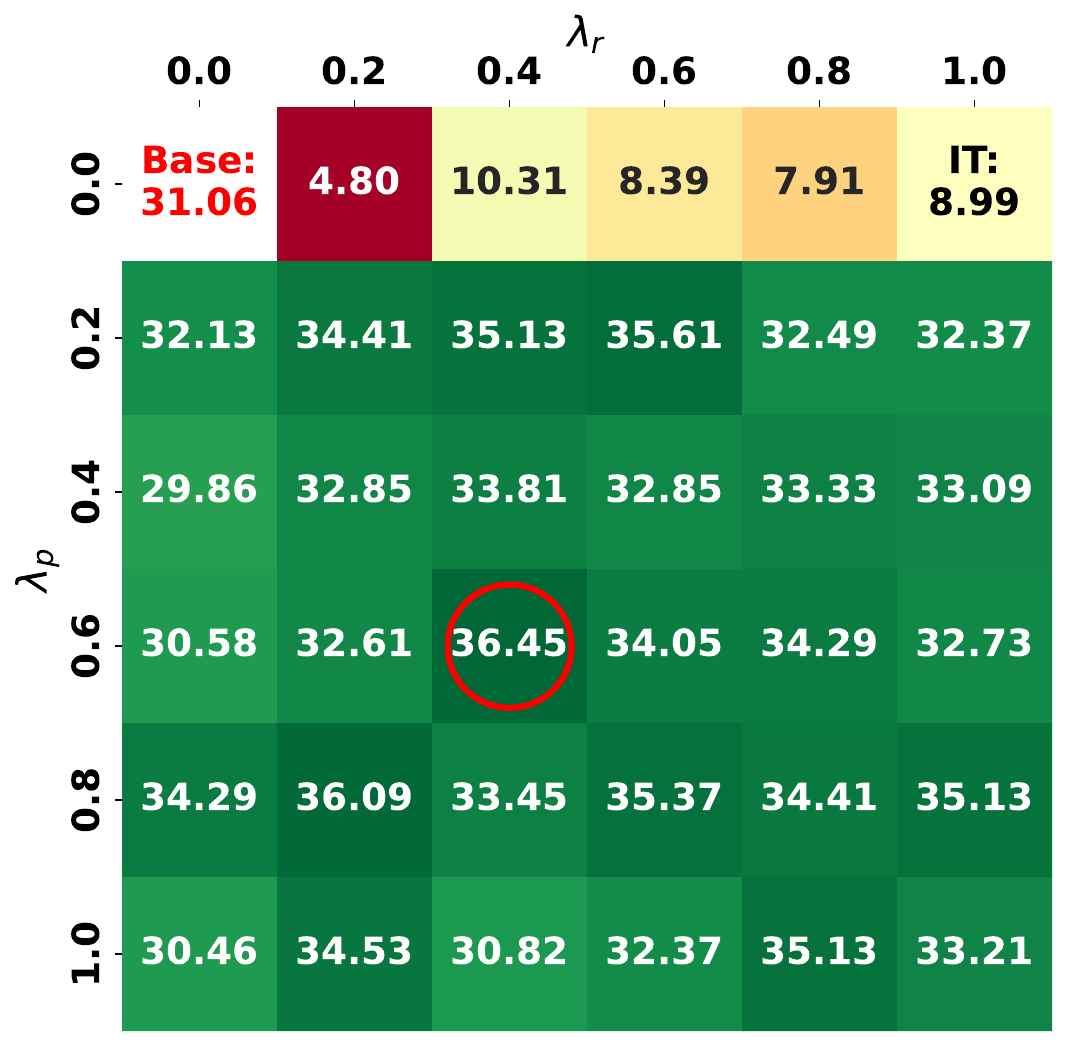} &
      \includegraphics[width=.19\textwidth]{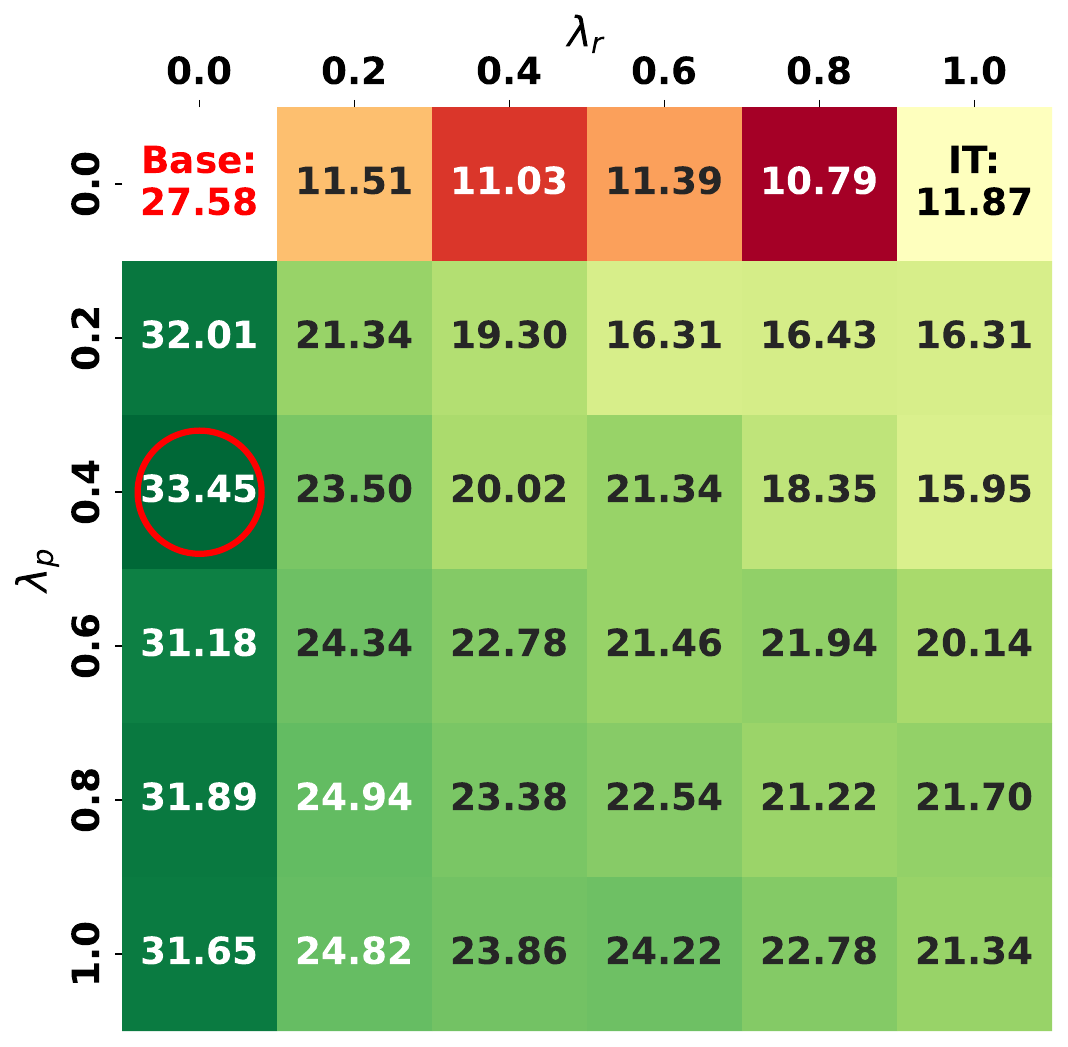} \\[4pt]

      \raisebox{1.3cm}[0pt][0pt]{\rotatebox{90}{\makebox[0pt][c]{MT-Bench}}} &
      \includegraphics[width=.19\textwidth]{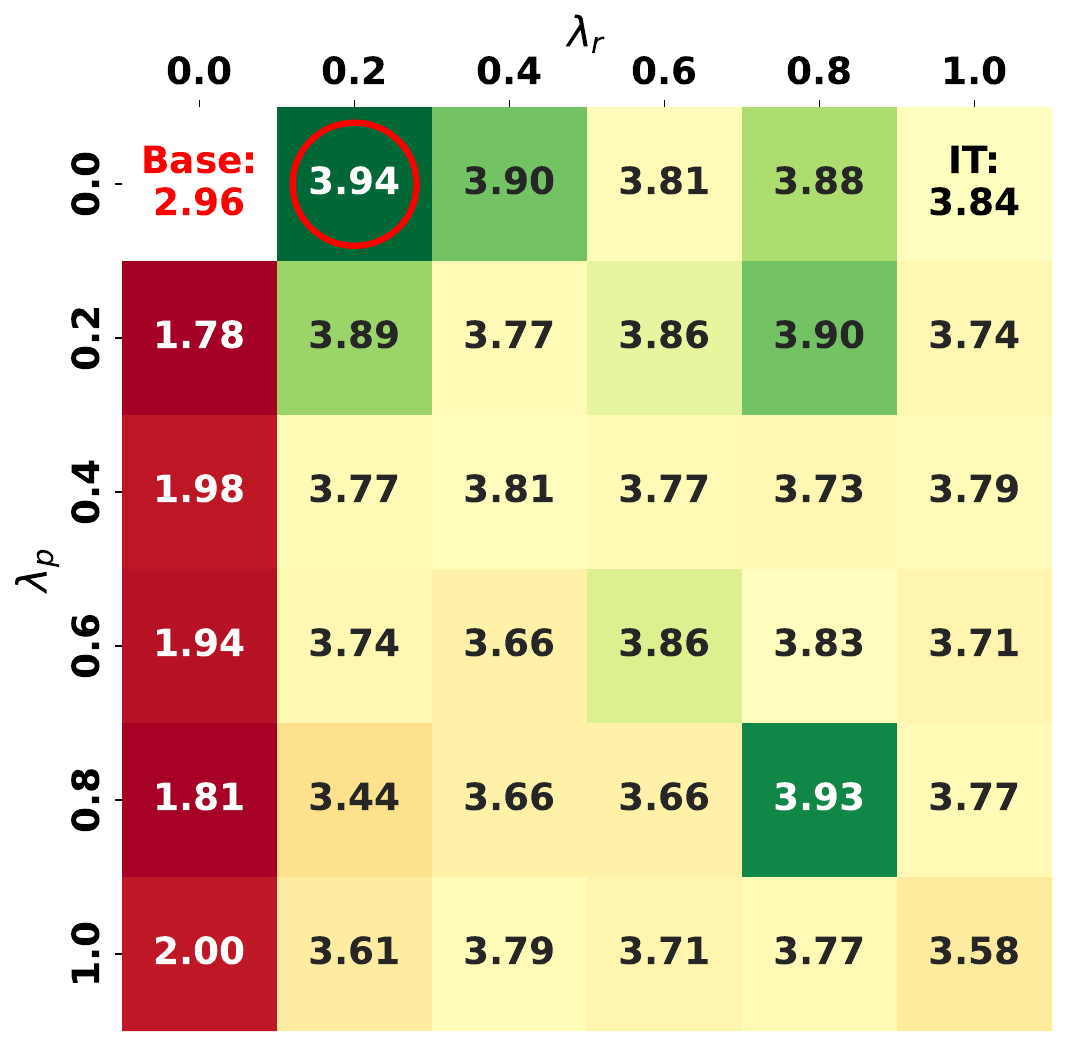} &
      \includegraphics[width=.19\textwidth]{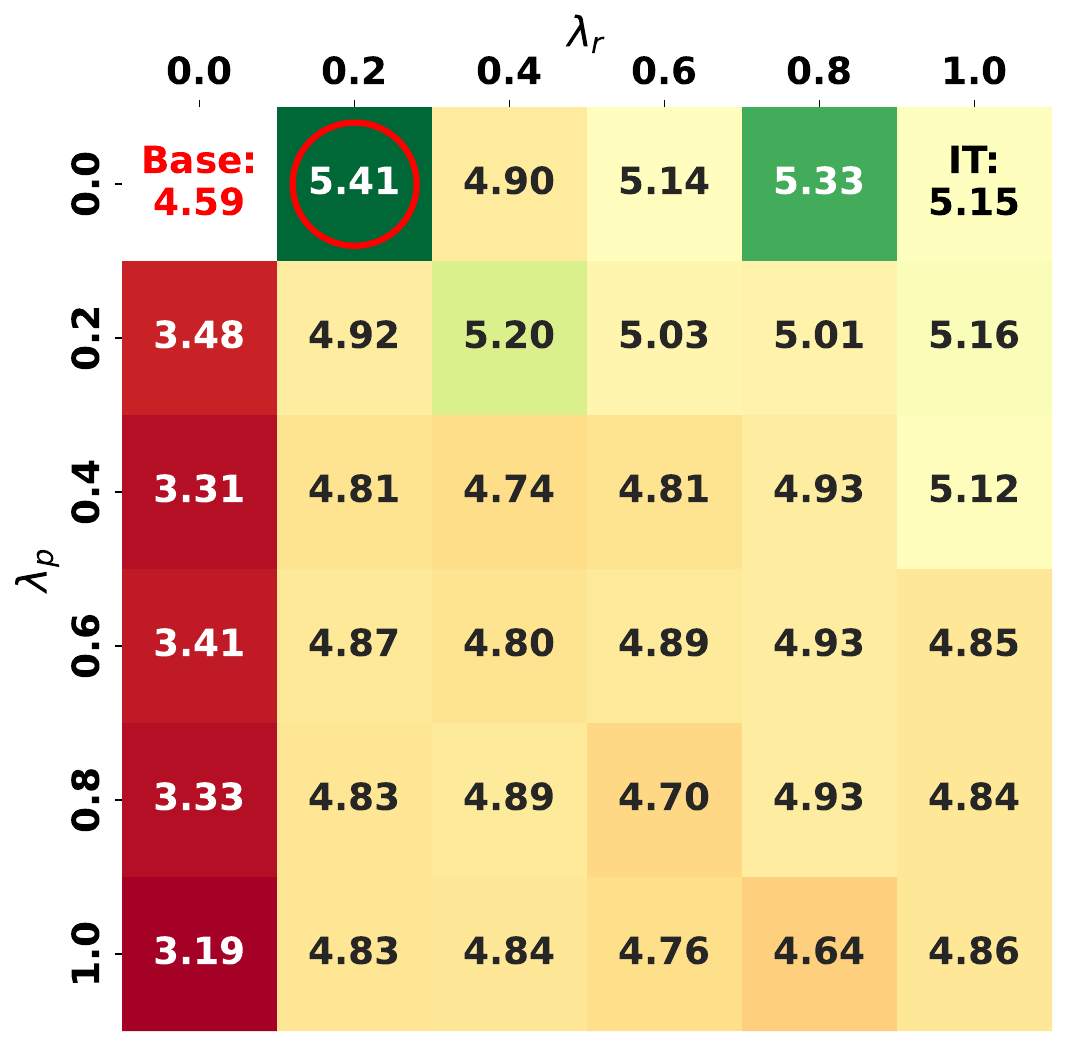} &
      \includegraphics[width=.19\textwidth]{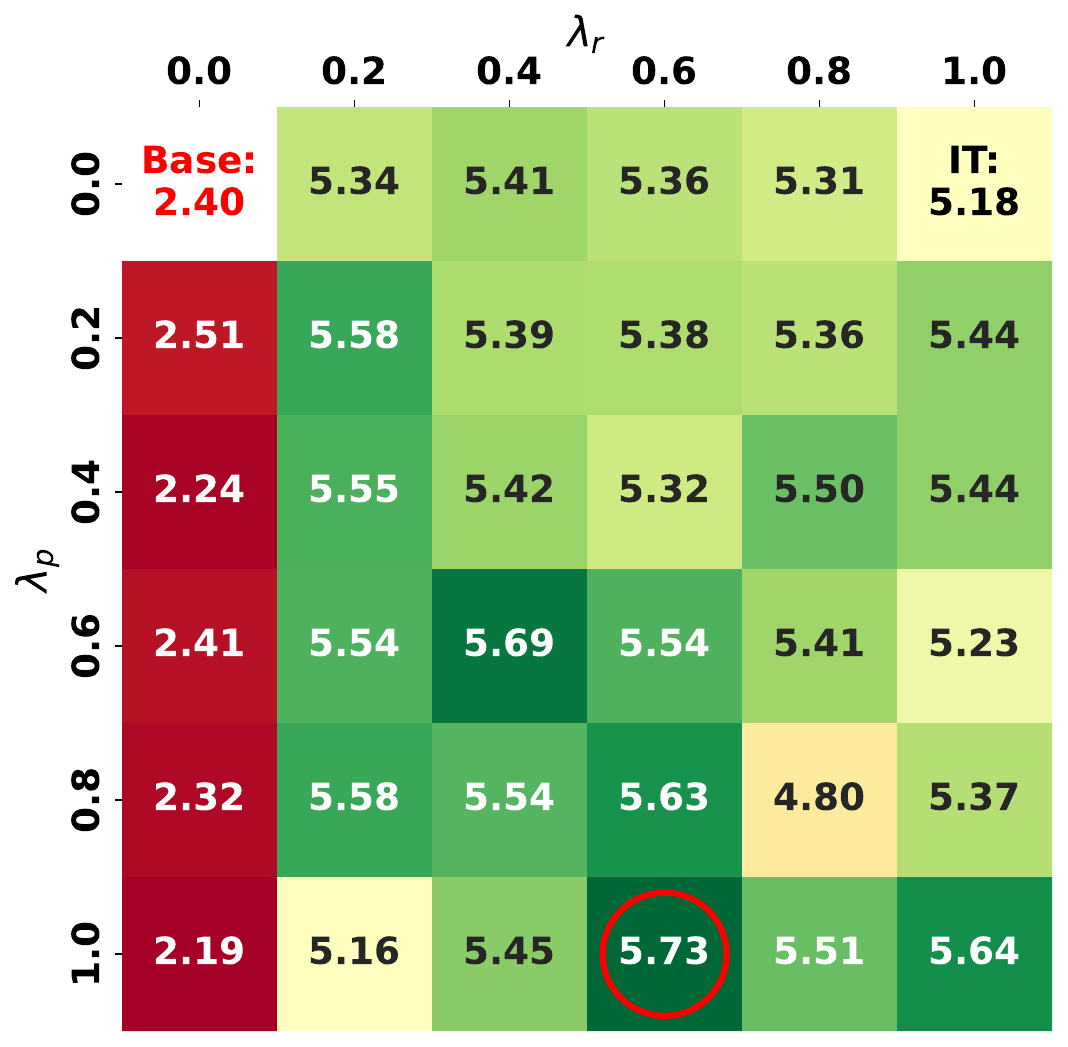} &
      \includegraphics[width=.19\textwidth]{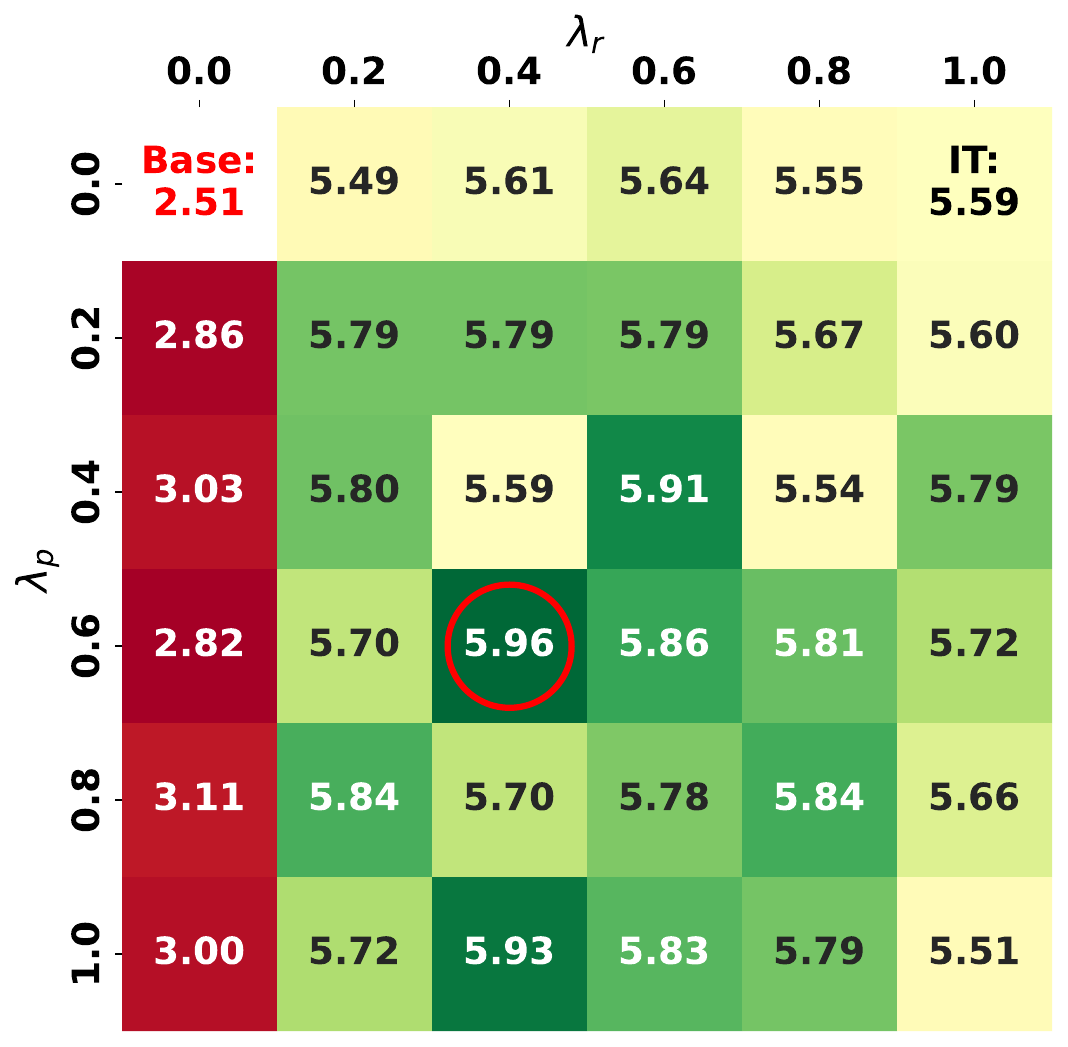} &
      \includegraphics[width=.19\textwidth]{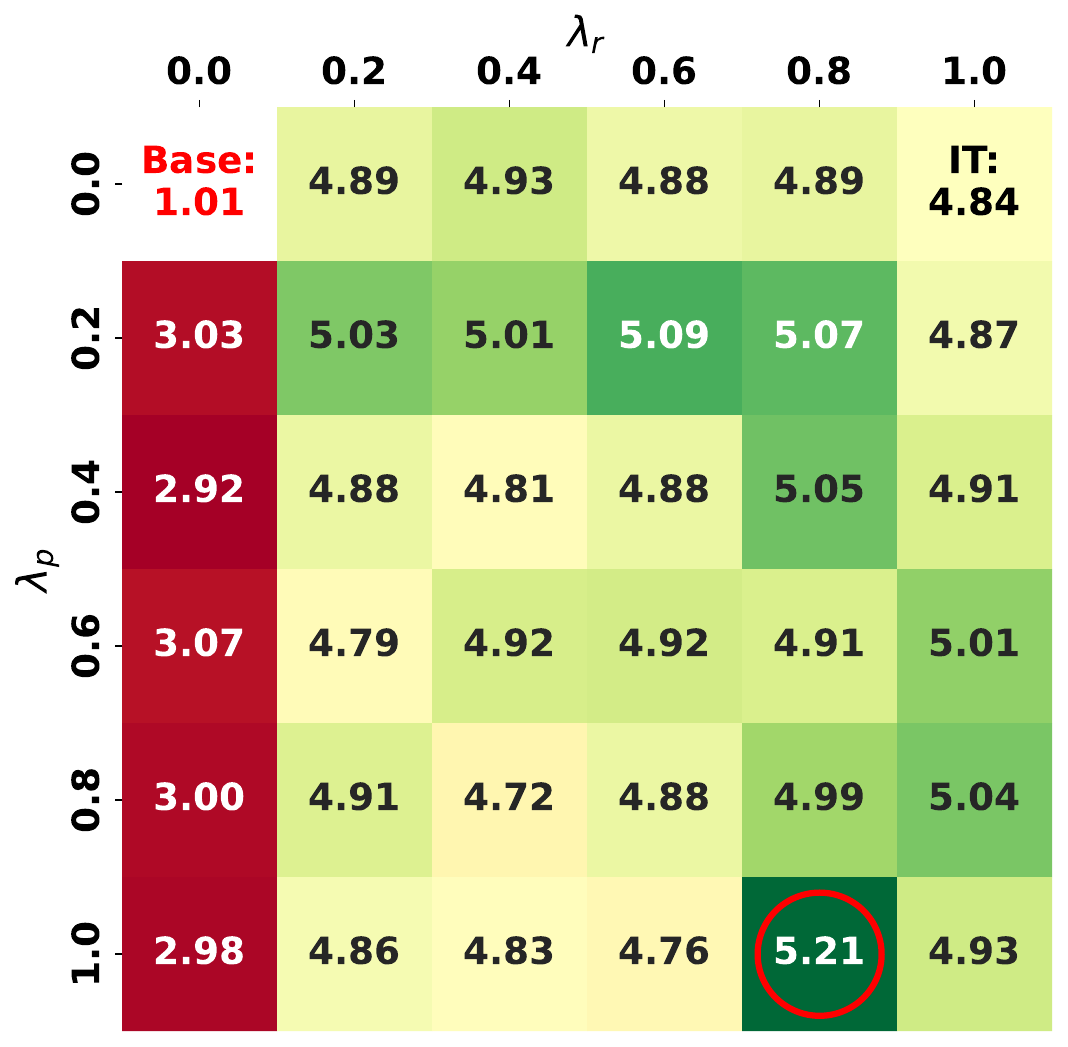} \\
  \end{tabular}

    \caption{Heatmaps depicting performance on MMLU (first row), BBH (second row), AlpacaEval (third row), IFEval (fourth row), and MT-Bench (fifth row) for different configurations of ($\lambda_p$, $\lambda_r$) and for different models finetuned on \textbf{Alpaca-Cleaned}. In each heatmap, the best performance is highlighted with a red circle. The color map is based on relative gain with respect to conventional instruction tuning. Each row of a heatmap corresponds to a prompt-token weight, and each column corresponds to a response-token weight. Conventional instruction tuning is marked with \texttt{IT}, and base model performance is marked with \texttt{Base}.}
    \label{fig:alpaca-cleaned_all}
\end{figure*}

\begin{figure*}[ht!]
  \centering
  \setlength{\tabcolsep}{2pt}      
  \renewcommand{\arraystretch}{1}  

  \begin{tabular}{@{} >{\centering\arraybackslash}m{2.1mm} *{5}{c} @{}}
      & Llama-3-1B & Llama-3-3B & Llama-3-8B & Mistral-7B & Gemma-2-2B \\[2pt]

      \raisebox{1.3cm}[0pt][0pt]{\rotatebox{90}{\makebox[0pt][c]{MMLU}}} &
      \includegraphics[width=.19\textwidth]{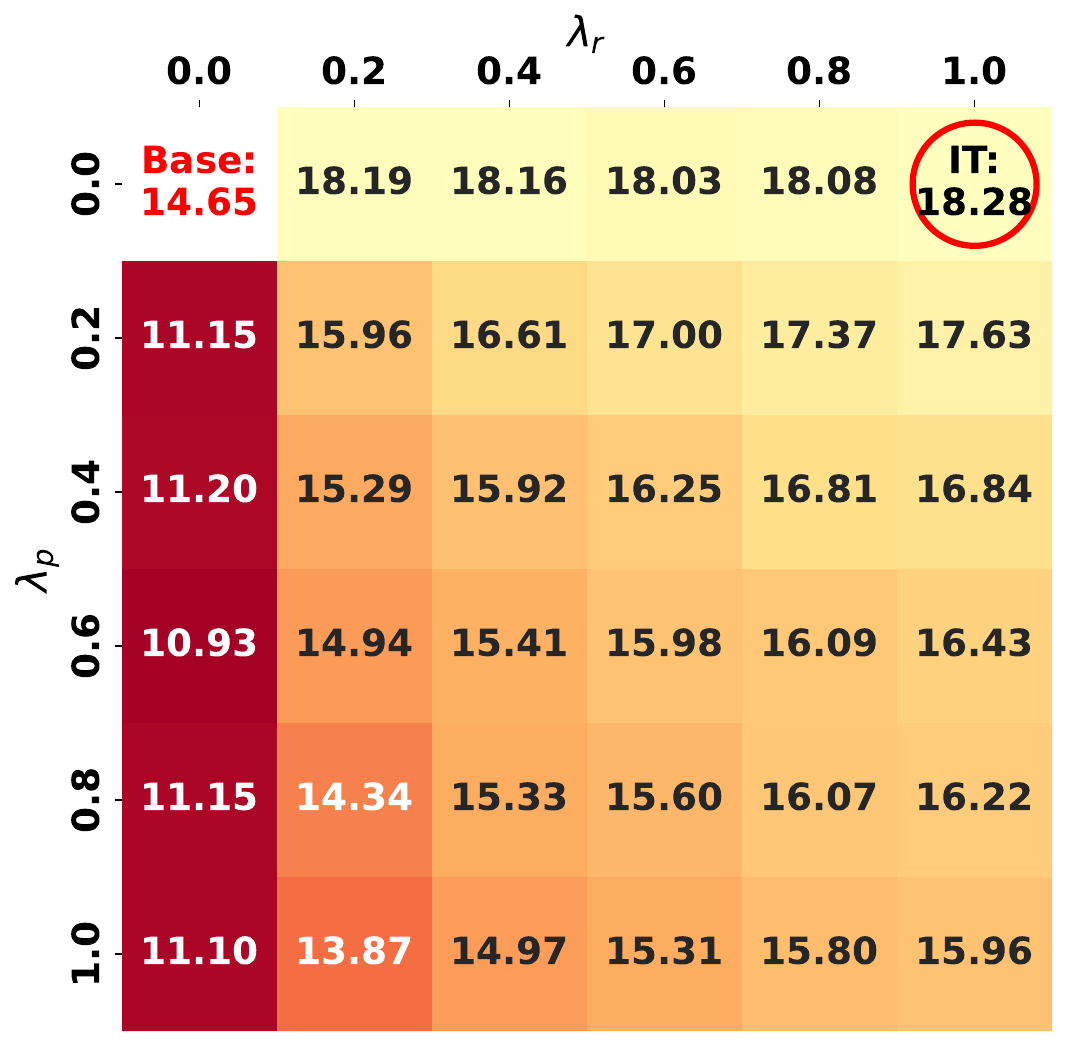} &
      \includegraphics[width=.19\textwidth]{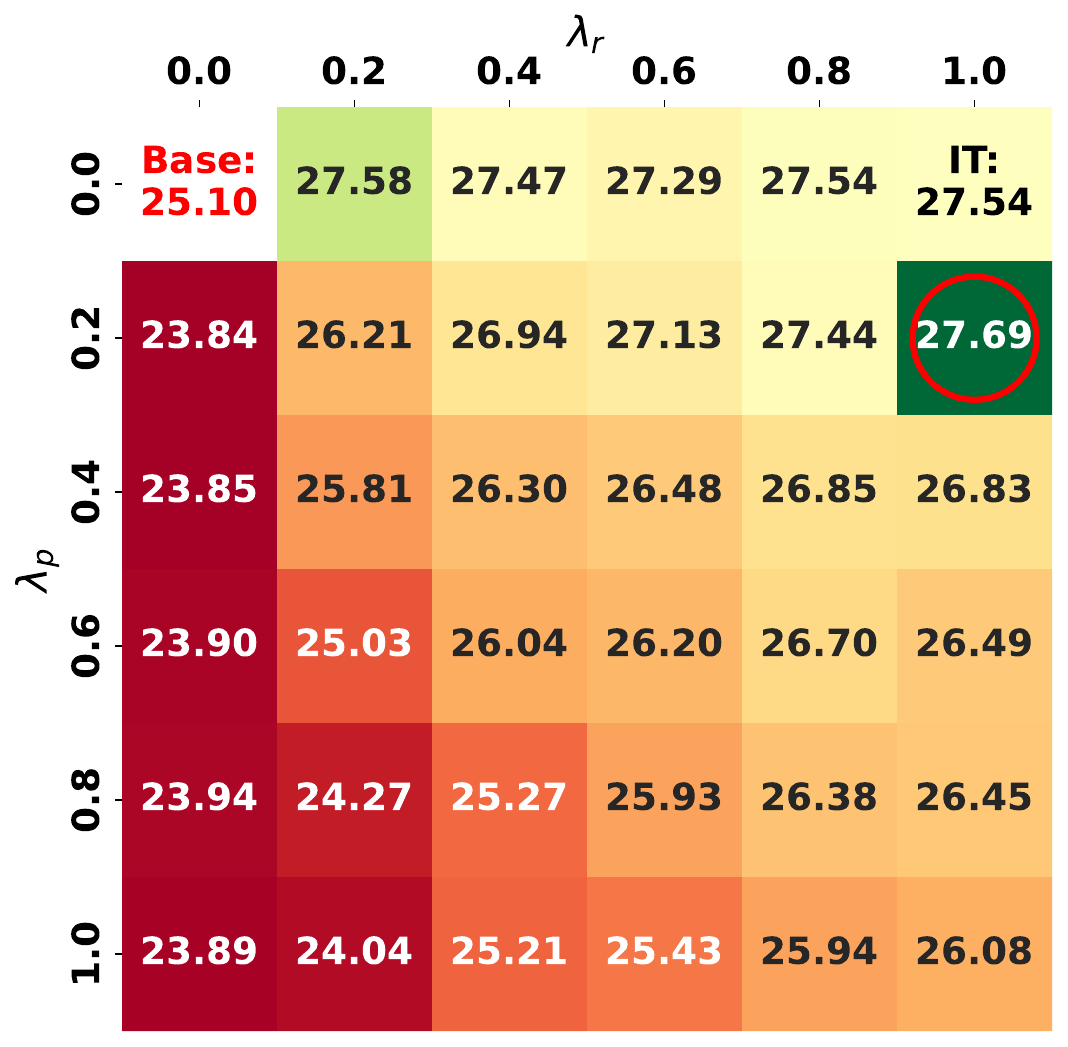} &
      \includegraphics[width=.19\textwidth]{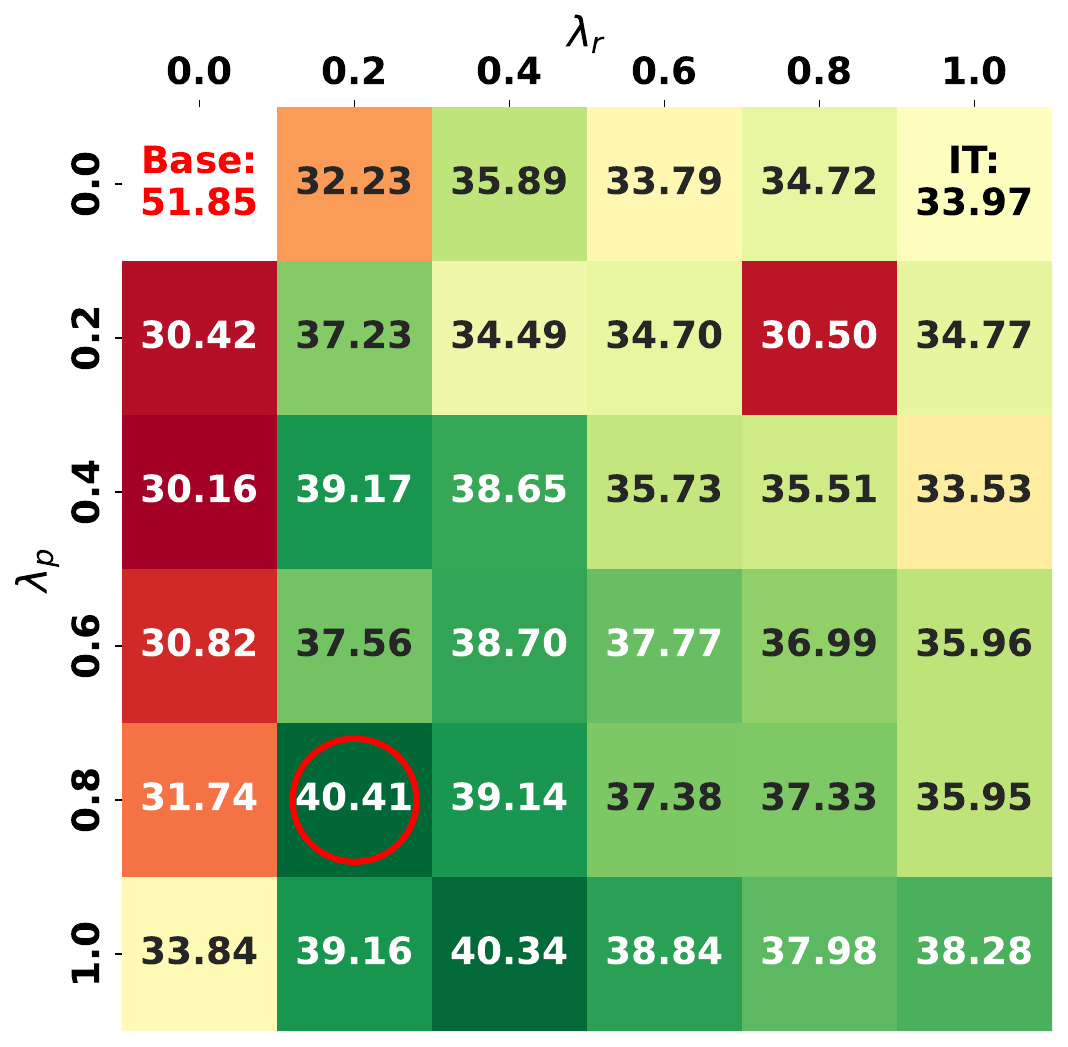} &
      \includegraphics[width=.19\textwidth]{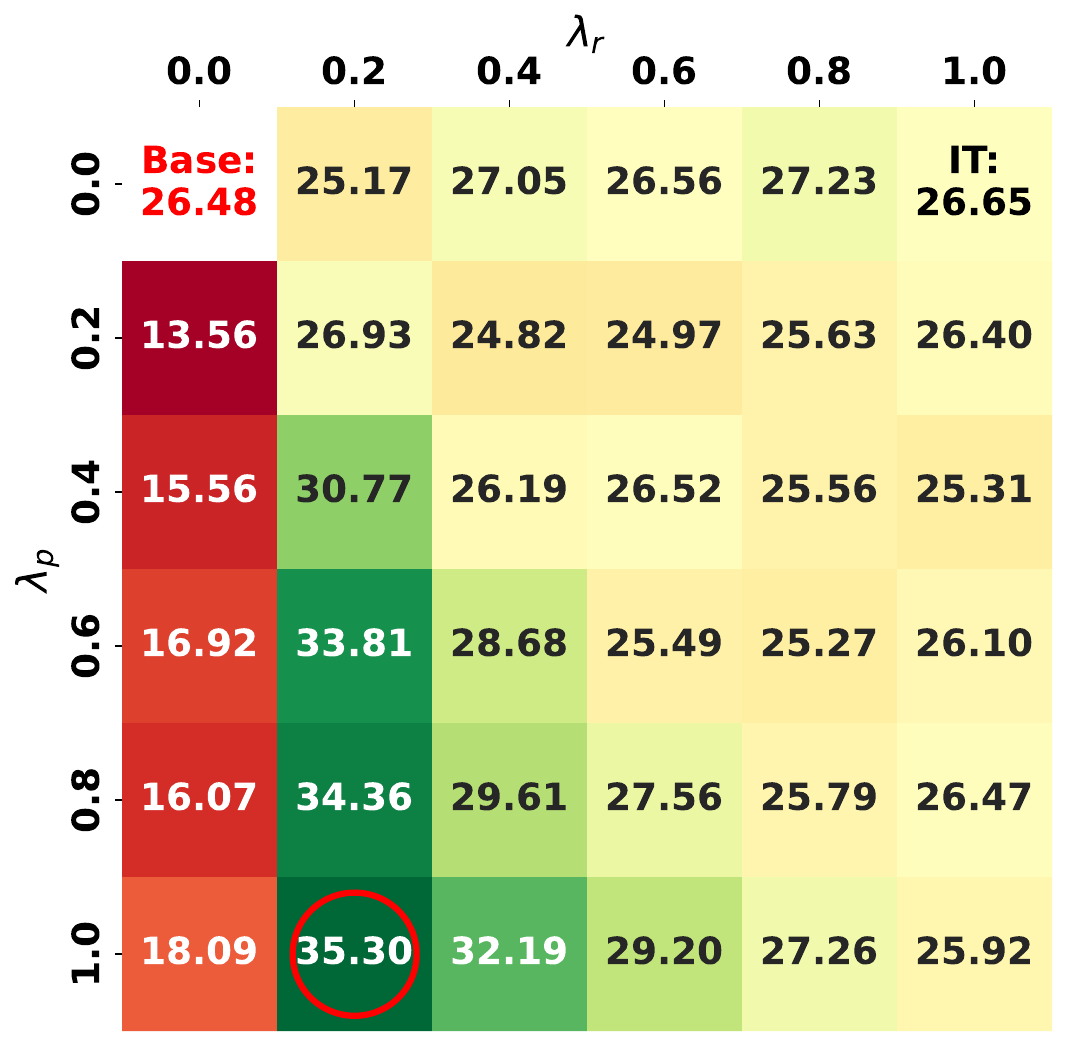} &
      \includegraphics[width=.19\textwidth]{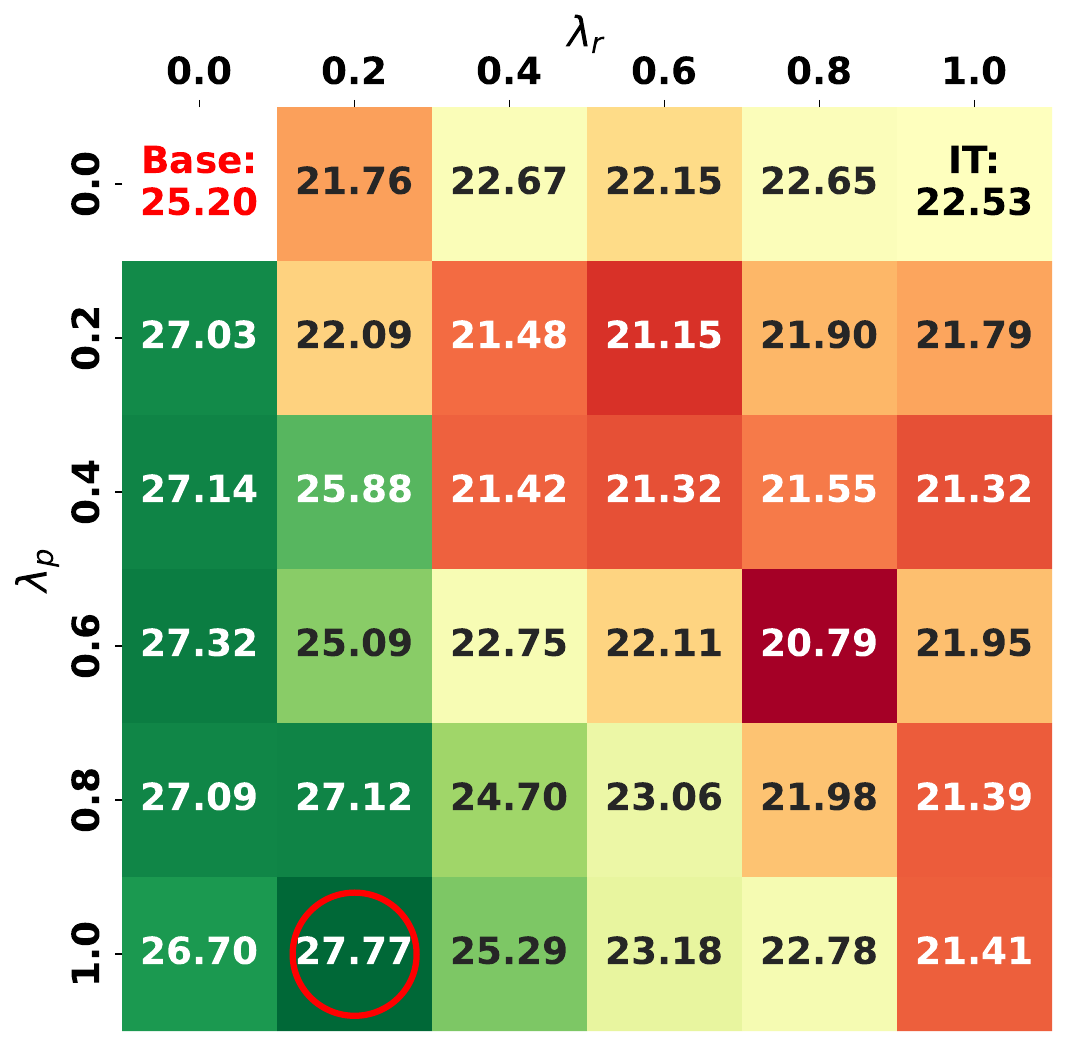} \\[4pt]

        \raisebox{1.3cm}[0pt][0pt]{\rotatebox{90}{\makebox[0pt][c]{BBH}}} &
      \includegraphics[width=.19\textwidth]{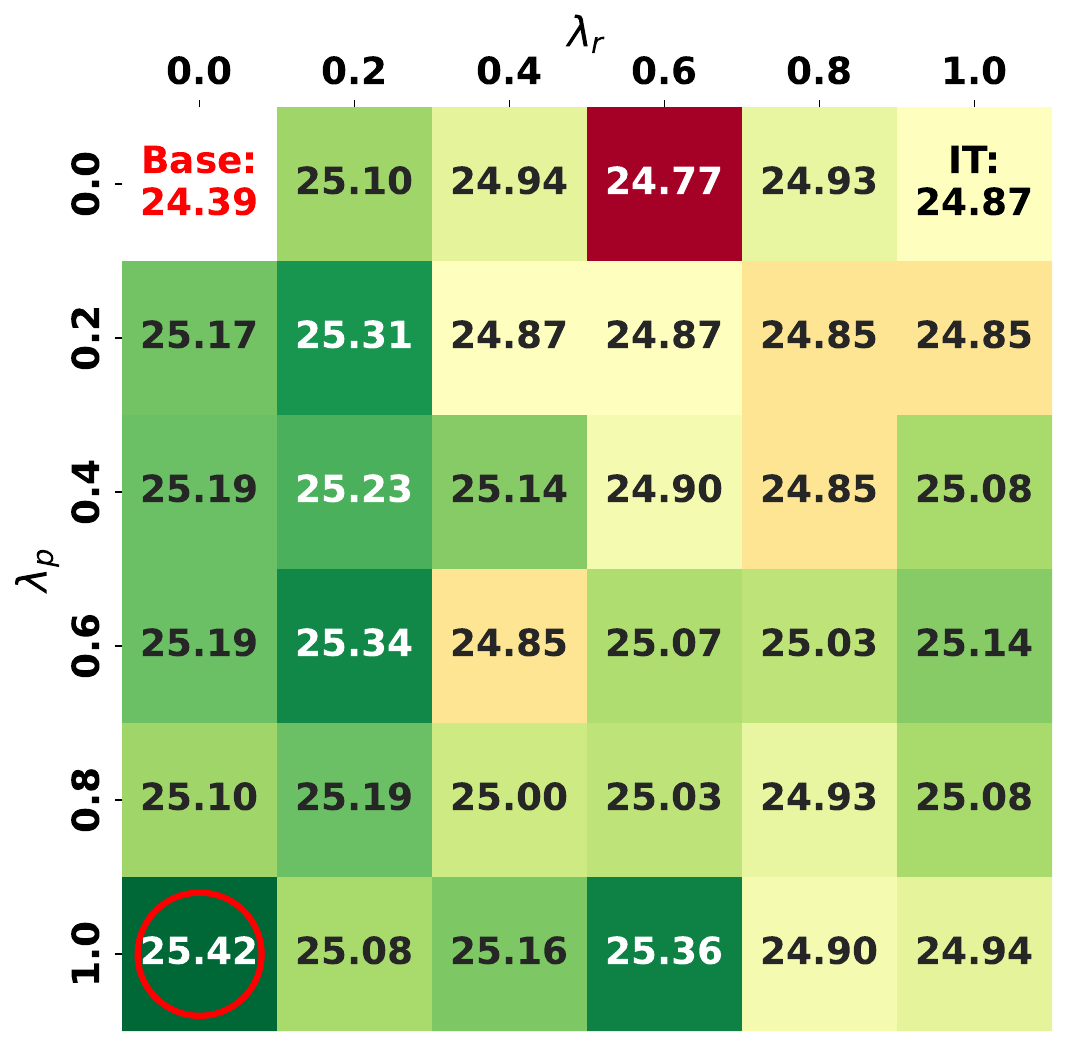} &
      \includegraphics[width=.19\textwidth]{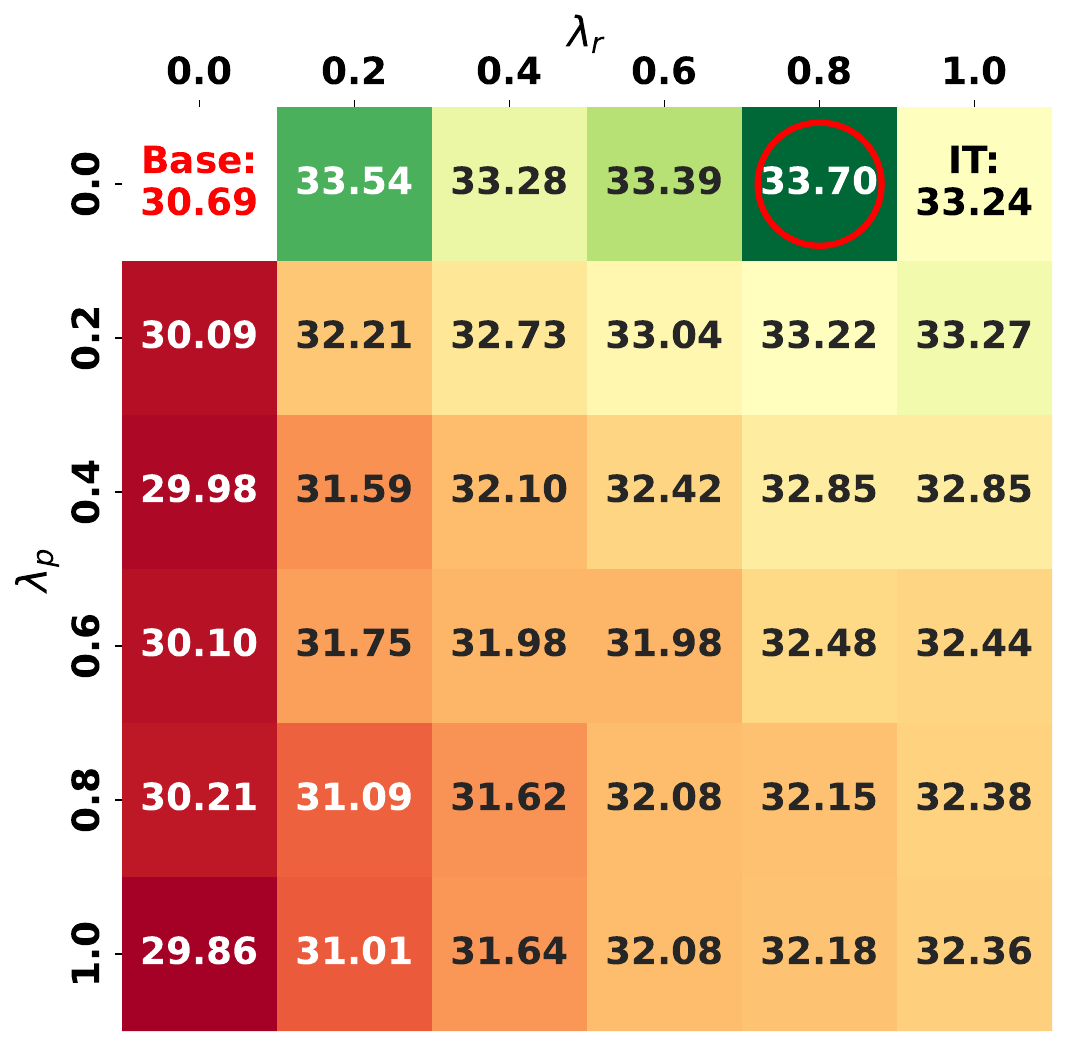} &
      \includegraphics[width=.19\textwidth]{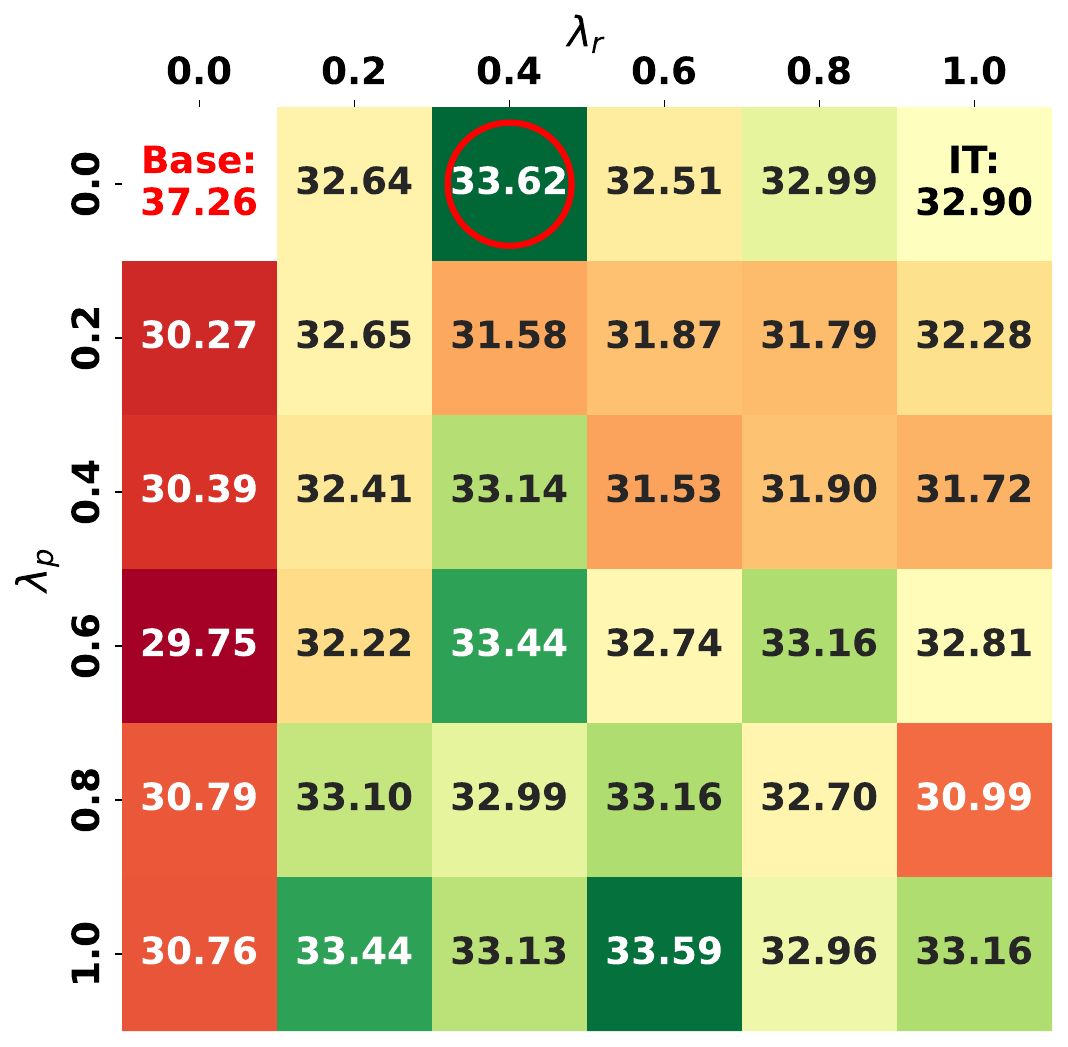} &
      \includegraphics[width=.19\textwidth]{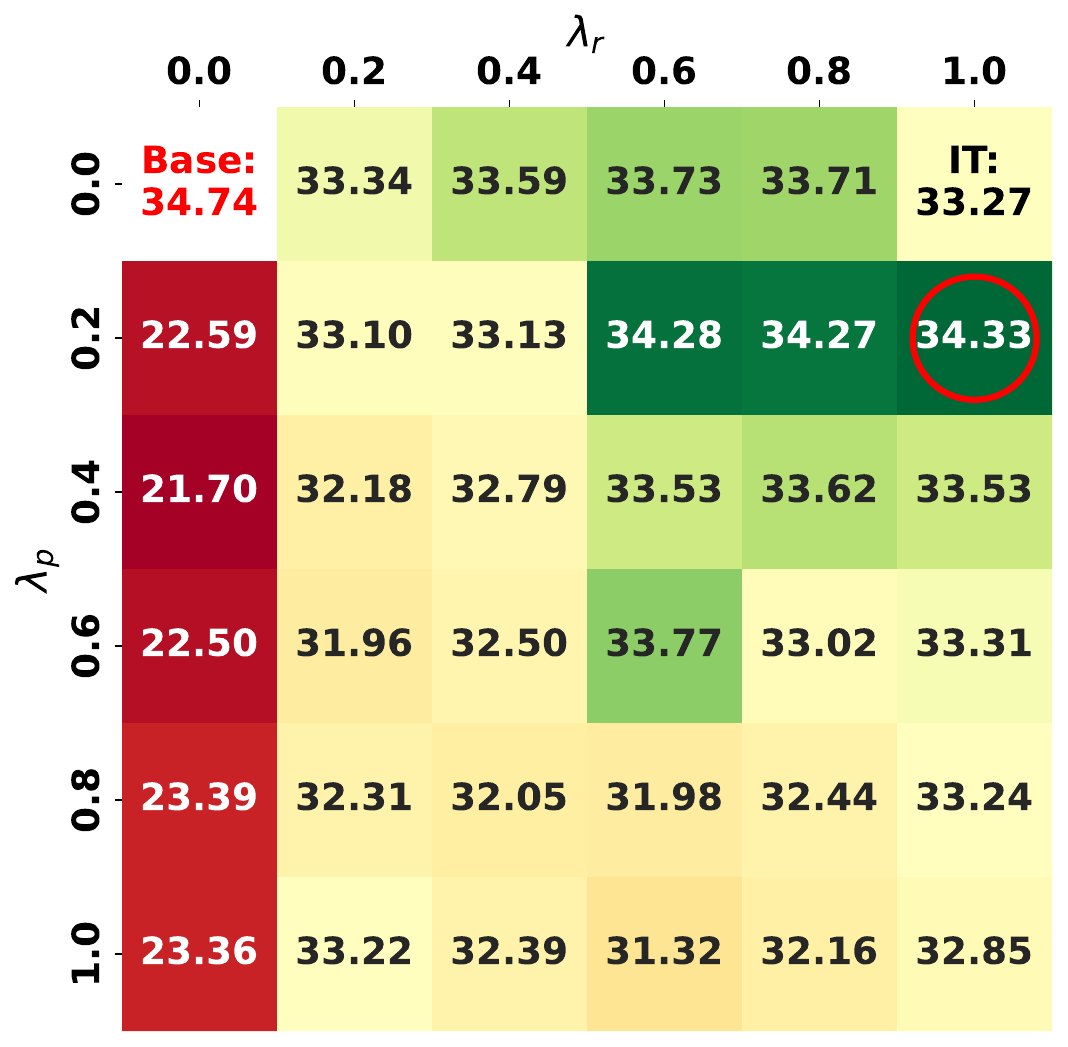} &
      \includegraphics[width=.19\textwidth]{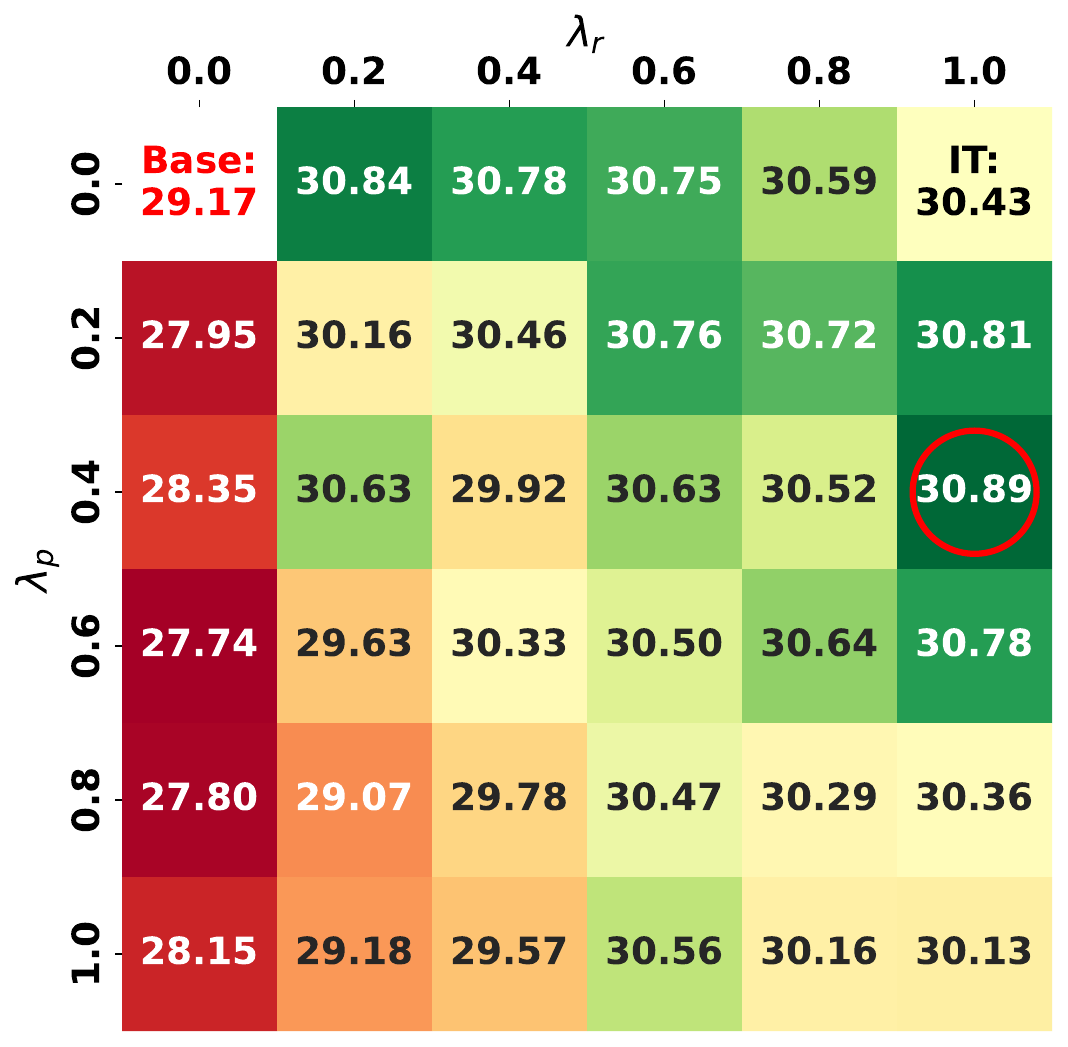} \\[4pt]

      \raisebox{1.3cm}[0pt][0pt]{\rotatebox{90}{\makebox[0pt][c]{AlpacaEval}}} &
      \includegraphics[width=.19\textwidth]{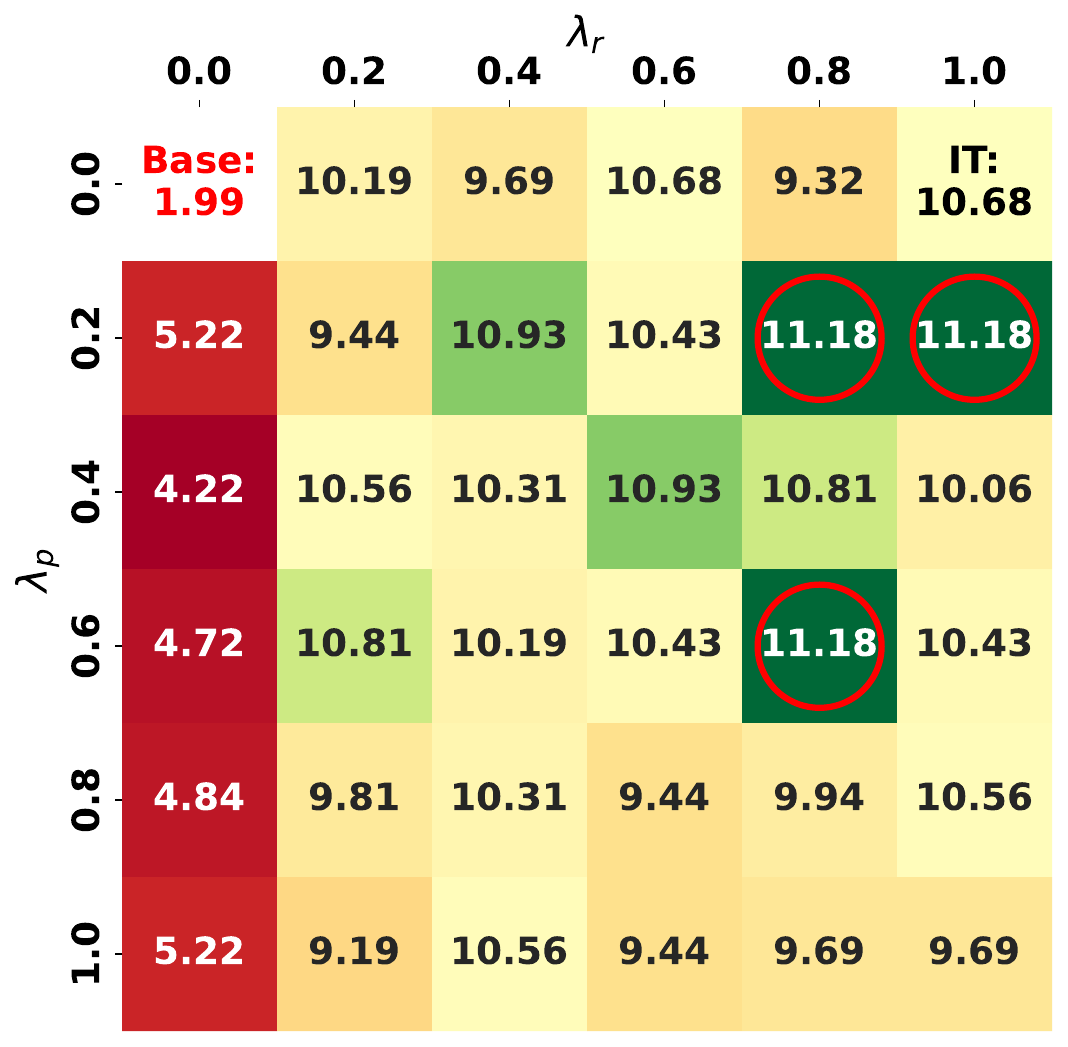} &
      \includegraphics[width=.19\textwidth]{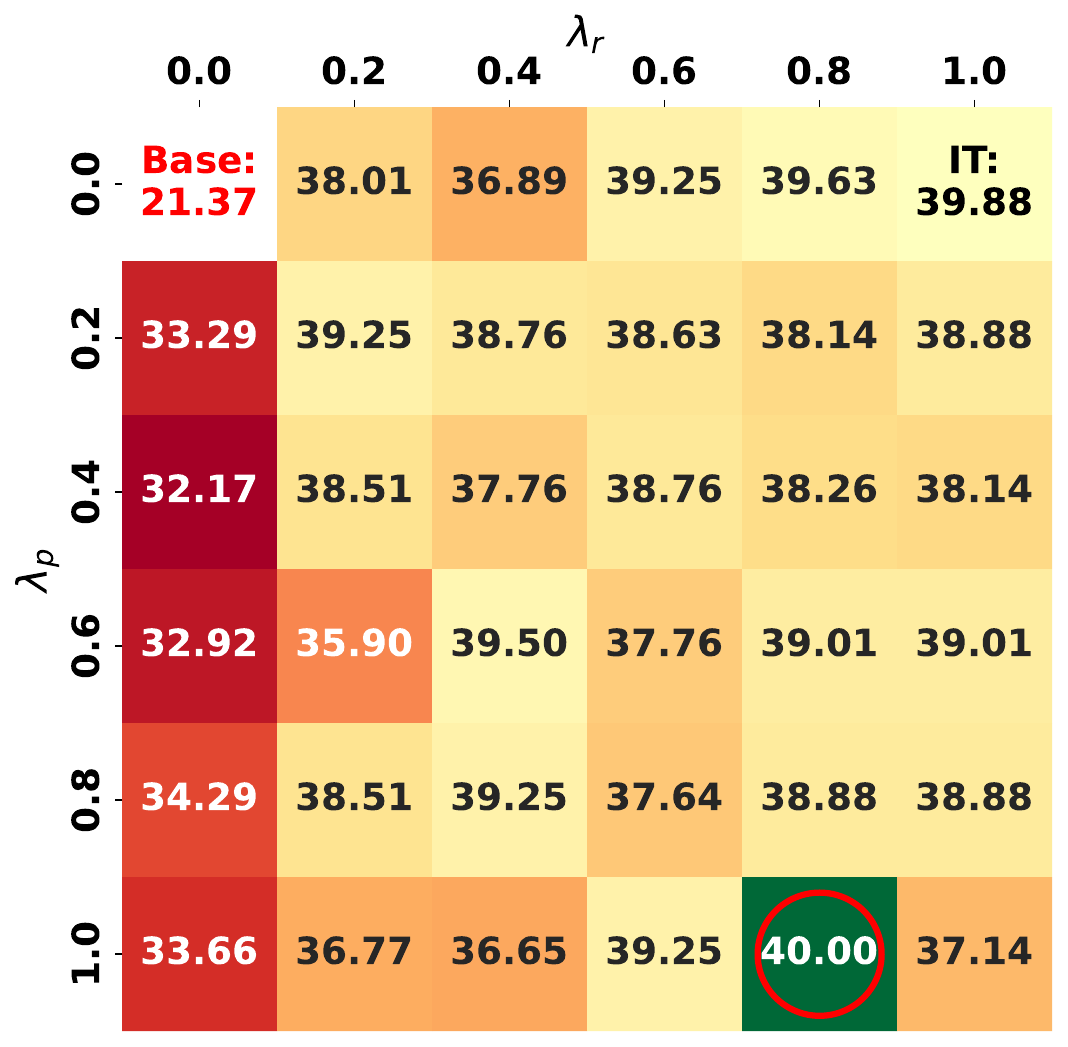} &
      \includegraphics[width=.19\textwidth]{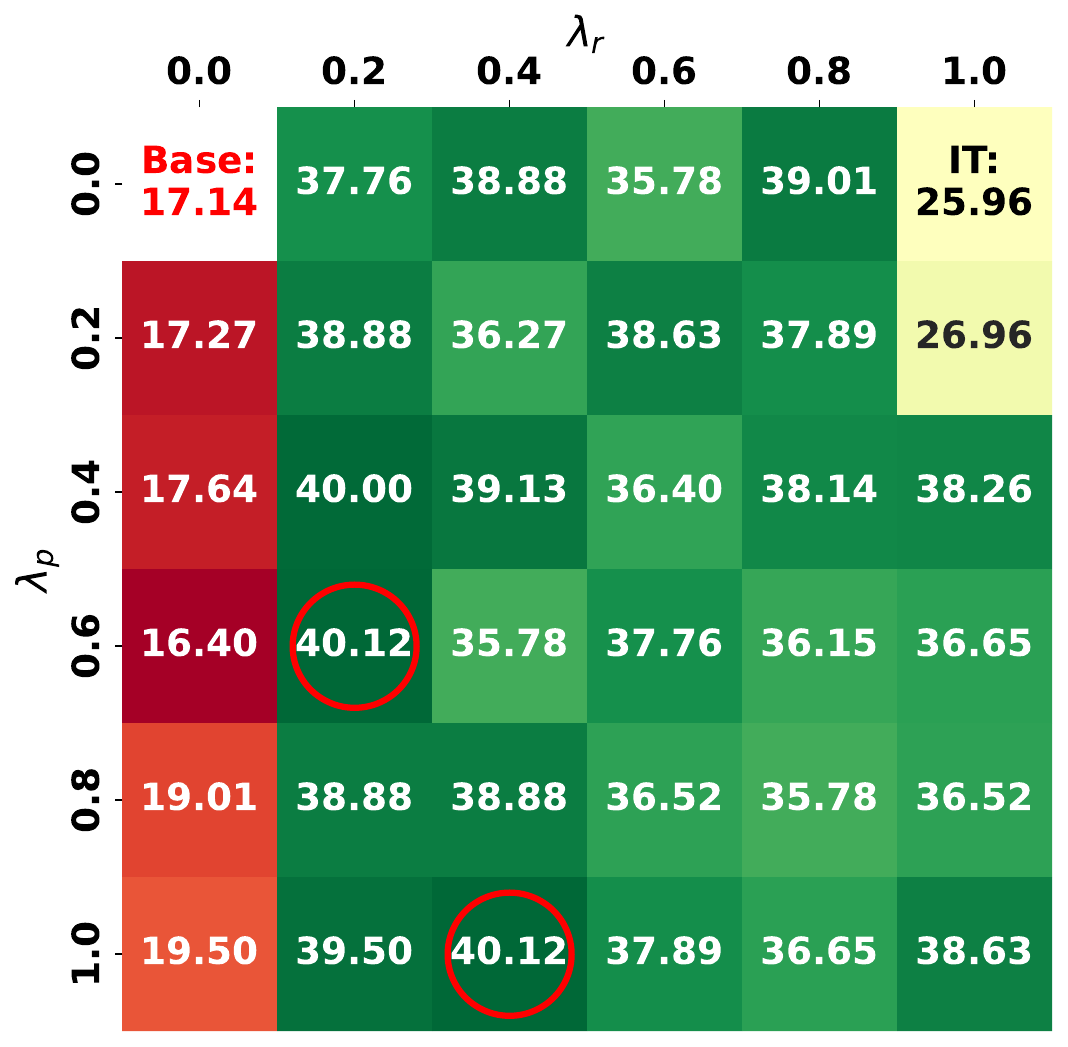} &
      \includegraphics[width=.19\textwidth]{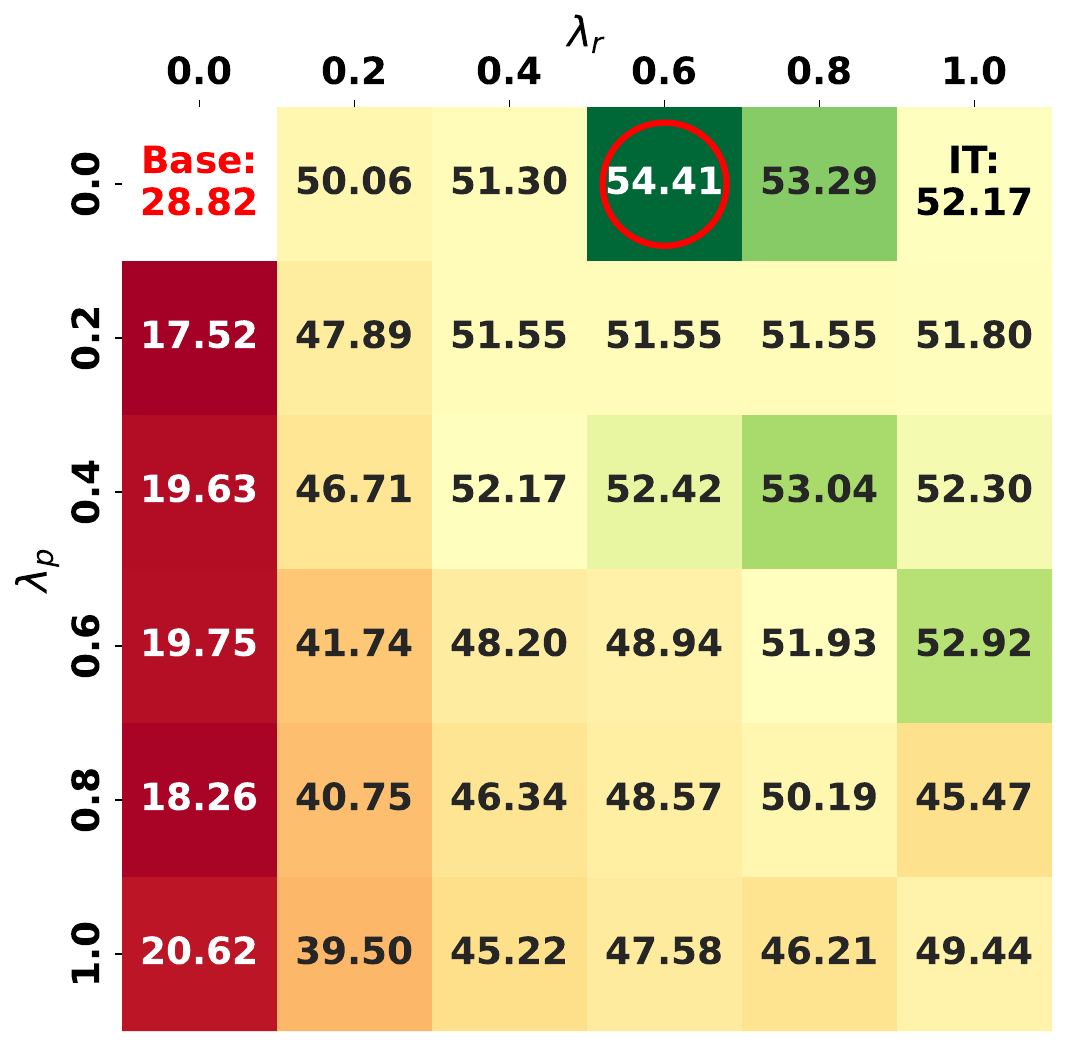} &
      \includegraphics[width=.19\textwidth]{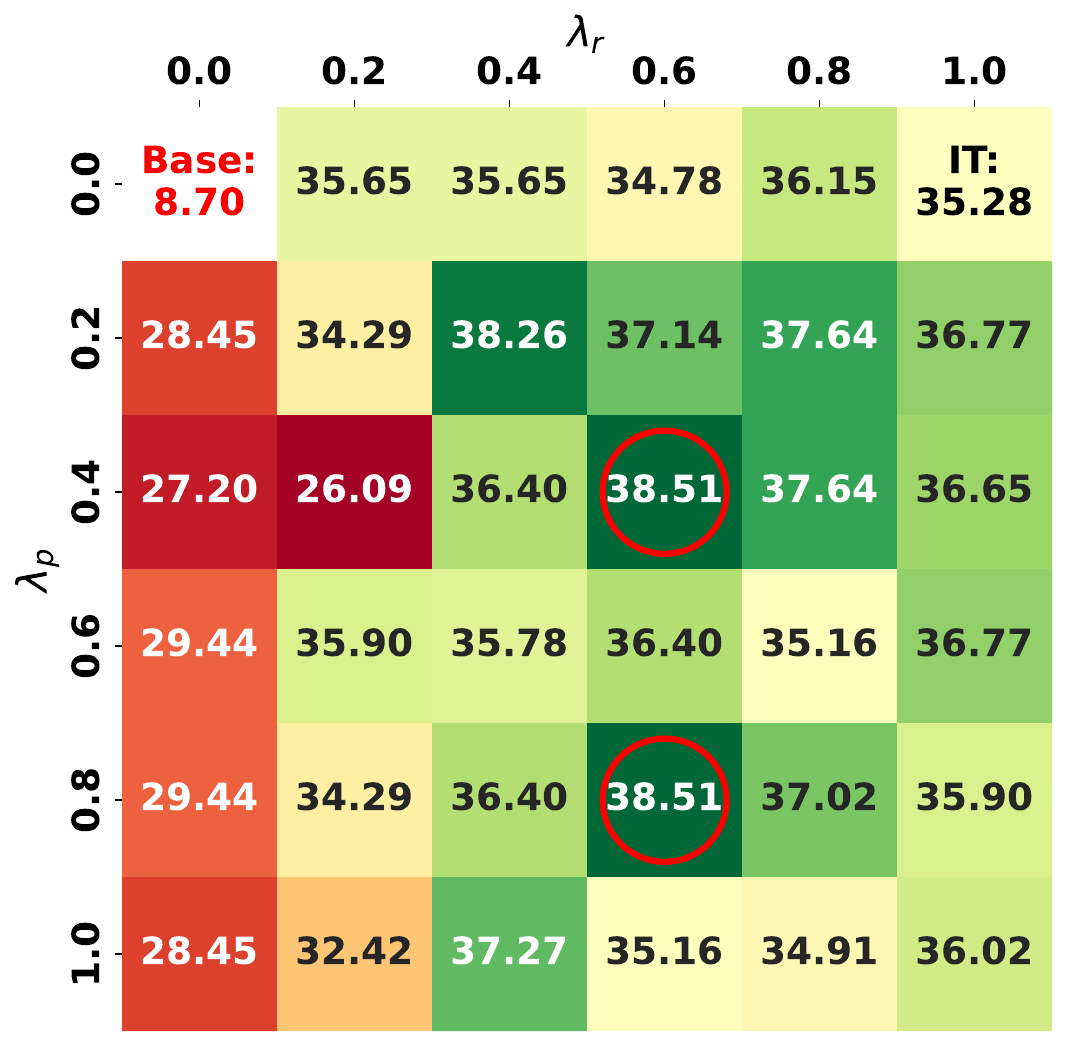} \\[4pt]

      \raisebox{1.3cm}[0pt][0pt]{\rotatebox{90}{\makebox[0pt][c]{IFEval}}} &
      \includegraphics[width=.19\textwidth]{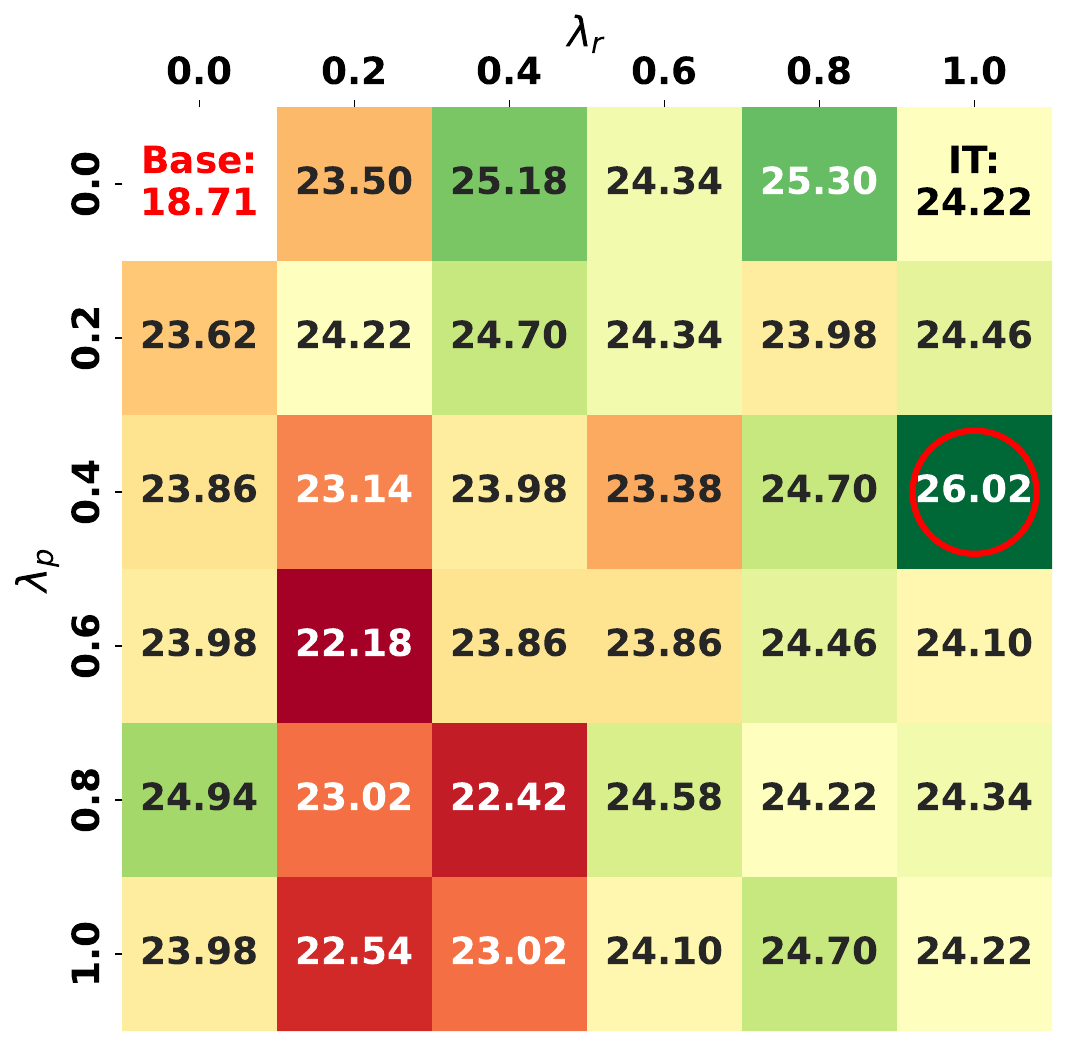} &
      \includegraphics[width=.19\textwidth]{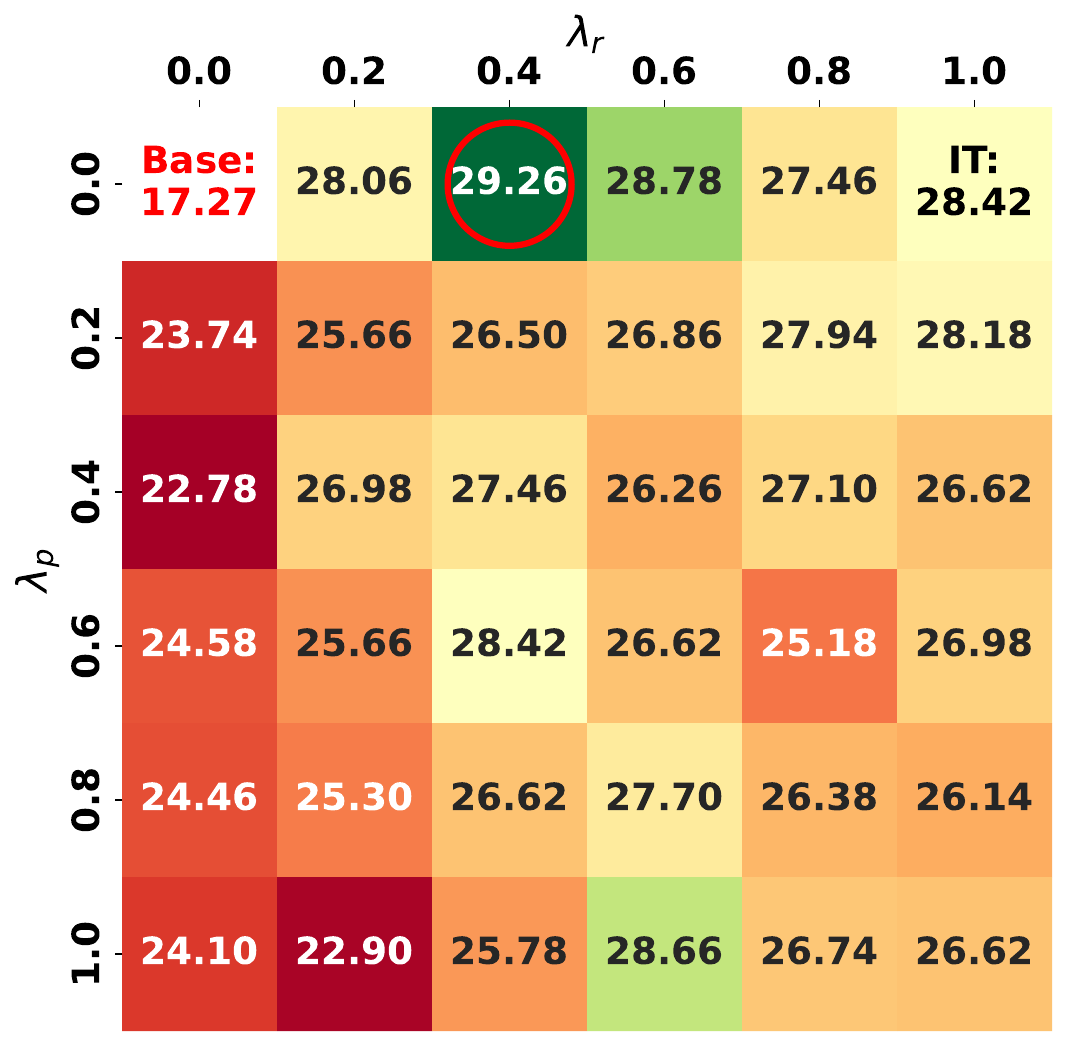} &
      \includegraphics[width=.19\textwidth]{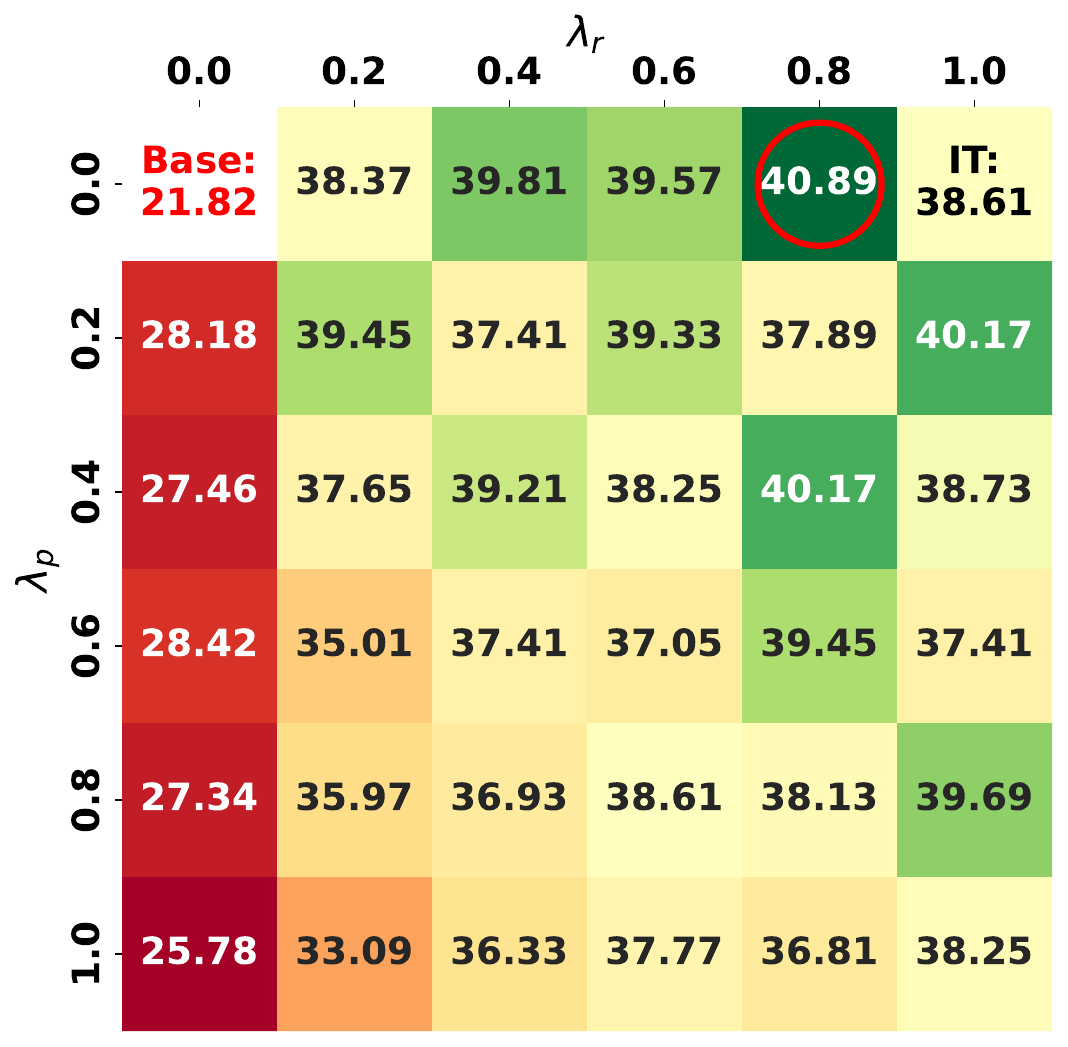} &
      \includegraphics[width=.19\textwidth]{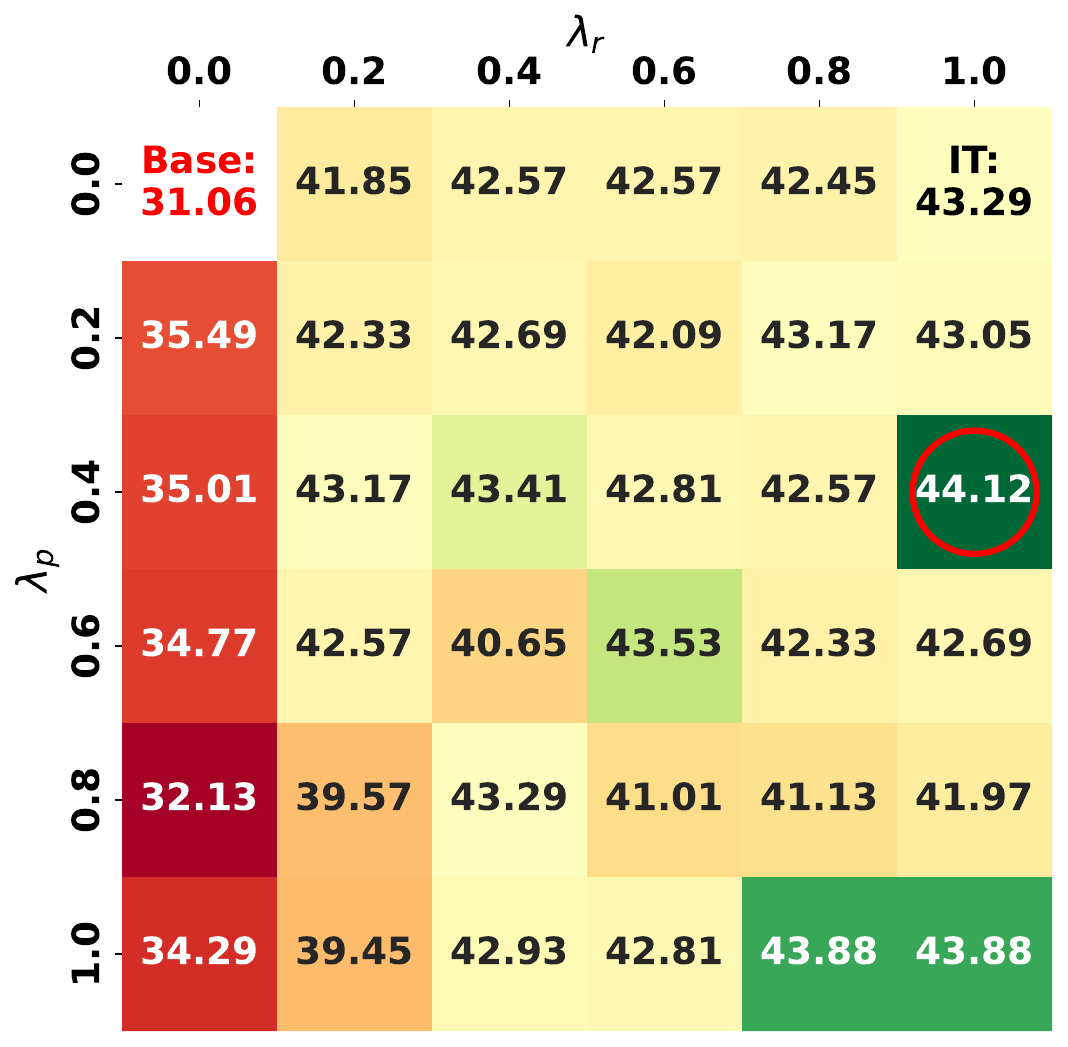} &
      \includegraphics[width=.19\textwidth]{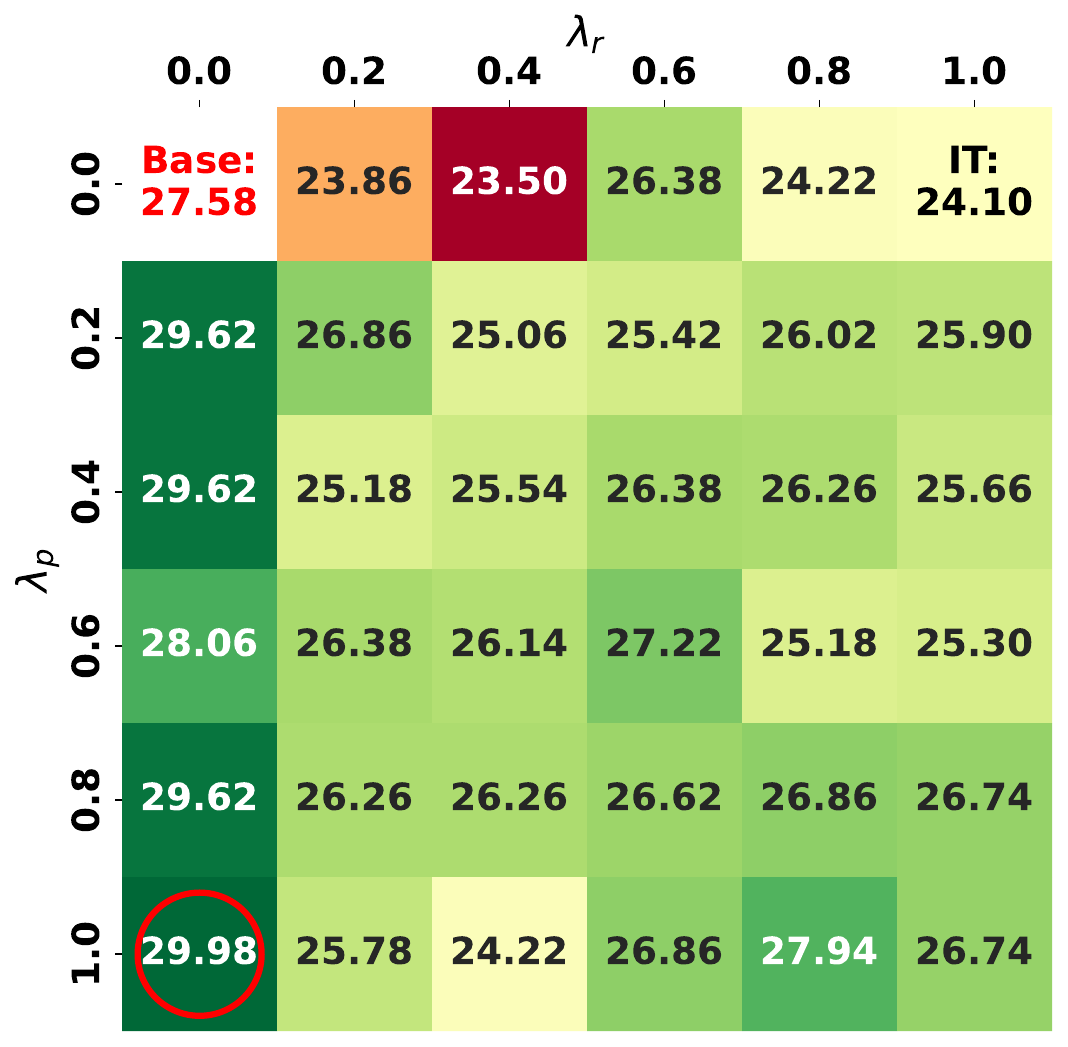} \\[4pt]

      \raisebox{1.3cm}[0pt][0pt]{\rotatebox{90}{\makebox[0pt][c]{MT-Bench}}} &
      \includegraphics[width=.19\textwidth]{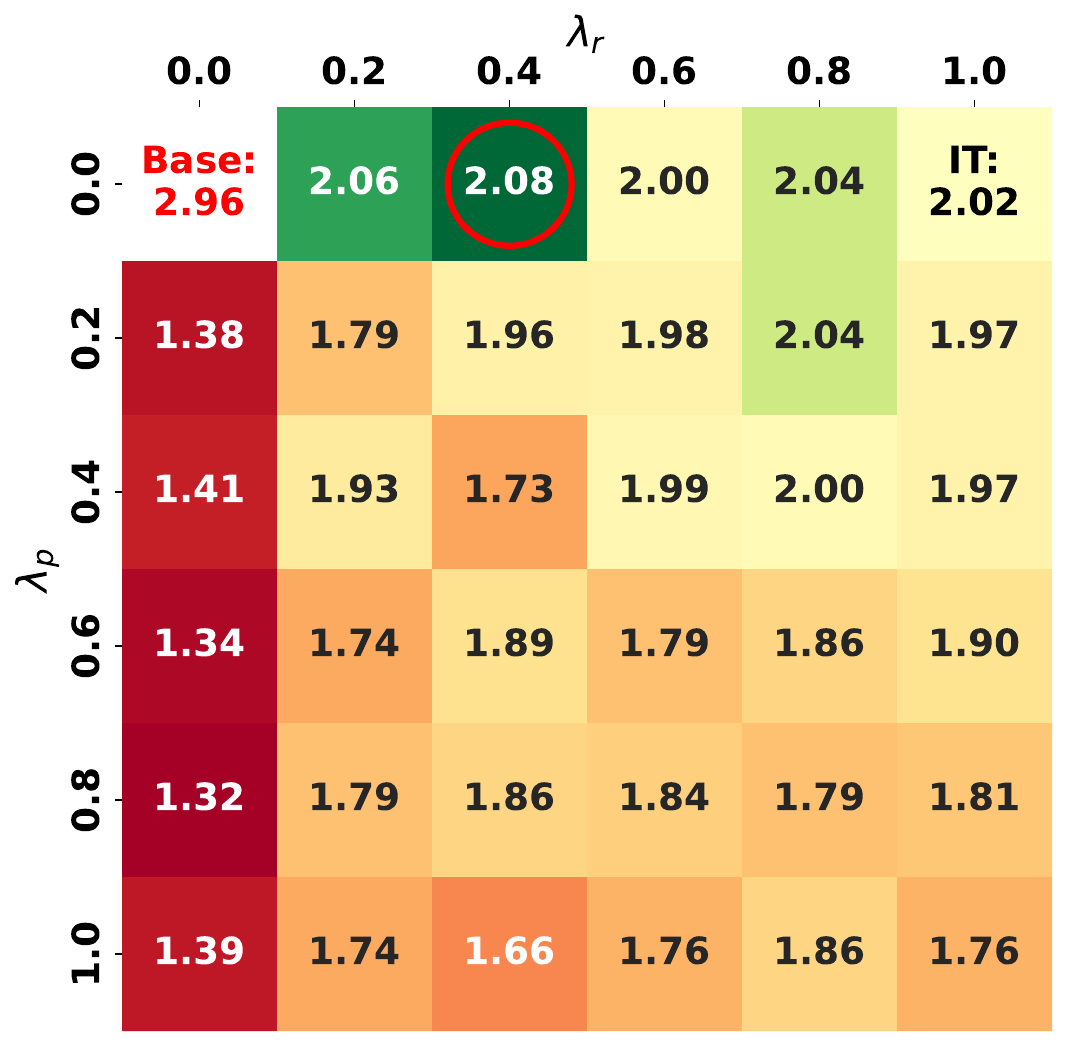} &
      \includegraphics[width=.19\textwidth]{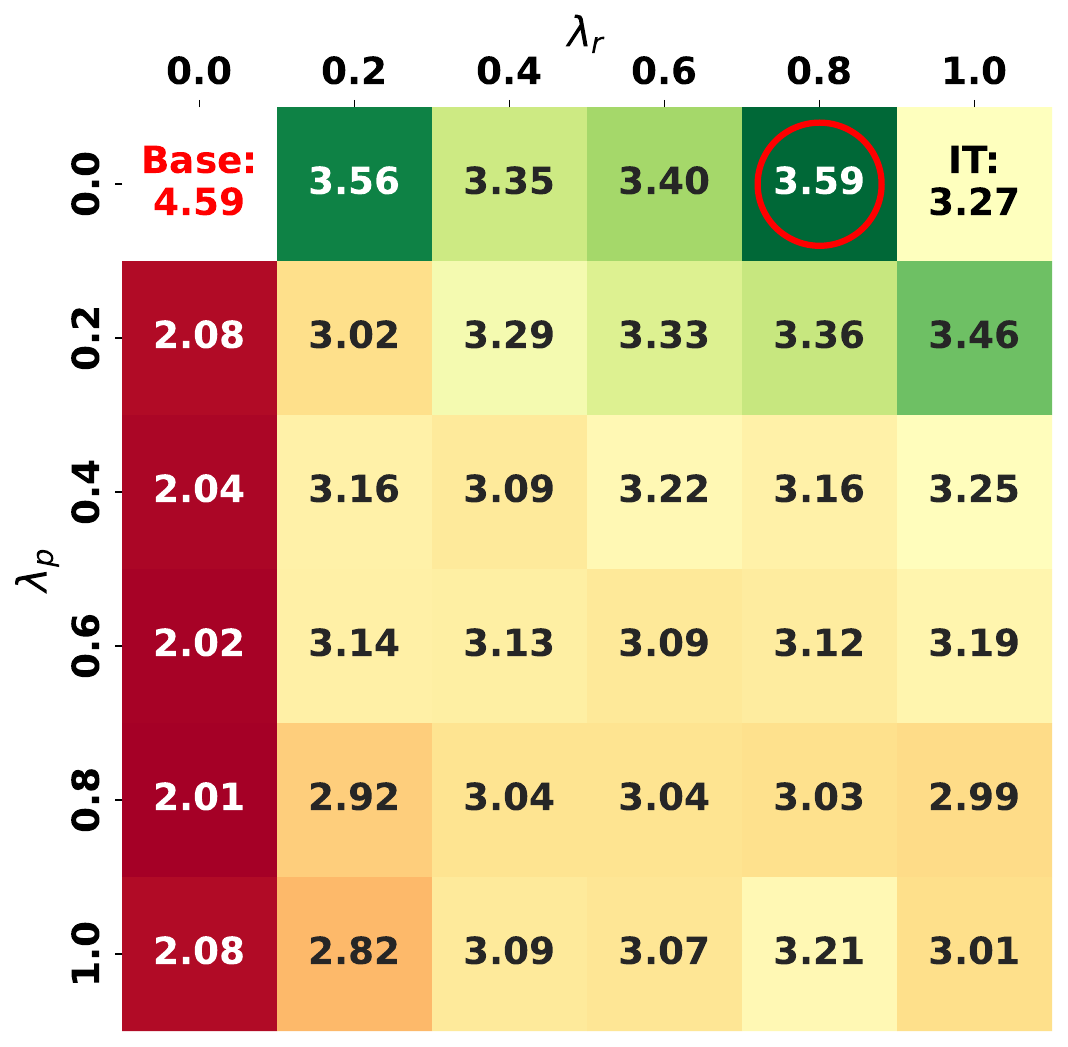} &
      \includegraphics[width=.19\textwidth]{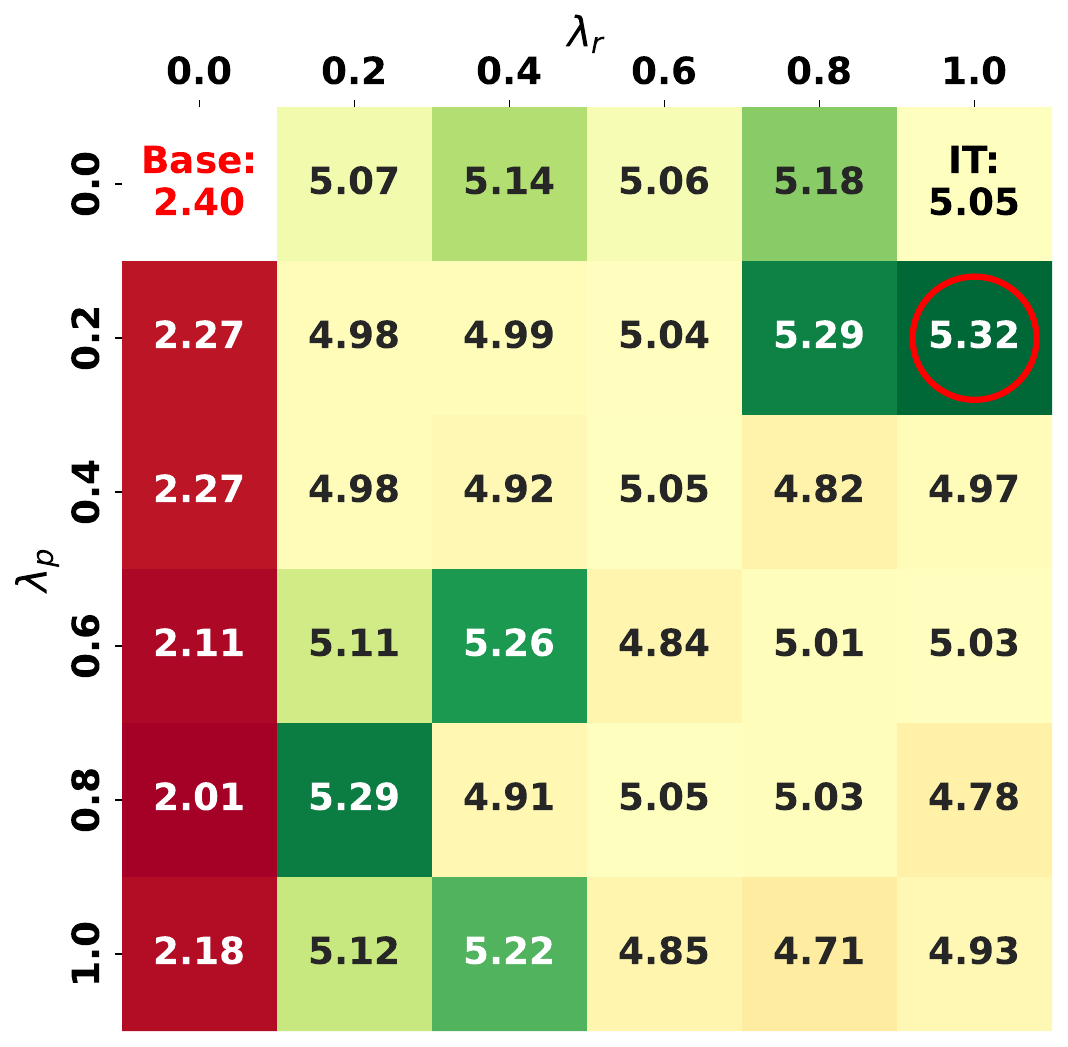} &
      \includegraphics[width=.19\textwidth]{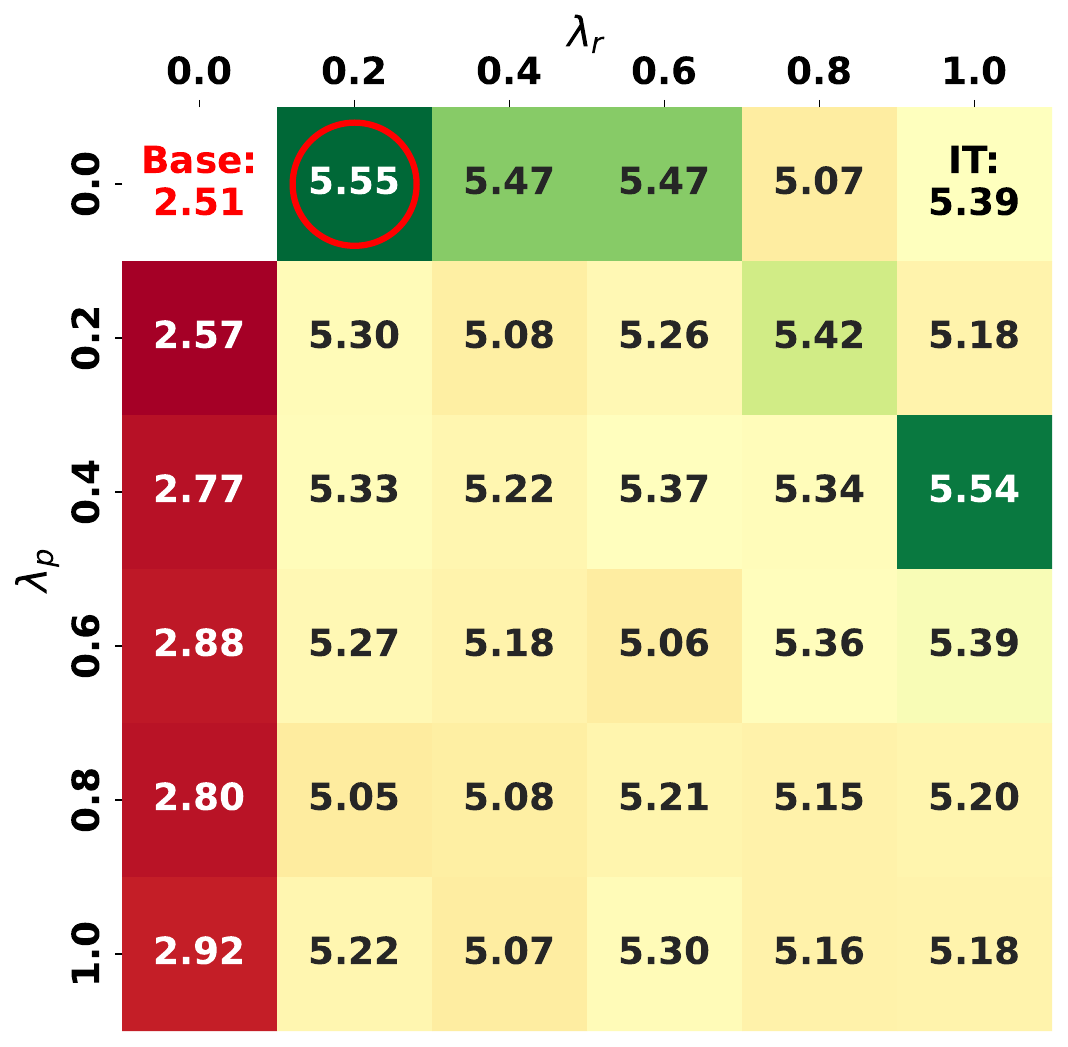} &
      \includegraphics[width=.19\textwidth]{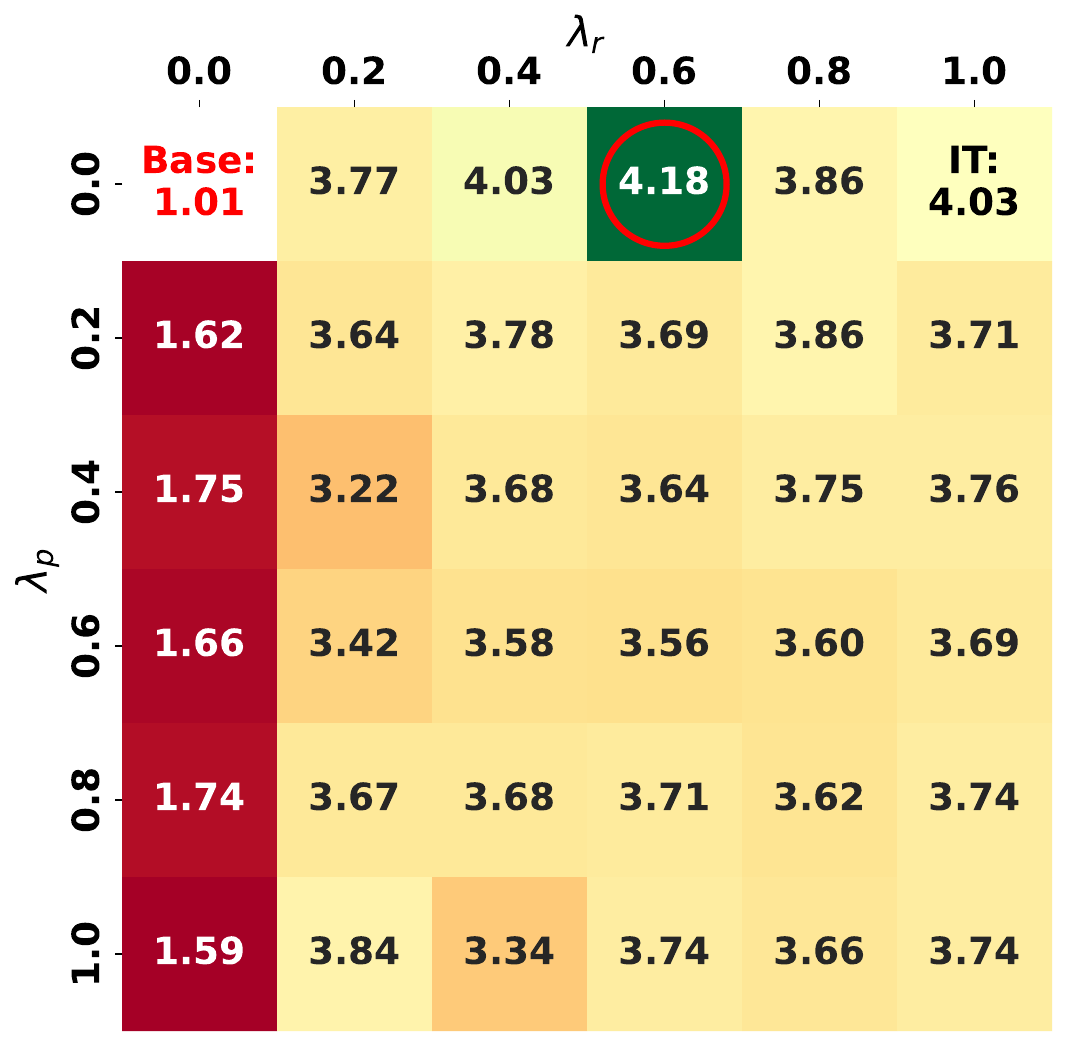} \\
  \end{tabular}

    \caption{Heatmaps depicting performance on MMLU (first row), BBH (second row), AlpacaEval (third row), IFEval (fourth row), and MT-Bench (fifth row) for different configurations of ($\lambda_p$, $\lambda_r$) and for different models finetuned on \textbf{LIMA}. In each heatmap, the best performance is highlighted with a red circle. The color map is based on relative gain with respect to conventional instruction tuning. Each row of a heatmap corresponds to a prompt-token weight, and each column corresponds to a response-token weight. Conventional instruction tuning is marked with \texttt{IT}, and base model performance is marked with \texttt{Base}.}
    \label{fig:lima_all}
\end{figure*}

\end{document}